\documentclass[10pt]{article} 
\usepackage[preprint]{tmlr}

\usepackage{amsmath,amsfonts,bm}

\def\eqref#1{equation~\ref{#1}}

\def\1{\bm{1}}

\DeclareMathAlphabet{\mathsfit}{\encodingdefault}{\sfdefault}{m}{sl}
\SetMathAlphabet{\mathsfit}{bold}{\encodingdefault}{\sfdefault}{bx}{n}

\DeclareMathOperator*{\argmax}{arg\,max}

\usepackage{hyperref}
\usepackage{url}
\usepackage{amsmath}
\usepackage{booktabs}
\usepackage{amsfonts}
\usepackage{mathrsfs} 
\usepackage{times}
\usepackage{latexsym}
\usepackage{graphicx}
\usepackage{tabularx}
\usepackage{makecell}
\usepackage{footmisc}
\usepackage{caption}
\usepackage{balance}
\usepackage{array}
\usepackage{adjustbox}
\usepackage{comment}
\usepackage{multicol}
\usepackage{diagbox}
\usepackage{multirow}
\usepackage{inconsolata}
\usepackage{graphicx}
\usepackage{color, colortbl}
\usepackage{threeparttable}

\usepackage{subcaption}
\usepackage{graphicx}

\usepackage[nolist,nohyperlinks]{acronym} 

\title{Analyzing the Effect of Noise in LLM Fine-tuning}

\def\month{MM}  
\def\year{YYYY} 
\def\openreview{\url{https://openreview.net/forum?id=XXXX}} 

\begin{document}
\begin{acronym}[nolist] 
\acro{LLMs}{Large Language Models}
\acro{MRR} {Mean Reciprocal Rank} 
\acro{CKA}{Centered Kernel Alignment}
\acro{LF} {Label Flip} 
\acro{TN}{Typographical Noise}
\acro{GN}{Grammatical Noise}
\acro{SC}{Sentiment Classification}
\acro{QA}{Question Answering}
\acro{MT}{Machine Translation}
\acro{KL}{Kullback-Leibler}
\acro{NLP}{Natural Language Processing}
\end{acronym}

\maketitle

\begin{abstract}
Fine-tuning is the dominant paradigm for adapting pretrained large language models (LLMs) to downstream \ac{NLP} tasks. In practice, fine-tuning datasets may contain various forms of noise arising from annotation errors, preprocessing artifacts, or automated data collection. While prior work has focused on designing robust learning algorithms to mitigate performance degradation under noisy conditions, comparatively little is known about how different types of noise affect the internal learning dynamics of LLMs during fine-tuning. In this work, we systematically study the impact of noise on model behavior across three pretrained model families (GPT-2, Qwen2 and Llama-2) and three diverse \ac{NLP} tasks. We introduce controlled perturbations corresponding to three common real-world noise types: label noise, grammatical noise, and typographical noise. Beyond task-level performance, we analyze layer-wise representation changes and attention patterns to understand how noise propagates through the network. Our results show that corrupting labels (i.e. label noise) consistently causes the largest performance degradation, whereas grammatical noise and typographical noise can occasionally yield mild regularization benefits. We further find that noise effects are localized primarily to task-specific layers, while attention structures remain comparatively stable. Our code is available here \footnote{\url{https://anonymous.4open.science/r/data-noise-influence-anonymous-383F}}. 
\end{abstract}

\section{Introduction} \label{sec:intro}
Fine-tuning has become a dominant paradigm for adapting pretrained language models to downstream \ac{NLP} tasks. Broadly speaking, fine-tuning implicitly assumes that the data used for training is reliable. In practice, training data is often noisy due to annotation errors, imperfect preprocessing pipelines, or automatic data collection methods such as web scraping and distant supervision \citep{label-noise-survey, ratner2017snorkel,zhang2025noisesurvey}.

For instance, classification datasets may contain incorrect labels, while generation or understanding tasks frequently include grammatical errors, spelling mistakes. Although a substantial body of research has studied robust learning under adversarial or corrupted data \citep{Patrini2017losscorrection,han2018coteaching,li2020dividemix}, most work focuses on designing algorithms that mitigate performance degradation. Much less is understood about how different types of noise influence the internal learning dynamics of LLMs during fine-tuning.

\begin{figure}[h]
    \centering
    \includegraphics[width=0.7\linewidth]{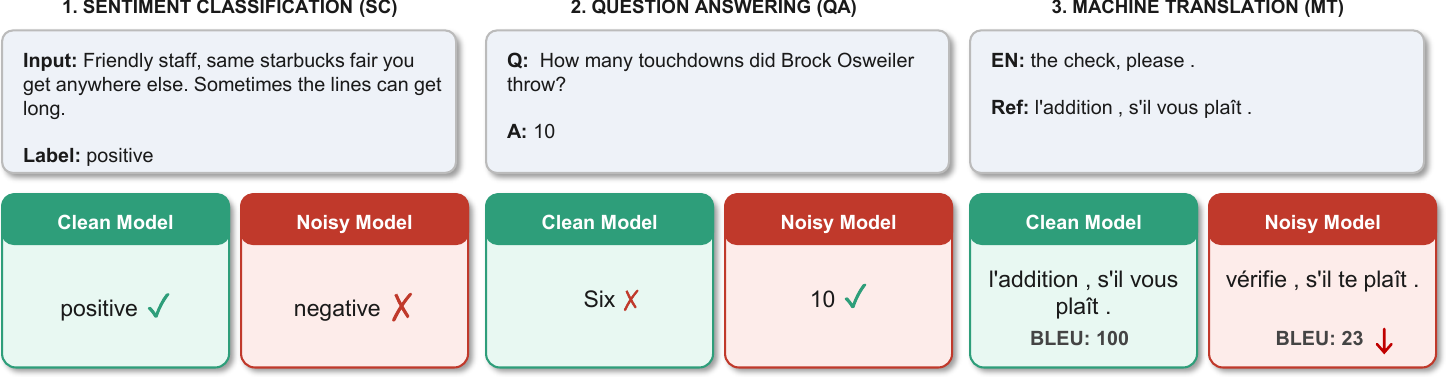}
    \caption{Examples of prediction changes in Llama-2 7B under different noise types at 40\% corruption ratio: label-flip (SC), typographical (QA), and grammatical (MT). }
    \label{fig:intropic}
\end{figure}
Modern language models generally have a large number of parameters, and fine-tuning often modifies only a small subset of representations responsible for task-specific behavior. Consequently, different noise sources may affect distinct parts of the model differently, potentially leading to performance degradation or unexpected improvements. 

To investigate the above-mentioned hypothesis, we systematically investigated the effect of noise during fine-tuning across multiple model families and tasks. We explored three widely used pretrained models GPT-2, Qwen2 and Llama-2, and evaluated them on three diverse \ac{NLP} tasks (i.e. \textbf{\ac{SC}}, \textbf{\ac{QA}} and \textbf{\ac{MT}}) to capture generalizable behavior.  Prior works   suggest that three categories of noise frequently arise in real-world text data \citep{survey1,zhang2025noisesurvey}. Consequently, we introduced controlled perturbations corresponding to three different noise categories: \textit{Label noise} \citep{label-noise-survey}, \textit{Typographical noise} \citep{karpukhin2019typo}, \textit{Grammatical noise} \citep{moradi2021grammar}. For all the above-mentioned noise types, we analyzed layer-wise representation changes and attention patterns to understand how noise propagates through the network. Our contributions can be summarized as follows.

\begin{itemize}
    \item \textbf{Label noise is the most harmful:} Across all tasks and models, label noise resulted in the largest performance degradation, whereas grammatical and typographical noise occasionally caused marginal performance improvements.

    \item \textbf{Noise impact is localized:} Layers that encode higher levels of task-specific information generally exhibit greater distortion when noise is introduced.
    \item  \textbf{Attention patterns remain stable:} Despite performance changes, token attention ordering changes only marginally.
\end{itemize}

Figure \ref{fig:intropic} illustrates the result of three different types of noise (i.e. label noise, typographical noise, grammatical noise) in different tasks (i.e. SC, QA, and MT). It can be seen from Figure \ref{fig:intropic} that although label noise and grammatical noise caused incorrect predictions, typographical noise actually improved the answer. The remainder of this paper is organized as follows. Section \ref{sec:related} reviews related work, Section \ref{sec:analysis} presents the methodology for analyzing the effects of noise, Section \ref{sec:setup} describes the experimental setup, and Section \ref{sec:results} reports the experimental results. Section \ref{sec:stratification} further stratifies the evaluation samples into robust and vulnerable groups to examine whether the representational changes observed under noise are uniform across all test samples. Finally, Section \ref{sec:conclusion} concludes the paper.


\section{Related Work}
\label{sec:related}
Prior research on learning in noisy settings can be broadly grouped into two areas: (a) the effects of noise in fine-tuning and (b) robust learning with noisy labels.

\subsection{Effect of Noise in Finetuning}
\label{sec:rw_mechanisms}

The study in \citet{liu2020early} identified the \emph{early-learning} phenomenon which says neural networks learn clean patterns before memorizing noisy labels, motivating early stopping as a regularizer. The work in \citet{tanzer2022memorisation} further refined this concept in BERT fine-tuning on noisy NER data, identifying three temporal phases fitting, settling, and memorization and showed that noisy samples drift in embedding space during the memorization phase. 
%
The work in \citet{chen2025basin} characterises the fine-tuning loss landscape as nested basins, where adversarial fine-tuning can escape the stability region of the pretrained model. The study in \citet{pac-bayesian} link flatter minima to greater noise resilience, and \citet{li2021improved} provides PAC-Bayes bounds relating layer-wise weight distance to generalization under noise. The study \citet{kim2024towards} found that parameter-efficient methods (LoRA, adapters, prompt tuning) are generally more robust than full fine-tuning under label noise at rates of $20-60\%$, attributing this to the low-rank bottleneck that limits memorization capacity. 

The above-mentioned works primarily analyze noisy learning from parameter \citep{kim2024towards} and loss-level perspectives \citep{pac-bayesian, chen2025basin}, typically focusing on classification tasks \citep{tanzer2022memorisation} and a single type of noise. In contrast, our work provides a complementary perspective by examining three types of noise at the representation level across widely used NLP tasks.


\subsection{Robust Learning with Noisy Labels}
\label{sec:rw_noisy_labels}

Classical approaches to robust learning under label noise can be broadly grouped into four categories. Loss correction methods model the noise transition matrix to correct the training objective \citep{Patrini2017losscorrection}. Robust loss design replaces cross-entropy with noise-tolerant alternatives such as MAE or generalized cross-entropy \citep{Ghosh_Kumar_Sastry_2017,NEURIPS2018_crossloss}. Sample selection methods leverage the memorization effect of deep networks, where clean samples consistently incur smaller losses early in training, to filter out unreliable examples \citep{han2018coteaching}. More recent work combines sample selection with semi-supervised learning, treating noisy-labelled instances as unlabelled data and jointly optimizing over both subsets \citep{li2020dividemix}. While effective for image classification, these methods focus on mitigating performance degradation, leaving largely unexplored the question of how different noise types reshape internal representations during fine-tuning. In a safety context, \citet{rosati2024representation} showed that harmful fine-tuning recovers latent harmful representations rather than creating new ones, consistent with the wrapper view.

Recently, with the advent of pretrained language models there has been a large body of work for training with noisy labels. The study \citet{zhu-etal-2022-bert} found that BERT is robust to synthetic label noise but degrades substantially under realistic, instance-dependent noise. \citet{wang2023laft} leveraged ChatGPT-generated rationales to separate clean from noisy samples during LLM fine-tuning. Their framework, LAFT, uses the agreement between the original noisy label and the LLM-predicted label as a confidence signal to partition training data into clean, ambiguous, and noisy subsets, each receiving different training strategies.
The study in \citet{luo2024robustft} extend noise-robust training to open-ended generation, moving beyond the classification setting.
Most relevant to our experimental design, \citet{MT-fine} show that in machine translation fine-tuning, \emph{target-side} noise (corrupted references) is substantially more damaging than \emph{source-side} noise (corrupted inputs) --- a finding that directly parallels our comparison of label corruption versus input-side (typographical, grammatical) noise. Similarly, the work in \citet{qi2024finetuning} demonstrate that as few as 10 adversarial examples can compromise safety alignment during fine-tuning. These studies focus on developing methods to \emph{resist} noise or on measuring its impact on \emph{output} performance.
Our work complements this line of research by asking a different question: not how to maintain accuracy under noise, but how noise reshapes the model's \emph{internal representations}.

\section{Methodology for Noise Analysis}
\label{sec:analysis}
A model that has similar, higher or lower task performance after being trained on noisy data may do so either by preserving its original task-specific representations (encoding obtained from trained on clean data) or by learning fundamentally different encoding strategies. Standard task-level metrics (e.g., accuracy, BLEU) cannot distinguish between these two scenarios. To disentangle these possibilities, we employ three complementary analysis methods that examine (i) how noise alters attention patterns, (ii) how noise affects task-relevant information encoded within the model, and (iii) how internal representations change before and after fine-tuning.

Throughout this section, we compare a clean model 
(fine-tuned on unperturbed data) with a noisy model (fine-tuned on corrupted data); noisy-model quantities are distinguished by a tilde. Subscripts index the layer~$\ell$ and sample~$s$; $\mathcal{S}$ denotes the evaluation set.


\begin{table}[t]
\caption{Summary of different metrics used to analyze the effect of noise. All metrics are computed at every layer~$\ell$.}
    \label{tab:analysis_summary}
\centering
\resizebox{0.5\columnwidth}{!}{
\begin{tabular}{l c}
    \toprule
    \textbf{Metric} & \textbf{Analysis Aspect} \\
    \midrule
    $\overline{D}_{KL}(\ell) $ & Attention value divergence  \\
    $\bar{\rho}_k(\ell)$ & Attention priority order  \\
    $\mathrm{Acc}_\ell$ & Task-aligned information (classification) \\
   $\mathrm{MRR}_\ell$ & Task-aligned information (generation) \\
  $\mathrm{Cos}_{s,\ell}$ & Per-sample directional shift  \\
   $\mathrm{CKA}(\mathbf{{H}_\ell}, \tilde{\mathbf{H}}_\ell)$ & Inter-sample structural change \\
    \bottomrule
    \end{tabular}}
\end{table}

 \subsection{Attention Matrix Analysis}
  \label{sec:attn_kl}
To examine whether noise alters the attention pattern, we conduct two complementary analyses. One focuses on  a) \textit{attention values} and the other focuses on b) \textit{the order of tokens} (i.e. tokens having the highest attention to the lowest attention). 

To understand the change in values within attention matrices, we computed the \ac{KL} divergence between the clean and noisy attention distributions, averaged over all samples, attention heads, and token positions at each layer. Let $\mathbf{a}_{h,t}^{(s)}$ and $\tilde{\mathbf{a}}_{h,t}^{(s)}$ denote the clean and noisy attention weight vectors for head~$h$, token position~$t$, and sample~$s$, with $H$ heads per layer and $T_s$ tokens per sample. Formally, 

\begin{equation}
\label{eq:attention kl}
\overline{D}_{\mathrm{KL}}(\ell) \;=\; 
  \frac{1}{|\mathcal{S}|} \sum_{s \in \mathcal{S}} \;
  \frac{1}{H} \sum_{h=1}^{H} \;
  \frac{1}{T_s} \sum_{t=1}^{T_s}
  D_{\mathrm{KL}}\!\bigl(\mathbf{a}_{h,t}^{(s)} \;\big\|\; \tilde{\mathbf{a}}_{h,t}^{(s)}\bigr)
\end{equation}


 A high $\overline{D}_{\mathrm{KL}}(\ell)$ indicates that noise has substantially altered how layer~$\ell$ distributes contextual importance across tokens.

While \ac{KL} divergence captures changes in attention magnitude,  it does not reveal how much the order of important tokens in a particular context have changed due to noise. To investigate the above mentioned phenomena, we also computed Spearman rank correlation coefficient between clean and noisy attention weights 
, restricted to the top-$k$ tokens for each token position ($t$), and averaged identically to Equation~\ref{eq:attention kl}.Formally,

\begin{equation}
\label{eq:rho}
\overline{\rho}_k(\ell) \;=\; 
  \frac{1}{|\mathcal{S}|} \sum_{s \in \mathcal{S}} \;
  \frac{1}{H} \sum_{h=1}^{H} \;
  \frac{1}{T_s} \sum_{t=1}^{T_s}
  \rho\!\bigl(\mathbf{a}_{h,t}^{(s)},\; 
              \tilde{\mathbf{a}}_{h,t}^{(s)}\bigr)
\end{equation}
High $\overline{\rho}_k$ with high \ac{KL} indicates attention value redistribution without priority change whereas high \ac{KL} with low $\overline{\rho}_k$ indicates major reordering of attention targets. 

The metrics used for attention matrix analysis complement the hidden-state analyses (outlined in Section \ref{sec:fixed_ruler} and \ref{sec:cosine}) by revealing whether
representational changes originate from altered attention patterns or from
transformations within the feed-forward sublayers.

\subsection{Probing}
\label{sec:fixed_ruler}

To estimate whether task-relevant information remains encoded within different layers of LLMs trained on noisy data in a similar way as the clean model, we employ two different probing strategies based on the task considered. 

\paragraph{Probing Strategy for SC} We utilize a linear classifier-based probing strategy similar to \citep{linearprobing}. We train a linear classifier on each layer of the fine-tuned LLM. The classifier uses the hidden layer representations of the fine-tuned LLMs as input. The accuracy of these classifiers ($\mathrm{Acc}_\ell$) can give an estimate of how much task-specific information is encoded into that layer. The objective of probing on each layer of LLMs is to investigate whether the distribution of task-aligned understanding in the noisy model is similar to the clean model. The classifier further indicates whether layer-wise representations are similar between clean and noisy models, providing insight into whether noise effects are localized. 

Linear classifier probing can only be applied to tasks where straightforward classification is applicable. However, for generative tasks (e.g., question answering and machine translation), this probing paradigm is not directly applicable, as the output space consists of token sequences rather than discrete class labels. Consequently, we implement Logit Lens \citep{nostalgebraist2020logitlens}  based prediction for probing in question answering and machine translation, which is similar to \citep{geva-etal-2023-dissecting} and \citep{jiang-etal-2024-large}. 

\paragraph{Probing Strategy for QA \& MT}
\label{sec:logit_lens}
%
%

For generative tasks, we apply the Logit Lens probe at each layer~$\ell$: the hidden-state representation is passed through the final layer normalization and the language model head to obtain a probability distribution over the vocabulary. We then compute 
the \ac{MRR} of the first target token. Formally,

\begin{equation}
\label{eq:mrr}
\mathrm{MRR}_\ell \;=\; 
  \frac{1}{|\mathcal{S}|} \sum_{s \in \mathcal{S}} 
  \frac{1}{\operatorname{rank}_\ell(y_s)}
\end{equation}

where $y_s$ is the first target token for sample~$s$ and $\operatorname{rank}_\ell(y_s)$ is its rank in the vocabulary distribution at layer~$\ell$. A high $\mathrm{MRR}_\ell$ indicates that the representation at that layer already encodes sufficient information 
to predict the correct output. Results using the average \ac{MRR} over the first five target tokens show consistent patterns in \autoref{app:five-mrr}.


\subsection{Similarity Based on Input Representations}
\label{sec:cosine}
To further quantify the change in input representations under noise, two different kinds of similarity measures were computed. The first one is the centered cosine similarity between the hidden states of the clean and noise-trained models for each evaluation sample at each layer (i.e. $\mathrm{Cos}_{s,\ell}$). 
Both the mean and standard deviation across all S evaluation samples are reported.

While cosine similarity measures individual vector alignment, it is insensitive to changes in encoding layer patterns across samples. To capture this, Linear \ac{CKA} \citep{cka} is computed between the representations of clean and noisy models.
\begin{table*}[t]
\caption{Supervised fine-tuning task-related performance of different models under different noise ratios (\ac{SC}: Accuracy; \ac{QA}: F1 score; \ac{MT}: BLEU score)}
\label{tab:sft-performance}
\centering
\resizebox{\textwidth}{!}{
\begin{tabular}{l l cc ccc ccc ccc}
\toprule
& & \multicolumn{2}{c}{\textbf{Baselines}} & \multicolumn{3}{c}{\textbf{Label Flip}} & \multicolumn{3}{c}{\textbf{Typo Error}} & \multicolumn{3}{c}{\textbf{Gramm. Error}} \\
\cmidrule(lr){3-4} \cmidrule(lr){5-7} \cmidrule(lr){8-10} \cmidrule(lr){11-13}
\textbf{Task} & \textbf{Model} & \textbf{Pretrained} & \textbf{Clean FT} & 20\% & 30\% & 40\% & 20\% & 30\% & 40\% & 20\% & 30\% & 40\% \\
\midrule
\multirow{3}{*}{Sentiment} & GPT-2 & 0.12 & 91.33 & 90.72 & 88.82 & 75.51 & 91.72 & 91.32 & 90.91  & 91.43 & 91.32 & 91.41 \\
& Qwen2 & 70.00 & 94.41 & 90.71 & 81.62 & 57.72 & 94.21 & 94.02 & 94.21 & 94.22 & 94.51 & 94.33 \\
& Llama-2 & 1.13 & 94.12 & 91.62 & 95.12 & 85.13 & 94.00 & 94.42 & 94.11 & 94.42 & 94.21 & 93.81 \\
\midrule
\multirow{3}{*}{QA} & GPT-2 & 8.11 & 35.51 & 33.68 & 32.64 & 31.58 & 35.28 & 34.93 & 43.97 & 35.84 & 35.72 & 35.72 \\
& Qwen2 & 41.02 & 68.00& 60.91& 51.9 & 49.72 & 70.01 & 69.63 & 69.62 & 70.71 & 70.51 & 70.63 \\
& Llama-2 & 20.00 & 85.72 & 79.00 & 52.2 & 73.93 & 86.91 & 86.12 & 83.71 & 85.21 & 86.03 & 86.53 \\
\midrule
\multirow{3}{*}{MT} & GPT-2 & 0.31 & 5.08 & 3.79 & 2.47 & 1.84  &4.99 & 4.91 & 4.79 & 5.06 & 5.01 & 5.00 \\
& Qwen2 & 18.34 & 44.80 & 43.52 & 42.44 & 42.68 & 45.25 & 45.38 & 44.48 &  44.01 & 44.92 & 44.18 \\
& Llama-2 & 16.57 & 51.95 & 51.86 & 49.95 & 50.37 &53.11 & 51.72 & 52.04 & 52.07 & 51.87 & 52.66 \\
\bottomrule
\end{tabular}}
\end{table*}

\begin{figure*}[t]
  \centering
\includegraphics[width=\textwidth]{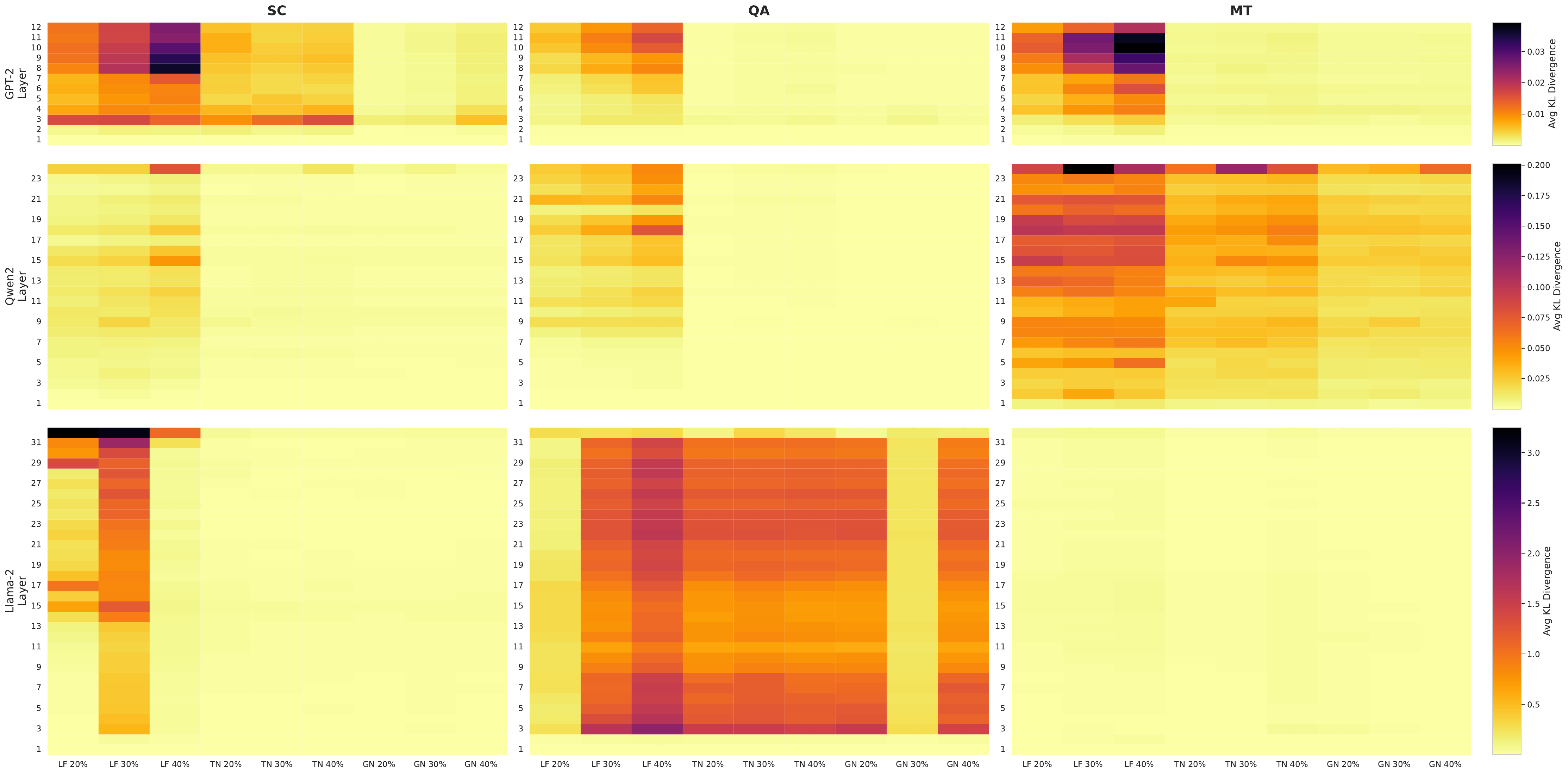}
  \caption{Layer-wise attention pattern divergence (\ac{KL} divergence)
  between clean and noise-trained models. Rows correspond to
  GPT-2, Qwen2, and Llama-2; columns correspond to SC, QA, and MT tasks.
  The x-axis denotes noise type and corruption ratio; the y-axis indicates the layer index of each model. Each cell shows the \ac{KL} divergence averaged across all attention
  heads at a given layer.  Each row uses an independent colour scale
  due to differing divergence magnitudes across architectures.}
  \label{fig:kl_attention}
\end{figure*}

For $N$ evaluation samples, let 
$\mathbf{H}_\ell \in \mathbb{R}^{N \times d}$ and 
$\tilde{\mathbf{H}}_\ell \in \mathbb{R}^{N \times d}$ denote the centered hidden-state matrices from the clean and noise-trained models at a given layer $\ell$, respectively. Linear CKA is defined as 
\begin{equation}
\scriptsize
    \mathrm{CKA}(\mathbf{{H}_\ell}, \tilde{\mathbf{H}}_\ell) \;=\; \frac{\bigl\|\tilde{\mathbf{H}}_\ell^\top \mathbf{H}_\ell\bigr\|_F^2}
    {\bigl\|\mathbf{H}_\ell^\top \mathbf{H}_\ell\bigr\|_F \;\cdot\; 
    \bigl\|\tilde{\mathbf{H}}_\ell^\top \tilde{\mathbf{H}}_\ell\bigr\|_F}
    \label{eq:cka}
\end{equation}

where $|\cdot|_F$ denotes the Frobenius norm \footnote{\url{https://en.wikipedia.org/wiki/Matrix_norm\#Frobenius_norm}}. Crucially, \ac{CKA} is invariant to orthogonal transformations and isotropic scaling.
If noise training merely rotated the representation space without altering its internal geometry, \ac{CKA} would remain near 1.0 regardless of how low the cosine similarity drops. Conversely, a low \ac{CKA} value provides definitive evidence that the inter-sample relational structure has been fundamentally altered, a change that cosine similarity alone cannot detect. 

\begin{figure*}[htb]
  \centering
\includegraphics[width=\textwidth]{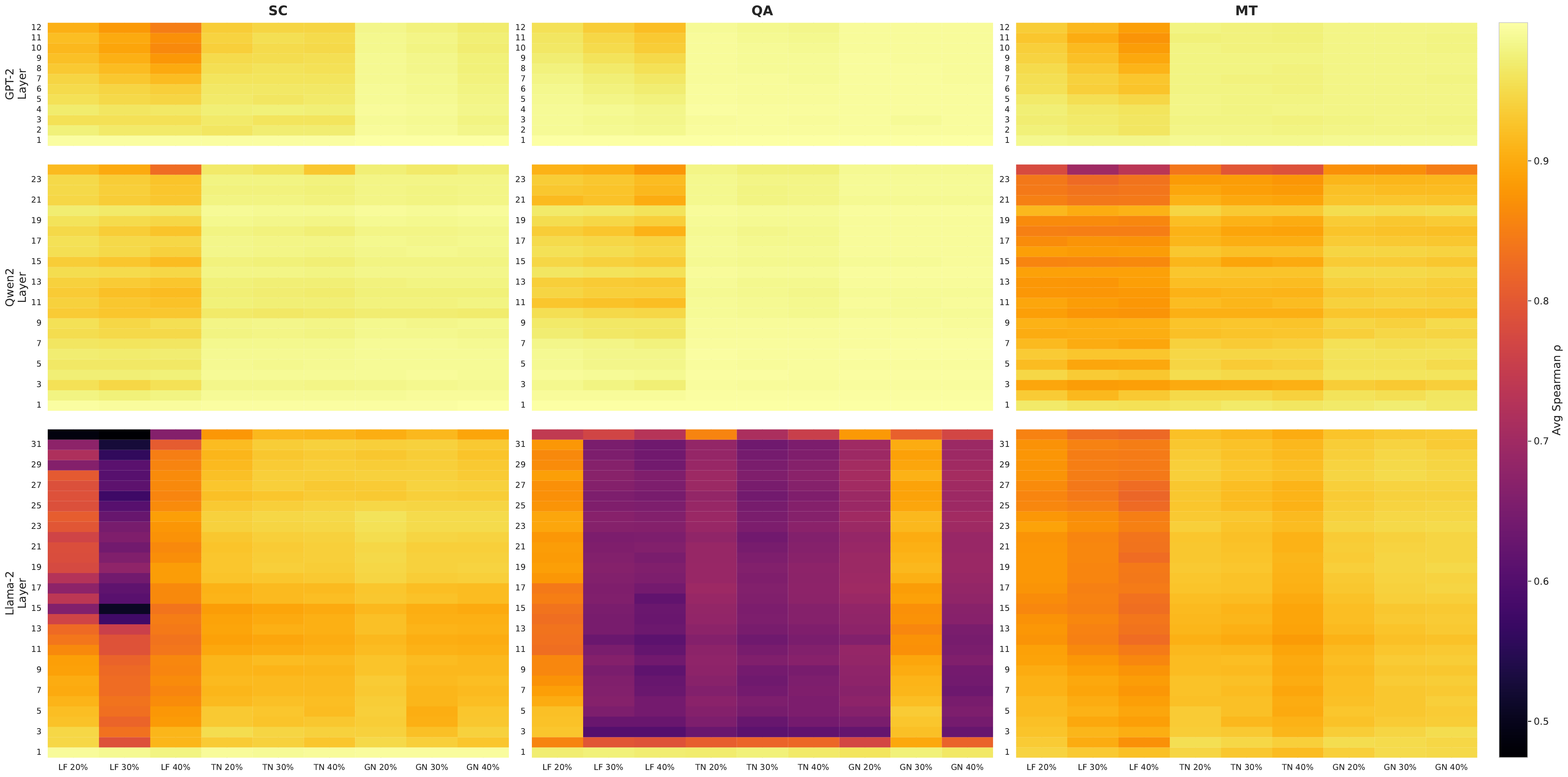}
  \caption{Layer-wise attention pattern stability measured by Spearman rank correlation ($\rho$) between clean and noise-trained models.} 
  \label{fig:spearman_attention}

\end{figure*}

Table~\ref{tab:analysis_summary} summarizes the metrics discussed above in analyzing noise effects across three tasks and three models.

\section{Experimental Setup}
\label{sec:setup}
Here we describe the details of dataset setup, followed by noise incorporation mechanism, fine-tuning setup and evaluation metrics.

\paragraph{Tasks \& Datasets} As described in Section \ref{sec:intro}, we explored three different NLP tasks: a) sentiment classification (SC), b) question answering (QA), and machine translation (MT), all framed as generative tasks. Existing
research \cite{nlpresearch1, nlpresearch2} showed that the three tasks mentioned above
are among the most widely used NLP tasks. 
For SC, we considered binary sentiment classification on movie reviews from Yelp Polarity
\citep{yelp_polarity} dataset. 
For QA the task is given a passage and a question, and the relevant portion is extracted from the passage as the answer. We used SQuAD v1.1 \citep{squad} dataset for this task. For MT, we focused on English-to-French translation from Tatoeba parallel
corpus \citep{tiedemann2020tatoeba}. 
Further dataset details (e.g. sample example from the dataset) are provided in \autoref{tab:dataset-stats}
(\autoref{app:prompts}).

\paragraph{Noise Types}
\label{sec:noise_types}
We explored three different noise types in our experiment setup a) \textit{Label Flip} (LN), b) \textit{Typographical Noise} (TN) and c) \textit{Grammatical Noise} (GN). Existing research \cite{survey1, unknown} shows that the above-mentioned three different types of noise primarily cover most of the widely observed noise in NLP tasks. Each one of them is described as follows.

\begin{itemize}
    \item \textbf{LF} Here, only the target output is corrupted.
For \ac{SC} this flips the polarity label (positive $\leftrightarrow$ negative). For QA the gold answer is replaced with a randomly sampled answer from the training pool, and for MT the reference translation is replaced with an unrelated target sentence.

\item \textbf{TN} Here, character-level perturbations (e.g. deletion, swap, insertion, or substitution of a single character) to approximately $10\%$ of words in the \emph{input} text (review, context, or source sentence), are applied to simulate common typing errors similar to \cite{gao2018blackboxgenerationadversarialtex}.

\item  \textbf{GN} Here, rule-based substitutions targeting subject--verb agreement
(\emph{is}$\leftrightarrow$\emph{are}, \emph{was}$\leftrightarrow$\emph{were},
\emph{has}$\leftrightarrow$\emph{have}) and article misuse (e.g., \emph{an apple}~$\to$~\emph{a apple}) into the input text at a rate of approximately $15\%$ word, similar to \cite{moradi2021grammar}.

\end{itemize}

 Prior research \cite{angluin1988learning,10.5555/2999611.2999745} shows that classical noise-robust learning algorithms require the noise rate to remain below 0.5. Following prior studies in noisy text classification \cite{liu-etal-2022-noise}, which find that many methods degrade significantly beyond 30\% noise, we evaluate three noise levels: 20\%, 30\%, and 40\%, covering the range from moderate to near-critical noise conditions.  Examples of the above-mentioned types of noise are given in Appendix \ref{app:noise-examples}. 

\paragraph{Fine-tuning Setup}
\label{sec:models} We fine-tuned three different LLMs (i.e. GPT-2 124 \citep{radford2019language}, Qwen2-0.5B \citep{qwen2}, Llama-2-7B \citep{touvron2023llama2openfoundation} ) in our experiment setup. Out of three models only GPT-2 fine-tuning was done on the full set of parameters. We applied QLoRA \citep{dettmers2023qlora} with bit NF4 quantization-based fine-tuning for both Qwen2-0.5B and Llama-2-7B. For both QLoRA models, LoRA adapters are applied to the attention and feed-forward projection layers. The specific LoRA rank, scaling factor $\alpha$, and target modules vary by task and are listed in detail in \autoref{tab:lora-config} (\autoref{app:hyperparameters}).

\begin{figure*}[t]
  \centering
  \begin{subfigure}[t]{0.32\textwidth}
    \includegraphics[width=\textwidth]{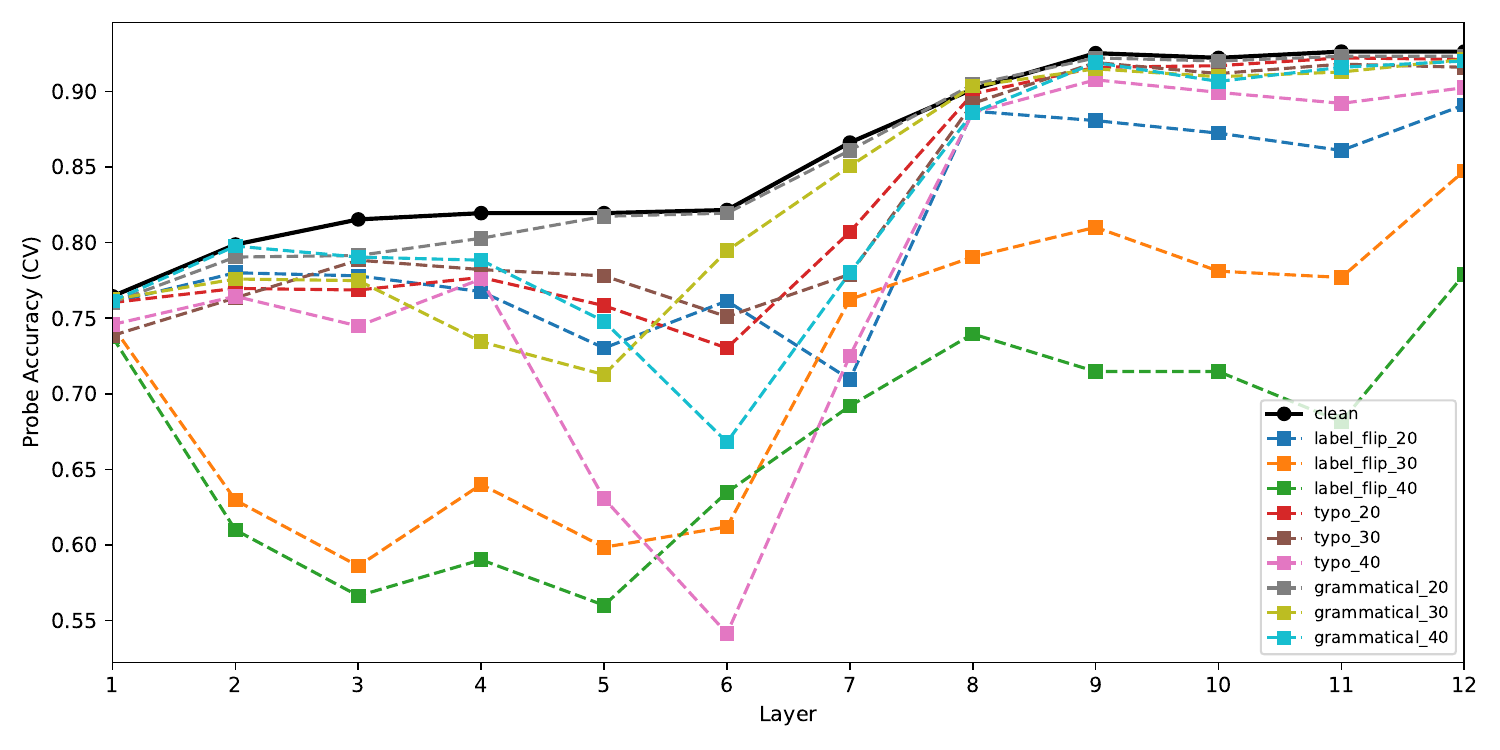}
    \caption{GPT2-SC}
  \end{subfigure}
  \hfill
  \begin{subfigure}[t]{0.32\textwidth}
    \includegraphics[width=\textwidth]{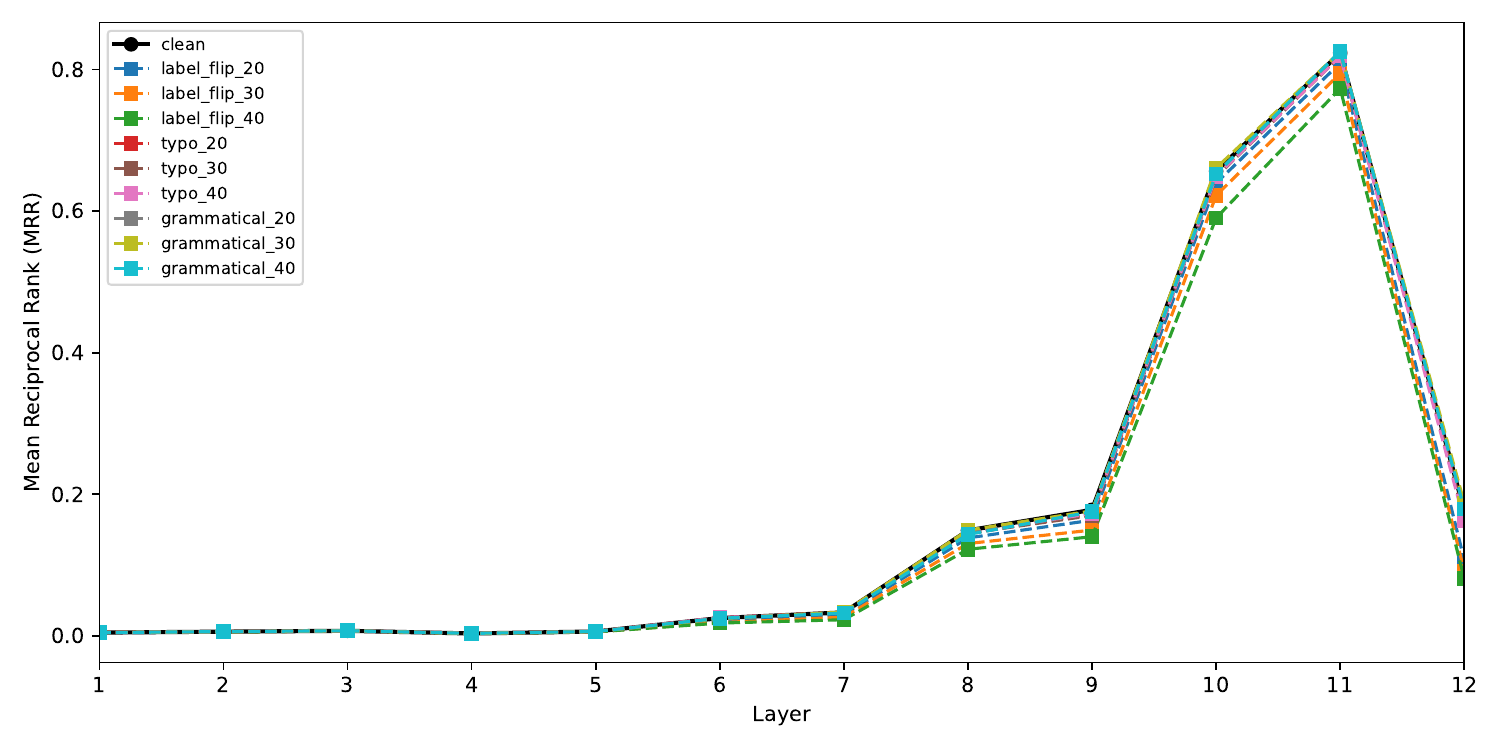}
    \caption{GPT2-QA}
  \end{subfigure}
  \hfill
  \begin{subfigure}[t]{0.32\textwidth}
    \includegraphics[width=\textwidth]{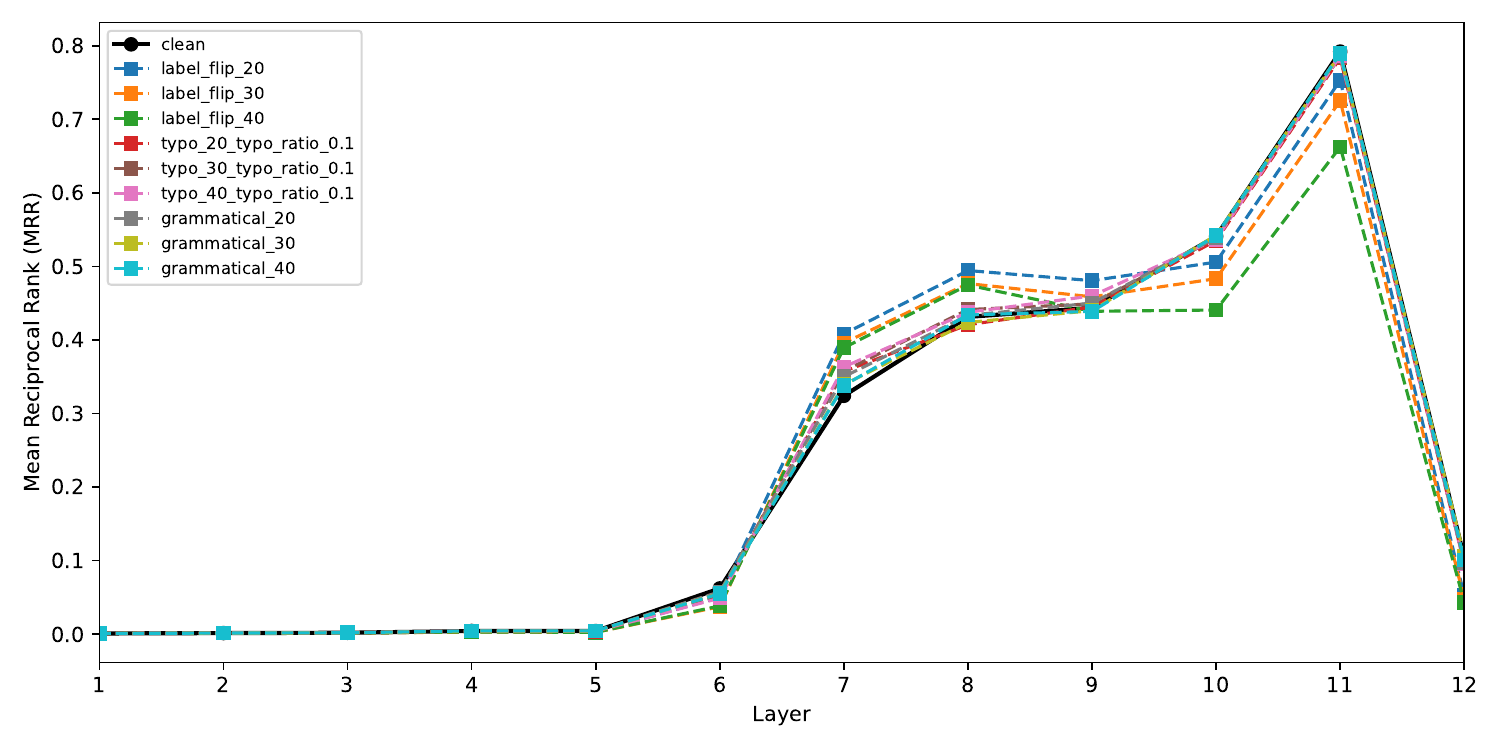}
    \caption{GPT2-MT}
      \end{subfigure}
    \begin{subfigure}[t]{0.32\textwidth}
    \includegraphics[width=\textwidth]{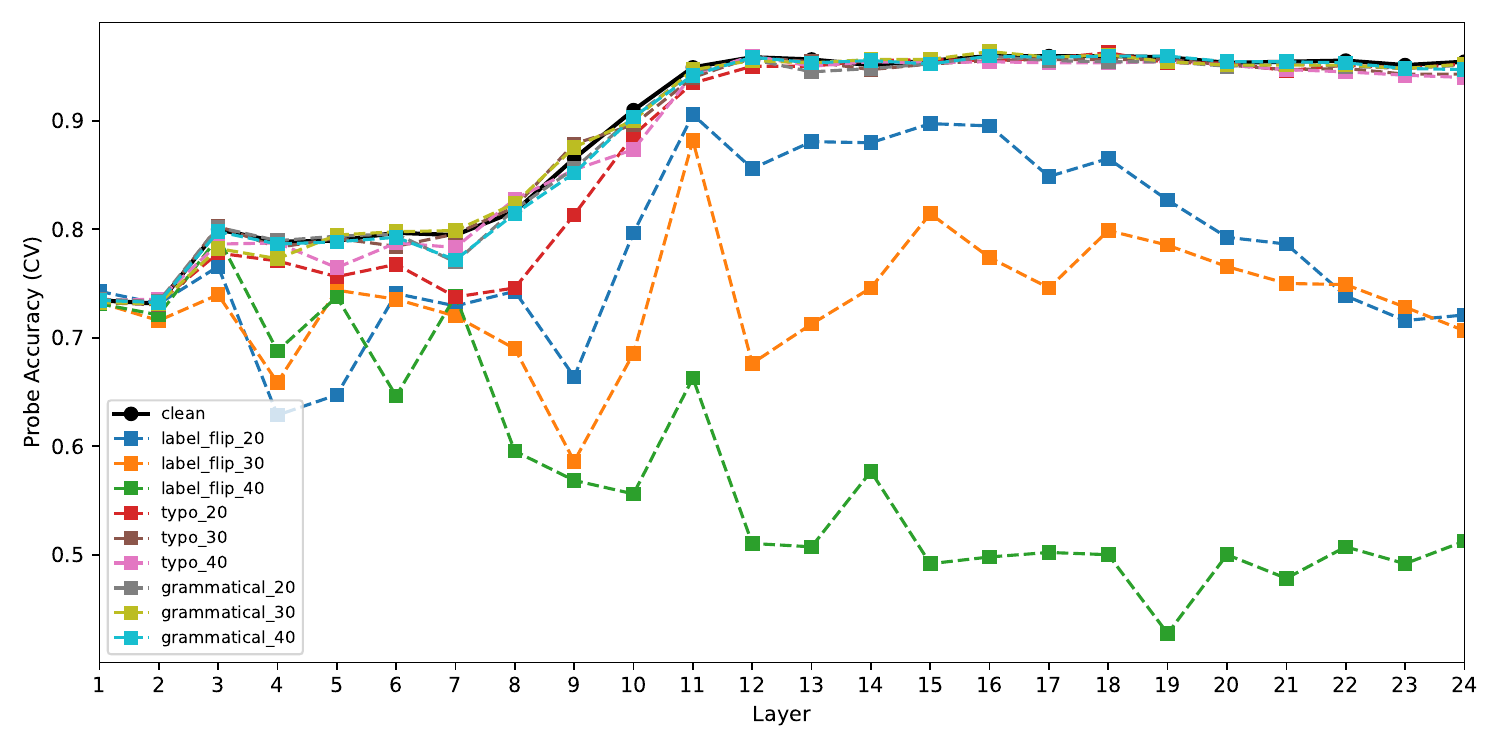}
    \caption{Qwen2-SC}
  \end{subfigure}
  \hfill
  \begin{subfigure}[t]{0.32\textwidth}
    \includegraphics[width=\textwidth]{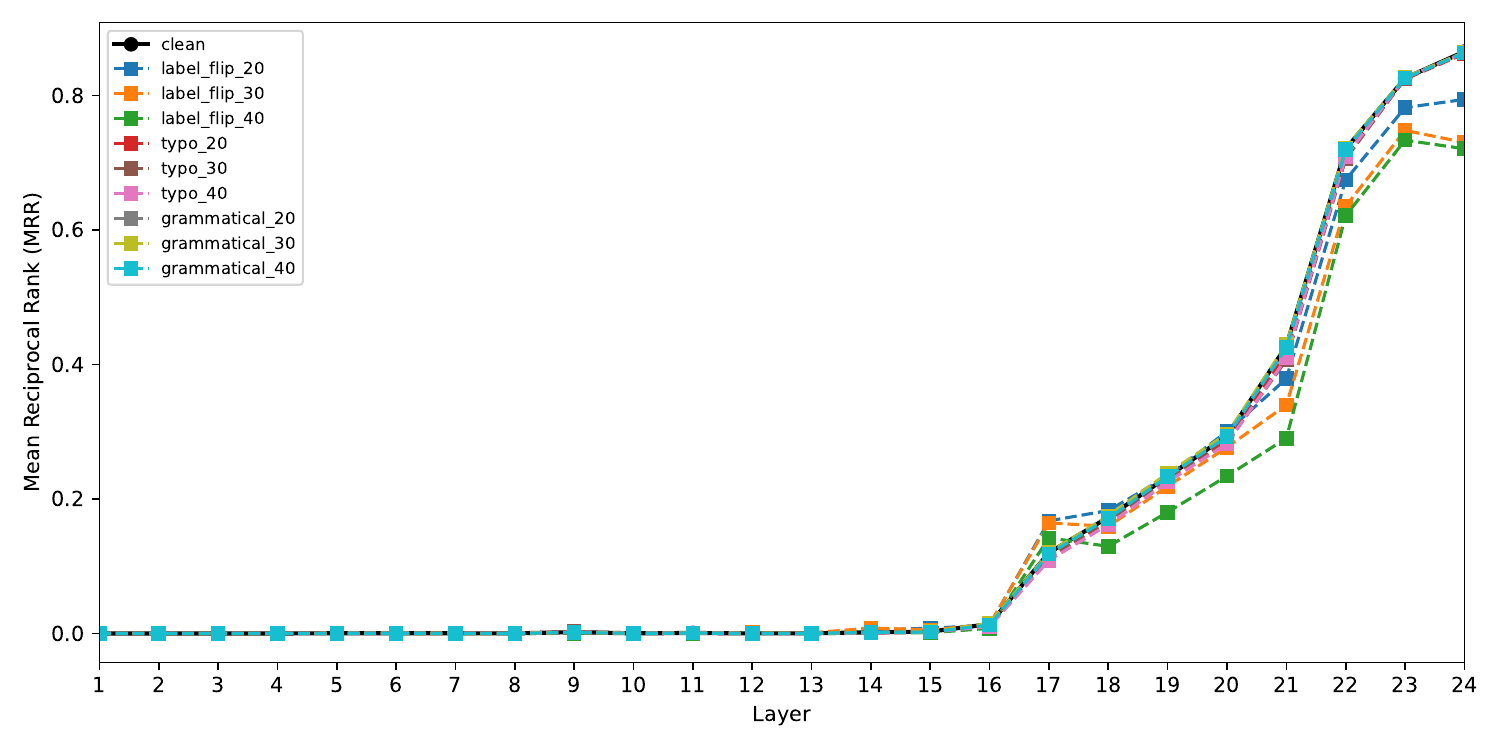}
    \caption{Qwen2-QA}
  \end{subfigure}
  \hfill
  \begin{subfigure}[t]{0.32\textwidth}
    \includegraphics[width=\textwidth]{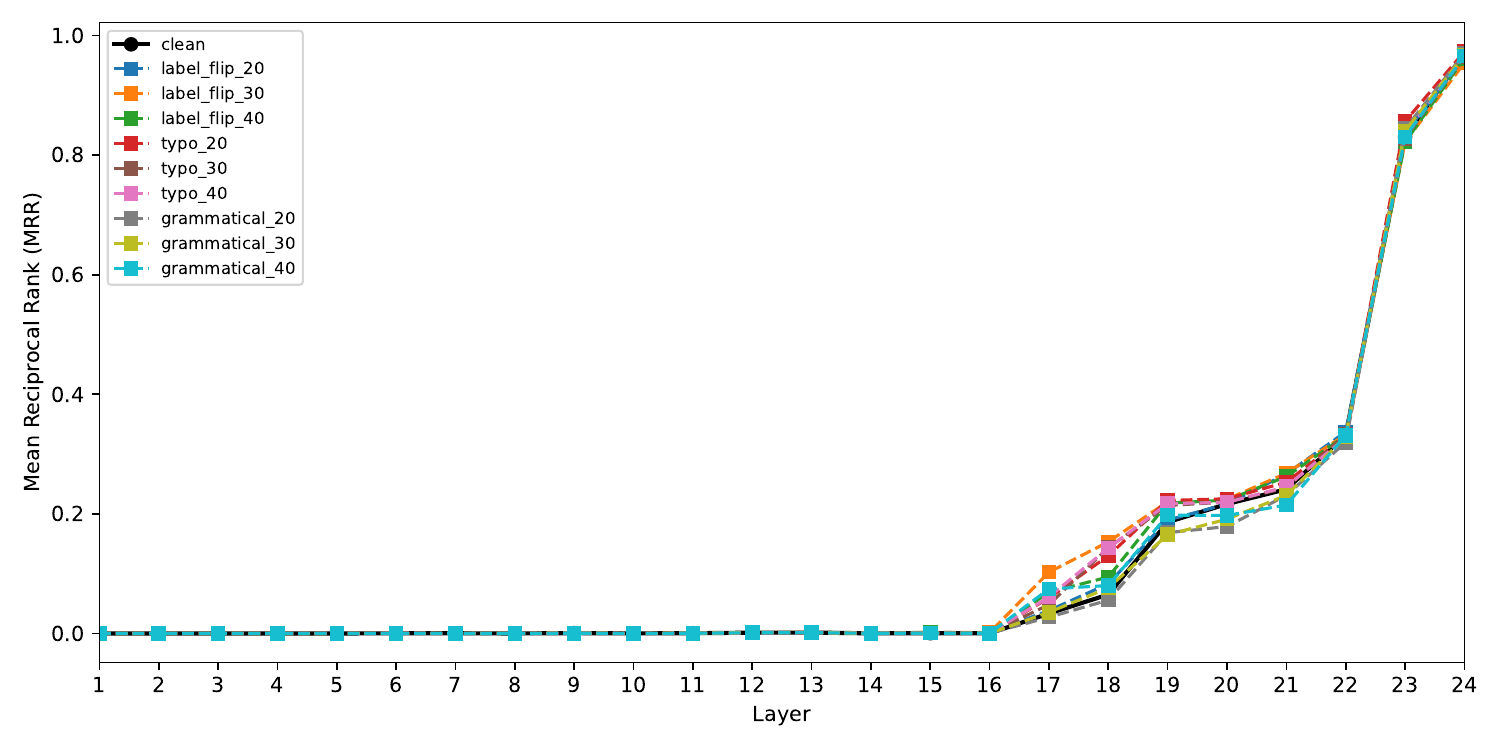}
    \caption{Qwen2-MT}
  \end{subfigure}
\begin{subfigure}[t]{0.32\textwidth}
    \includegraphics[width=\textwidth]{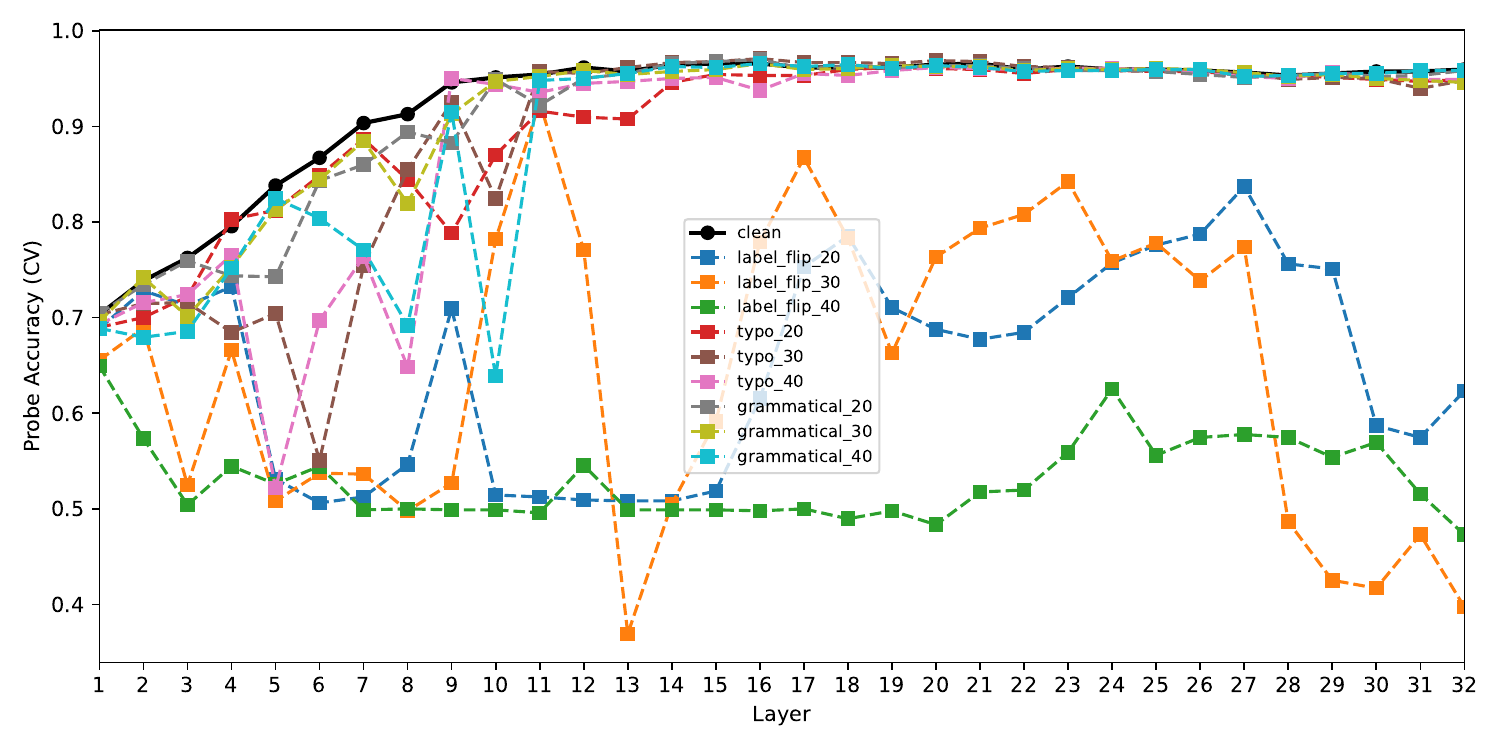}
    \caption{Llama2-SC}
  \end{subfigure}
  \hfill
  \begin{subfigure}[t]{0.32\textwidth}
    \includegraphics[width=\textwidth]{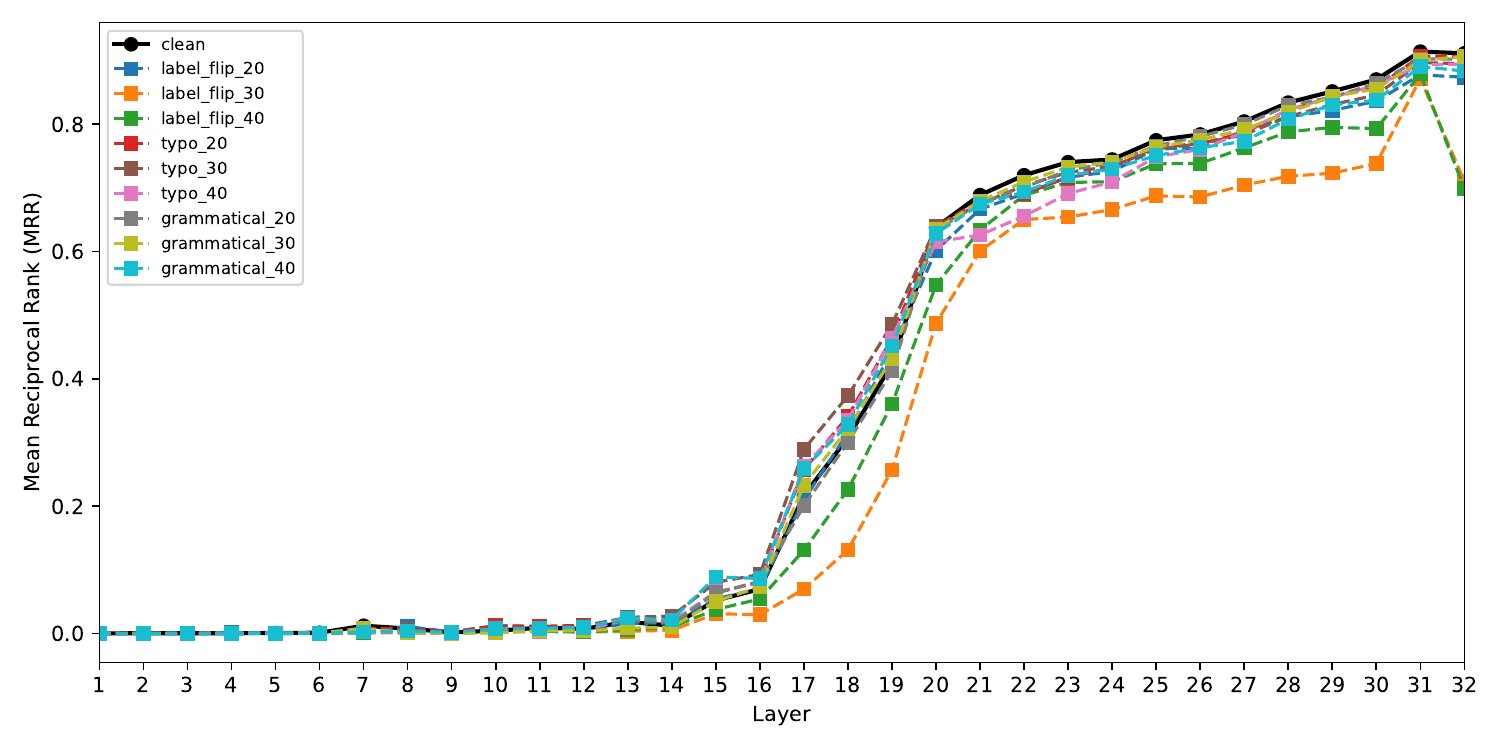}
    \caption{Llama2-QA}
  \end{subfigure}
  \hfill
  \begin{subfigure}[t]{0.32\textwidth}
    \includegraphics[width=\textwidth]{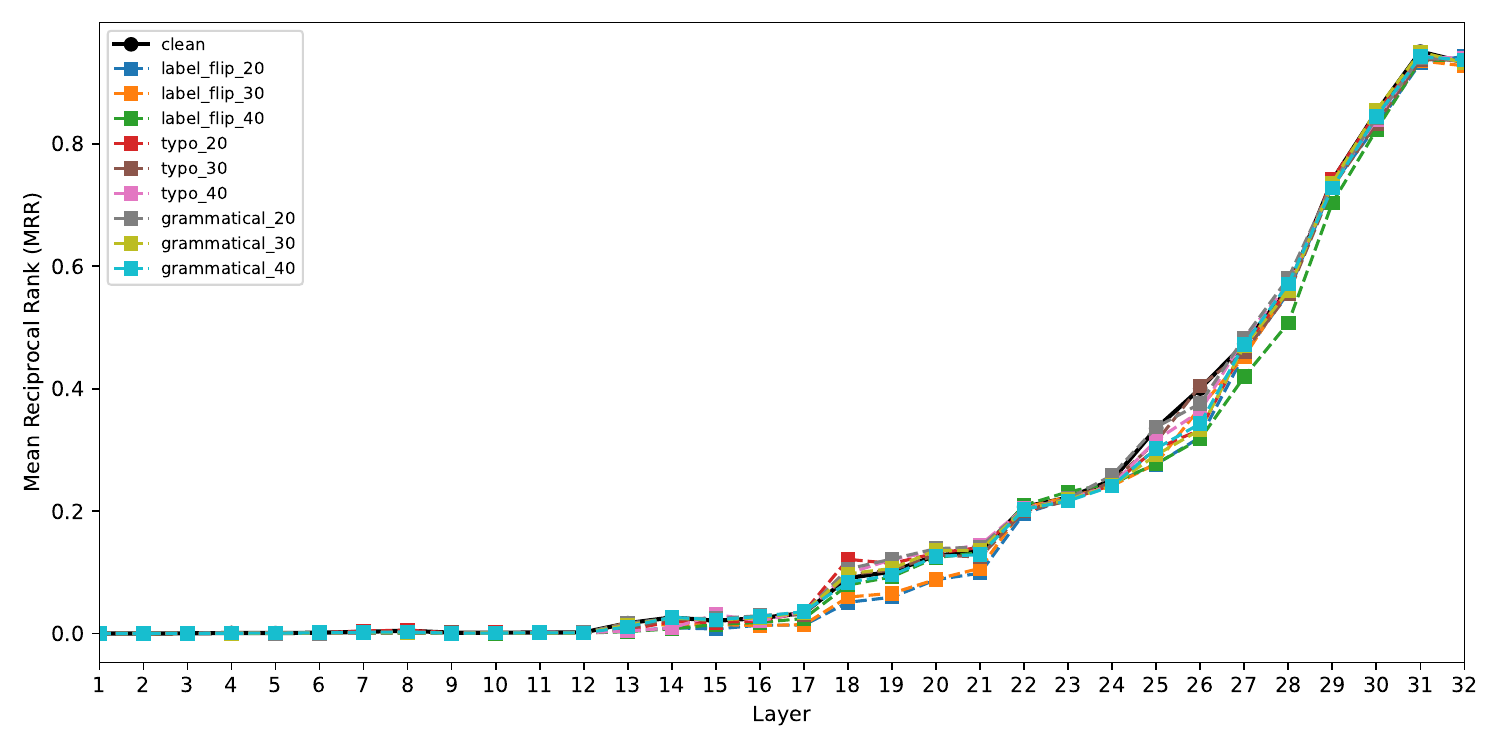}
    \caption{Llama2-MT}
  \end{subfigure}
  \caption{Layer-wise task information analysis for GPT-2 (124M), Qwen2-0.5B and Llama2-7B under all noise conditions. (a,d,g) Probing accuracy for \ac{SC}. (b,c,e,f,h,i) Logit Lens based MRR (first token) for QA and MT.} 
  \label{fig:layerwiseprobing}
\end{figure*}
Since GPT-2 uses full fine-tuning while the larger models use QLoRA, observed differences could in principle reflect the fine-tuning paradigm rather than model scale. To disentangle these factors, we conducted a LoRA ablation on GPT-2 for \ac{SC} under LF noise. The results ($75.51\%$ for full fine-tuning vs.\ $73.92\%$ for LoRA at $40\%$ corruption) are nearly identical, confirming that the choice of fine-tuning paradigm does not confound our results. We report full ablation results in \autoref{app:lora_ablation}.

All models were trained using the SFT-MASK protocol implemented via the \texttt{SFTTrainer} from the TRL library \citep{vonwerra2020trl}, the cross-entropy loss is computed only over completion tokens, with prompt tokens excluded from the loss computation. 
A greedy decoding strategy was used for evaluation for all models. To verify the stability of our findings, we trained Llama-2 sentiment models with three additional random seeds under label-flip $40\%$ noise; we also did a similar seed-based analysis for CKA to investigate seed-induced representational variance (\autoref{app:seed_stability}). Implementation details are provided in \autoref{app:hyperparameters}.

\paragraph{Evaluation}
\label{sec:evaluation} For each task, we used a corresponding task-specific metric for evaluating the overall performance on the task. For SC we used accuracy (whether the generated completion matches the gold label), for QA we used token-level F1, and for MT we used BLEU score (computed with SacreBLEU; \citealp{post2018call}).
All noise conditions for a given model-task pair are evaluated on the \emph{same} clean test set, ensuring that performance differences reflect model behavior rather than evaluation set variation.

\section{Results} \label{sec:results}

\subsection{Task Performance Under Noise}
\autoref{tab:sft-performance} displays the model performance in different tasks under different types of noise, the main findings from it are as follows: \textbf{a)} For all types of tasks and across all the models, LF noise ($40\%$) has caused the most amount of damage. In general, LF caused more damage compared to other forms of noise. On average, the amount of noise caused by label noise is $21\%$, which is far greater than both grammatical and typo errors. \textbf{b)} Noise doesn't always cause damage to the model. As can be seen from \autoref{tab:sft-performance}, TN has mostly improved the task accuracy across all the models and tasks to a certain extent. One potential explanation for this phenomenon is that increasing \ac{TN} has helped to build a robust model. \textbf{c)} For \ac{SC} and \ac{MT} tasks, noise has affected the smaller models more compared to the larger models. However, for \ac{QA} models, the same amount of noise has affected the larger model more compared to the smaller models. \textbf{d)} For the same task and similar types of noise, how the model will be affected depends on its architecture (e.g. First three rows in \autoref{tab:sft-performance}).

\subsection{Attention Pattern Shifts}
To further investigate the cause of damage by noise, we observed how attention patterns have changed across different models and different tasks in \autoref{fig:kl_attention}. The primary observations from \autoref{fig:kl_attention} is as follows. \textbf{a)} Since LF noise is causing the most amount of damage, correspondingly, we see larger changes in attention patterns for LF (Only exception is Llama2 QA). This is observed for all tasks and all models. \textbf{b)} For GPT-2 and Qwen2 (smaller models) most of the changes are concentrated within initial layers. For Llama-2 changes are concentrated form the initial to the middle layers (except only QA typo 40\%).   \textbf{c)} Another important thing we have observed is that the magnitude of change is very small overall across all the models and all the task types. Broadly speaking, the attention matrices are not that susceptible to noise. \textbf{d)} As we increase the noise generally, with the increase in the amount of noise the change is stronger no matter for better or worse performance. Observations from Figure \ref{fig:spearman_attention} is very similar to Figure \ref{fig:kl_attention}. It shows that order and magnitude of attention values follow the same pattern.

\begin{figure*}[t]
  \centering
  \begin{subfigure}[t]{0.32\textwidth}
    \includegraphics[width=\textwidth]{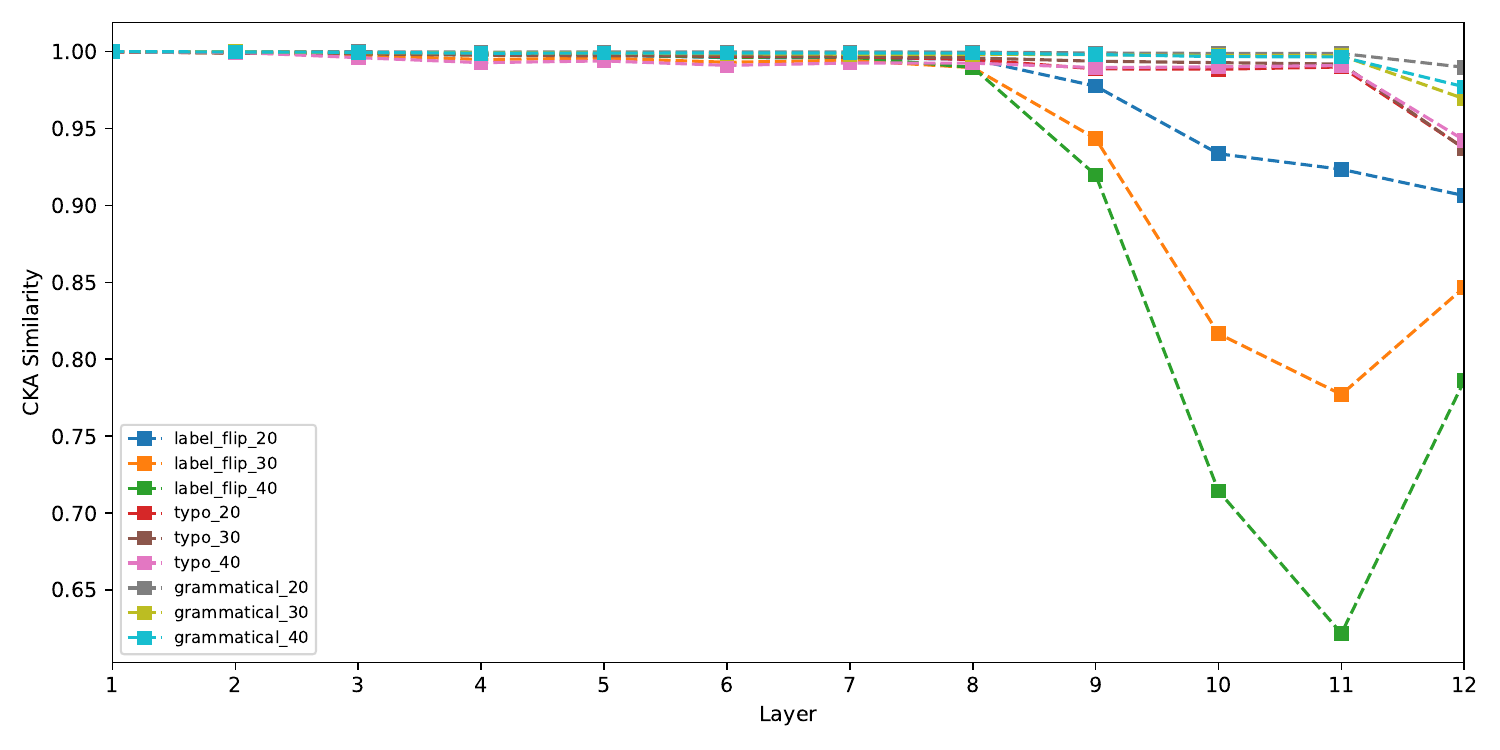}
    \caption{GPT2-SC}
  \end{subfigure}
  \hfill
  \begin{subfigure}[t]{0.32\textwidth}
    \includegraphics[width=\textwidth]{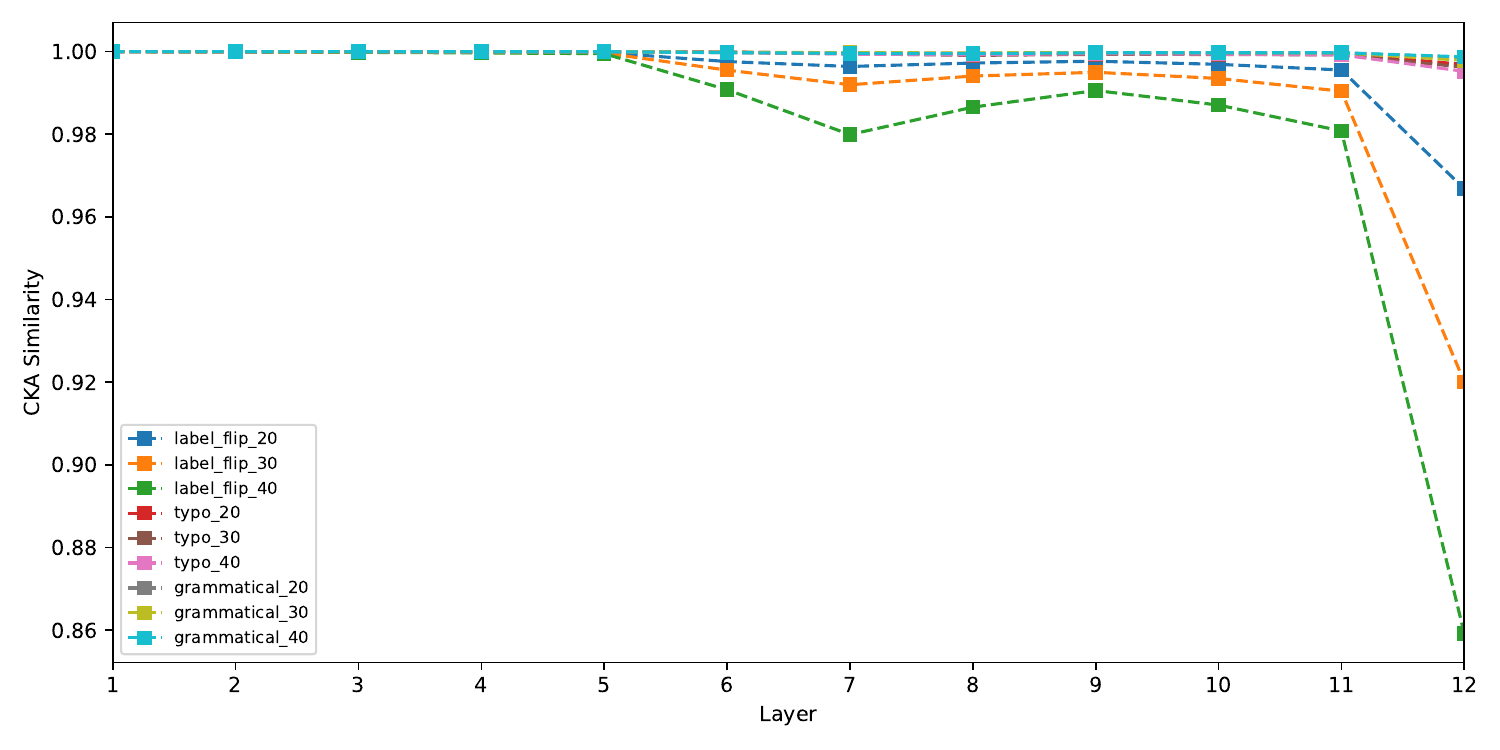}
    \caption{GPT2-QA}
  \end{subfigure}
  \hfill
  \begin{subfigure}[t]{0.32\textwidth}
    \includegraphics[width=\textwidth]{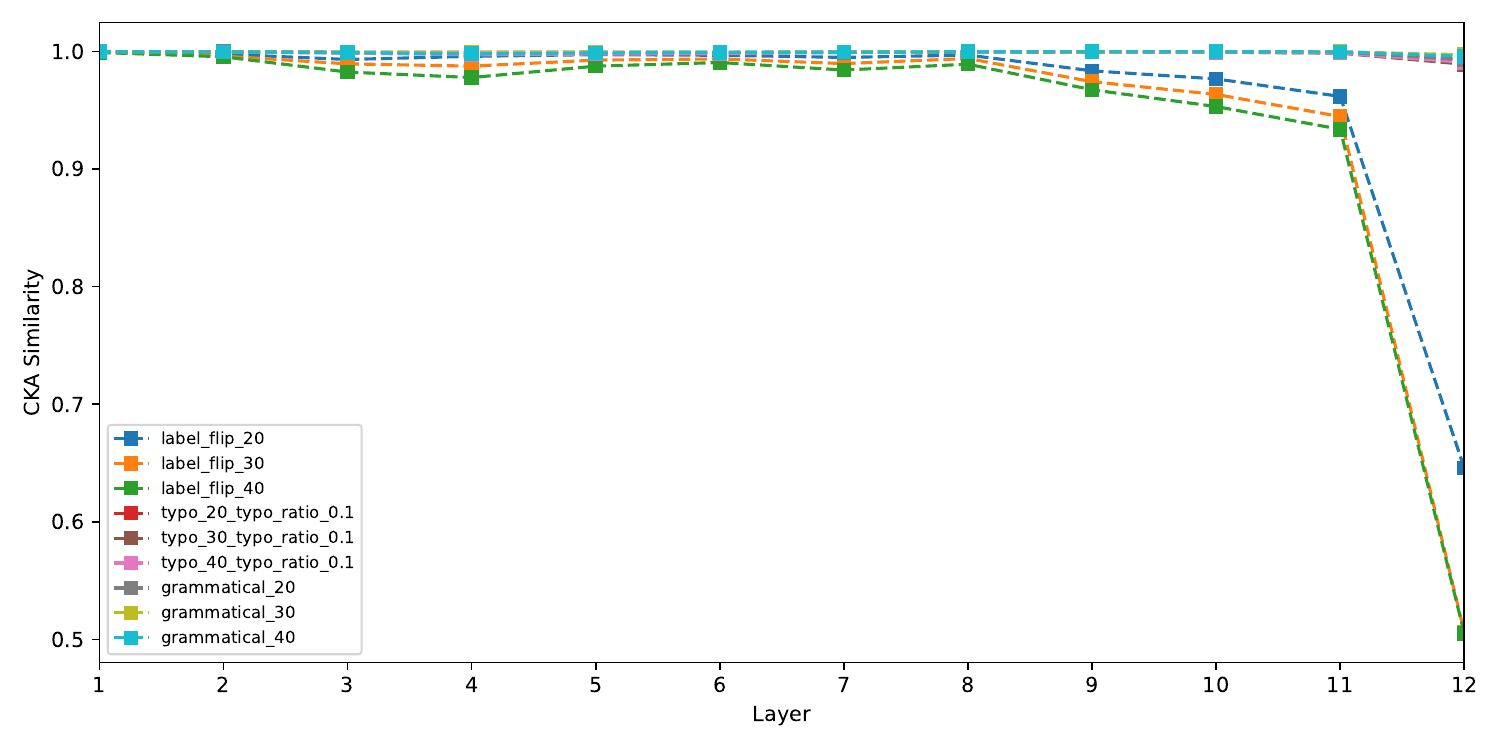}
    \caption{GPT2-MT}
  \end{subfigure}
  \begin{subfigure}[t]{0.32\textwidth}
    \includegraphics[width=\textwidth]{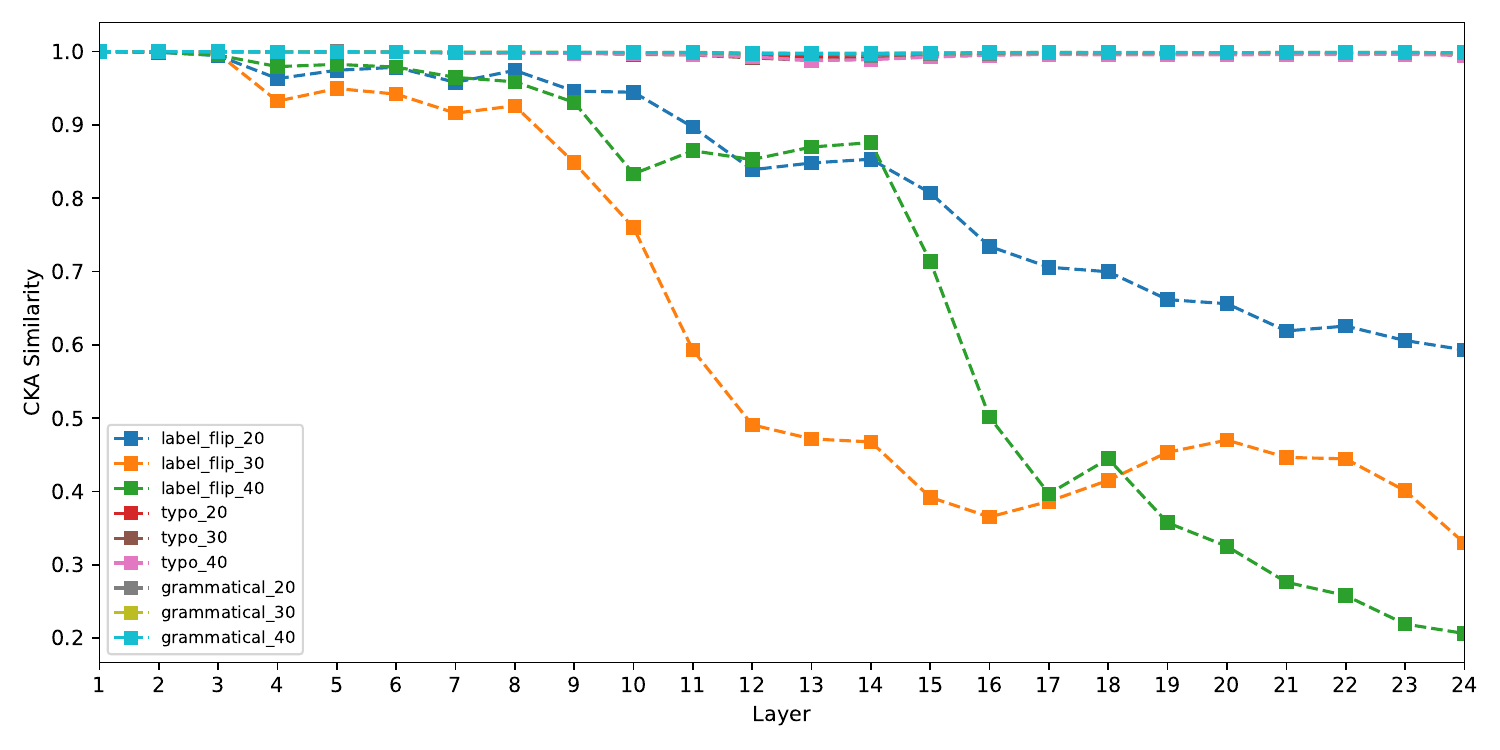}
    \caption{Qwen2-SC}
  \end{subfigure}
  \hfill
  \begin{subfigure}[t]{0.32\textwidth}
    \includegraphics[width=\textwidth]{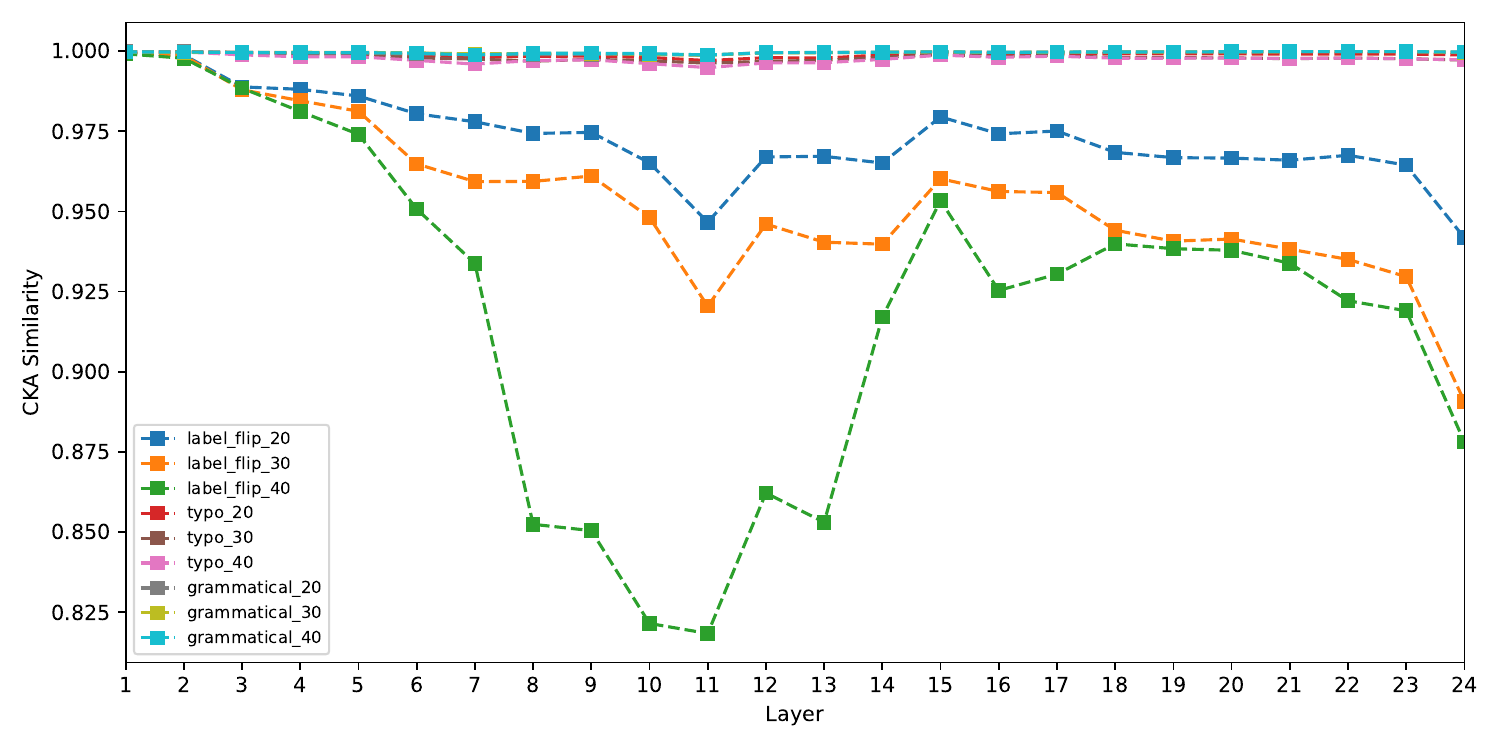}
    \caption{Qwen2-QA}
  \end{subfigure}
  \hfill
  \begin{subfigure}[t]{0.32\textwidth}
    \includegraphics[width=\textwidth]{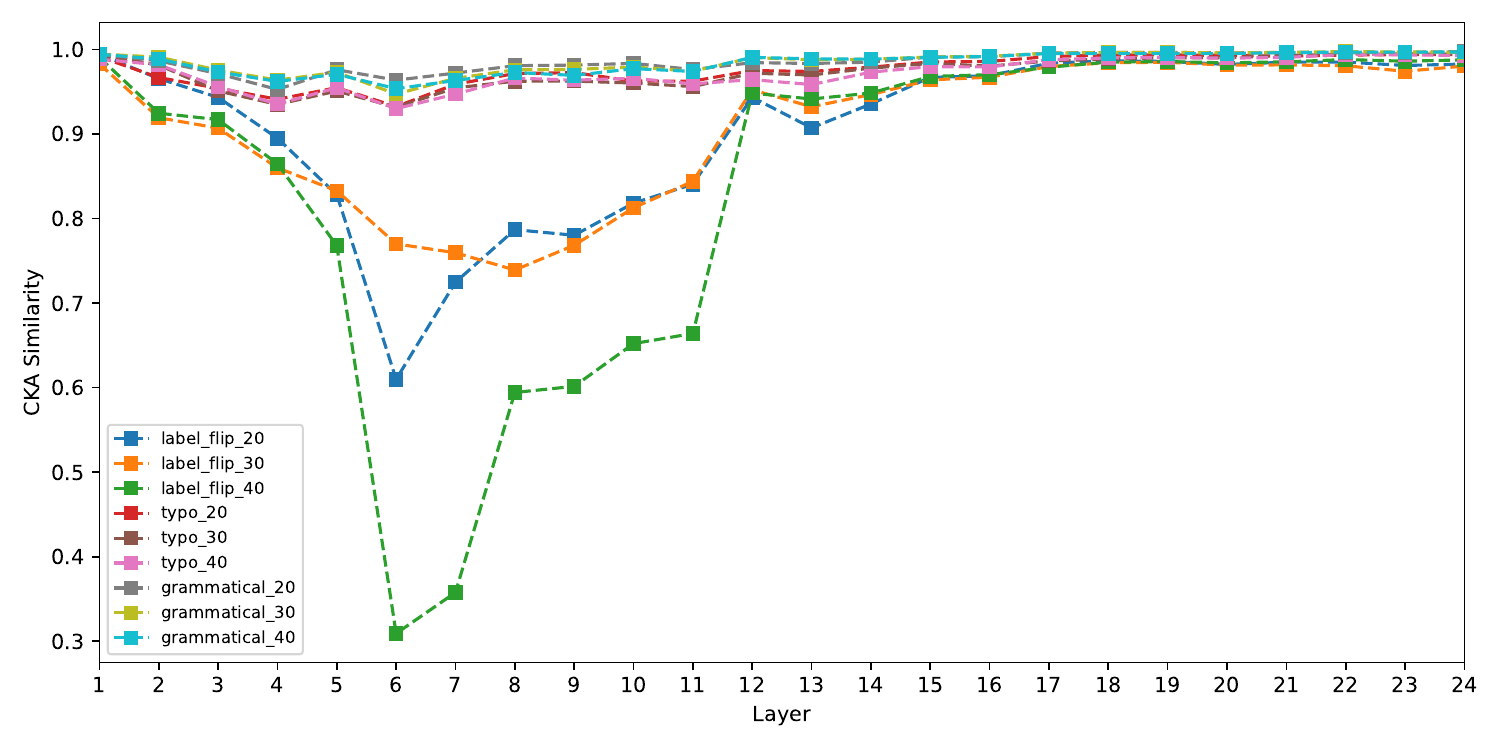}
    \caption{Qwen2-MT}
  \end{subfigure}
   \begin{subfigure}[t]{0.32\textwidth}
    \includegraphics[width=\textwidth]{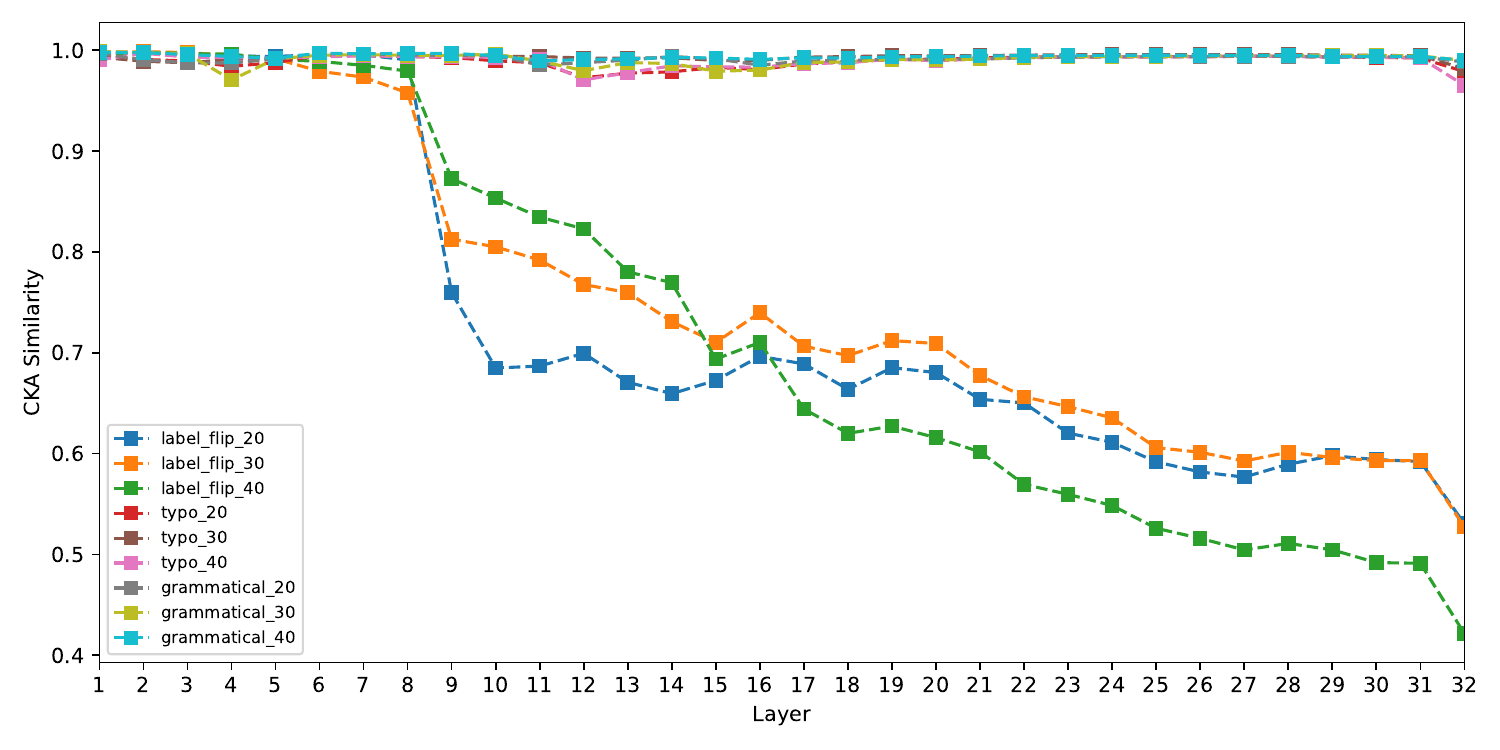}
    \caption{Llama2-SC}
  \end{subfigure}
  \hfill
  \begin{subfigure}[t]{0.32\textwidth}
    \includegraphics[width=\textwidth]{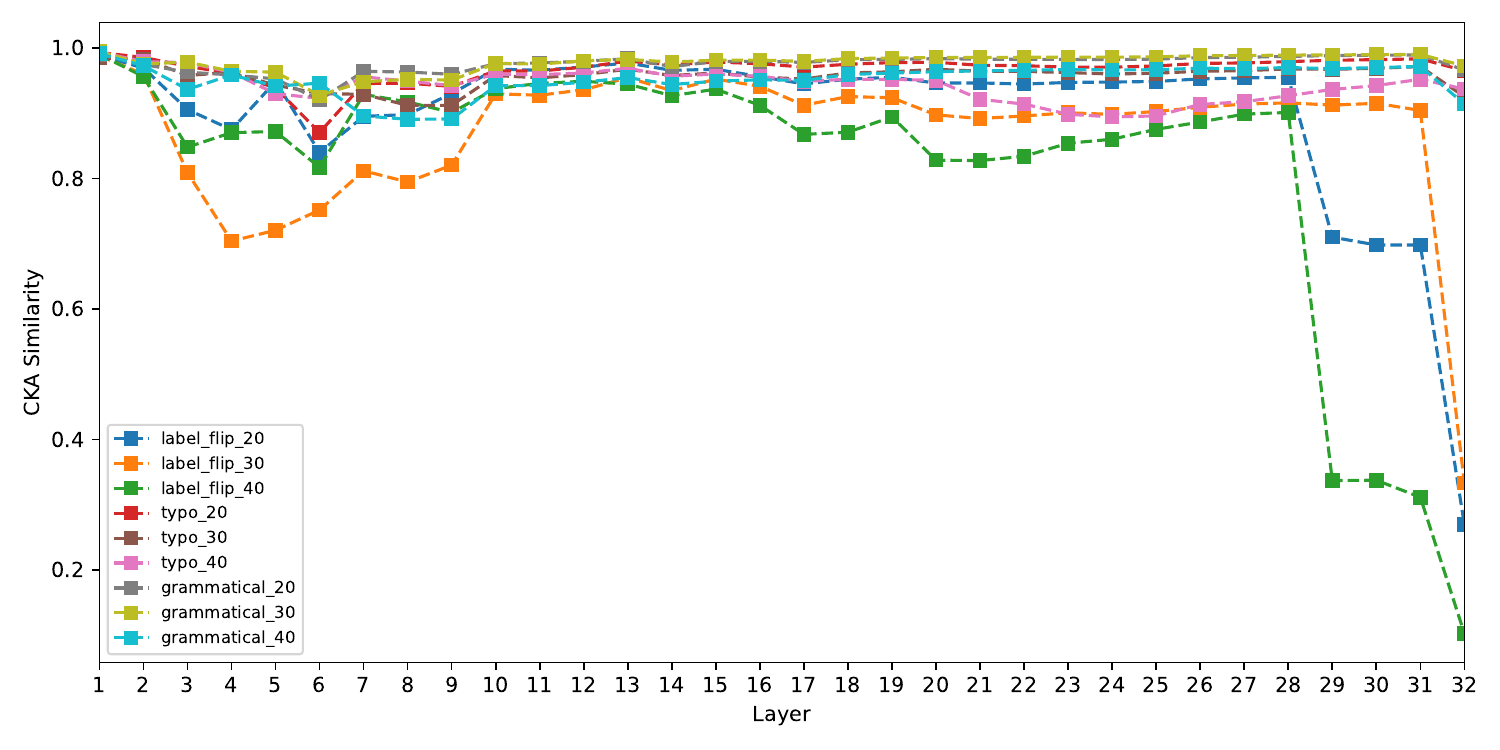}
    \caption{Llama2-QA}
  \end{subfigure}
  \hfill
  \begin{subfigure}[t]{0.32\textwidth}
    \includegraphics[width=\textwidth]{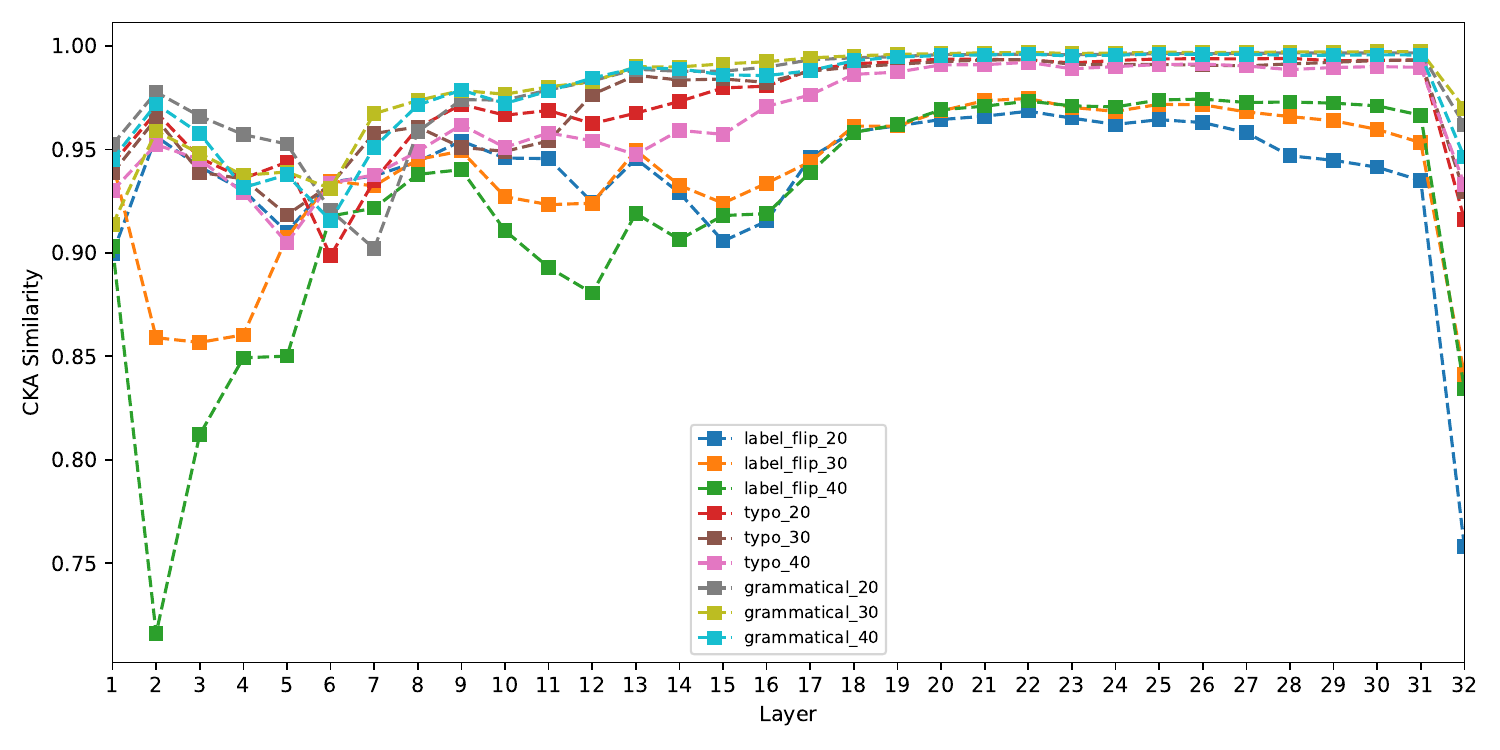}
    \caption{Llama2-MT}
  \end{subfigure}
  \caption{Layer-wise Linear CKA similarity between clean and 
  noise-trained model representations across three tasks. Each row corresponds to a different backbone: (a–c) GPT-2 Small (12 layers), (d–f) Qwen2-0.5B (24 layers), and (g–i) LLaMA-2-7B (32 layers). }
  \label{fig:gpt2_cka}
\end{figure*}
\subsection{Layer-wise Task Information Under Noise}
\label{sec:layerwise}
To further dig deeper into understanding the cause of damage due to noise, we observed probing accuracy for individual layers across all models and all tasks. \autoref{fig:layerwiseprobing} shows the performance of probing. The key findings are as follows.  \textbf{a)} Apart from Sentiment analysis there is mostly a consistent pattern in the performance of different layers across different models and different types of noise. The generic pattern is that the initial layers have a poor performance and then later layers are performing better showing that later layers have better task-performing ability compared to initial layers.  \textbf{b)} Table  \ref{tab:sft-performance} showed that the most damage was caused by label flip in sentiment analysis compared to any other configurations. This is visible from \autoref{fig:layerwiseprobing} where we can see that the sentiment analysis level flip probing curve is maintaining the most distance from the clean model curve compared to other tasks. From \autoref{fig:layerwiseprobing}  we know that initial layers did not have any task-specific information. That's why the probing accuracies for initial layers are almost identical (approx to 0). This raised the question of whether the representations in the initial layers are also similar between clean and noisy models. Because two poor representations can yield similar poor downstream performance without providing any indication of how similar they are. In \autoref{fig:gpt2_cka} and \autoref{fig:centered_cosine_grid} (\autoref{app:centered_cosine}), we have done centered cosine similarity / CKA similarity analysis across all the layers, and it can be observed that the similarities are very high in the initial layers, indicating that task-specific noise primarily targets layers with more task information.

Results using the average \ac{MRR} over the first five target tokens are provided
  in \autoref{app:five-mrr} and show consistent patterns. We additionally report teacher-forced token accuracy in \autoref{app:token_acc}, where the model receives the ground-truth prefix at each position; this complementary binary metric confirms the same trends.


\subsection{Representational Similarity}
\autoref{fig:gpt2_cka} shows the layerwise linear CKA between the clean model and different types of noise. The main findings from \autoref{fig:gpt2_cka} are as follows. \textbf{a)} It can be observed that for most cases increasing more noise has created more distortion in the corresponding layer representation (e.g. for \ac{LF} the green curves are mostly at the lowest similarity point at each layer).
\textbf{b)} As can be observed in \autoref{tab:sft-performance} that LF has caused more damage compared to other noise, similarly the largest distortion (lowest cosine similarity) is observed in LF noise for all the tasks across all the models in \autoref{fig:gpt2_cka}. \textbf{c)} As it can be seen in \autoref{fig:layerwiseprobing} that for most of the tasks the task-specific information was encoded in the later layers. Similar things can also be observed in \autoref{fig:gpt2_cka}. The cosine similarity of the initial layers are generally higher than the later layers except MT in Qwen2-0.5B. Based on the above-mentioned observation, it can be said that generally the noise affects the layers that have more task-specific information more compared to the ones where there was not that much task-specific information. The result for centered cosine similarity is shown in \autoref{fig:centered_cosine_grid} (\autoref{app:centered_cosine}). The patterns are consistent with the CKA-based similarity results, confirming that noise primarily affects layers with more task-specific information.

\section{Robust vs. Vulnerable Stratification}
\label{sec:stratification}
We initially investigated the overall changes in models through the methods described in Section \ref{sec:attn_kl}, \ref{sec:fixed_ruler} and \ref{sec:cosine}. However, the effect on noisy data fine-tuning may not be uniform on all the test samples. There are test samples for which the model's prediction remains unchanged after fine-tuning with noisy data. Broadly speaking, they are robust samples with respect to a task and a model. Similarly, there are samples for which predictions changed due to using a model fine-tuned on noisy data. Broadly speaking, these are vulnerable samples.

To examine the effect of robust and vulnerable samples separately, we stratified the evaluation samples into two groups: \emph{robust} samples and \emph{vulnerable} samples. We then applied all the above mentioned analysis approaches on the different types of dataset separately. The objective was to observe whether aggregate representational metrics mask heterogeneous effects across subpopulations with different type of outcomes.

From Figure \ref{fig:stratification_sent_centered_cosine}, \ref{fig:stratification_sent_cka} and \ref{fig:stratification_qa_mrr_first} in Appendix \ref{ap:stratification} we observed that in most cases the damage is more for the data points for which the prediction was wrong due to noise compared to the ones for which the prediction didn't change in spite of noise.  It is interesting to note that in spite of no change in prediction, there was still distortion in the internal representation.

\section{Conclusion and Future Work} \label{sec:conclusion}
This work presents a systematic analysis of the effects of three types of noise—label noise, typographical noise, and grammatical noise—across three widely used NLP tasks and three different language models. Through a set of complementary analyses, we examine how these noise sources affect model behavior at both the prediction and representation levels. Our results show that the impact of noise tends to be largely localized within specific layers of the model rather than uniformly affecting the entire network. Furthermore, among the three noise types considered, label noise consistently leads to the most significant degradation in model performance, highlighting the sensitivity of LLMs to incorrect supervision signals during training. Despite these representational changes, attention structures remain comparatively stable across all noise types and models, suggesting that noise primarily reshapes feed-forward representations rather than altering how the model distributes contextual importance across tokens. A further stratification of test samples into robust and vulnerable groups reveals that while vulnerable samples consistently show greater representational distortion, even robust samples whose predictions remain unchanged exhibit non-trivial internal representation shifts, indicating that task-level performance alone underestimates the true extent of noise-induced representational change.

These findings offer insights into how various forms of noise impact internal representations and task performance in LLMs. In future work, we plan to leverage these insights to design fine-tuning strategies that explicitly account for noise during training. In particular, we aim to develop robust fine-tuning approaches that can mitigate the adverse effects of noisy data while preserving task-relevant representations, thereby improving the reliability of LLMs in real-world noisy environments.

%

\bibliography{main}

@inproceedings{karpukhin2019typo,
    title = "Training on Synthetic Noise Improves Robustness to Natural Noise in Machine Translation",
    author = "Karpukhin, Vladimir  and
      Levy, Omer  and
      Eisenstein, Jacob  and
      Ghazvininejad, Marjan",
    editor = "Xu, Wei  and
      Ritter, Alan  and
      Baldwin, Tim  and
      Rahimi, Afshin",
    booktitle = "Proceedings of the 5th Workshop on Noisy User-generated Text (W-NUT 2019)",
    month = nov,
    year = "2019",
    pages = "42--47",
}

@inproceedings{nlpresearch2,
    title = "It Takes Two to Tango: Navigating Conceptualizations of {NLP} Tasks and Measurements of Performance",
    author = "Subramonian, Arjun  and
      Yuan, Xingdi  and
      Daum{\'e} III, Hal  and
      Blodgett, Su Lin",
    editor = "Rogers, Anna  and
      Boyd-Graber, Jordan  and
      Okazaki, Naoaki",
    booktitle = "Findings of the Association for Computational Linguistics: ACL 2023",
    month = jul,
    year = "2023",
    pages = "3234--3279",
}

@inproceedings{nlpresearch1,
    title = "Which Domains, Tasks and Languages are in the Focus of {NLP} Research on the Languages of {E}urope?",
    author = "Alves, Diego  and
      Tadi{\'c}, Marko  and
      Rehm, Georg",
    editor = "Gaspari, Federico  and
      Moorkens, Joss  and
      Aldabe, Itziar  and
      Farwell, Aritz  and
      Altuna, Begona  and
      Piperidis, Stelios  and
      Rehm, Georg  and
      Rigau, German",
    booktitle = "Proceedings of the Second International Workshop Towards Digital Language Equality (TDLE): Focusing on Sustainability @ LREC-COLING 2024",
    month = may,
    year = "2024",
    pages = "18--32",
}

@inproceedings{survey1,
author = {Subramaniam, L. Venkata and Roy, Shourya and Faruquie, Tanveer A. and Negi, Sumit},
title = {A survey of types of text noise and techniques to handle noisy text},
year = {2009},
booktitle = {Proceedings of The Third Workshop on Analytics for Noisy Unstructured Text Data},
pages = {115–122},
}

@inproceedings{moradi2021grammar,
  title={Evaluating the robustness of neural language models to input perturbations},
  author={Moradi, Milad and Samwald, Matthias},
  booktitle={Proceedings of the 2021 Conference on Empirical Methods in Natural Language Processing},
  pages={1558--1570},
  year={2021}
}

@misc{gao2018blackboxgenerationadversarialtex,
      title={Black-box Generation of Adversarial Text Sequences to Evade Deep Learning Classifiers}, 
      author={Ji Gao and Jack Lanchantin and Mary Lou Soffa and Yanjun Qi},
      year={2018},
      eprint={1801.04354},
      archivePrefix={arXiv},
      primaryClass={cs.CL},
      url={https://arxiv.org/abs/1801.04354}, 
}

@inproceedings{MT-fine,
    title = "Fine-Tuning Large Language Models to Translate: Will a Touch of Noisy Data in Misaligned Languages Suffice?",
    author = "Zhu, Dawei  and
      Chen, Pinzhen  and
      Zhang, Miaoran  and
      Haddow, Barry  and
      Shen, Xiaoyu  and
      Klakow, Dietrich",
    editor = "Al-Onaizan, Yaser  and
      Bansal, Mohit  and
      Chen, Yun-Nung",
    booktitle = "Proceedings of the 2024 Conference on Empirical Methods in Natural Language Processing",
    month = nov,
    year = "2024",
    address = "Miami, Florida, USA",
    publisher = "Association for Computational Linguistics",
    url = "https://aclanthology.org/2024.emnlp-main.24/",
    doi = "10.18653/v1/2024.emnlp-main.24",
    pages = "388--409"
}

@misc{luo2024robustft,
      title={RobustFT: Robust Supervised Fine-tuning for Large Language Models under Noisy Response}, 
      author={Junyu Luo and Xiao Luo and Kaize Ding and Jingyang Yuan and Zhiping Xiao and Ming Zhang},
      year={2024},
      eprint={2412.14922},
      archivePrefix={arXiv},
      primaryClass={cs.CL},
      url={https://arxiv.org/abs/2412.14922}, 
}

@inproceedings{linearprobing,
  title={What do neural machine translation models learn about morphology?},
  author={Belinkov, Yonatan and Durrani, Nadir and Dalvi, Fahim and Sajjad, Hassan and Glass, James},
  booktitle={Proceedings of the 55th Annual Meeting of the Association for Computational Linguistics (Volume 1: Long Papers)},
  pages={861--872},
  year={2017}
}

@inproceedings{geva-etal-2023-dissecting,
    title = "Dissecting Recall of Factual Associations in Auto-Regressive Language Models",
    author = "Geva, Mor  and
      Bastings, Jasmijn  and
      Filippova, Katja  and
      Globerson, Amir",
    editor = "Bouamor, Houda  and
      Pino, Juan  and
      Bali, Kalika",
    booktitle = "Proceedings of the 2023 Conference on Empirical Methods in Natural Language Processing",
    month = dec,
    year = "2023",
    address = "Singapore",
    publisher = "Association for Computational Linguistics",
    url = "https://aclanthology.org/2023.emnlp-main.751/",
    doi = "10.18653/v1/2023.emnlp-main.751",
    pages = "12216--12235"
}

@inproceedings{jiang-etal-2024-large,
    title = "On Large Language Models' Hallucination with Regard to Known Facts",
    author = "Jiang, Che  and
      Qi, Biqing  and
      Hong, Xiangyu  and
      Fu, Dayuan  and
      Cheng, Yang  and
      Meng, Fandong  and
      Yu, Mo  and
      Zhou, Bowen  and
      Zhou, Jie",
    editor = "Duh, Kevin  and
      Gomez, Helena  and
      Bethard, Steven",
    booktitle = "Proceedings of the 2024 Conference of the North American Chapter of the Association for Computational Linguistics: Human Language Technologies (Volume 1: Long Papers)",
    month = jun,
    year = "2024",
    address = "Mexico City, Mexico",
    publisher = "Association for Computational Linguistics",
    url = "https://aclanthology.org/2024.naacl-long.60/",
    doi = "10.18653/v1/2024.naacl-long.60",
    pages = "1041--1053"
}

@InProceedings{Patrini2017losscorrection,
author = {Patrini, Giorgio and Rozza, Alessandro and Krishna Menon, Aditya and Nock, Richard and Qu, Lizhen},
title = {Making Deep Neural Networks Robust to Label Noise: A Loss Correction Approach},
booktitle = {Proceedings of the IEEE Conference on Computer Vision and Pattern Recognition (CVPR)},
month = {July},
year = {2017}
}

@article{Ghosh_Kumar_Sastry_2017, title={Robust Loss Functions under Label Noise for Deep Neural Networks}, volume={31}, url={https://ojs.aaai.org/index.php/AAAI/article/view/10894}, DOI={10.1609/aaai.v31i1.10894}, abstractNote={ &lt;p&gt; In many applications of classifier learning, training data suffers from label noise. Deep networks are learned using huge training data where the problem of noisy labels is particularly relevant. The current techniques proposed for learning deep networks under label noise focus on modifying the network architecture and on algorithms for estimating true labels from noisy labels. An alternate approach would be to look for loss functions that are inherently noise-tolerant. For binary classification there exist theoretical results on loss functions that are robust to label noise. In this paper, we provide some sufficient conditions on a loss function so that risk minimization under that loss function would be inherently tolerant to label noise for multiclass classification problems. These results generalize the existing results on noise-tolerant loss functions for binary classification. We study some of the widely used loss functions in deep networks and show that the loss function based on mean absolute value of error is inherently robust to label noise. Thus standard back propagation is enough to learn the true classifier even under label noise. Through experiments, we illustrate the robustness of risk minimization with such loss functions for learning neural networks. &lt;/p&gt; }, number={1}, journal={Proceedings of the AAAI Conference on Artificial Intelligence}, author={Ghosh, Aritra and Kumar, Himanshu and Sastry, P. S.}, year={2017}, month={Feb.} }

@inproceedings{NEURIPS2018_crossloss,
 author = {Zhang, Zhilu and Sabuncu, Mert},
 booktitle = {Advances in Neural Information Processing Systems},
 editor = {S. Bengio and H. Wallach and H. Larochelle and K. Grauman and N. Cesa-Bianchi and R. Garnett},
 pages = {},
 publisher = {Curran Associates, Inc.},
 title = {Generalized Cross Entropy Loss for Training Deep Neural Networks with Noisy Labels},
 url = {https://proceedings.neurips.cc/paper_files/paper/2018/file/f2925f97bc13ad2852a7a551802feea0-Paper.pdf},
 volume = {31},
 year = {2018}
}

@article{han2018coteaching,
  title={Co-teaching: Robust training of deep neural networks with extremely noisy labels},
  author={Han, Bo and Yao, Quanming and Yu, Xingrui and Niu, Gang and Xu, Miao and Hu, Weihua and Tsang, Ivor and Sugiyama, Masashi},
  journal={Advances in neural information processing systems},
  volume={31},
  year={2018}
}

@inproceedings{liu-etal-2022-noise,
    title = "Noise Learning for Text Classification: A Benchmark",
    author = "Liu, Bo  and
      Xu, Wandi  and
      Xiang, Yuejia  and
      Wu, Xiaojun  and
      He, Lejian  and
      Zhang, Bowen  and
      Zhu, Li",
    booktitle = "Proceedings of the 29th International Conference on Computational Linguistics",
    month = oct,
    year = "2022",
    pages = "4557--4567",
}

@inproceedings{10.5555/2999611.2999745,
author = {Natarajan, Nagarajan and Dhillon, Inderjit S. and Ravikumar, Pradeep and Tewari, Ambuj},
title = {Learning with noisy labels},
year = {2013},
booktitle = {Proceedings of the 27th International Conference on Neural Information Processing Systems - Volume 1},
pages = {1196–1204},
}

@article{li2020dividemix,
  title={Dividemix: Learning with noisy labels as semi-supervised learning},
  author={Li, Junnan and Socher, Richard and Hoi, Steven CH},
  journal={arXiv preprint arXiv:2002.07394},
  year={2020}
}

@inproceedings{zhu-etal-2022-bert,
    title = "Is {BERT} Robust to Label Noise? A Study on Learning with Noisy Labels in Text Classification",
    author = "Zhu, Dawei  and
      Hedderich, Michael A.  and
      Zhai, Fangzhou  and
      Adelani, David Ifeoluwa  and
      Klakow, Dietrich",
    editor = "Tafreshi, Shabnam  and
      Sedoc, Jo{\~a}o  and
      Rogers, Anna  and
      Drozd, Aleksandr  and
      Rumshisky, Anna  and
      Akula, Arjun",
    booktitle = "Proceedings of the Third Workshop on Insights from Negative Results in NLP",
    month = may,
    year = "2022",
    address = "Dublin, Ireland",
    publisher = "Association for Computational Linguistics",
    url = "https://aclanthology.org/2022.insights-1.8/",
    doi = "10.18653/v1/2022.insights-1.8",
    pages = "62--67"
}

@inproceedings{
wang2023laft,
title={Noise-Robust Fine-Tuning of Pretrained Language Models via External Guidance},
author={Song Wang and Zhen Tan and Ruocheng Guo and Jundong Li},
booktitle={The 2023 Conference on Empirical Methods in Natural Language Processing},
year={2023},
url={https://openreview.net/forum?id=DSmHC8bi3j}
}

@inproceedings{
qi2024finetuning,
title={Fine-tuning Aligned Language Models Compromises Safety, Even When Users Do Not Intend To!},
author={Xiangyu Qi and Yi Zeng and Tinghao Xie and Pin-Yu Chen and Ruoxi Jia and Prateek Mittal and Peter Henderson},
booktitle={The Twelfth International Conference on Learning Representations},
year={2024},
url={https://openreview.net/forum?id=hTEGyKf0dZ}
}

@inproceedings{liu2020early,
author = {Liu, Sheng and Niles-Weed, Jonathan and Razavian, Narges and Fernandez-Granda, Carlos},
title = {Early-learning regularization prevents memorization of noisy labels},
year = {2020},
isbn = {9781713829546},
publisher = {Curran Associates Inc.},
address = {Red Hook, NY, USA},
booktitle = {Proceedings of the 34th International Conference on Neural Information Processing Systems},
articleno = {1707},
numpages = {12},
location = {Vancouver, BC, Canada},
series = {NIPS '20}
}

@inproceedings{tanzer2022memorisation,
    title = "Memorisation versus Generalisation in Pre-trained Language Models",
    author = {T{\"a}nzer, Michael  and
      Ruder, Sebastian  and
      Rei, Marek},
    editor = "Muresan, Smaranda  and
      Nakov, Preslav  and
      Villavicencio, Aline",
    booktitle = "Proceedings of the 60th Annual Meeting of the Association for Computational Linguistics (Volume 1: Long Papers)",
    month = may,
    year = "2022",
    address = "Dublin, Ireland",
    publisher = "Association for Computational Linguistics",
    url = "https://aclanthology.org/2022.acl-long.521/",
    doi = "10.18653/v1/2022.acl-long.521",
    pages = "7564--7578"
}

@inproceedings{
rosati2024representation,
title={Representation Noising: A Defence Mechanism Against Harmful Finetuning},
author={Domenic Rosati and Jan Wehner and Kai Williams and Lukasz Bartoszcze and Robie Gonzales and carsten maple and Subhabrata Majumdar and Hassan Sajjad and Frank Rudzicz},
booktitle={The Thirty-eighth Annual Conference on Neural Information Processing Systems},
year={2024},
url={https://openreview.net/forum?id=eP9auEJqFg}
}

@article{chen2025basin,
  title={Understanding pre-training and fine-tuning from loss landscape perspectives},
  author={Chen, Huanran and Dong, Yinpeng and Wei, Zeming and Huang, Yao and Zhang, Yichi and Su, Hang and Zhu, Jun},
  journal={arXiv e-prints},
  pages={arXiv--2505},
  year={2025}
}

@ARTICLE{label-noise-survey,
  author={Frenay, Benoit and Verleysen, Michel},
  journal={IEEE Transactions on Neural Networks and Learning Systems}, 
  title={Classification in the Presence of Label Noise: A Survey}, 
  year={2014},
  volume={25},
  number={5},
  pages={845-869},
  doi={10.1109/TNNLS.2013.2292894}}

@article{zhang2025noisesurvey,
title = {A survey on learning with noisy labels in Natural Language Processing: How to train models with label noise},
journal = {Engineering Applications of Artificial Intelligence},
volume = {146},
pages = {110157},
year = {2025},
issn = {0952-1976},
doi = {https://doi.org/10.1016/j.engappai.2025.110157},
url = {https://www.sciencedirect.com/science/article/pii/S0952197625001575},
author = {Han Zhang and Yazhou Zhang and Jiajun Li and Junxiu Liu and Lixia Ji}
}

@article{ratner2017snorkel,
author = {Ratner, Alexander and Bach, Stephen H. and Ehrenberg, Henry and Fries, Jason and Wu, Sen and R\'{e}, Christopher},
title = {Snorkel: rapid training data creation with weak supervision},
year = {2017},
issue_date = {November 2017},
publisher = {VLDB Endowment},
volume = {11},
number = {3},
issn = {2150-8097},
url = {https://doi.org/10.14778/3157794.3157797},
doi = {10.14778/3157794.3157797},
journal = {Proc. VLDB Endow.},
month = nov,
pages = {269–282},
numpages = {14}
}

@InProceedings{pac-bayesian,
  title = 	 {Robust Fine-Tuning of Deep Neural Networks with Hessian-based Generalization Guarantees},
  author =       {Ju, Haotian and Li, Dongyue and Zhang, Hongyang R},
  booktitle = 	 {Proceedings of the 39th International Conference on Machine Learning},
  pages = 	 {10431--10461},
  year = 	 {2022},
  editor = 	 {Chaudhuri, Kamalika and Jegelka, Stefanie and Song, Le and Szepesvari, Csaba and Niu, Gang and Sabato, Sivan},
  volume = 	 {162},
  series = 	 {Proceedings of Machine Learning Research},
  month = 	 {17--23 Jul},
  publisher =    {PMLR},
  url = 	 {https://proceedings.mlr.press/v162/ju22a.html}
}

@unknown{unknown,
author = {Bryant, Christopher and Yuan, Zheng and Qorib, Muhammad and Cao, Hannan and Ng, Hwee and Briscoe, Ted},
year = {2022},
month = {11},
pages = {},
title = {Grammatical Error Correction: A Survey of the State of the Art},
doi = {10.48550/arXiv.2211.05166}
}

@inproceedings{
li2021improved,
title={Improved Regularization and Robustness for Fine-tuning in Neural Networks},
author={Dongyue Li and Hongyang R. Zhang},
booktitle={Advances in Neural Information Processing Systems},
editor={A. Beygelzimer and Y. Dauphin and P. Liang and J. Wortman Vaughan},
year={2021},
url={https://openreview.net/forum?id=j7buX9nsfis}
}

@article{angluin1988learning,
  title={Learning from noisy examples},
  author={Angluin, Dana and Laird, Philip},
  journal={Machine learning},
  volume={2},
  number={4},
  pages={343--370},
  year={1988},
  publisher={Springer}
}

@inproceedings{kim2024towards,
  title={Towards robust and generalized parameter-efficient fine-tuning for noisy label learning},
  author={Kim, Yeachan and Kim, Junho and Lee, SangKeun},
  booktitle={Proceedings of the 62nd Annual Meeting of the Association for Computational Linguistics (Volume 1: Long Papers)},
  pages={5922--5936},
  year={2024}
}

@inproceedings{ethayarajh2019contextual,
    title = "How Contextual are Contextualized Word Representations? {C}omparing the Geometry of {BERT}, {ELM}o, and {GPT}-2 Embeddings",
    author = "Ethayarajh, Kawin",
    editor = "Inui, Kentaro  and
      Jiang, Jing  and
      Ng, Vincent  and
      Wan, Xiaojun",
    booktitle = "Proceedings of the 2019 Conference on Empirical Methods in Natural Language Processing and the 9th International Joint Conference on Natural Language Processing (EMNLP-IJCNLP)",
    month = nov,
    year = "2019",
    address = "Hong Kong, China",
    publisher = "Association for Computational Linguistics",
    url = "https://aclanthology.org/D19-1006/",
    doi = "10.18653/v1/D19-1006",
    pages = "55--65"
}

@inproceedings{
choe2025logix,
title={What is Your Data Worth to {GPT}? {LLM}-Scale Data Valuation with Influence Functions},
author={Sang Keun Choe and Hwijeen Ahn and Juhan Bae and Kewen Zhao and Youngseog Chung and Adithya Pratapa and Willie Neiswanger and Emma Strubell and Teruko Mitamura and Jeff Schneider and Eduard Hovy and Roger Baker Grosse and Eric P. Xing},
booktitle={The Thirty-ninth Annual Conference on Neural Information Processing Systems},
year={2025},
url={https://openreview.net/forum?id=zPKeJAEo27}
}

@inproceedings{geva2021transformer,
    title = "Transformer Feed-Forward Layers Are Key-Value Memories",
    author = "Geva, Mor  and
      Schuster, Roei  and
      Berant, Jonathan  and
      Levy, Omer",
    editor = "Moens, Marie-Francine  and
      Huang, Xuanjing  and
      Specia, Lucia  and
      Yih, Scott Wen-tau",
    booktitle = "Proceedings of the 2021 Conference on Empirical Methods in Natural Language Processing",
    month = nov,
    year = "2021",
    address = "Online and Punta Cana, Dominican Republic",
    publisher = "Association for Computational Linguistics",
    url = "https://aclanthology.org/2021.emnlp-main.446/",
    doi = "10.18653/v1/2021.emnlp-main.446",
    pages = "5484--5495"
}

@inproceedings{
meng2023massediting,
title={Mass-Editing Memory in a Transformer},
author={Kevin Meng and Arnab Sen Sharma and Alex J Andonian and Yonatan Belinkov and David Bau},
booktitle={The Eleventh International Conference on Learning Representations },
year={2023},
url={https://openreview.net/forum?id=MkbcAHIYgyS}
}

@article{yelp_polarity,
  title={Character-level convolutional networks for text classification},
  author={Zhang, Xiang and Zhao, Junbo and LeCun, Yann},
  journal={Advances in neural information processing systems},
  volume={28},
  year={2015}
}

@inproceedings{squad,
    title = "{SQ}u{AD}: 100,000+ Questions for Machine Comprehension of Text",
    author = "Rajpurkar, Pranav  and
      Zhang, Jian  and
      Lopyrev, Konstantin  and
      Liang, Percy",
    editor = "Su, Jian  and
      Duh, Kevin  and
      Carreras, Xavier",
    booktitle = "Proceedings of the 2016 Conference on Empirical Methods in Natural Language Processing",
    month = nov,
    year = "2016",
    address = "Austin, Texas",
    publisher = "Association for Computational Linguistics",
    url = "https://aclanthology.org/D16-1264",
    doi = "10.18653/v1/D16-1264",
    pages = "2383--2392",
    eprint={1606.05250},
    archivePrefix={arXiv},
    primaryClass={cs.CL},
}

@inproceedings{tiedemann2020tatoeba,
  title={The tatoeba translation challenge--realistic data sets for low resource and multilingual MT},
  author={Tiedemann, J{\"o}rg},
  booktitle={Proceedings of the fifth conference on machine translation},
  pages={1174--1182},
  year={2020}
}

@article{grosse2023studying,
  title={Studying large language model generalization with influence functions},
  author={Grosse, Roger and Bae, Juhan and Anil, Cem and Elhage, Nelson and Tamkin, Alex and Tajdini, Amirhossein and Steiner, Benoit and Li, Dustin and Durmus, Esin and Perez, Ethan and others},
  journal={arXiv preprint arXiv:2308.03296},
  year={2023}
}

@article{radford2019language,
  title={Language models are unsupervised multitask learners},
  author={Radford, Alec and Wu, Jeffrey and Child, Rewon and Luan, David and Amodei, Dario and Sutskever, Ilya and others},
  journal={OpenAI blog},
  volume={1},
  number={8},
  pages={9},
  year={2019}
}

@article{qwen2,
    title   = {Qwen2 Technical Report}, 
    author  = {An Yang and Baosong Yang and Binyuan Hui and Bo Zheng and Bowen Yu and Chang Zhou and Chengpeng Li and Chengyuan Li and Dayiheng Liu and Fei Huang and Guanting Dong and Haoran Wei and Huan Lin and Jialong Tang and Jialin Wang and Jian Yang and Jianhong Tu and Jianwei Zhang and Jianxin Ma and Jin Xu and Jingren Zhou and Jinze Bai and Jinzheng He and Junyang Lin and Kai Dang and Keming Lu and Keqin Chen and Kexin Yang and Mei Li and Mingfeng Xue and Na Ni and Pei Zhang and Peng Wang and Ru Peng and Rui Men and Ruize Gao and Runji Lin and Shijie Wang and Shuai Bai and Sinan Tan and Tianhang Zhu and Tianhao Li and Tianyu Liu and Wenbin Ge and Xiaodong Deng and Xiaohuan Zhou and Xingzhang Ren and Xinyu Zhang and Xipin Wei and Xuancheng Ren and Yang Fan and Yang Yao and Yichang Zhang and Yu Wan and Yunfei Chu and Yuqiong Liu and Zeyu Cui and Zhenru Zhang and Zhihao Fan},
    journal = {arXiv preprint arXiv:2407.10671},
    year    = {2024}
}

@misc{touvron2023llama2openfoundation,
      title={Llama 2: Open Foundation and Fine-Tuned Chat Models}, 
      author={Hugo Touvron and Louis Martin and Kevin Stone and Peter Albert and Amjad Almahairi and Yasmine Babaei and Nikolay Bashlykov and Soumya Batra and Prajjwal Bhargava and Shruti Bhosale and Dan Bikel and Lukas Blecher and Cristian Canton Ferrer and Moya Chen and Guillem Cucurull and David Esiobu and Jude Fernandes and Jeremy Fu and Wenyin Fu and Brian Fuller and Cynthia Gao and Vedanuj Goswami and Naman Goyal and Anthony Hartshorn and Saghar Hosseini and Rui Hou and Hakan Inan and Marcin Kardas and Viktor Kerkez and Madian Khabsa and Isabel Kloumann and Artem Korenev and Punit Singh Koura and Marie-Anne Lachaux and Thibaut Lavril and Jenya Lee and Diana Liskovich and Yinghai Lu and Yuning Mao and Xavier Martinet and Todor Mihaylov and Pushkar Mishra and Igor Molybog and Yixin Nie and Andrew Poulton and Jeremy Reizenstein and Rashi Rungta and Kalyan Saladi and Alan Schelten and Ruan Silva and Eric Michael Smith and Ranjan Subramanian and Xiaoqing Ellen Tan and Binh Tang and Ross Taylor and Adina Williams and Jian Xiang Kuan and Puxin Xu and Zheng Yan and Iliyan Zarov and Yuchen Zhang and Angela Fan and Melanie Kambadur and Sharan Narang and Aurelien Rodriguez and Robert Stojnic and Sergey Edunov and Thomas Scialom},
      year={2023},
      eprint={2307.09288},
      archivePrefix={arXiv},
      primaryClass={cs.CL},
      url={https://arxiv.org/abs/2307.09288}, 
}

@article{dettmers2023qlora,
  title={Qlora: Efficient finetuning of quantized llms},
  author={Dettmers, Tim and Pagnoni, Artidoro and Holtzman, Ari and Zettlemoyer, Luke},
  journal={Advances in neural information processing systems},
  volume={36},
  pages={10088--10115},
  year={2023}
}

@software{vonwerra2020trl,
  title   = {{TRL: Transformers Reinforcement Learning}},
  author  = {von Werra, Leandro and Belkada, Younes and Tunstall, Lewis and Beeching, Edward and Thrush, Tristan and Lambert, Nathan and Huang, Shengyi and Rasul, Kashif and Gallouédec, Quentin},
  license = {Apache-2.0},
  url     = {https://github.com/huggingface/trl},
  year    = {2020}
}

@inproceedings{post2018call,
    title = "A Call for Clarity in Reporting {BLEU} Scores",
    author = "Post, Matt",
    editor = "Bojar, Ond{\v{r}}ej  and
      Chatterjee, Rajen  and
      Federmann, Christian  and
      Fishel, Mark  and
      Graham, Yvette  and
      Haddow, Barry  and
      Huck, Matthias  and
      Yepes, Antonio Jimeno  and
      Koehn, Philipp  and
      Monz, Christof  and
      Negri, Matteo  and
      N{\'e}v{\'e}ol, Aur{\'e}lie  and
      Neves, Mariana  and
      Post, Matt  and
      Specia, Lucia  and
      Turchi, Marco  and
      Verspoor, Karin",
    booktitle = "Proceedings of the Third Conference on Machine Translation: Research Papers",
    month = oct,
    year = "2018",
    address = "Brussels, Belgium",
    publisher = "Association for Computational Linguistics",
    url = "https://aclanthology.org/W18-6319/",
    doi = "10.18653/v1/W18-6319",
    pages = "186--191"
}

@inproceedings{cka,
  title={Similarity of neural network representations revisited},
  author={Kornblith, Simon and Norouzi, Mohammad and Lee, Honglak and Hinton, Geoffrey},
  booktitle={International conference on machine learning},
  pages={3519--3529},
  year={2019},
  organization={PMlR}
}

@misc{nostalgebraist2020logitlens,
  author       = {nostalgebraist},
  title        = {Interpreting GPT: The Logit Lens},
  howpublished = {\url{https://www.lesswrong.com/posts/AcKRB8wDpdaN6v6ru/interpreting-gpt-the-logit-lens}},
  year         = {2020},
  note         = {LessWrong blog post},
}
\bibliographystyle{tmlr}

\appendix

\section{Dataset Statistics and Prompt Templates}
  \label{app:prompts}

  \paragraph{Dataset statistics.}
  Table~\ref{tab:dataset-stats} summarises the datasets and split sizes used for each task.
  All models share the same training, validation, and test samples for a given task.

  \begin{table}[h]
  \caption{Dataset statistics for each task.}
  \label{tab:dataset-stats}
  \centering
  \small
  \begin{tabular}{llrrr}
  \toprule
  \textbf{Task} & \textbf{Source} & \textbf{Train} & \textbf{Val} & \textbf{Test} \\
  \midrule
  Sentiment & Yelp Polarity & 10\,000 & 1\,000 & 1\,000 \\
  QA        & SQuAD v1.1    & 10\,000 & 1\,000 & 1\,000 \\
  MT        & Tatoeba EN--FR & 20\,000 & 1\,000 & 1\,000 \\
  \bottomrule
  \end{tabular}
  \end{table}

  \paragraph{Prompt templates.}
  All tasks are formatted as causal language modelling with a prompt--completion structure.
  Table~\ref{tab:prompt-templates} lists the prompt template used for each model--task combination.
  During training under the SFT-MASK protocol, the loss is computed only over the completion tokens (shown after the final colon). Prompt templates were selected per model to match each architecture's
  pre-training conventions. Crucially, the prompt is held constant across
  all noise conditions for a given model--task pair, ensuring that
  observed performance differences reflect only the effect of training
  data corruption.

  \begin{table}[h]
  \caption{Prompt templates by model and task. The completion begins after the final colon in each template. \texttt{\textbackslash n} denotes a newline character.}
  \label{tab:prompt-templates}
  \centering
  \small
  \begin{tabular}{lll}
  \toprule
  \textbf{Task} & \textbf{Model} & \textbf{Template} \\
  \midrule
  \multirow{1}{*}{Sentiment} & All
    & \texttt{Review: \{text\}\textbackslash nSentiment: \{label\}} \\
  \addlinespace
  \multirow{2}{*}{QA}
    & GPT-2
    & \texttt{Context: \{c\}\textbackslash nQuestion: \{q\}\textbackslash nAnswer: \{a\}} \\
    & LLaMA-2, Qwen-2
    & \texttt{\#\#\# Context:\textbackslash n\{c\}\textbackslash n\textbackslash n\#\#\# Question:\textbackslash n\{q\}\textbackslash n\textbackslash n\#\#\# Answer: \{a\}} \\
  \addlinespace
  \multirow{2}{*}{MT}
    & GPT-2, LLaMA-2
    & \texttt{English: \{eng\}\textbackslash nFrench: \{fra\}} \\
    & Qwen-2
    & \texttt{Translate English to French.\textbackslash n\textbackslash n\#\#\# English:\textbackslash n\{eng\}\textbackslash n\textbackslash n\#\#\# French: \{fra\}} \\
  \bottomrule
  \end{tabular}
  \end{table}

\subsection{Noise Type Examples}
  \label{app:noise-examples}

  Tables~\ref{tab:noise-label-flip}--\ref{tab:noise-grammatical} show concrete before-and-after examples for each noise type applied to the three tasks.
  Corrupted portions are shown in \textbf{bold}.

  \begin{table}[h]
  \caption{Label flip noise examples. The input text remains unchanged; only the target label/output is replaced.}
  \label{tab:noise-label-flip}
  \centering
  \small
  \begin{tabularx}{\textwidth}{l X X}
  \toprule
  \textbf{Task} & \textbf{Original} & \textbf{After Label Flip} \\
  \midrule
  Sentiment
  & \textit{Text:} ``The food was terrible and the service was even worse.'' \newline \textit{Label:} Negative
  & \textit{Text:} ``The food was terrible and the service was even worse.'' \newline \textit{Label:} \textbf{Positive} \\
  \addlinespace
  QA
  & \textit{Context:} ``The Eiffel Tower was built in 1889 for the World's Fair. It is located in Paris, France.'' \newline \textit{Question:} ``When was the Eiffel Tower built?'' \newline \textit{Answer:} ``1889''
  & \textit{Context:} (unchanged) \newline \textit{Question:} (unchanged) \newline \textit{Answer:} \textbf{``the 10th century''} \\
  \addlinespace
  MT
  & \textit{English:} ``Swimming at night is dangerous.'' \newline \textit{French:} ``Il est dangereux de nager de nuit.''
  & \textit{English:} (unchanged) \newline \textit{French:} \textbf{``C'est la saison des fraises.''} \\
  \bottomrule
  \end{tabularx}

  \end{table}

  \begin{table}[h]
  \caption{Typo noise examples. Character-level perturbations (deletion, swap, insertion, or substitution) are applied to randomly selected words at a rate of 10\% of words per sample.}
  \label{tab:noise-typo}
  \centering
  \small
  \begin{tabularx}{\textwidth}{l X X}
  \toprule
  \textbf{Task} & \textbf{Original} & \textbf{After Typo Injection} \\
  \midrule
  Sentiment
  & ``The food was terrible and the service was even worse.''
  & ``The \textbf{fodo} was terrible and the service was even \textbf{wrse}.'' \\
  \addlinespace
  QA
  & \textit{Context:} ``The Eiffel Tower was built in 1889 for the World's Fair.''
  & \textit{Context:} ``The Eiffel \textbf{Towr} was built in 1889 for the World's Fair.'' \\
  \addlinespace
  MT
  & \textit{English:} ``Swimming at night is dangerous.''
  & \textit{English:} ``\textbf{Swimmiing} at night is dangerous.'' \\
  \bottomrule
  \end{tabularx}
  \end{table}

  \begin{table}[h]
  \caption{Grammatical noise examples. Rule-based substitutions target verb conjugation (\textit{is}$\leftrightarrow$\textit{are}, \textit{was}$\leftrightarrow$\textit{were}, \textit{has}$\leftrightarrow$\textit{have}) and
   article usage (\textit{a}$\leftrightarrow$\textit{an}) at a rate of 15\% of words per sample.}
  \label{tab:noise-grammatical}
  \centering
  \small
  \begin{tabularx}{\textwidth}{l X X}
  \toprule
  \textbf{Task} & \textbf{Original} & \textbf{After Grammatical Errors} \\
  \midrule
  Sentiment
  & ``The food was terrible and the service was even worse.''
  & ``The food \textbf{were} terrible and the service \textbf{were} even worse.'' \\
  \addlinespace
  QA
  & \textit{Context:} ``The Eiffel Tower was built in 1889. It is located in Paris.''
  & \textit{Context:} ``The Eiffel Tower \textbf{were} built in 1889. It \textbf{are} located in Paris.'' \\
  \addlinespace
  MT
  & \textit{English:} ``Swimming at night is dangerous.''
  & \textit{English:} ``Swimming at night \textbf{are} dangerous.'' \\
  \bottomrule
  \end{tabularx}
  \end{table}

 \section{Training Hyperparameters}
  \label{app:hyperparameters}

  All models are trained with the AdamW optimiser and a cosine learning-rate
  schedule. GPT-2 is fully fine-tuned; LLaMA-2 and Qwen-2 use QLoRA
  (4-bit NF4 quantisation) with LoRA adapters.
  Table~\ref{tab:lora-config} lists the LoRA configuration for each
  model--task pair. Table~\ref{tab:training-config} lists the remaining
  training hyperparameters.

  \begin{table}[h]
  \caption{QLoRA adapter configuration. ``All~7'' denotes
  \texttt{q\_proj}, \texttt{k\_proj}, \texttt{v\_proj}, \texttt{o\_proj},
  \texttt{gate\_proj}, \texttt{up\_proj}, \texttt{down\_proj}.}
  \label{tab:lora-config}
  \centering
  \small
  \begin{tabular}{llcccc}
  \toprule
  \textbf{Model} & \textbf{Task} & $r$ & $\alpha$ & \textbf{Dropout} &
  \textbf{Target Modules} \\
  \midrule
  \multirow{3}{*}{Qwen-2}
    & Sentiment & 8   & 32  & 0.10 & All 7 \\
    & QA        & 16  & 32  & 0.05 & All 7 \\
    & MT        & 32  & 64  & 0.05 & All 7 \\
  \addlinespace
  \multirow{3}{*}{LLaMA-2}
    & Sentiment & 32  & 64  & 0.05 & All 7 \\
    & QA        & 16  & 32  & 0.05 & All 7 \\
    & MT        & 128 & 256 & 0.05 & All 7 \\
  \bottomrule
  \end{tabular}
  \end{table}

  \begin{table}[h]
  \caption{Training hyperparameters. All configurations use cosine LR scheduling.
  Weight decay is 0 except where noted.}
  \label{tab:training-config}
  \centering
  \small
  \begin{tabular}{llccccc}
  \toprule
  \textbf{Model} & \textbf{Task} & \textbf{LR} & \textbf{Eff.\ Batch} &
  \textbf{Epochs} & \textbf{Warmup} & \textbf{Max Len} \\
  \midrule
  \multirow{3}{*}{GPT-2}
    & Sentiment & 5e-5 & 32 & 3  & 0.03 & 256  \\
    & QA        & 2e-5 & 32 & 3  & 0.03 & 1024 \\
    & MT        & 5e-5 & 32 & 30 & 0.03 & 256  \\
  \addlinespace
  \multirow{3}{*}{Qwen-2}
    & Sentiment & 5e-5 & 80 & 3  & 0.03 & 256  \\
    & QA        & 2e-5 & 32 & 2  & 0.03 & 1024 \\
    & MT        & 2e-4 & 32 & 3  & 0.03 & 256  \\
  \addlinespace
  \multirow{3}{*}{LLaMA-2}
    & Sentiment & 1e-4 & 16 & 3  & 0.05 & 256  \\
    & QA        & 2e-4 & 8  & 2  & 0.03 & 1024 \\
    & MT        & 1e-4 & 32 & 10 & 0.05 & 128  \\
  \bottomrule
  \end{tabular}
  \end{table}

\section{Layer-wise Centered Cosine Similarity}
\label{app:centered_cosine}

\begin{figure*}[t]
    \centering
  \begin{subfigure}[t]{0.32\textwidth}
    \includegraphics[width=\textwidth]{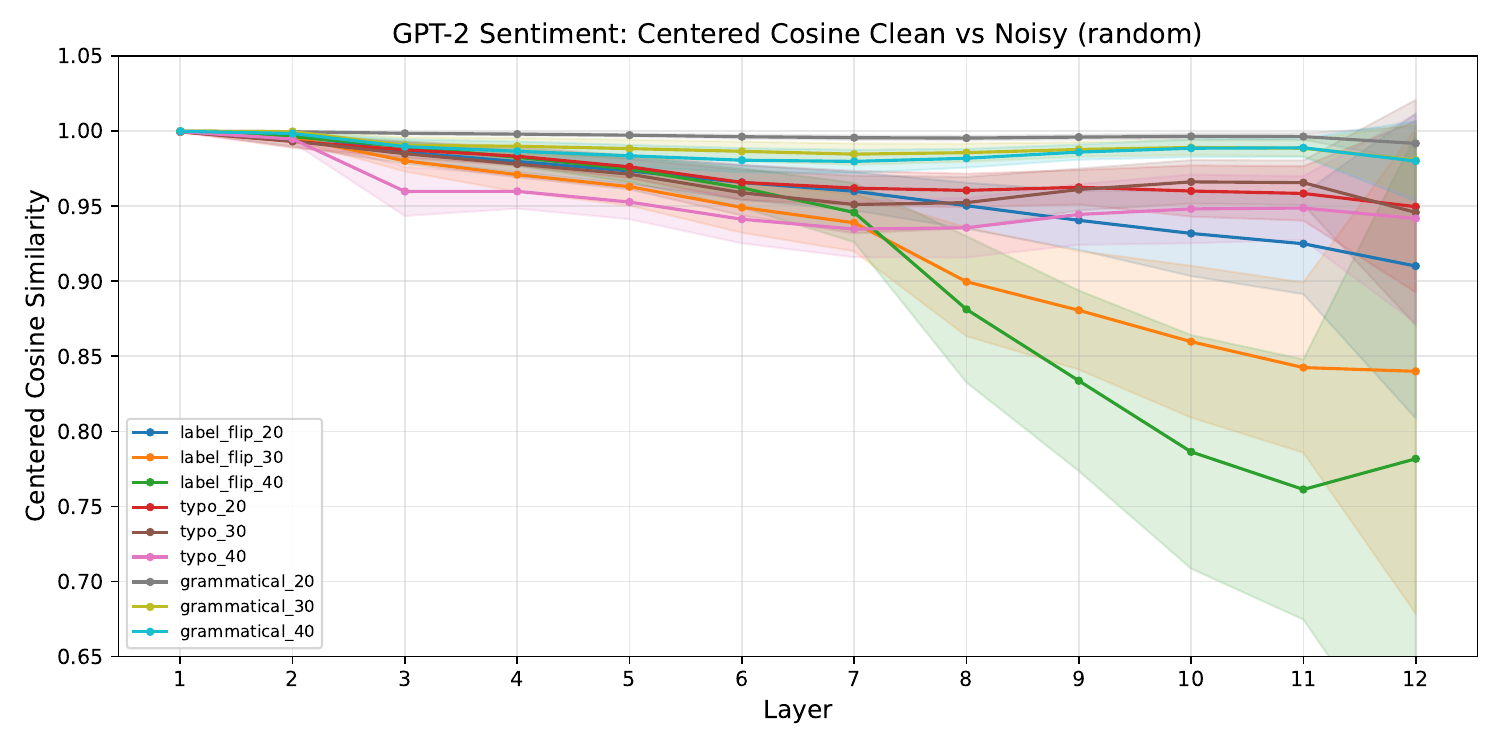}
  \end{subfigure}
  \hfill
  \begin{subfigure}[t]{0.32\textwidth}
    \includegraphics[width=\textwidth]{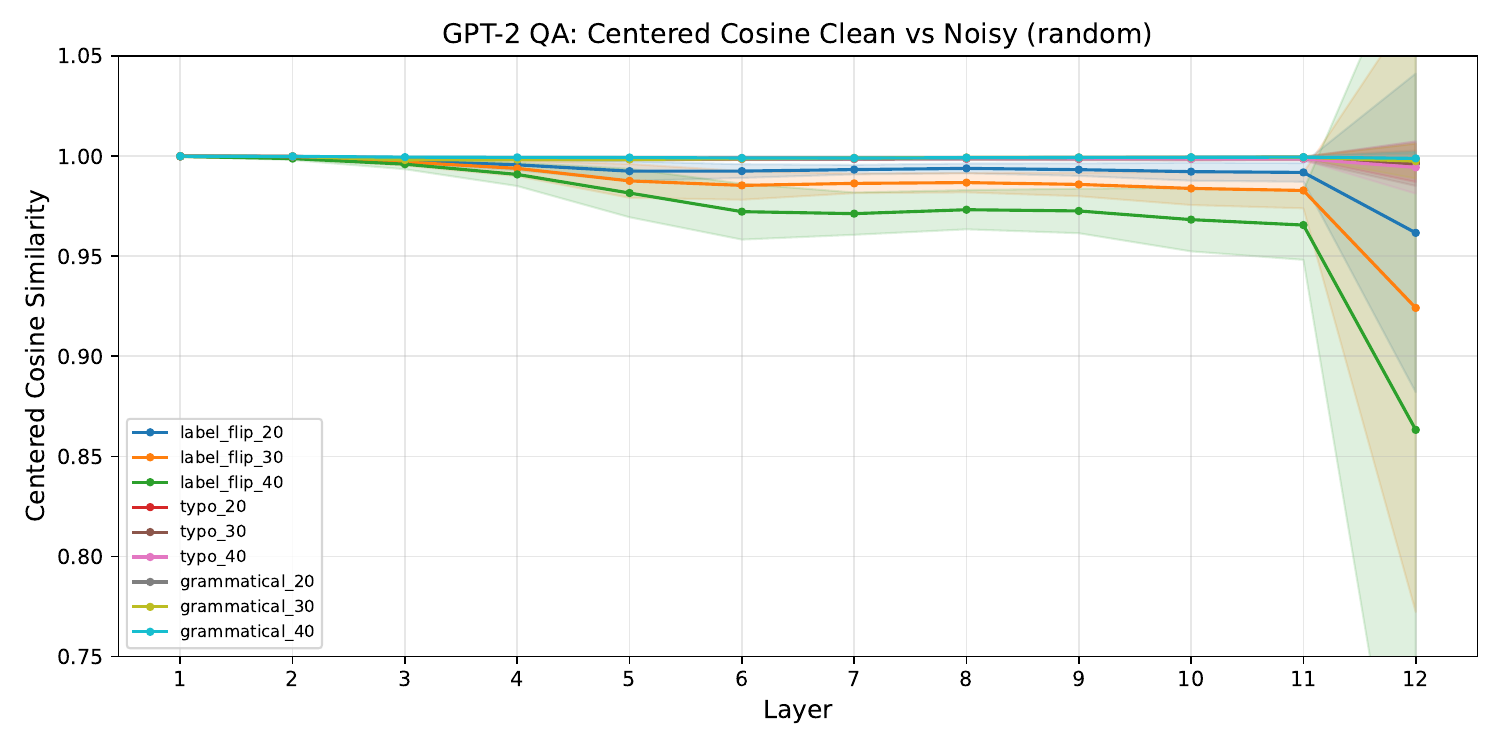}
  \end{subfigure}
  \hfill
  \begin{subfigure}[t]{0.32\textwidth}
    \includegraphics[width=\textwidth]{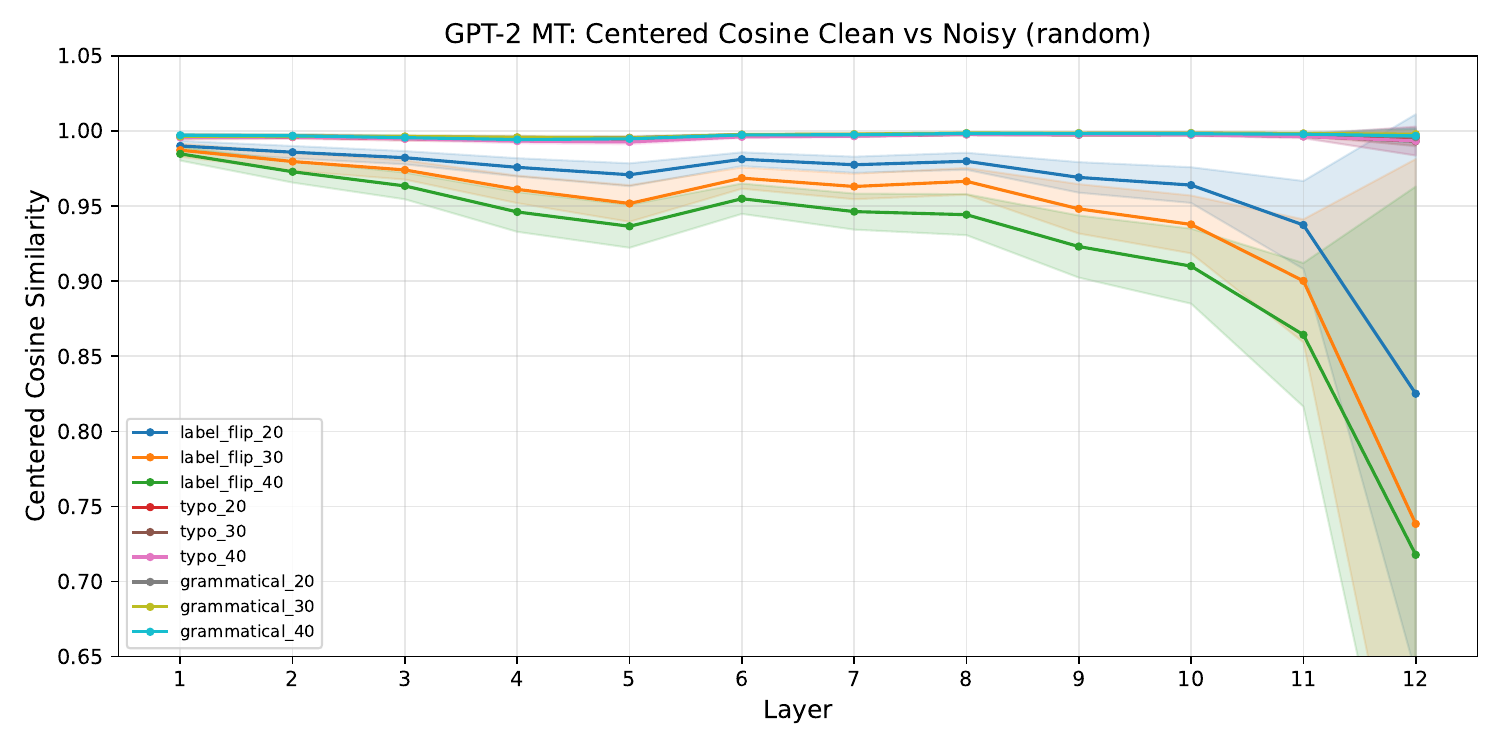}
  \end{subfigure}
    \begin{subfigure}[t]{0.32\textwidth}
    \includegraphics[width=\textwidth]{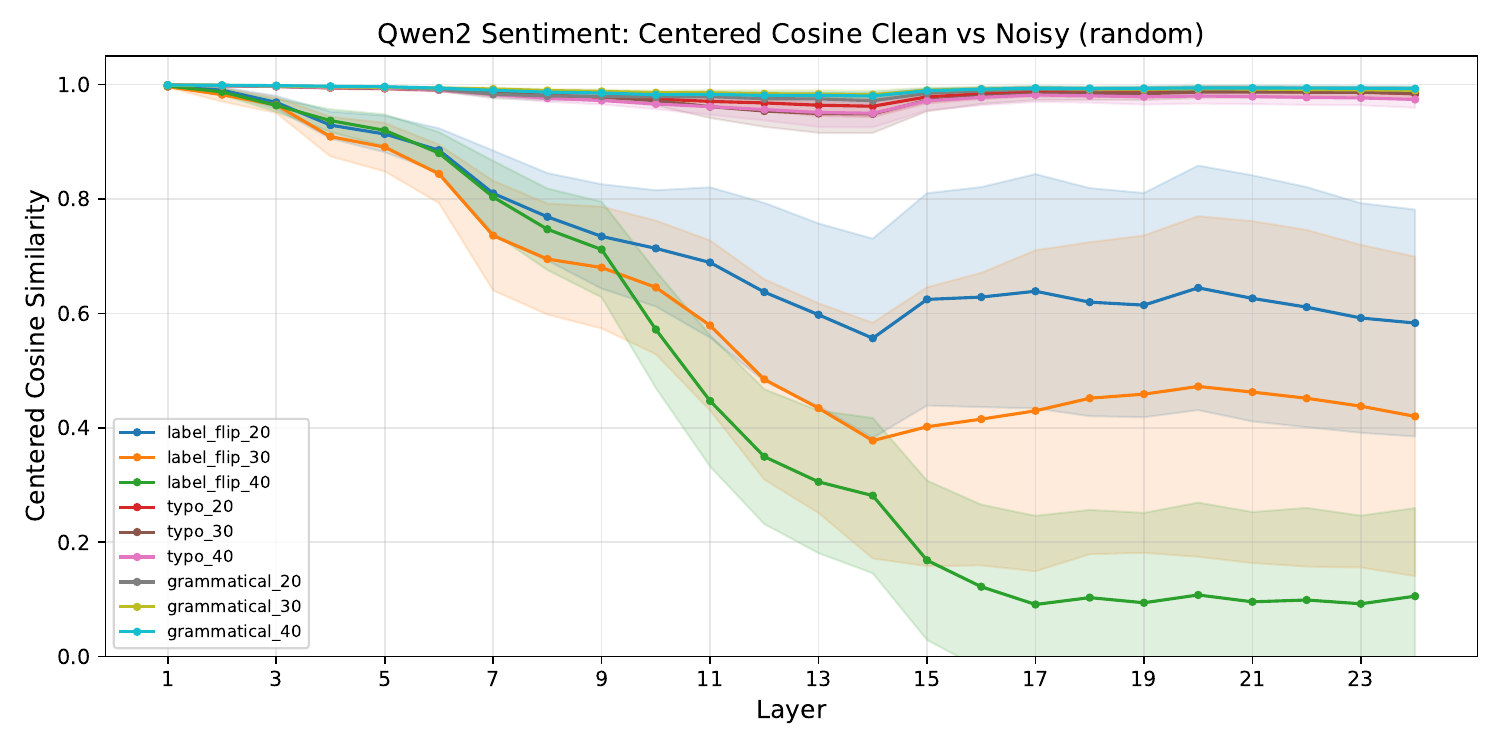}
  \end{subfigure}
  \hfill
  \begin{subfigure}[t]{0.32\textwidth}
    \includegraphics[width=\textwidth]{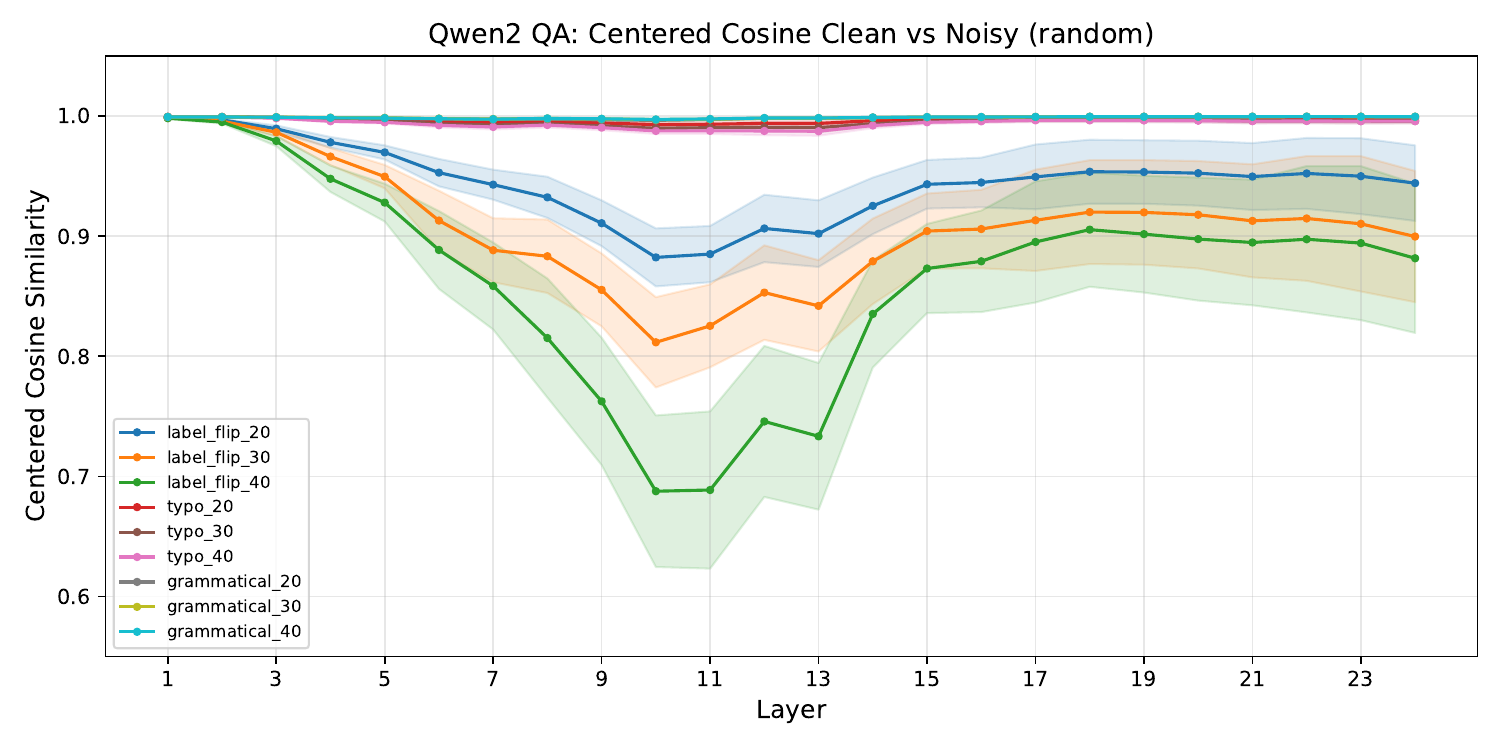}
  \end{subfigure}
  \hfill
  \begin{subfigure}[t]{0.32\textwidth}
    \includegraphics[width=\textwidth]{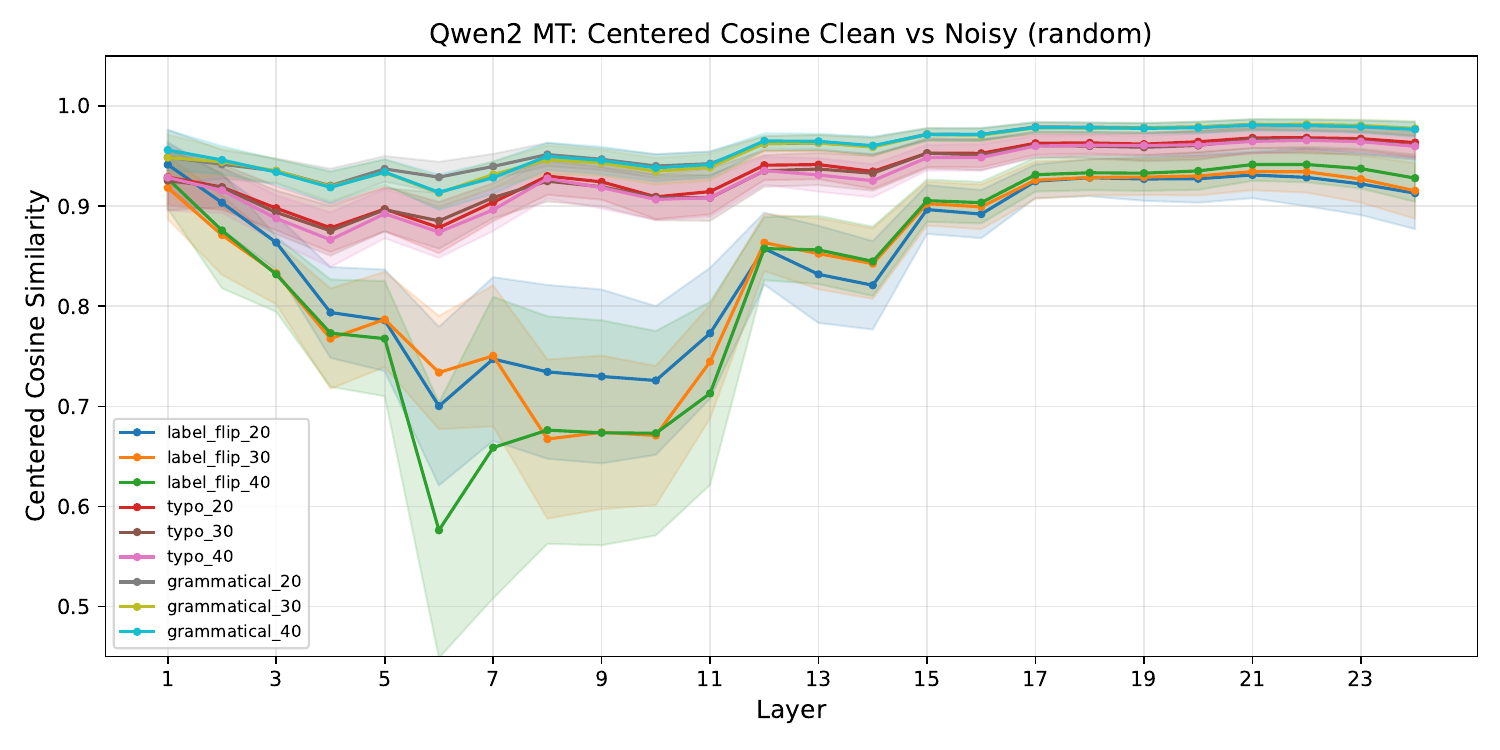}
  \end{subfigure}
    \begin{subfigure}[t]{0.32\textwidth}
    \includegraphics[width=\textwidth]{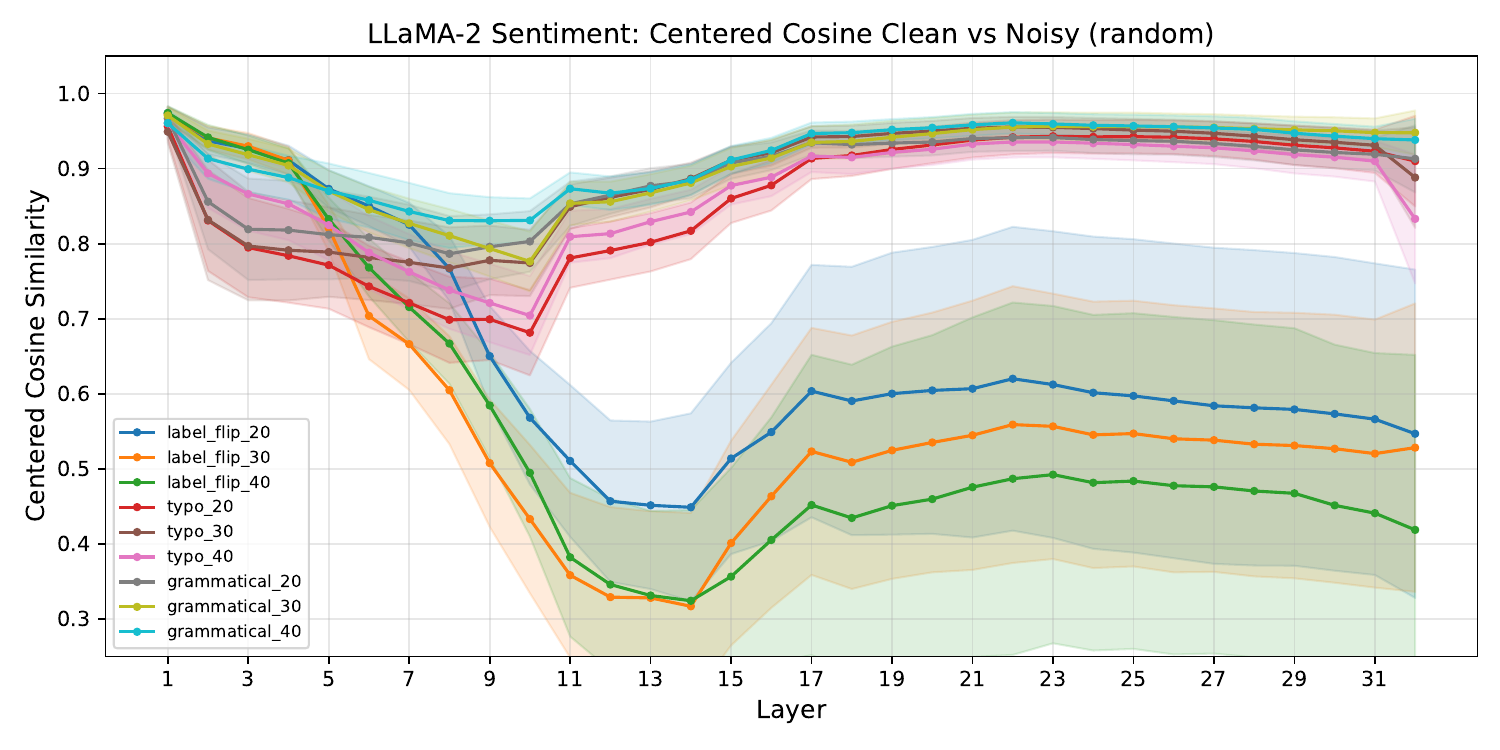}
  \end{subfigure}
  \hfill
  \begin{subfigure}[t]{0.32\textwidth}
    \includegraphics[width=\textwidth]{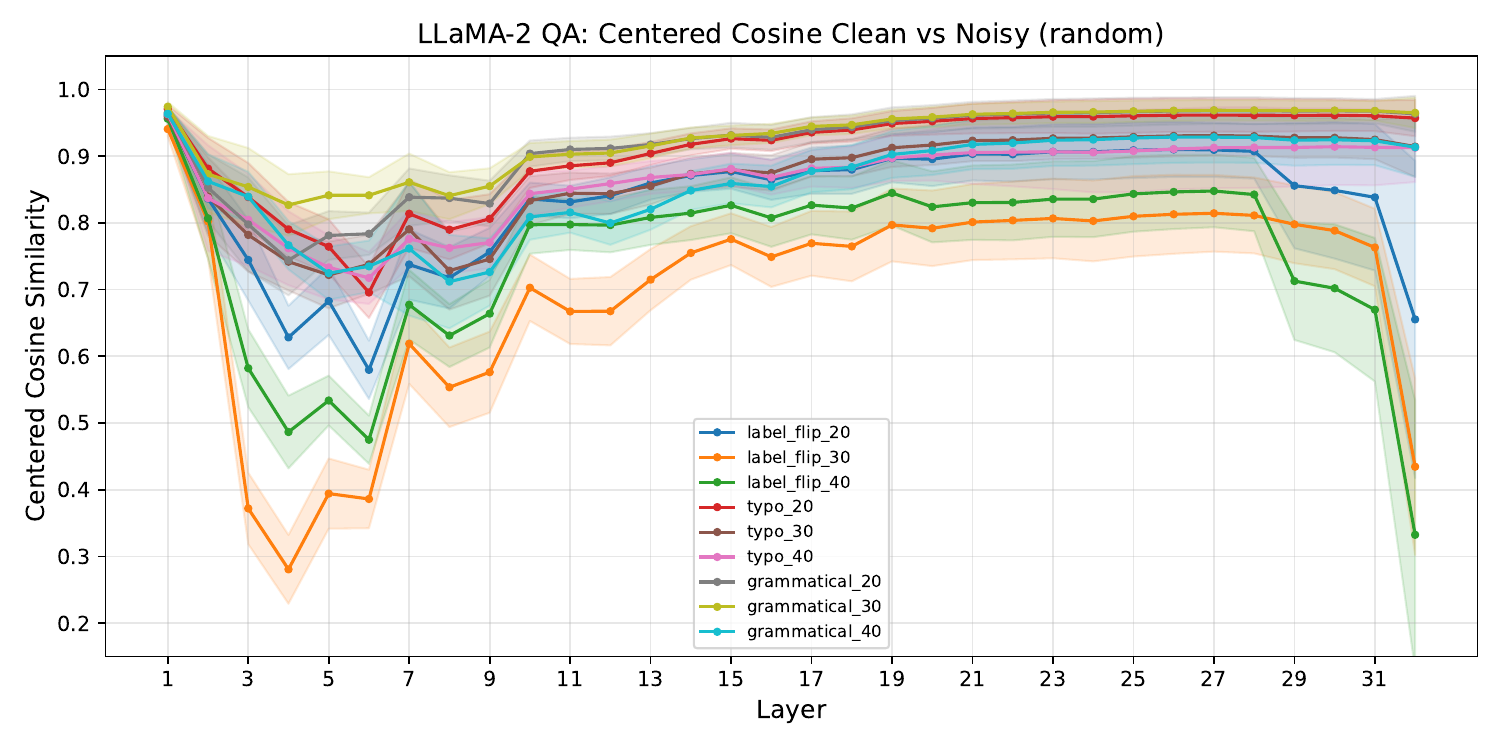}
  \end{subfigure}
  \hfill
  \begin{subfigure}[t]{0.32\textwidth}
    \includegraphics[width=\textwidth]{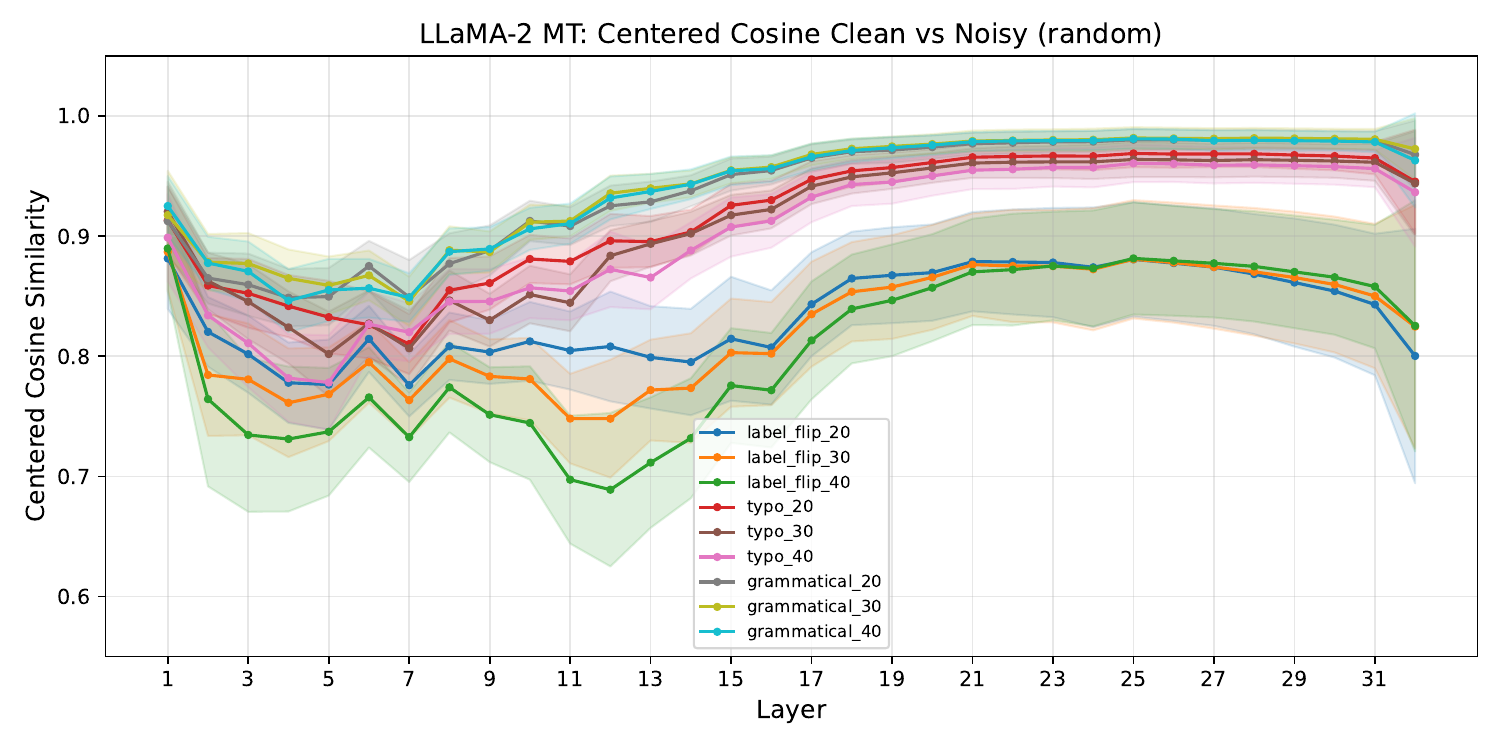}
  \end{subfigure}
    \caption{Layer-wise centered cosine similarity between clean and noise-trained model representations across all nine model--task combinations. Rows correspond to models (GPT-2, Qwen-2, LLaMA-2); columns correspond to tasks (SC, QA, MT). Centered cosine removes the shared mean direction before computing similarity, correcting for the anisotropy of contextualised representations. }
    \label{fig:centered_cosine_grid}
\end{figure*}

Figure~\ref{fig:centered_cosine_grid} presents the full layer-wise centered cosine similarity between clean and noise-trained models across all nine model--task combinations.
Centered cosine similarity is computed after subtracting the sample-wise mean from each representation matrix, removing the anisotropy-induced bias that inflates raw cosine similarity~\citep{ethayarajh2019contextual}.

\section{Robust Vs. Vulnerable Stratification} \label{ap:stratification}

\begin{figure*}[p]
  \centering
  \begin{subfigure}[t]{0.32\textwidth}
    \includegraphics[width=\textwidth]{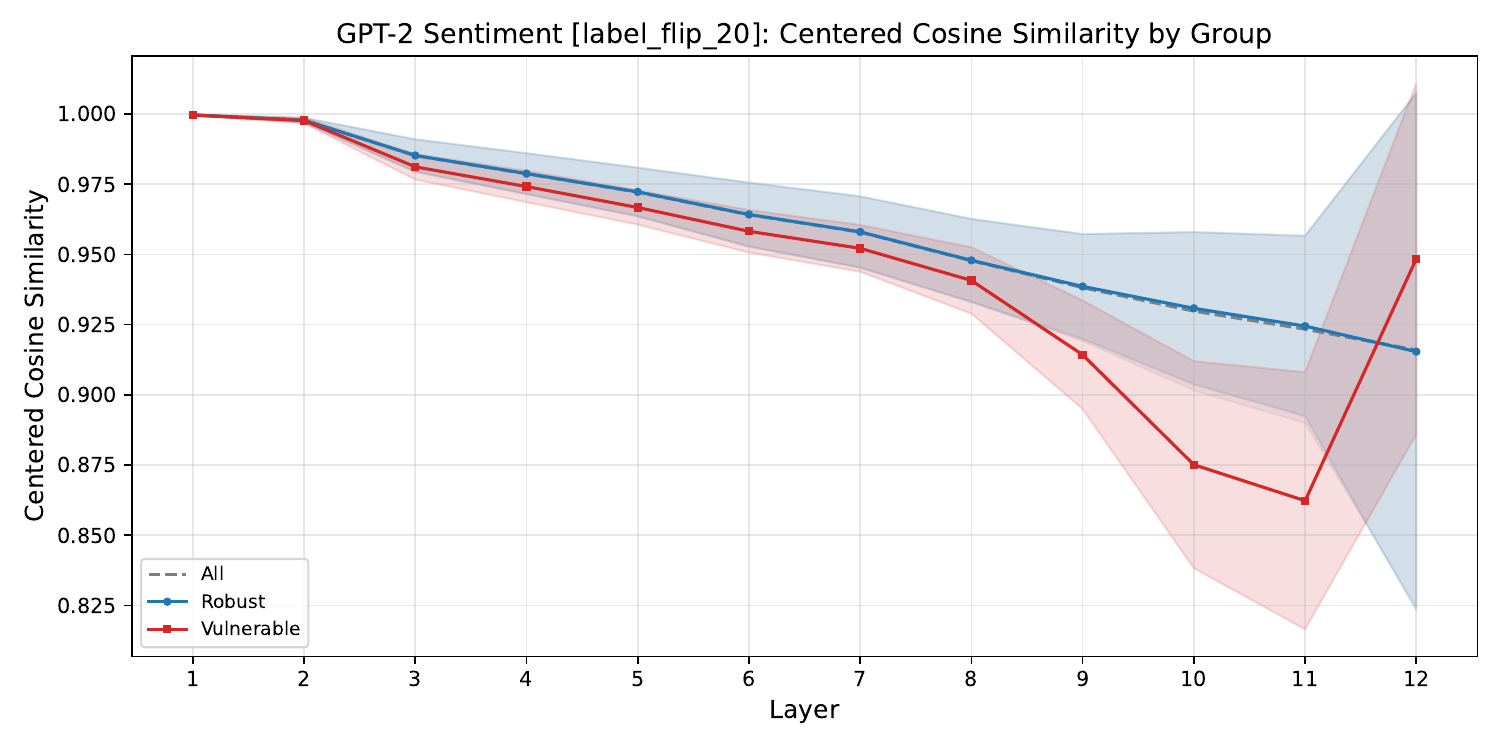}
  \end{subfigure}\hfill
  \begin{subfigure}[t]{0.32\textwidth}
    \includegraphics[width=\textwidth]{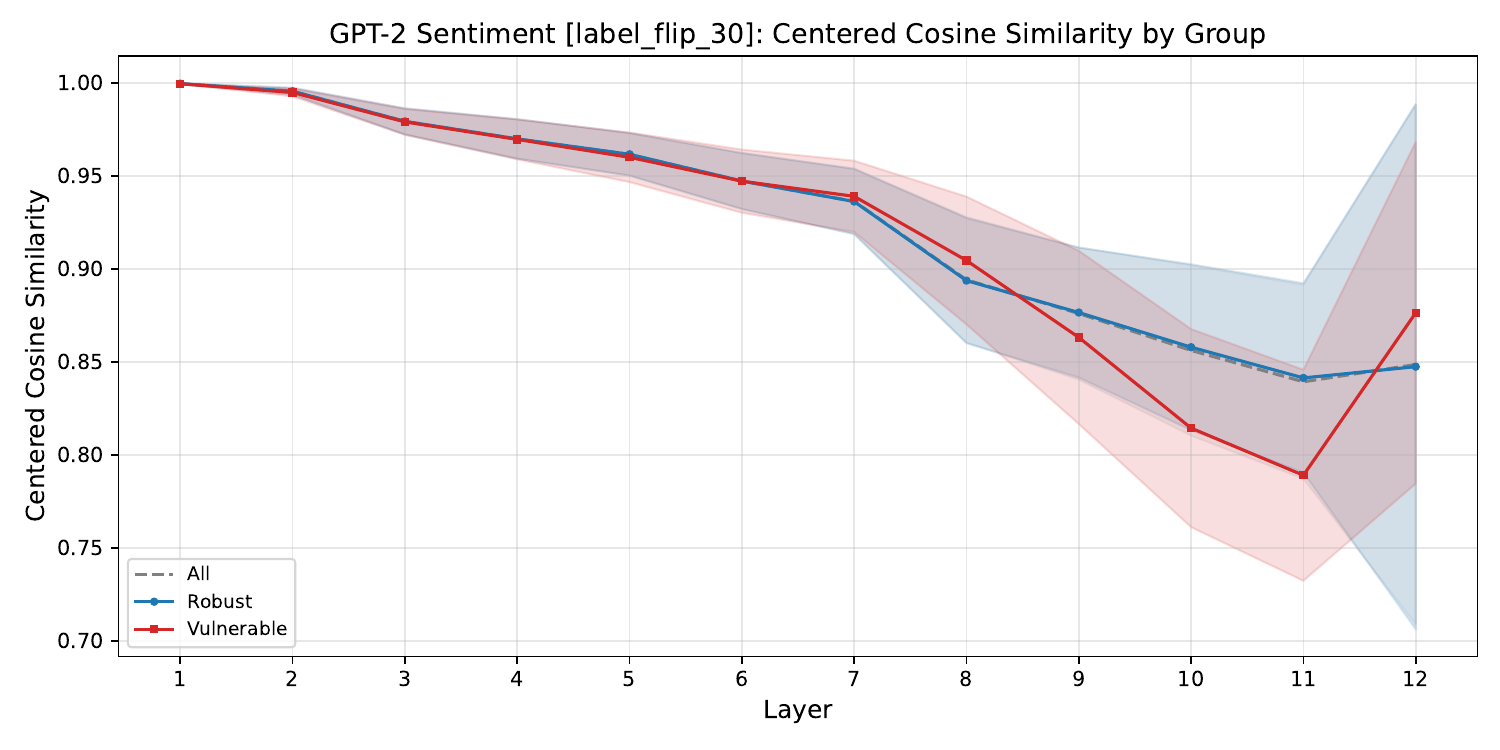}
  \end{subfigure}\hfill
  \begin{subfigure}[t]{0.32\textwidth}
    \includegraphics[width=\textwidth]{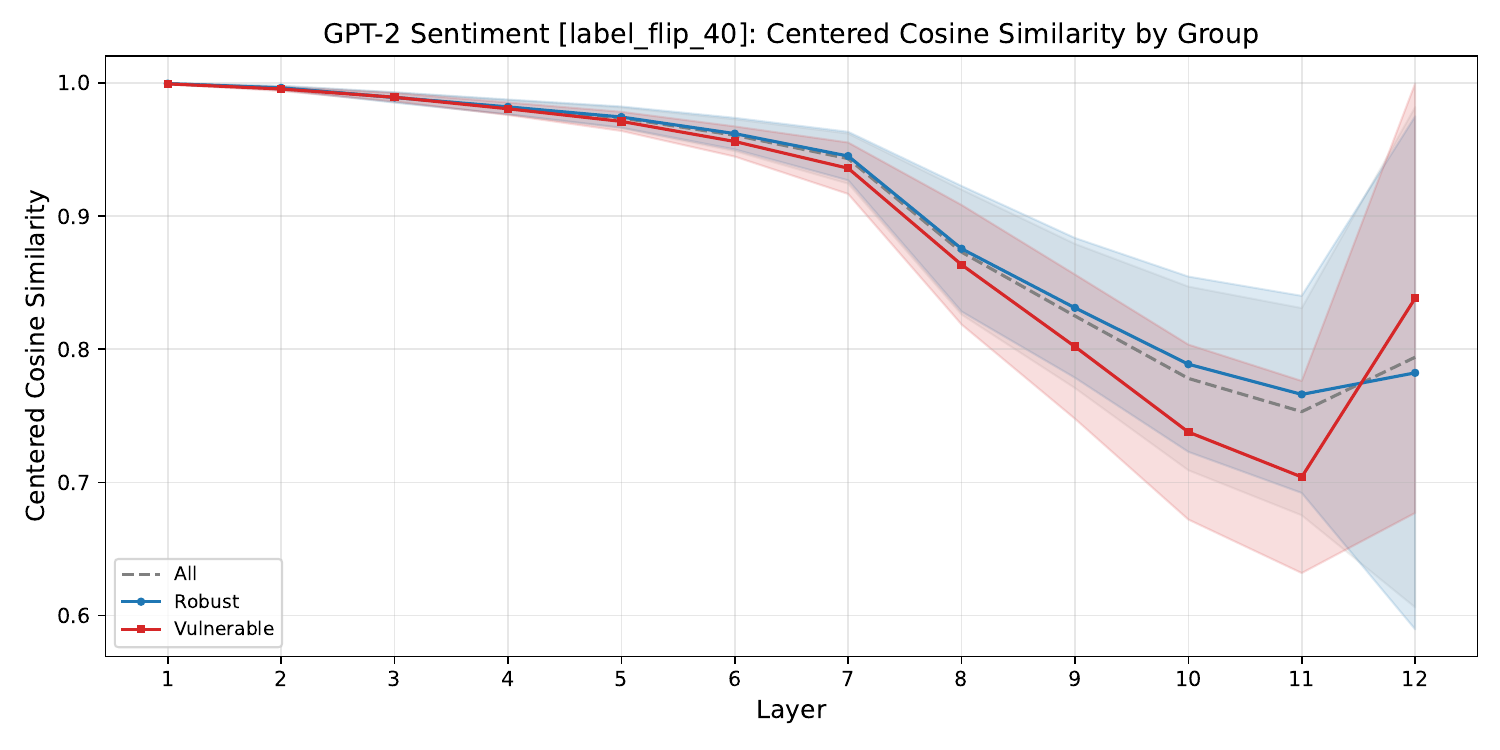}
  \end{subfigure}
  \begin{subfigure}[t]{0.32\textwidth}
    \includegraphics[width=\textwidth]{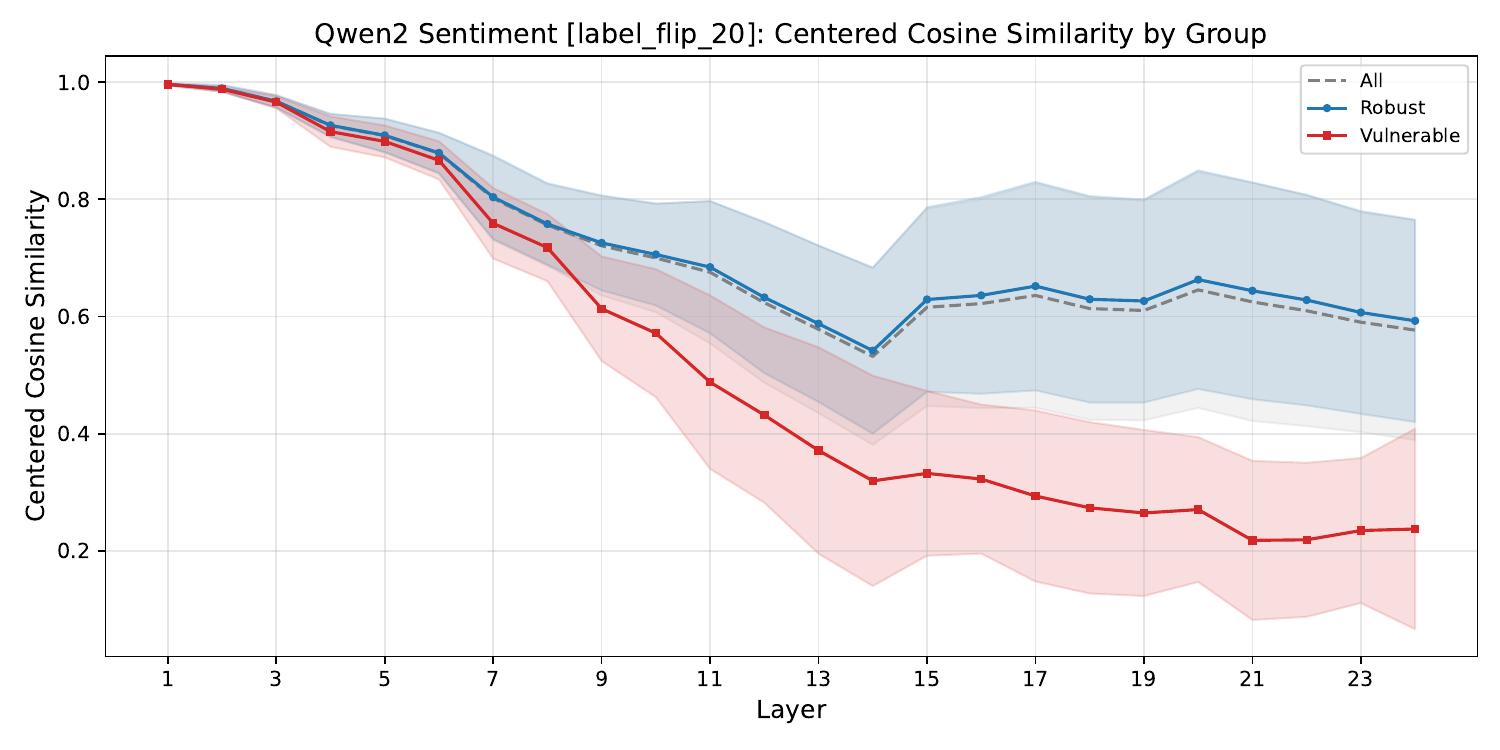}
  \end{subfigure}\hfill
  \begin{subfigure}[t]{0.32\textwidth}
    \includegraphics[width=\textwidth]{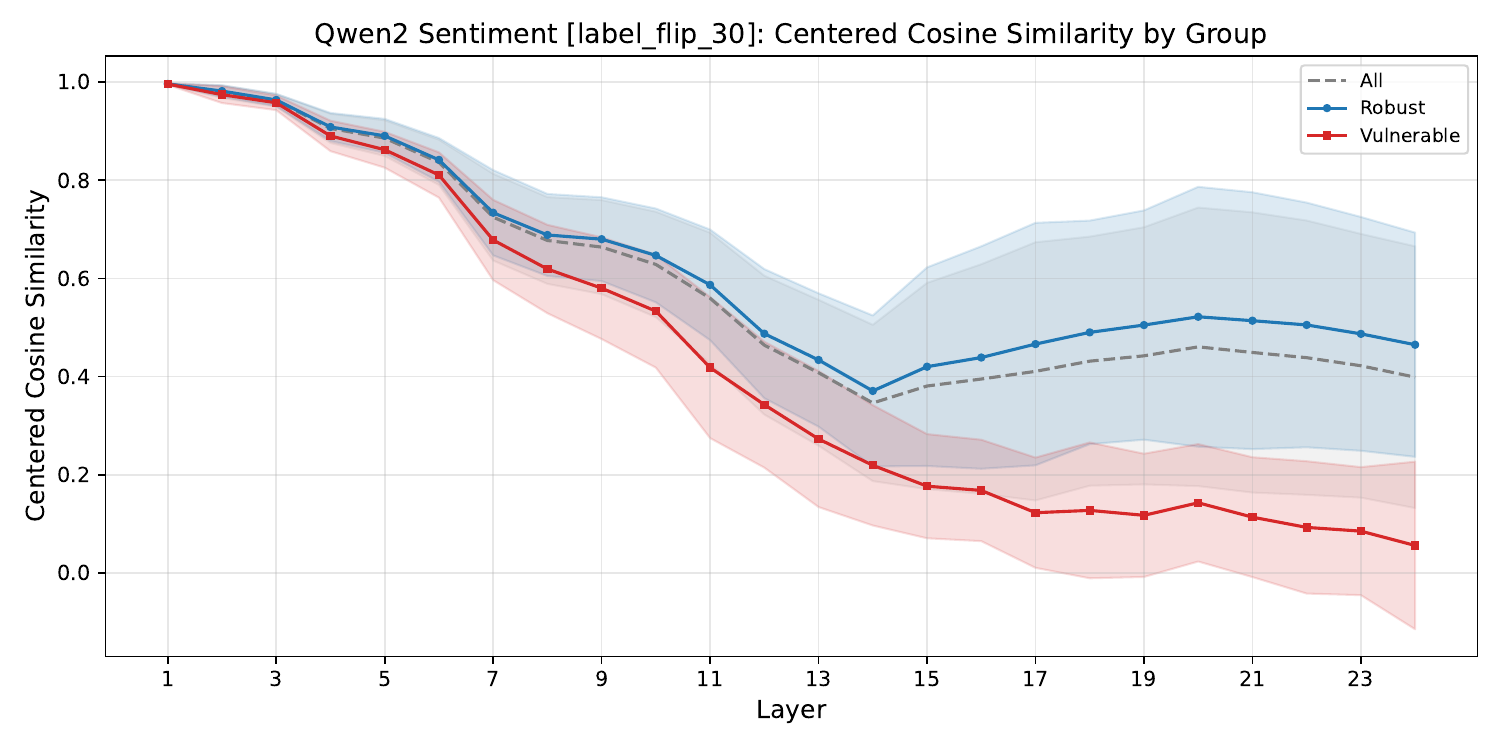}
  \end{subfigure}\hfill
  \begin{subfigure}[t]{0.32\textwidth}
    \includegraphics[width=\textwidth]{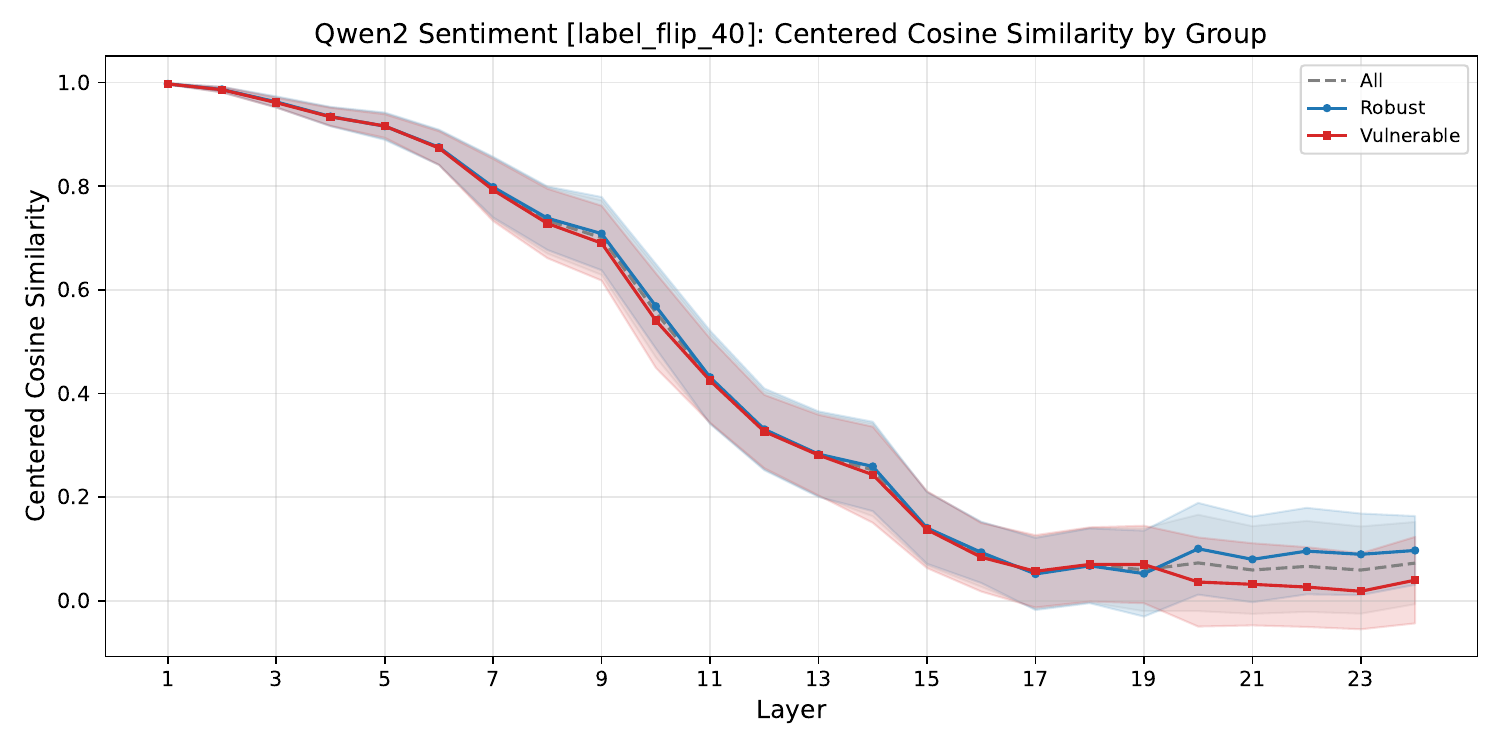}
  \end{subfigure}
  \begin{subfigure}[t]{0.32\textwidth}
    \includegraphics[width=\textwidth]{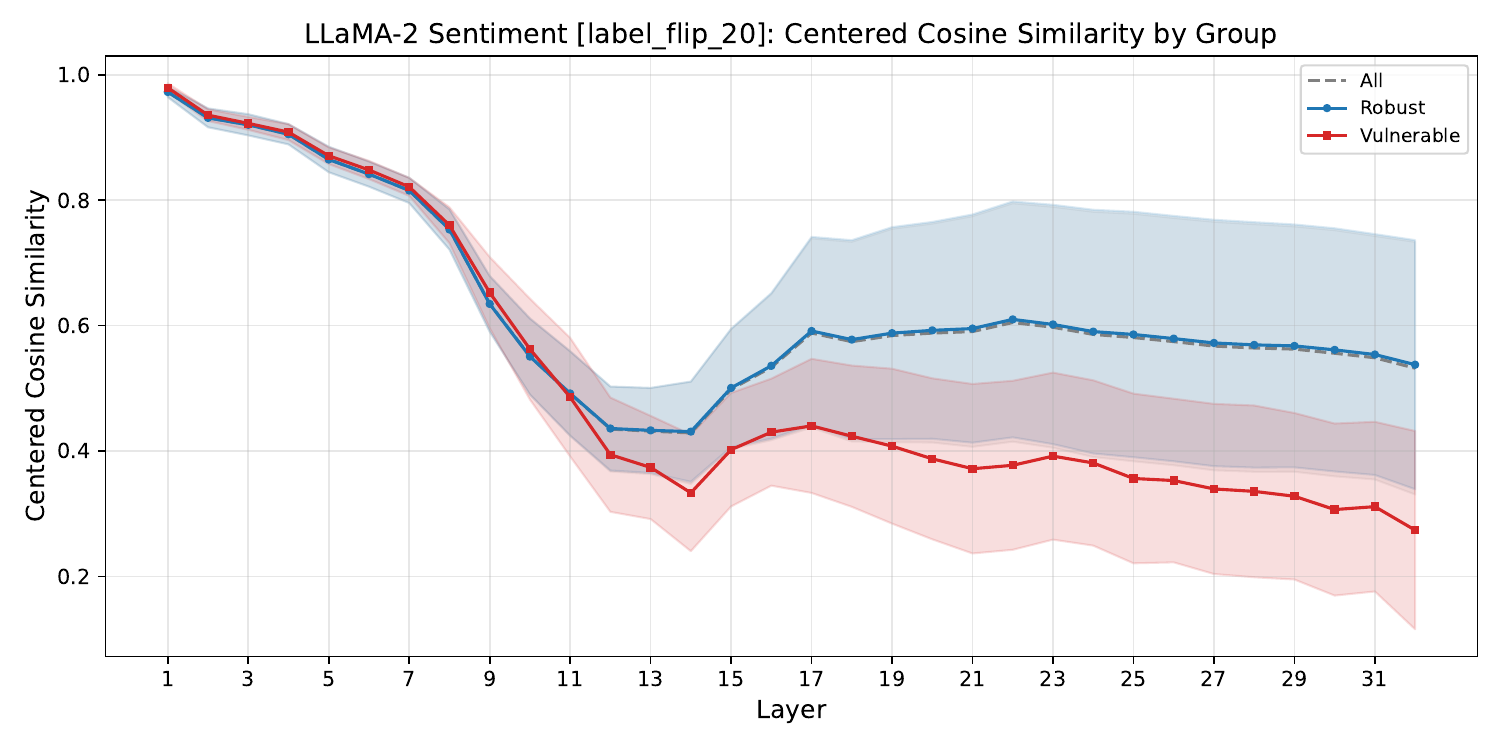}
  \end{subfigure}\hfill
  \begin{subfigure}[t]{0.32\textwidth}
    \includegraphics[width=\textwidth]{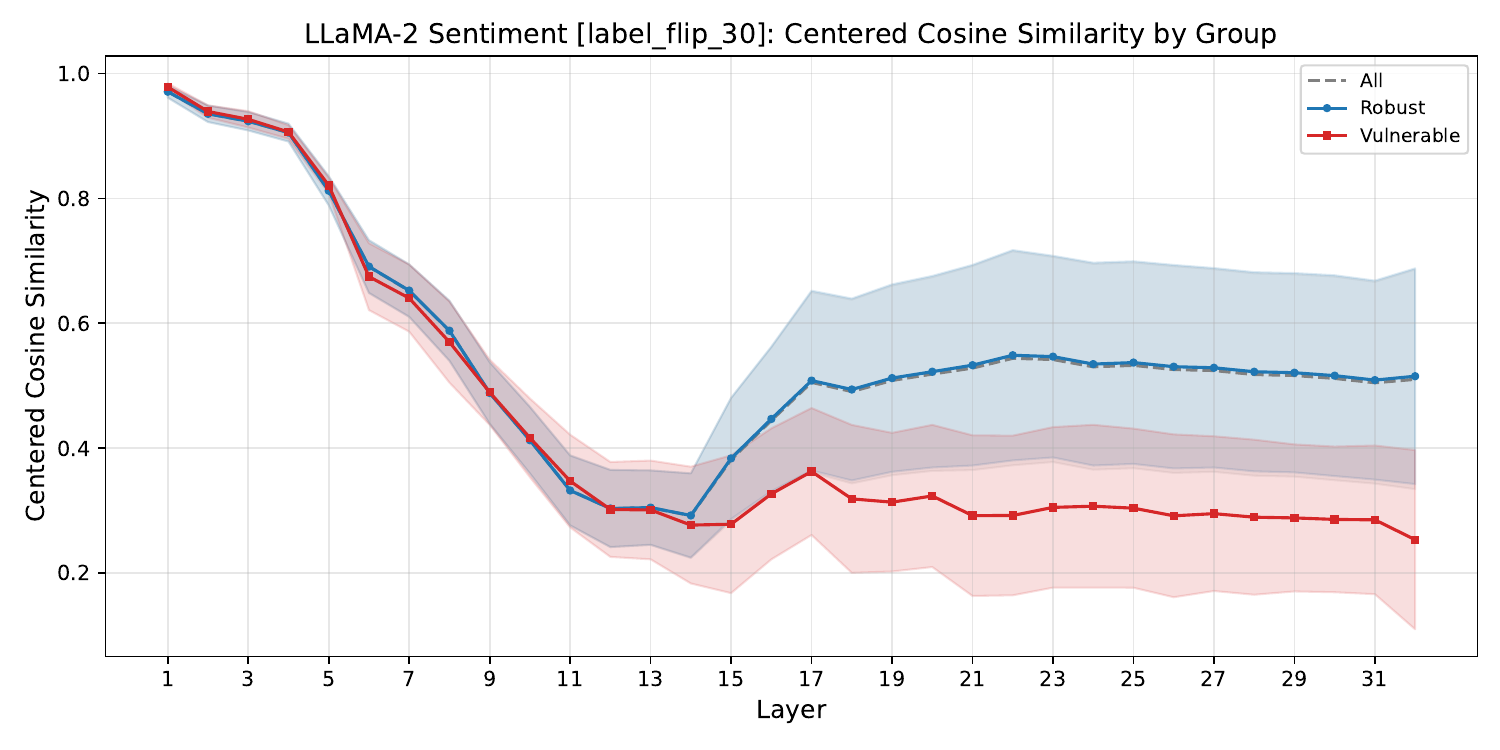}
  \end{subfigure}\hfill
  \begin{subfigure}[t]{0.32\textwidth}
    \includegraphics[width=\textwidth]{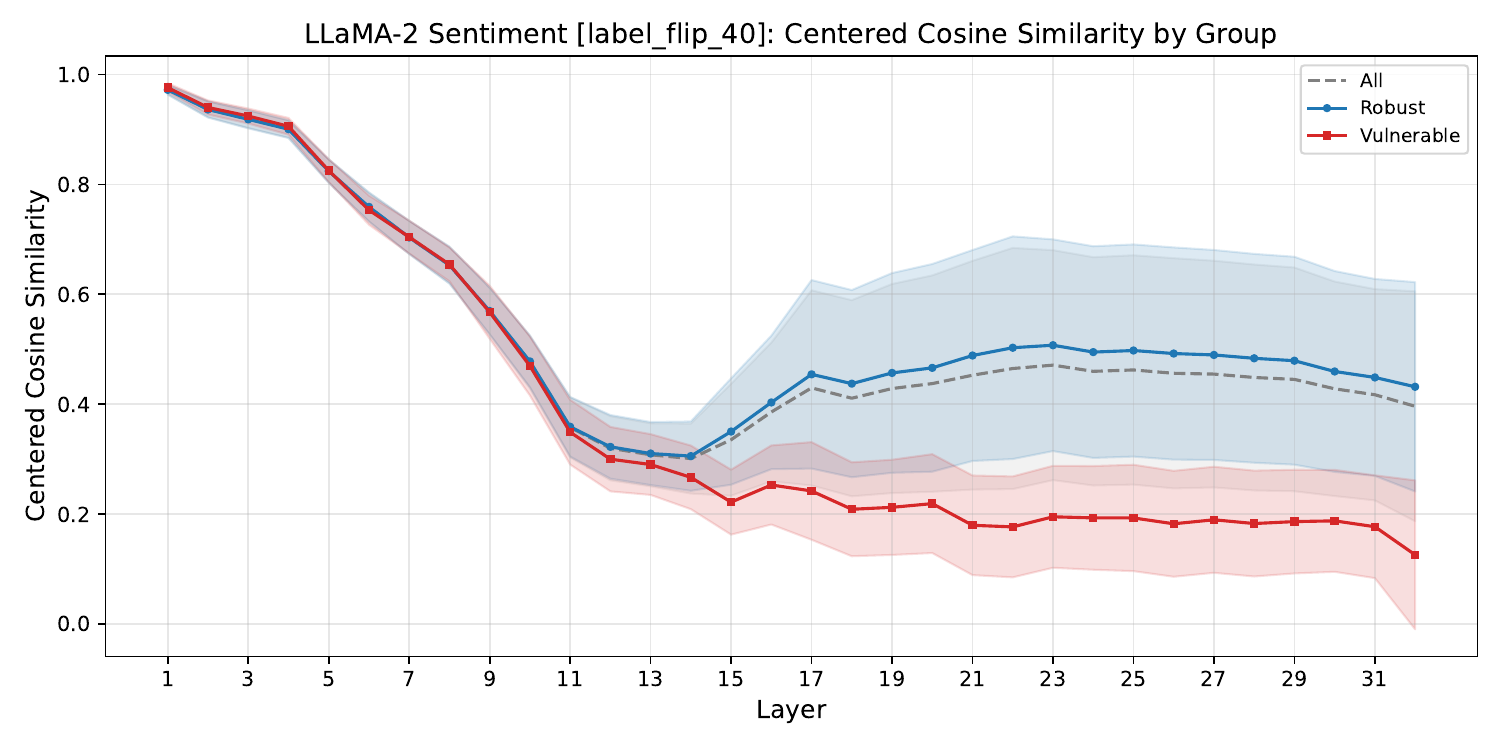}
  \end{subfigure}
    \begin{subfigure}[t]{0.32\textwidth}
    \includegraphics[width=\textwidth]{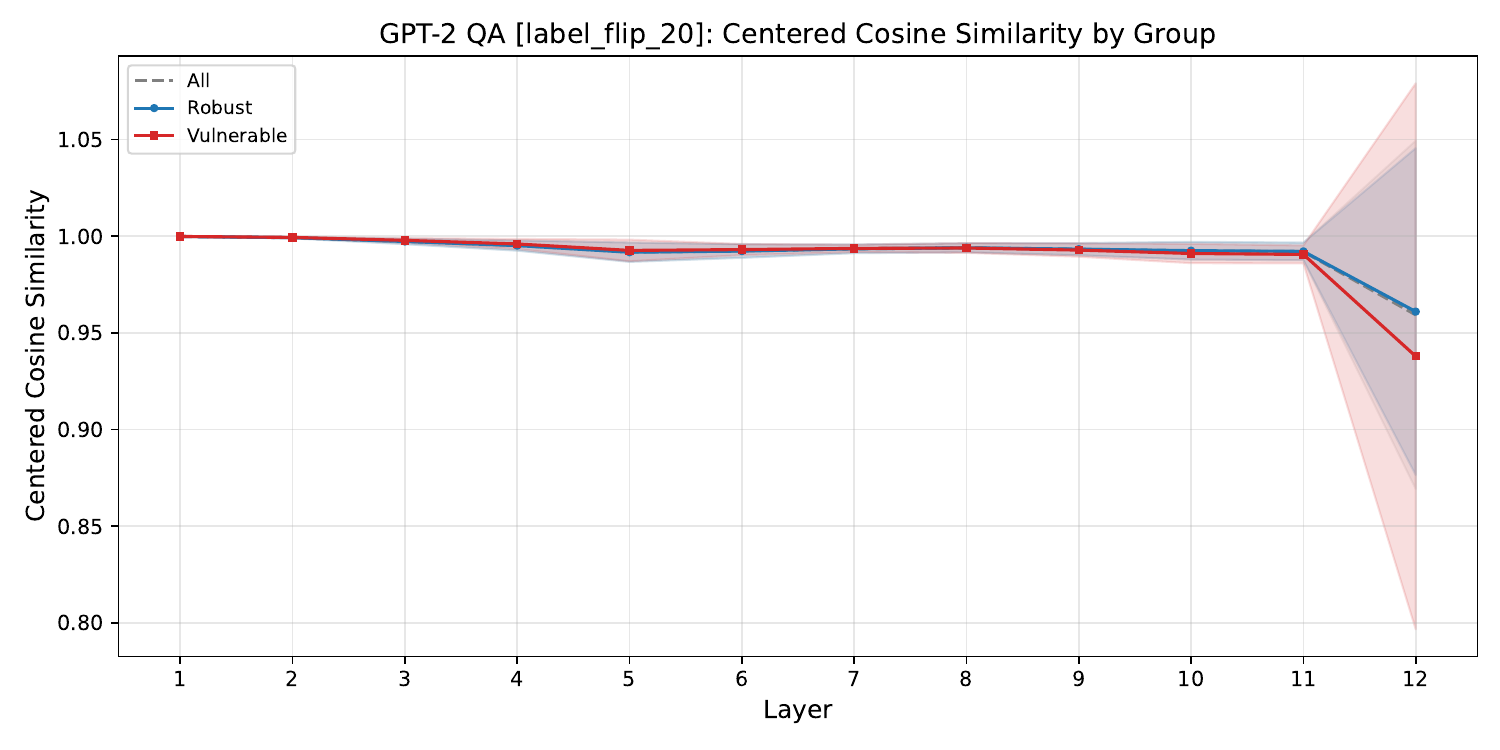}
  \end{subfigure}\hfill
  \begin{subfigure}[t]{0.32\textwidth}
    \includegraphics[width=\textwidth]{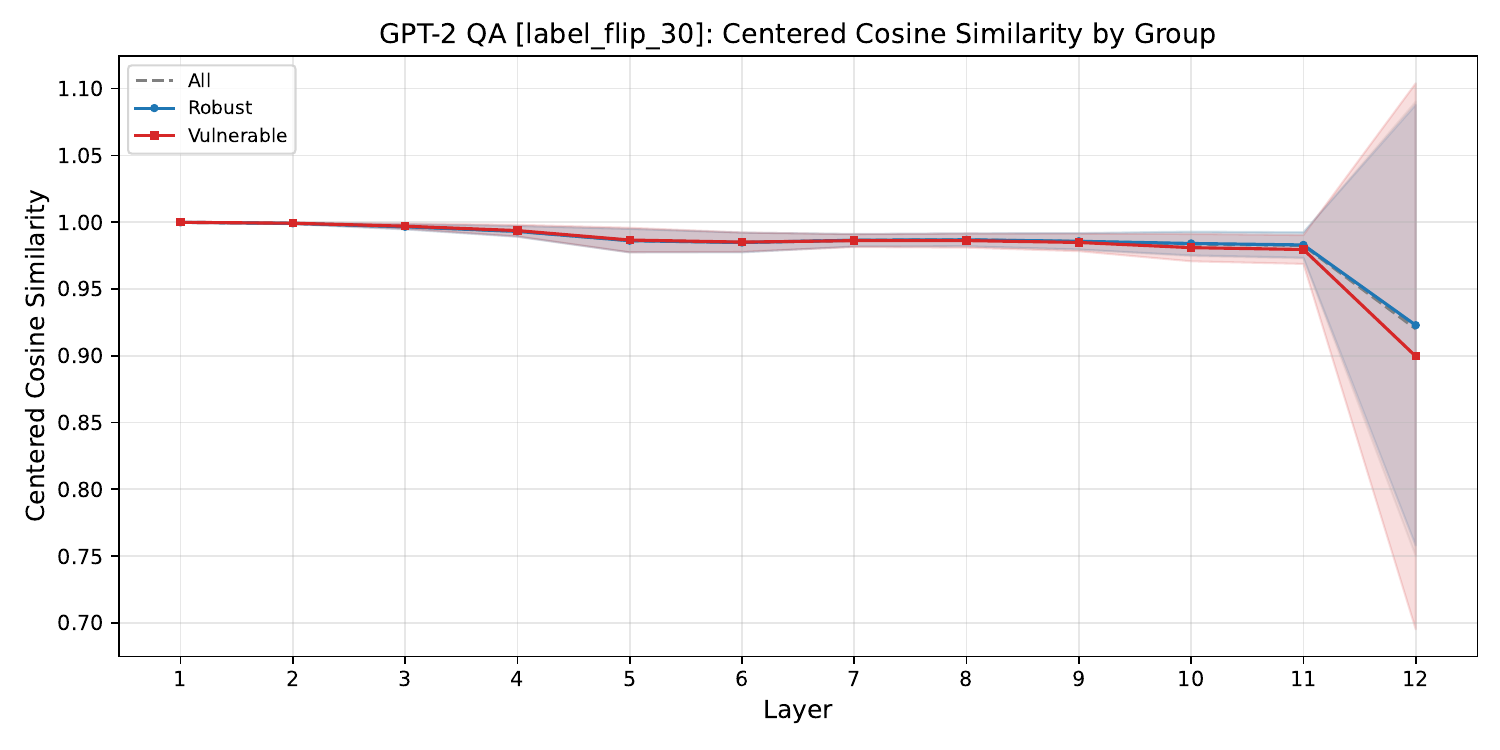}
  \end{subfigure}\hfill
  \begin{subfigure}[t]{0.32\textwidth}
    \includegraphics[width=\textwidth]{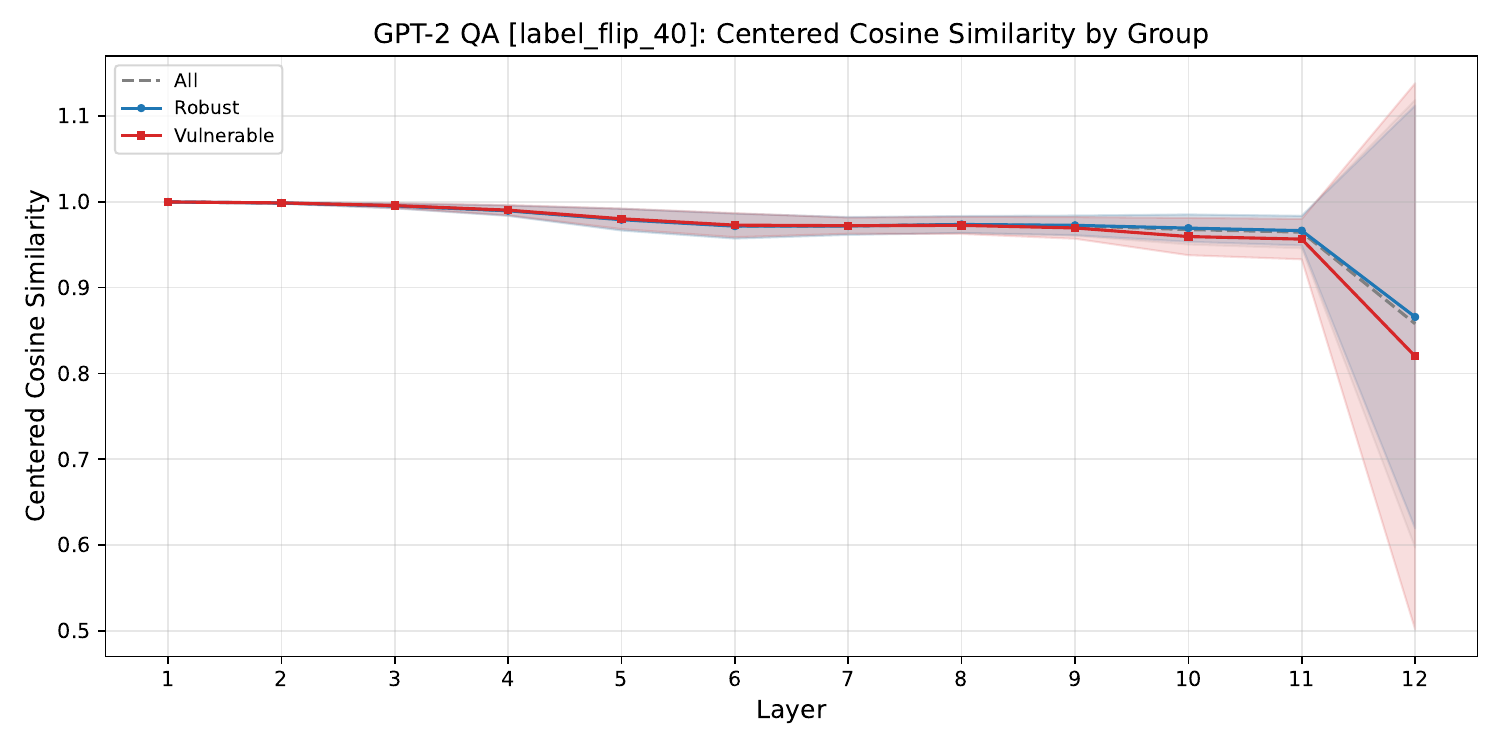}
  \end{subfigure}
  \begin{subfigure}[t]{0.32\textwidth}
    \includegraphics[width=\textwidth]{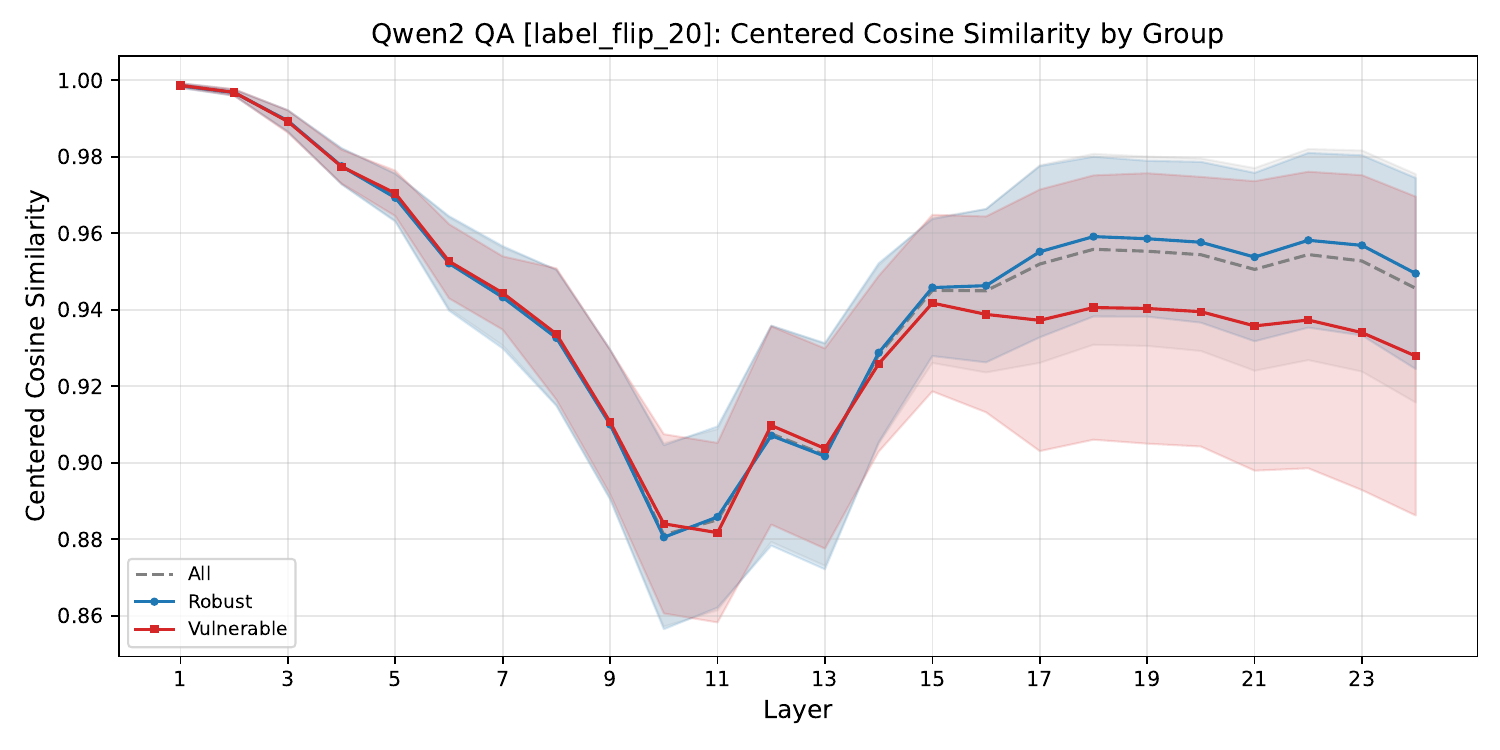}
  \end{subfigure}\hfill
  \begin{subfigure}[t]{0.32\textwidth}
    \includegraphics[width=\textwidth]{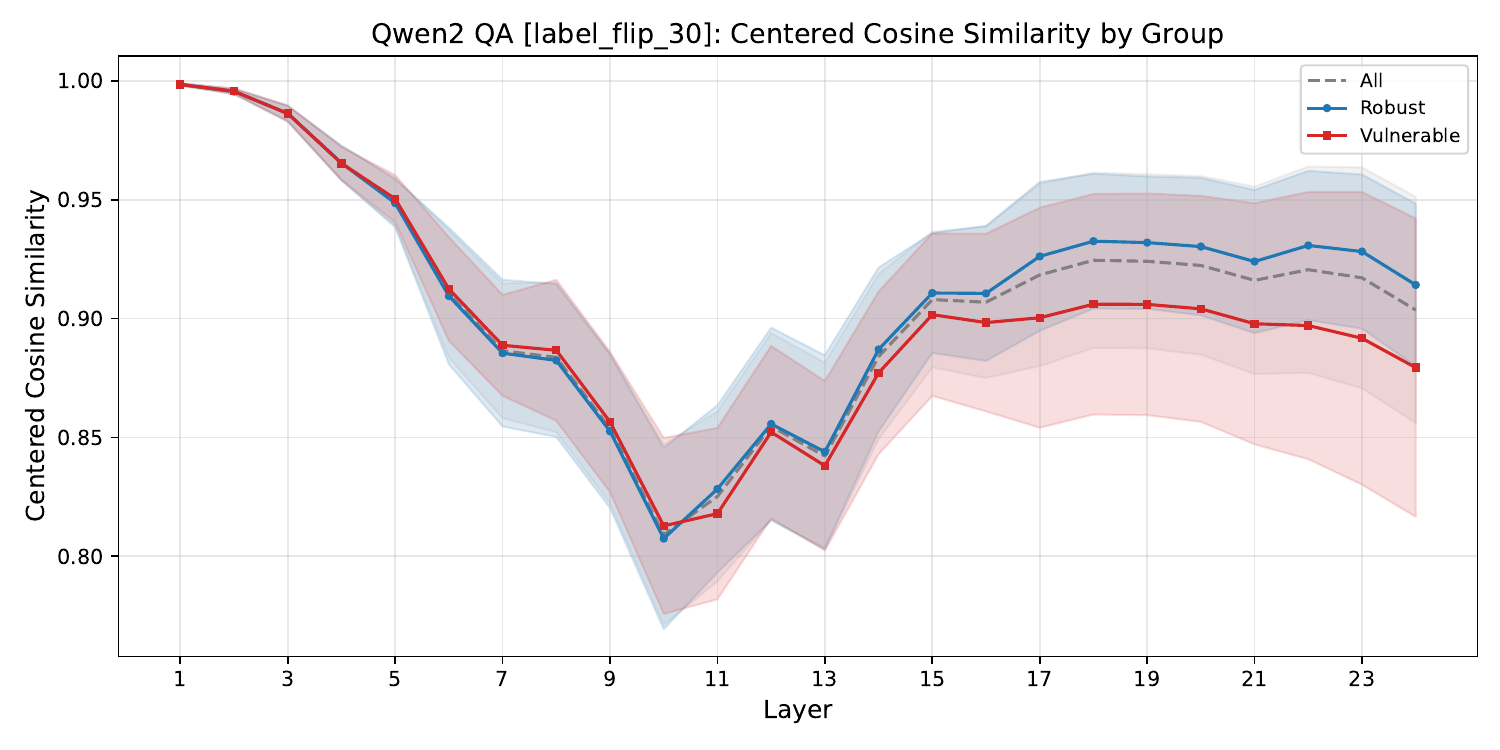}
  \end{subfigure}\hfill
  \begin{subfigure}[t]{0.32\textwidth}
    \includegraphics[width=\textwidth]{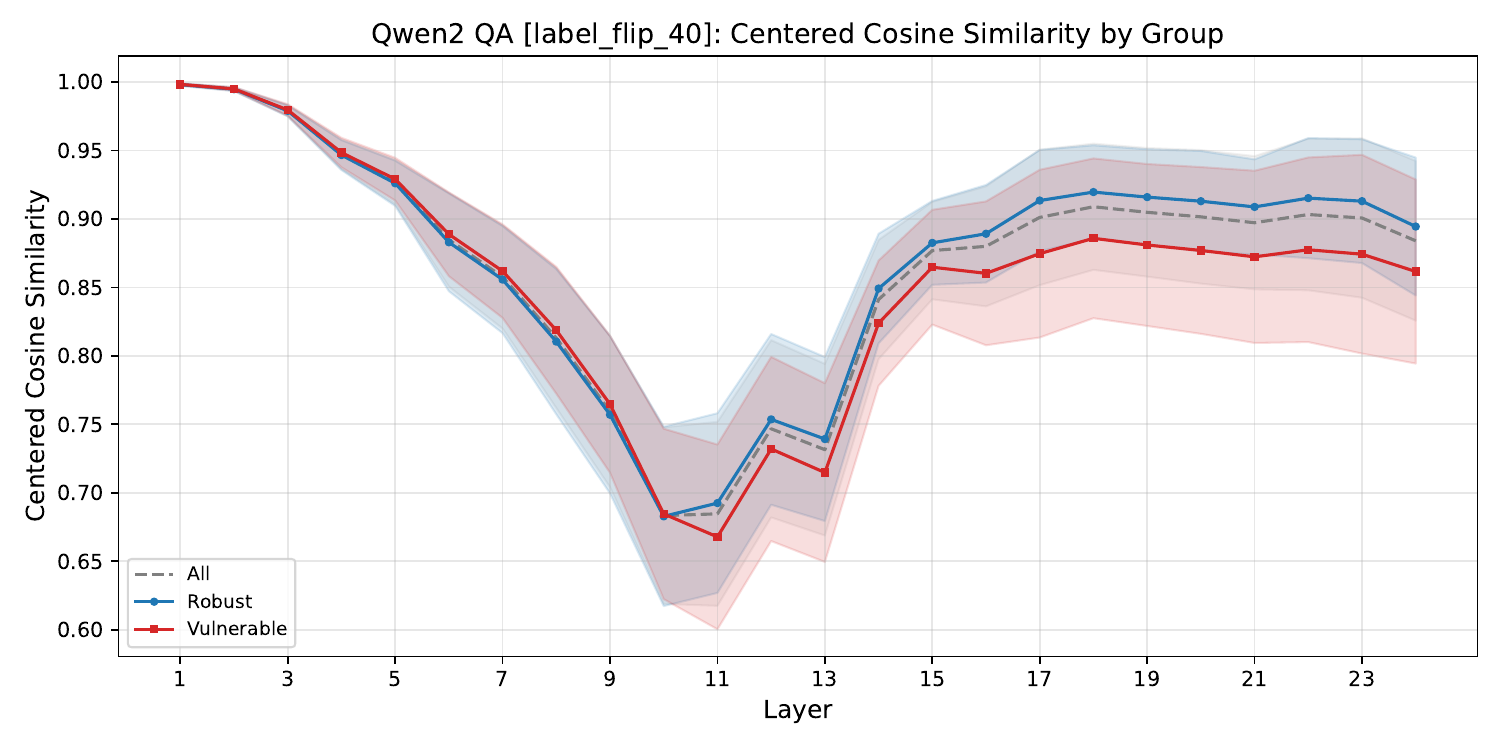}
  \end{subfigure}
  \begin{subfigure}[t]{0.32\textwidth}
    \includegraphics[width=\textwidth]{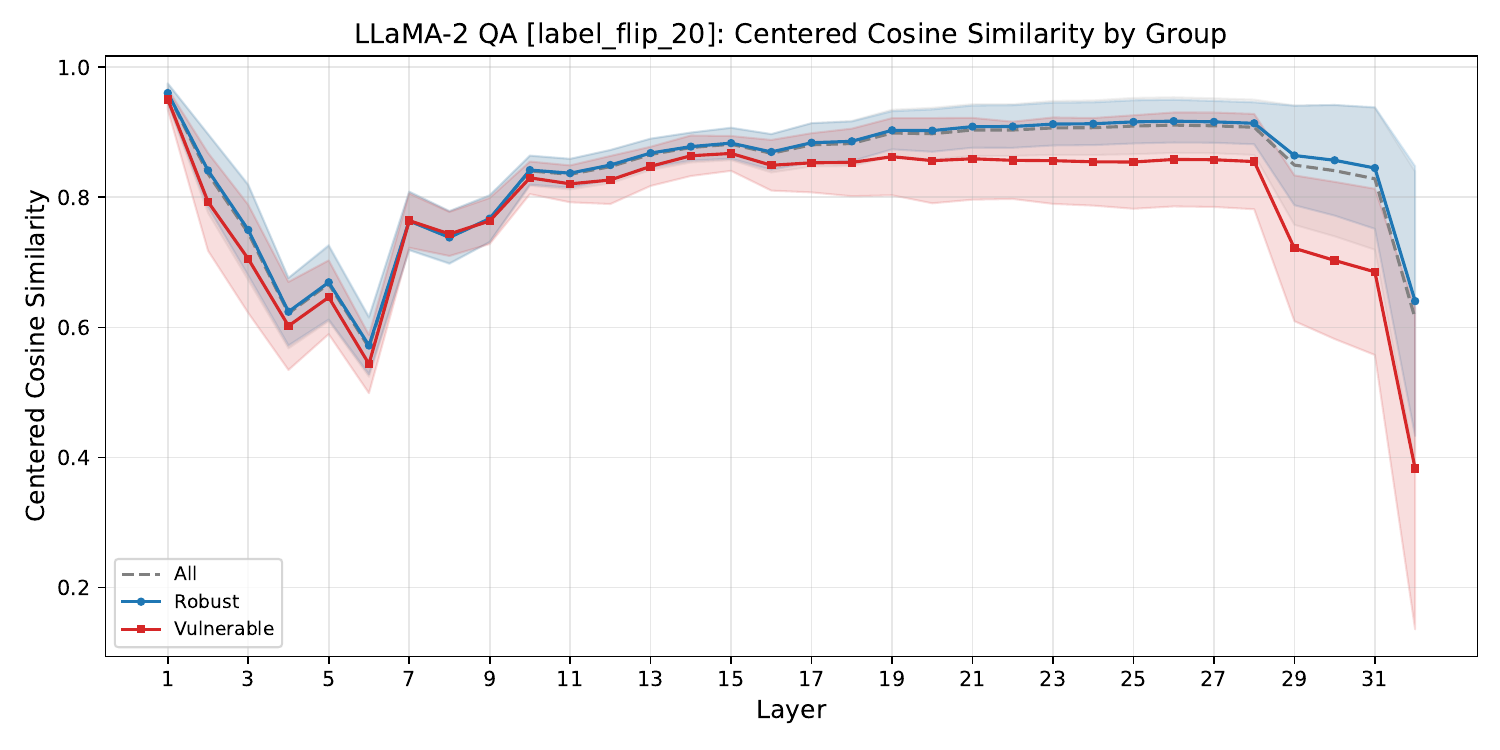}
  \end{subfigure}\hfill
  \begin{subfigure}[t]{0.32\textwidth}
    \includegraphics[width=\textwidth]{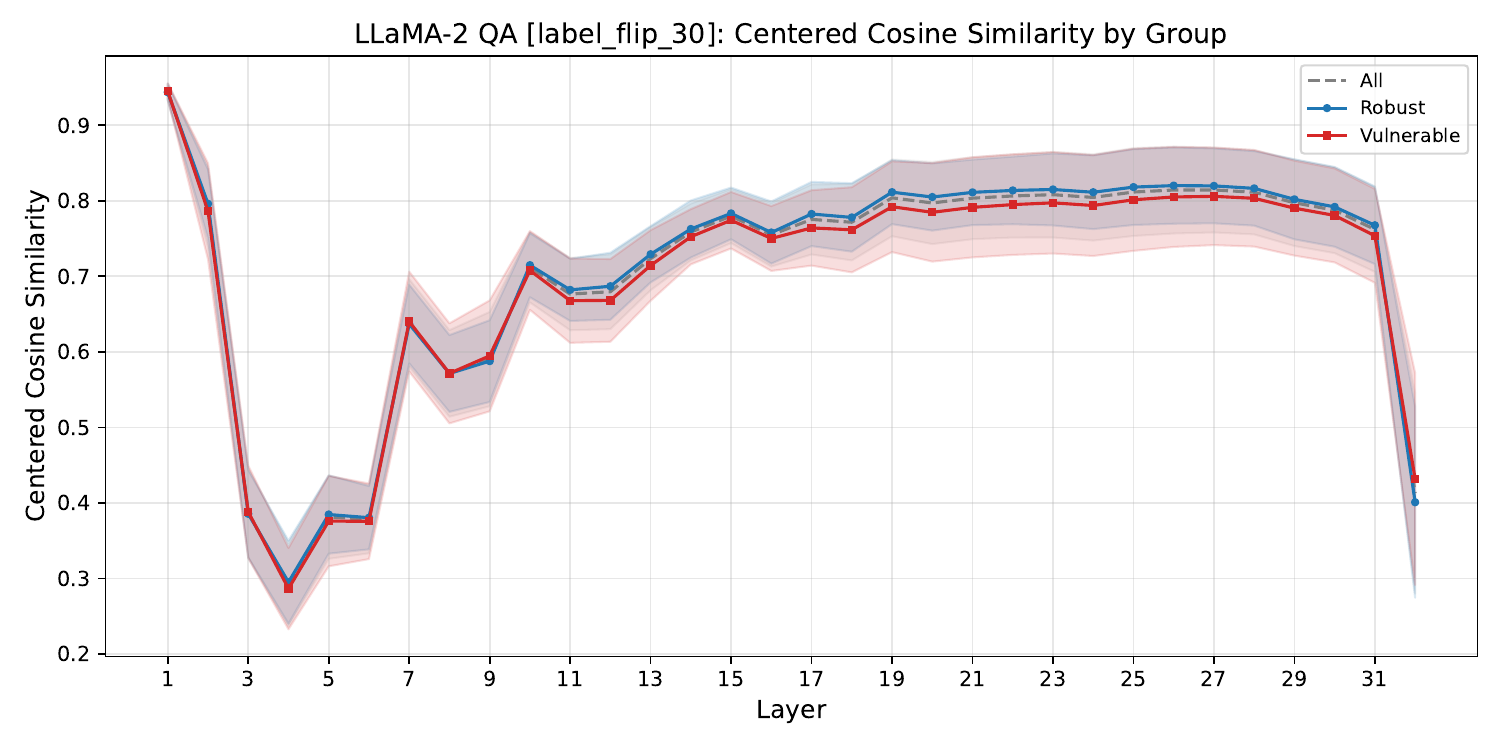}
  \end{subfigure}\hfill
  \begin{subfigure}[t]{0.32\textwidth}
    \includegraphics[width=\textwidth]{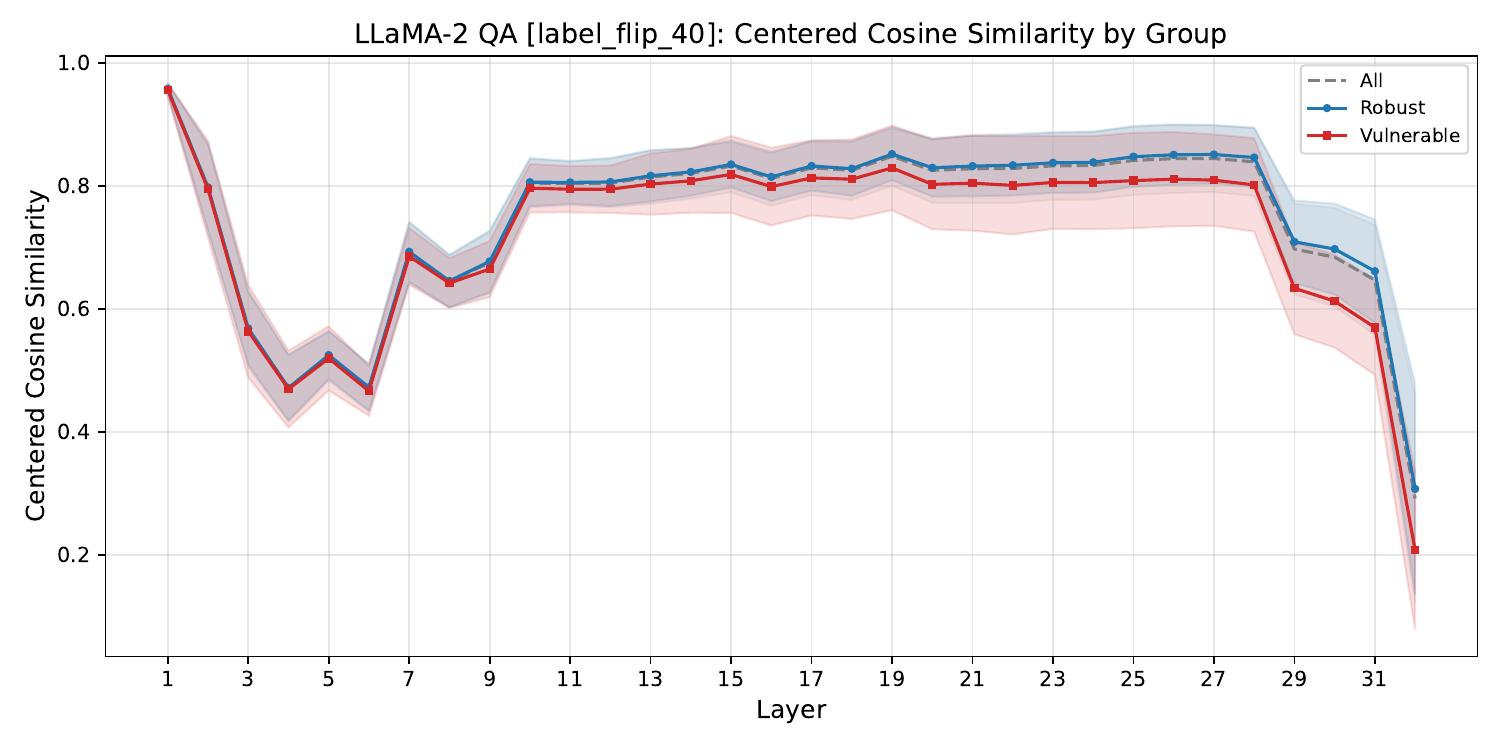}
  \end{subfigure}
    \begin{subfigure}[t]{0.32\textwidth}
    \includegraphics[width=\textwidth]{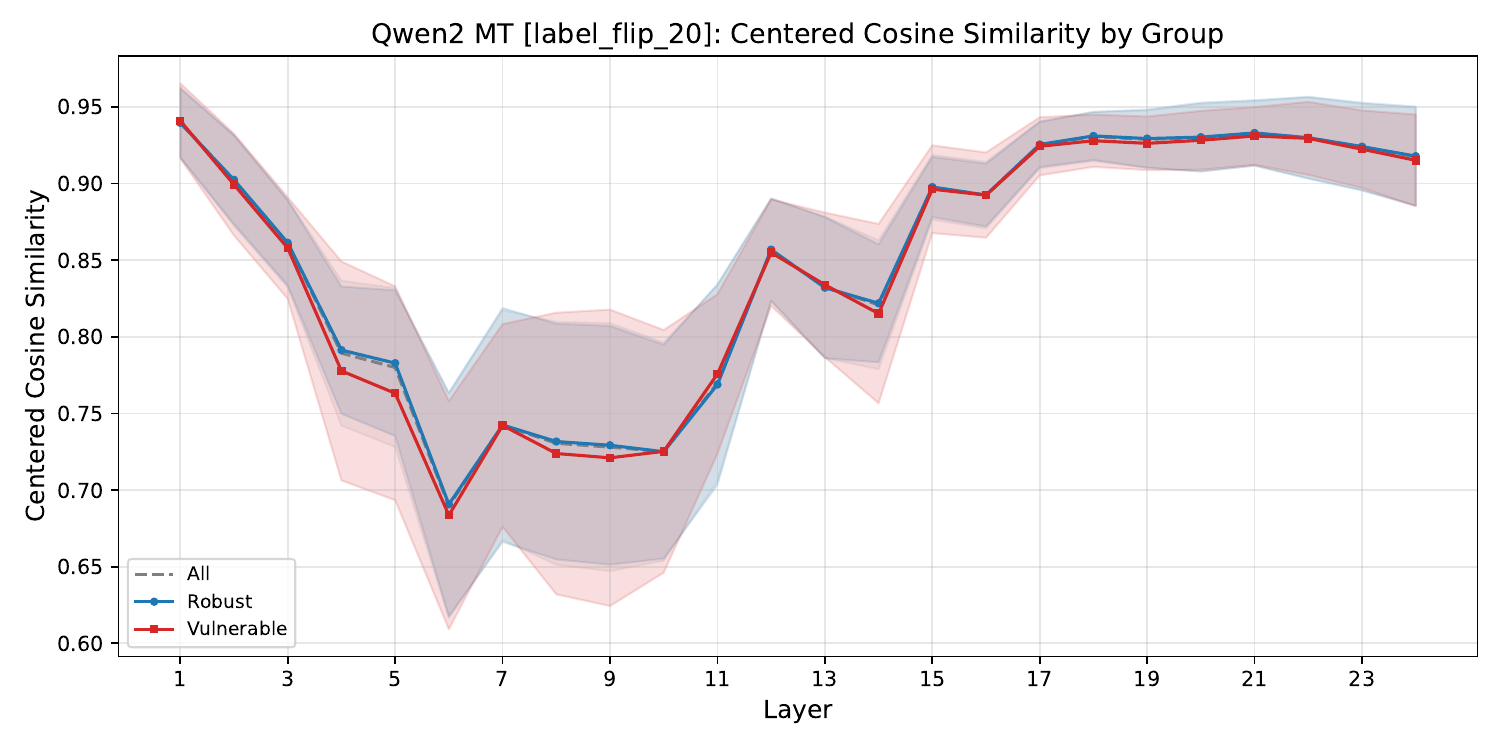}
  \end{subfigure}\hfill
  \begin{subfigure}[t]{0.32\textwidth}
    \includegraphics[width=\textwidth]{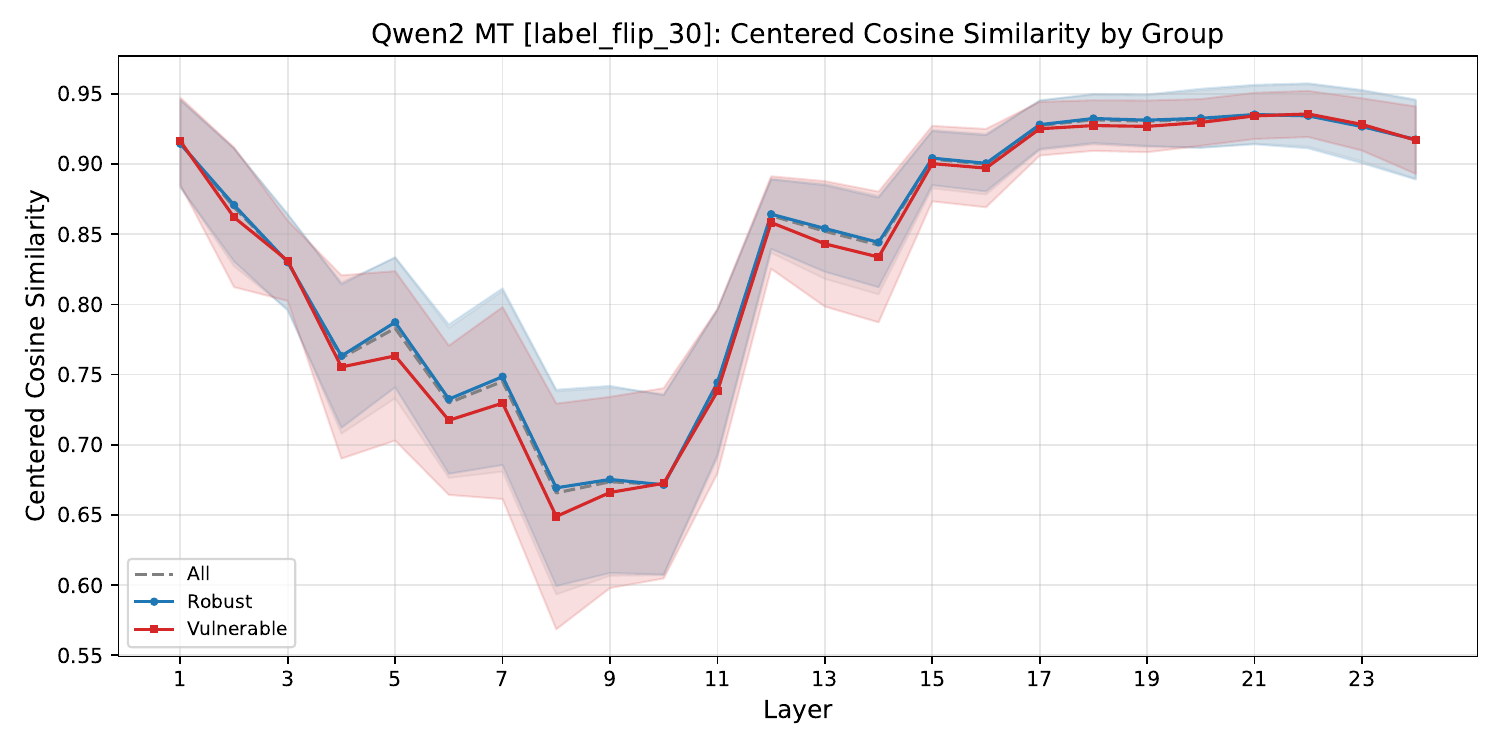}
  \end{subfigure}\hfill
  \begin{subfigure}[t]{0.32\textwidth}
    \includegraphics[width=\textwidth]{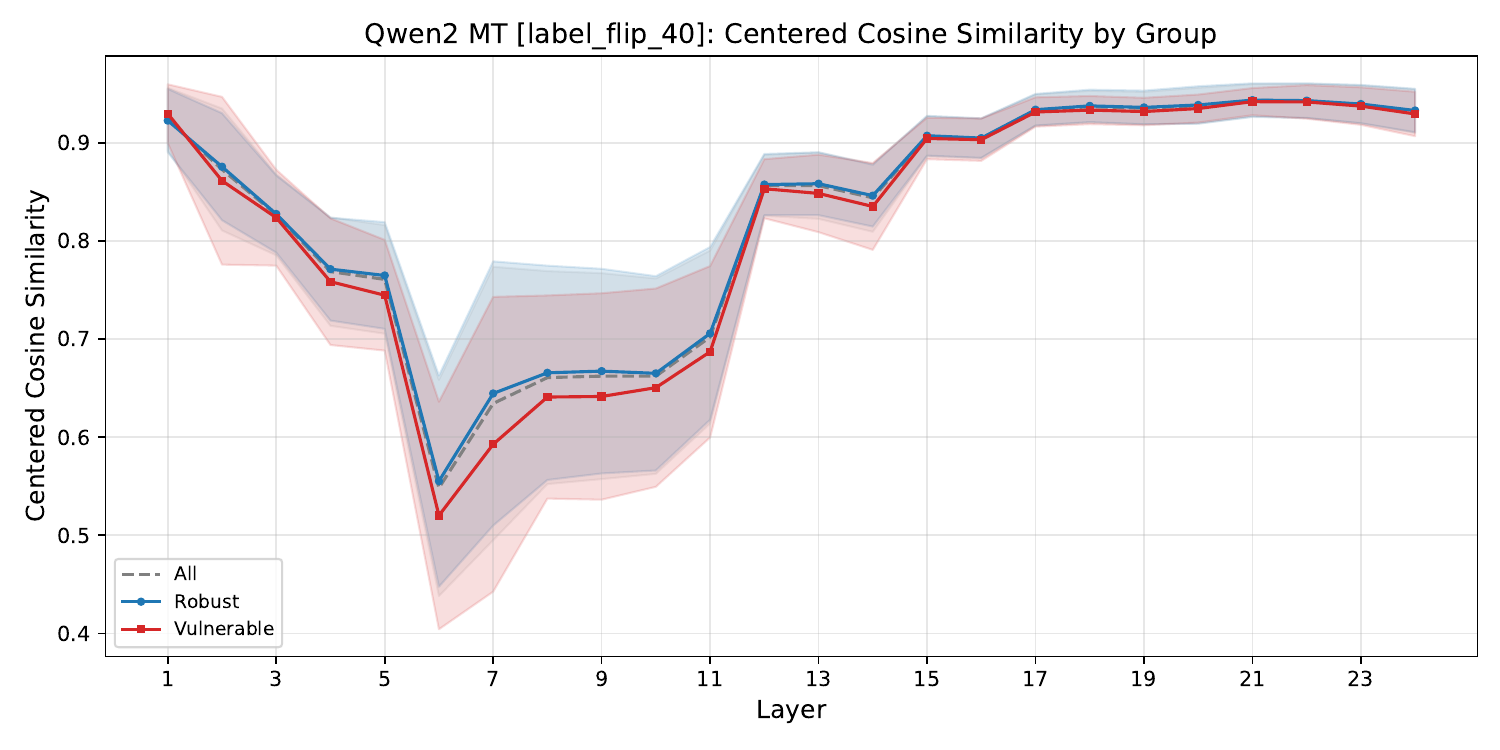}
  \end{subfigure}
  \begin{subfigure}[t]{0.32\textwidth}
    \includegraphics[width=\textwidth]{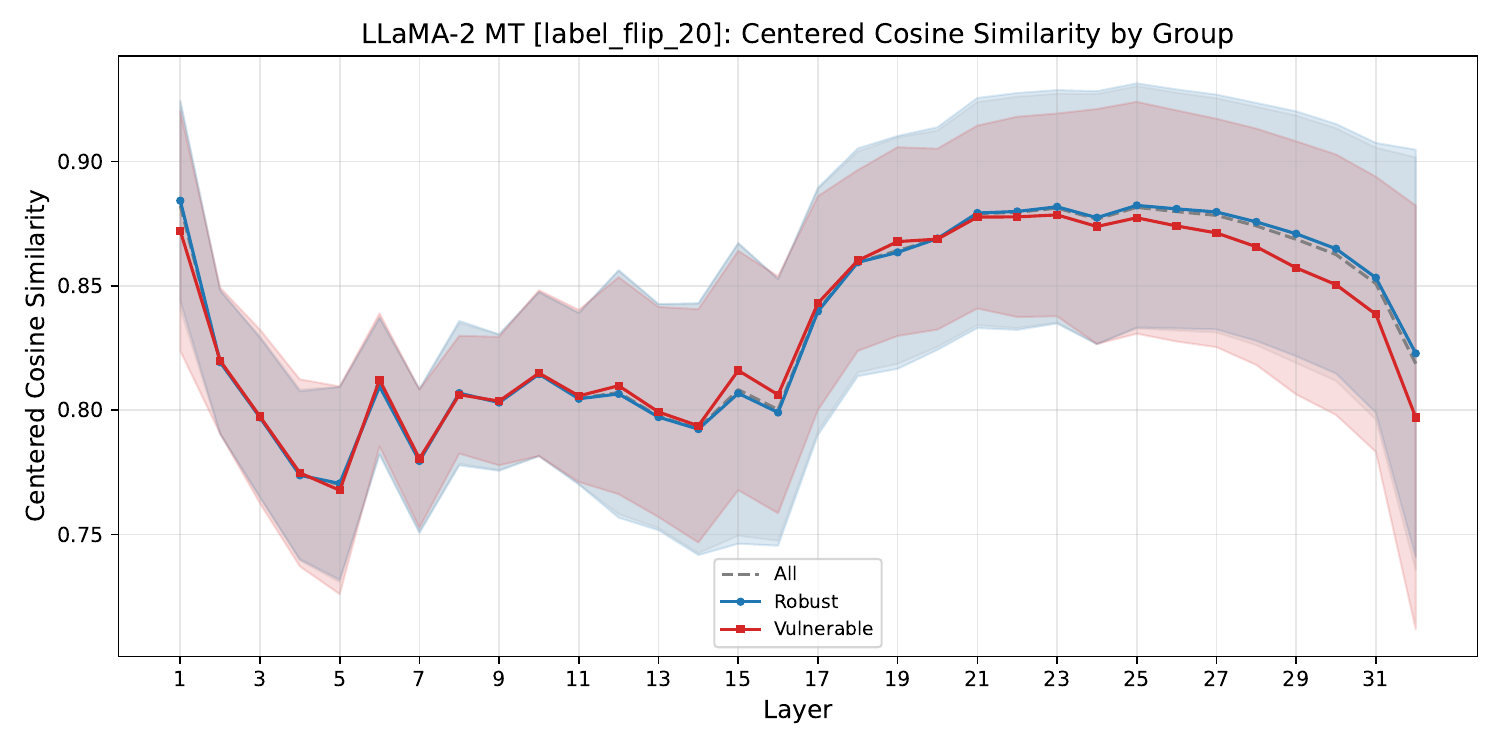}
  \end{subfigure}\hfill
  \begin{subfigure}[t]{0.32\textwidth}
    \includegraphics[width=\textwidth]{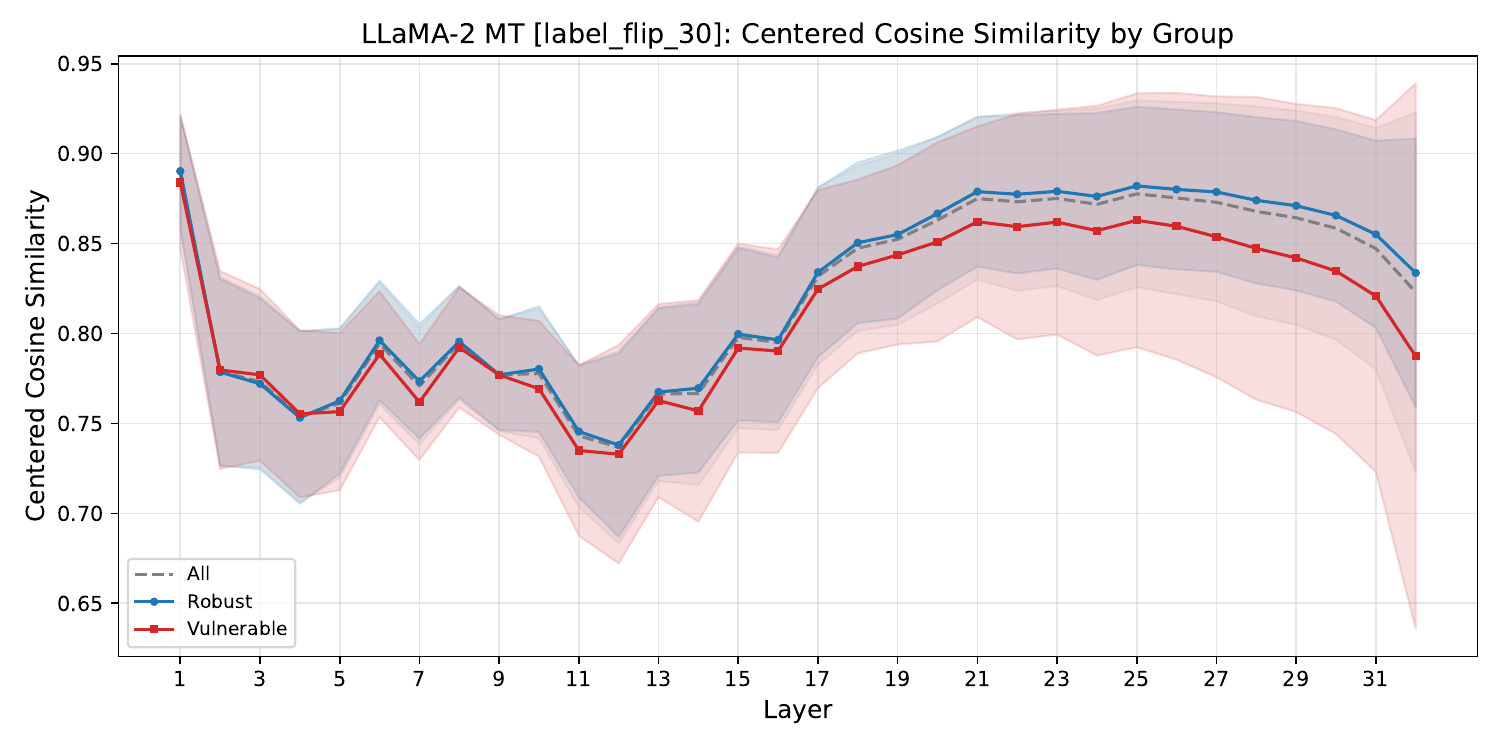}
  \end{subfigure}\hfill
  \begin{subfigure}[t]{0.32\textwidth}
    \includegraphics[width=\textwidth]{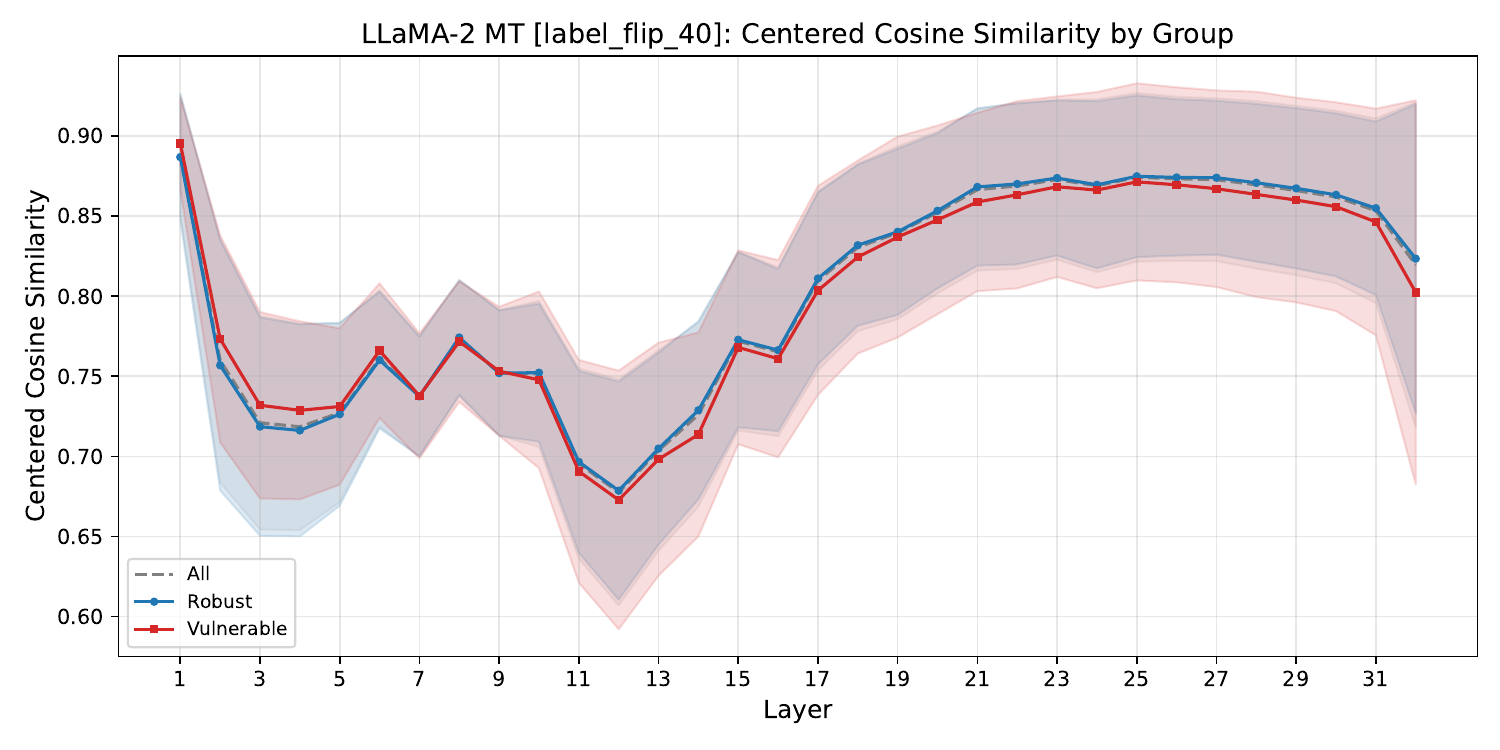}
  \end{subfigure}
  \caption{Robust vs.\ vulnerable stratification: \textbf{centered cosine similarity} for \textbf{\ac{SC}}, \textbf{\ac{QA}} and \textbf{\ac{MT}} under label-flip noise. Centered cosine removes the shared mean direction before computing similarity, correcting for anisotropy.}
  \label{fig:stratification_sent_centered_cosine}
\end{figure*}

\begin{figure*}[p]
  \centering
  \begin{subfigure}[t]{0.32\textwidth}
    \includegraphics[width=\textwidth]{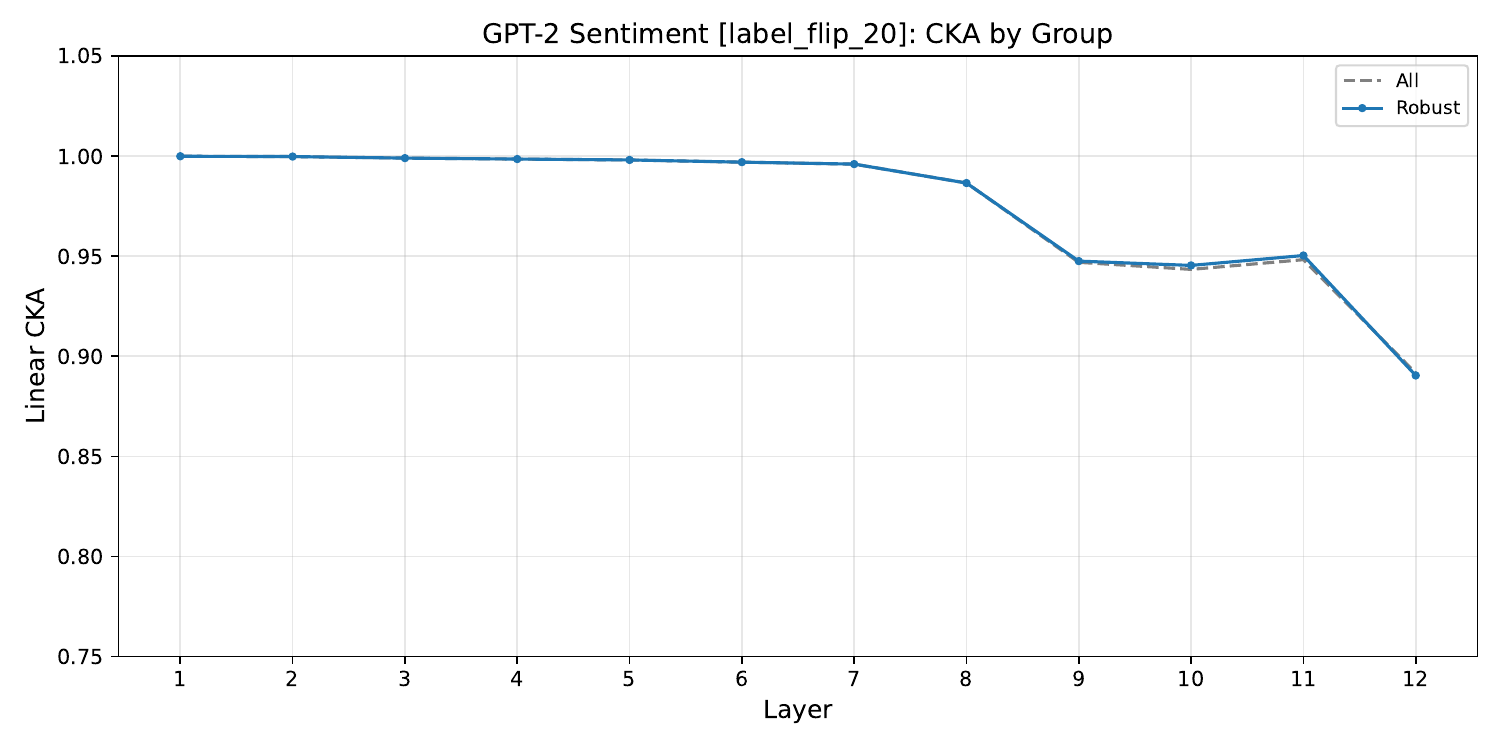}
  \end{subfigure}\hfill
  \begin{subfigure}[t]{0.32\textwidth}
    \includegraphics[width=\textwidth]{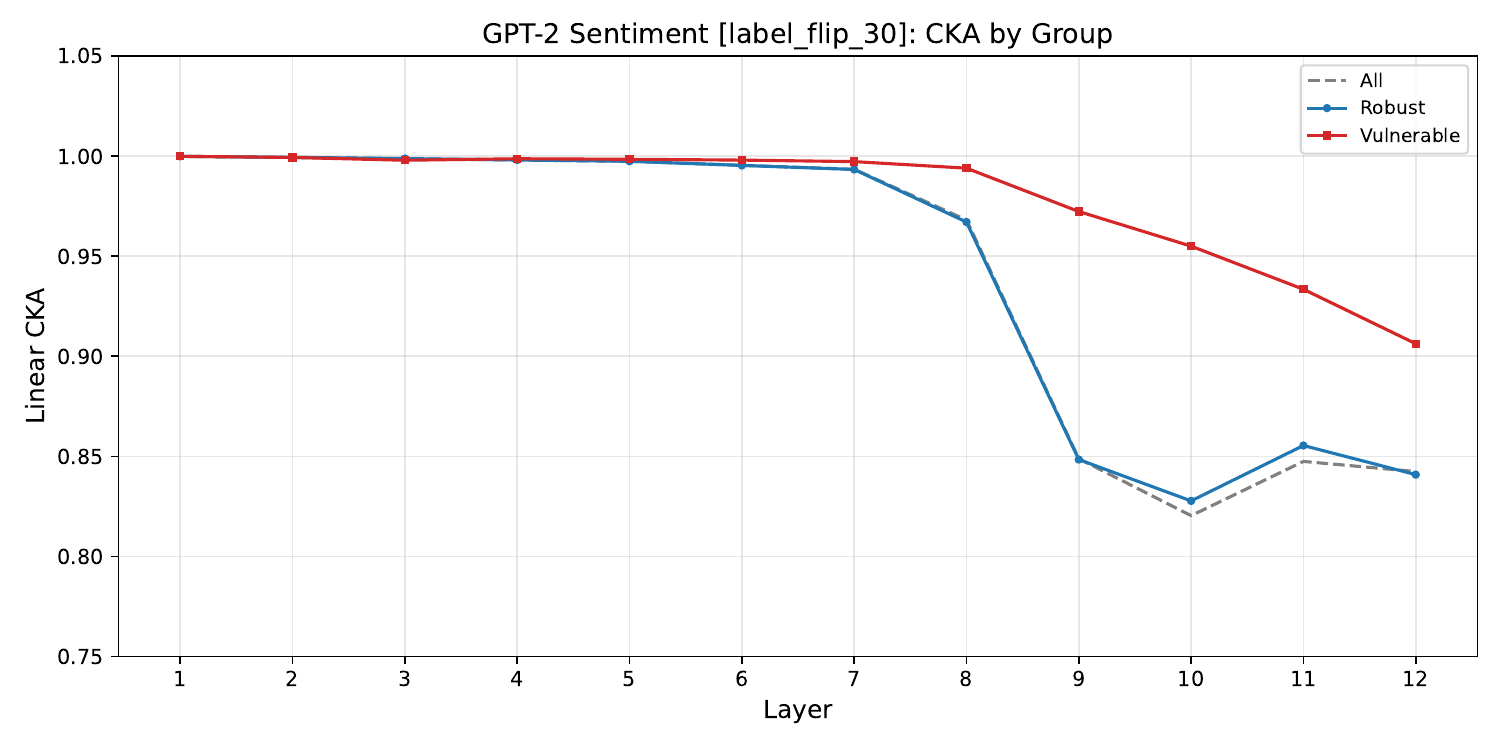}
  \end{subfigure}\hfill
  \begin{subfigure}[t]{0.32\textwidth}
    \includegraphics[width=\textwidth]{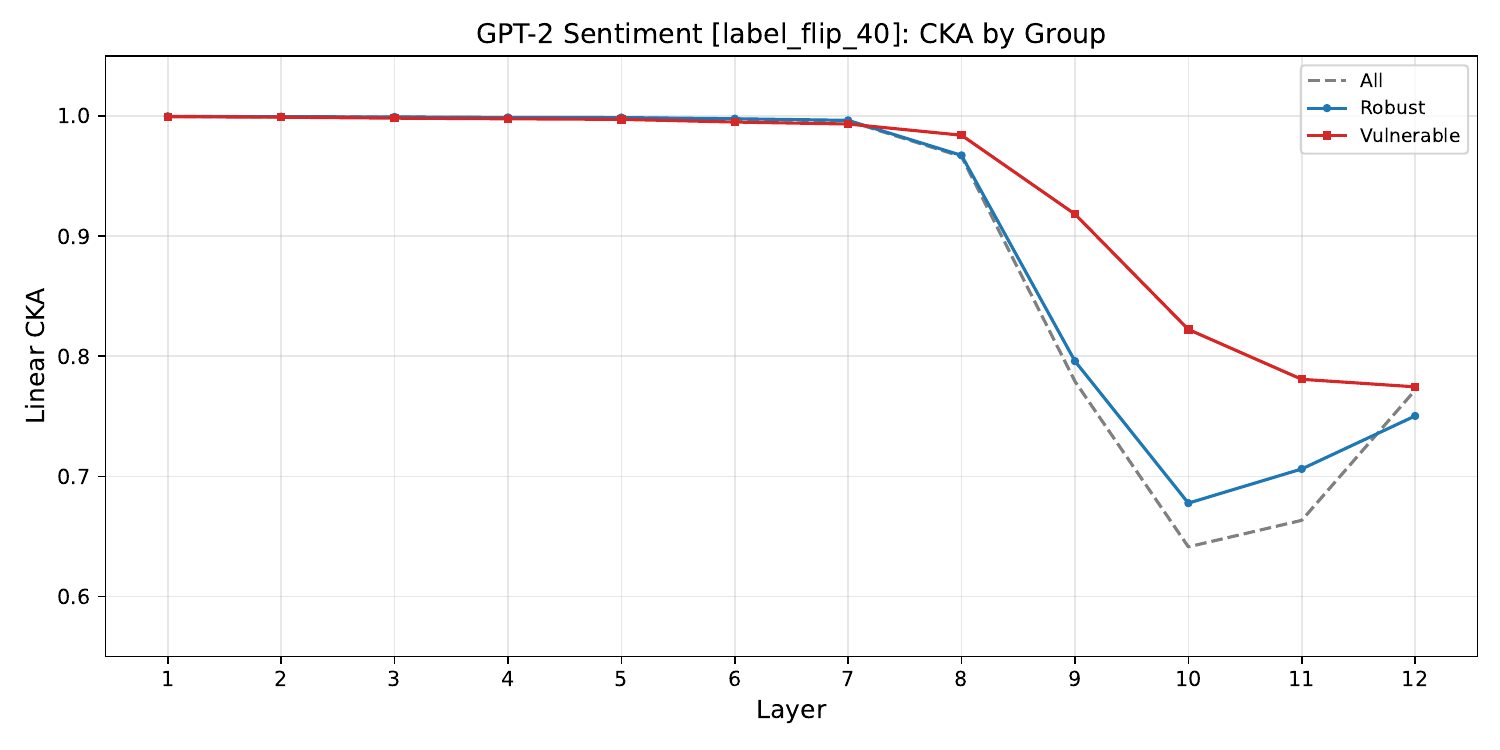}
  \end{subfigure}
  \begin{subfigure}[t]{0.32\textwidth}
    \includegraphics[width=\textwidth]{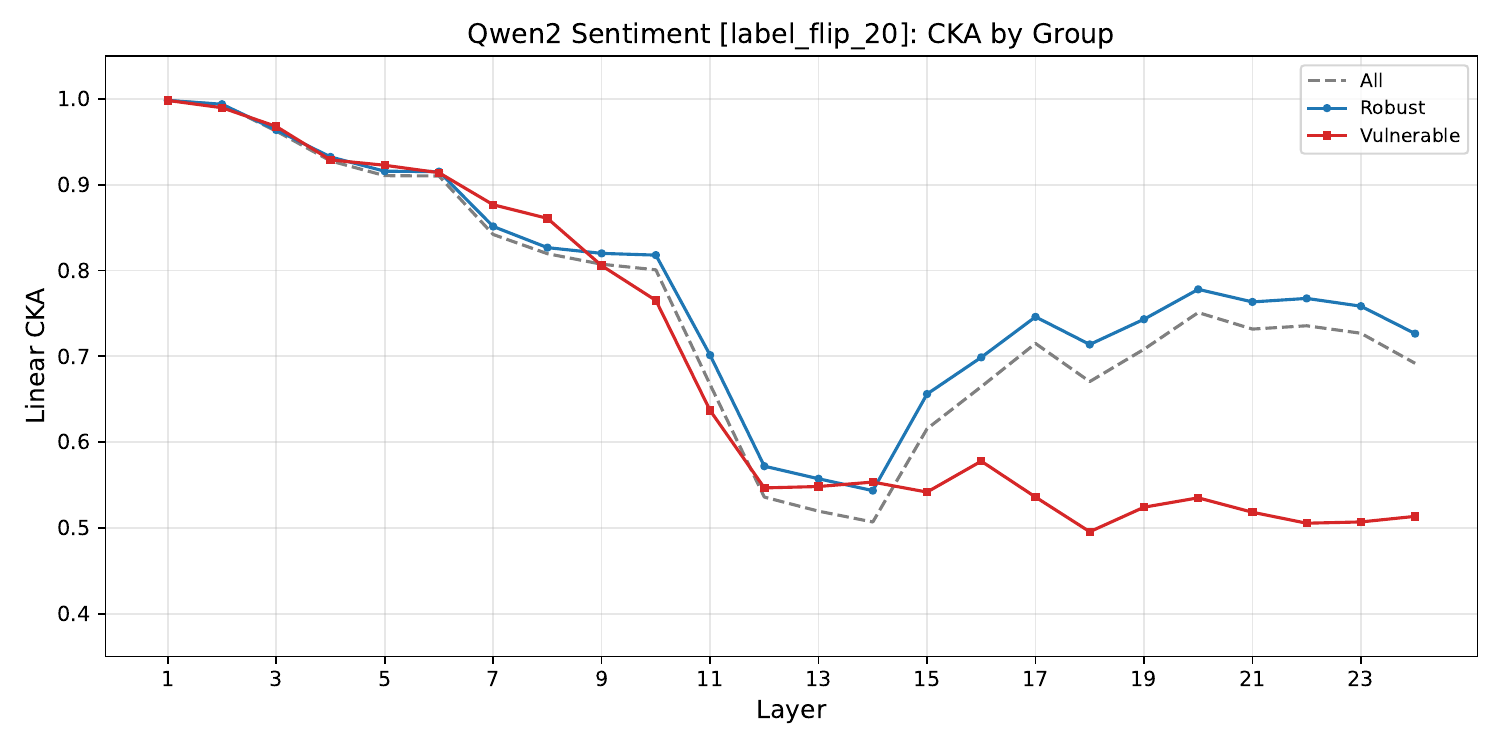}
  \end{subfigure}\hfill
  \begin{subfigure}[t]{0.32\textwidth}
    \includegraphics[width=\textwidth]{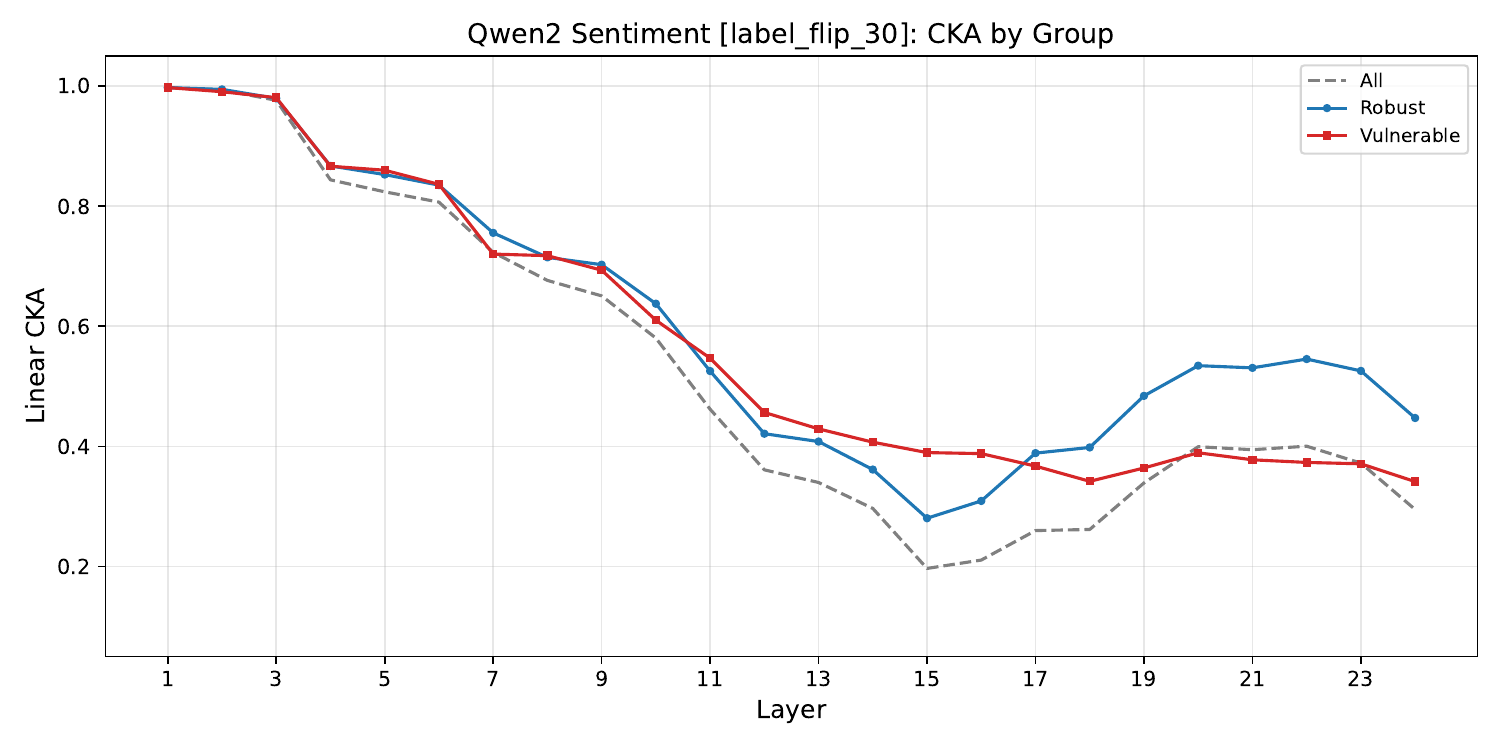}
  \end{subfigure}\hfill
  \begin{subfigure}[t]{0.32\textwidth}
    \includegraphics[width=\textwidth]{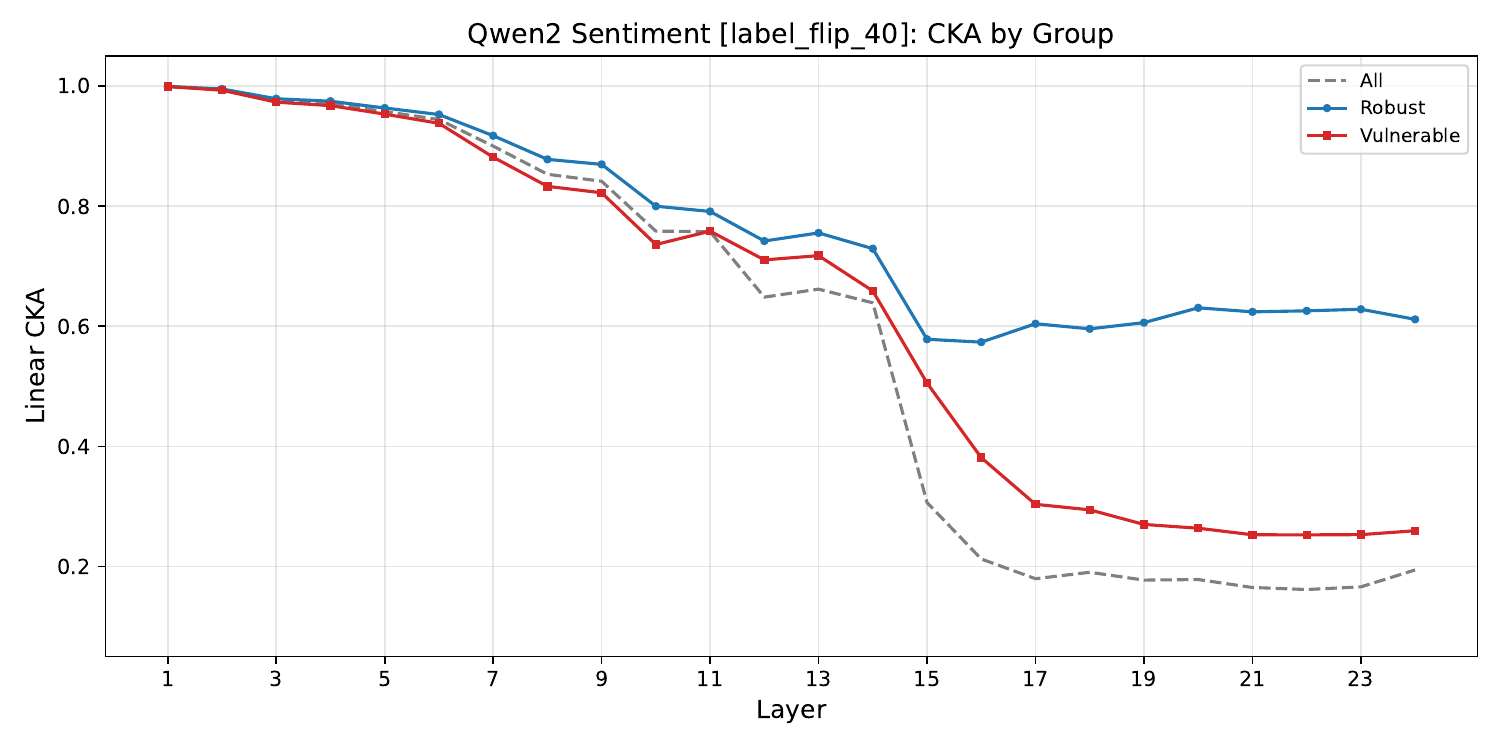}
  \end{subfigure}
  \begin{subfigure}[t]{0.32\textwidth}
    \includegraphics[width=\textwidth]{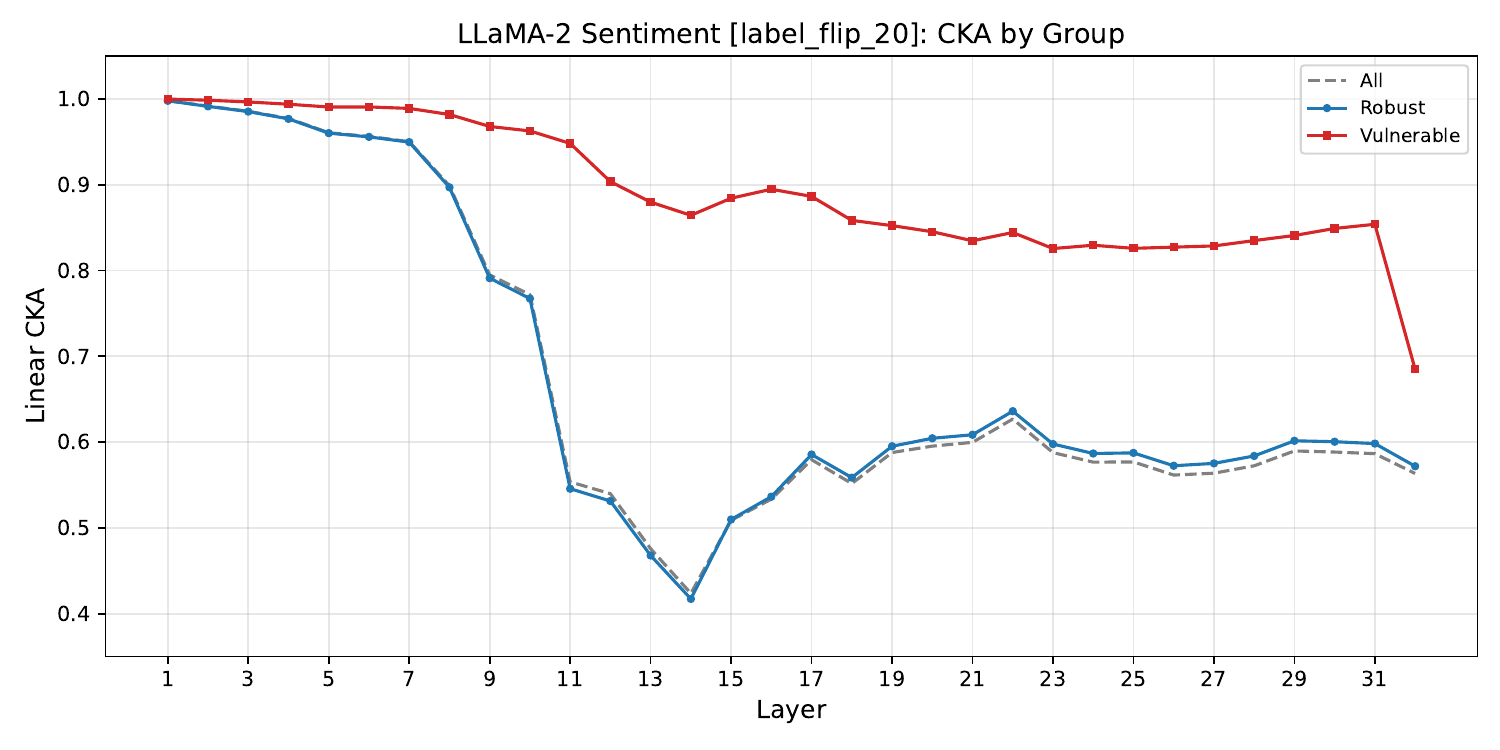}
  \end{subfigure}\hfill
  \begin{subfigure}[t]{0.32\textwidth}
    \includegraphics[width=\textwidth]{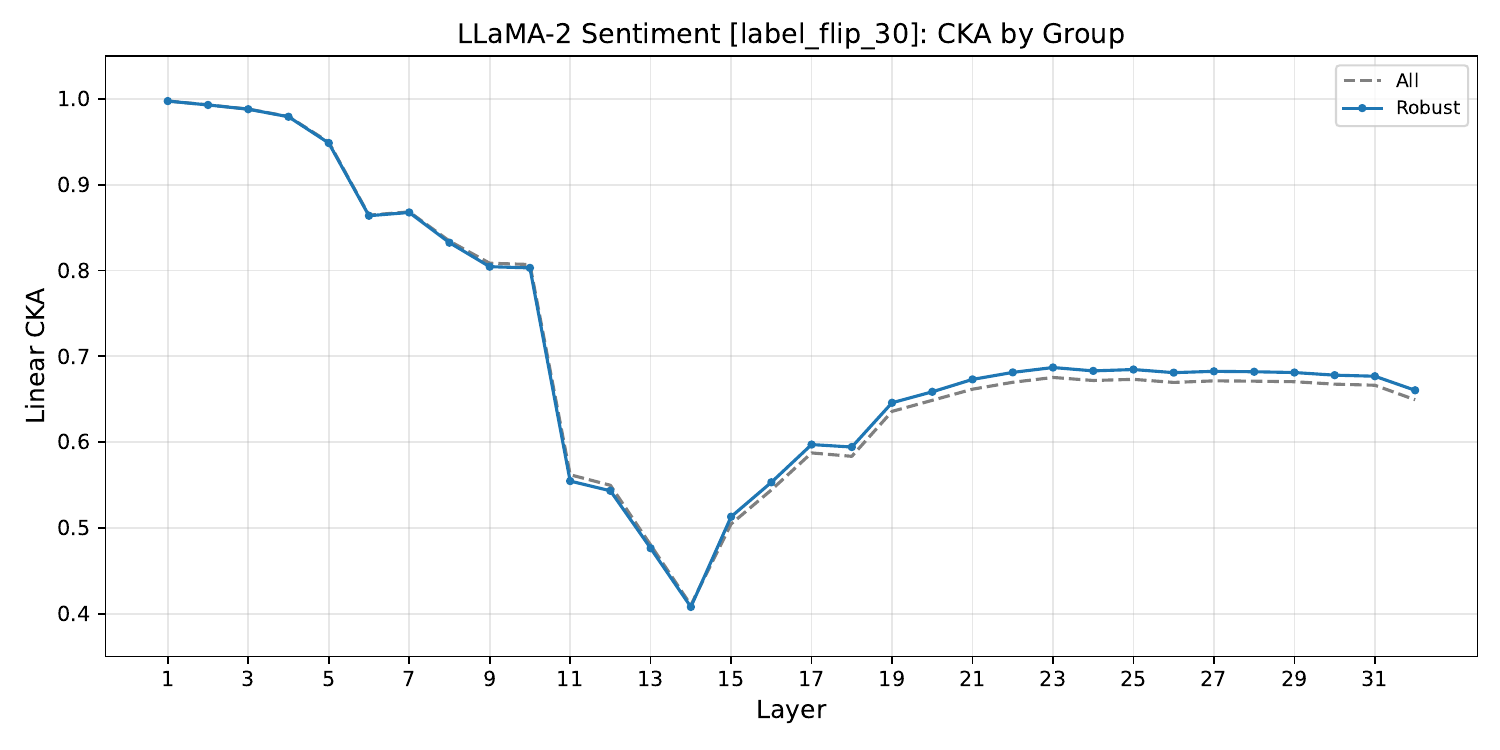}
  \end{subfigure}\hfill
  \begin{subfigure}[t]{0.32\textwidth}
    \includegraphics[width=\textwidth]{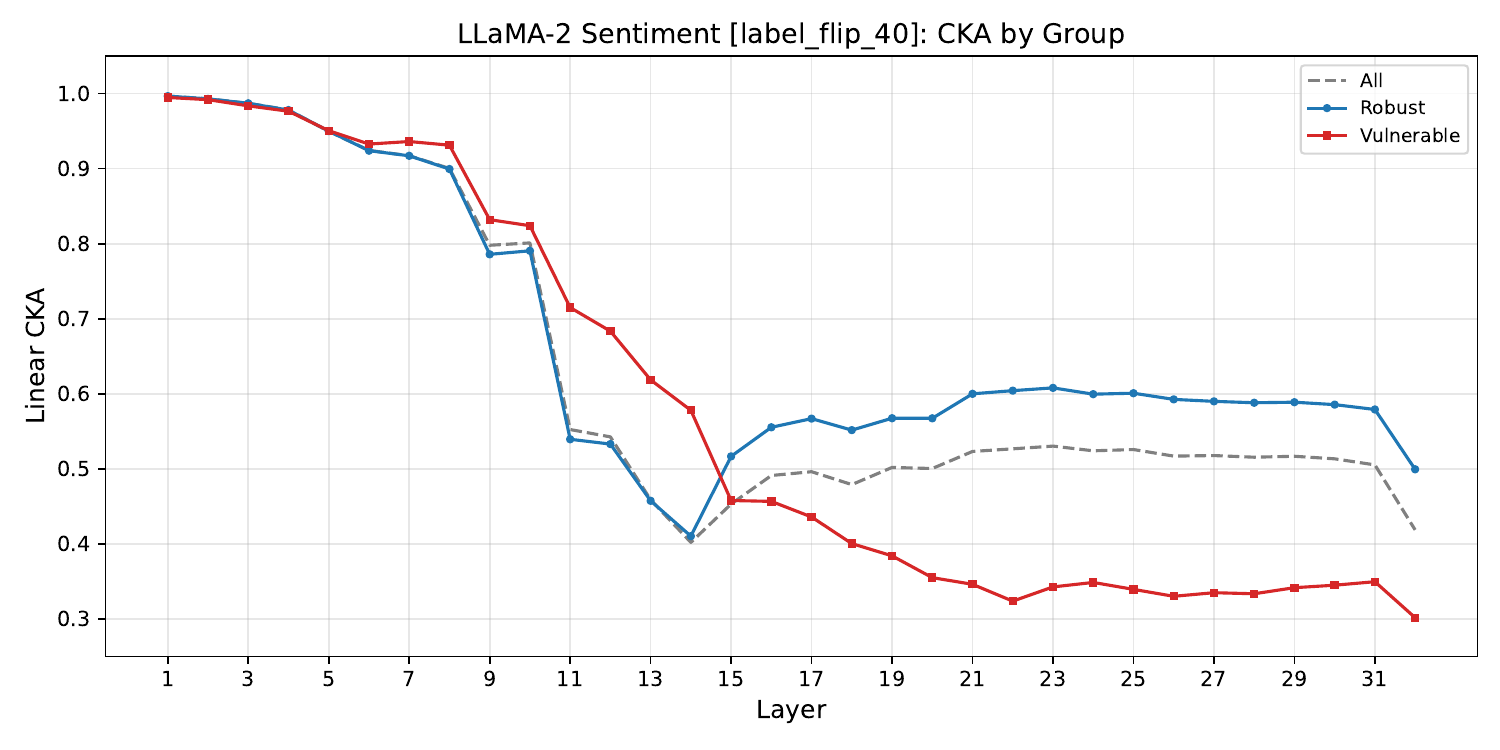}
  \end{subfigure}
    \begin{subfigure}[t]{0.32\textwidth}
    \includegraphics[width=\textwidth]{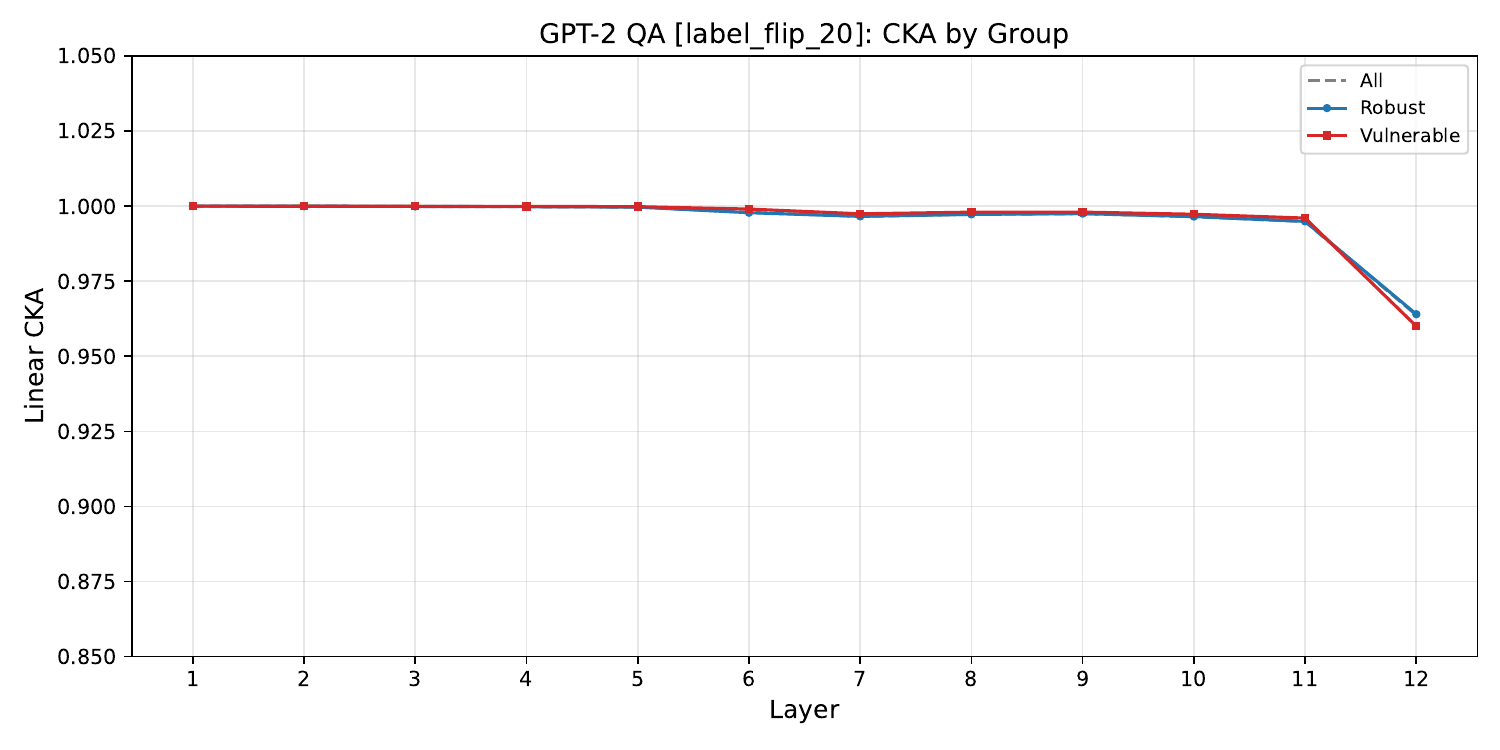}
  \end{subfigure}\hfill
  \begin{subfigure}[t]{0.32\textwidth}
    \includegraphics[width=\textwidth]{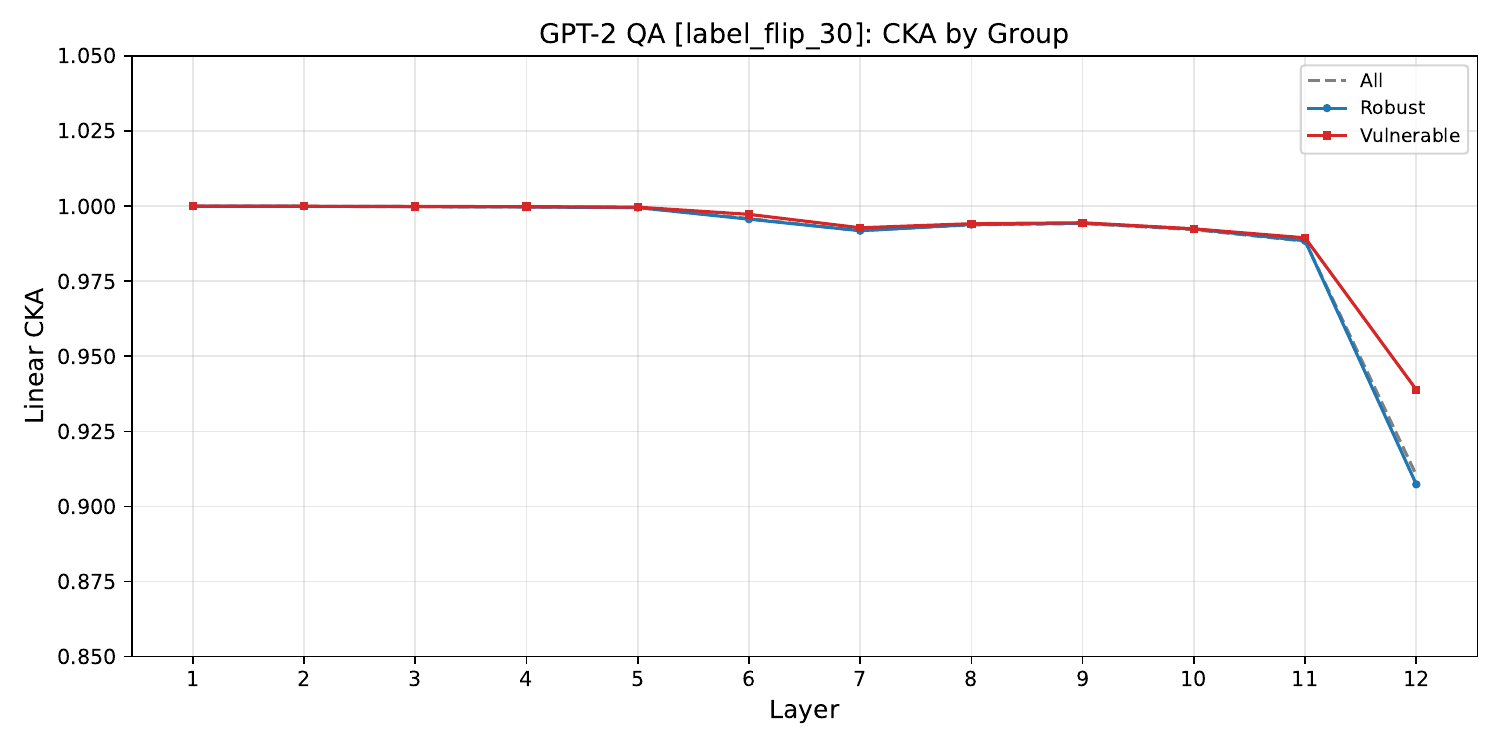}
  \end{subfigure}\hfill
  \begin{subfigure}[t]{0.32\textwidth}
    \includegraphics[width=\textwidth]{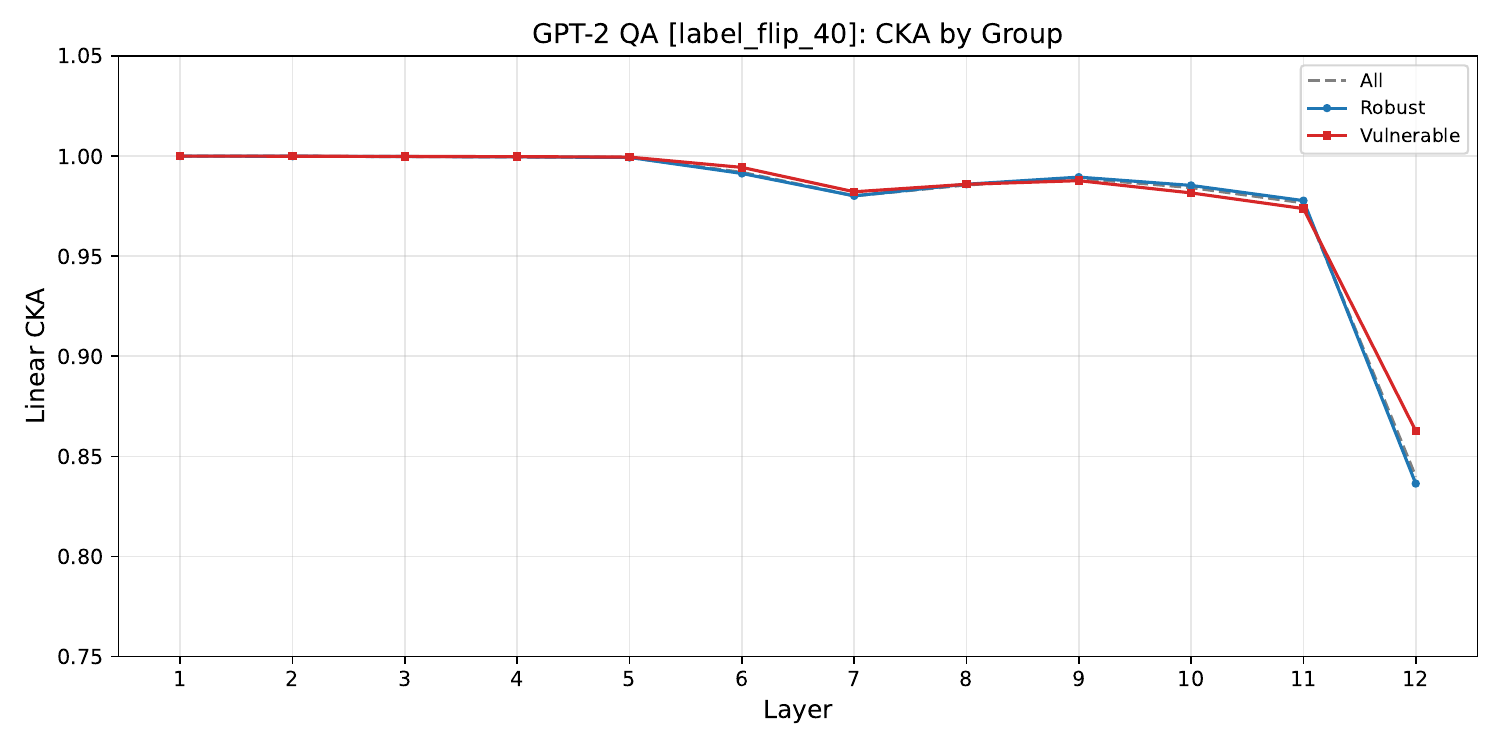}
  \end{subfigure}
  \begin{subfigure}[t]{0.32\textwidth}
    \includegraphics[width=\textwidth]{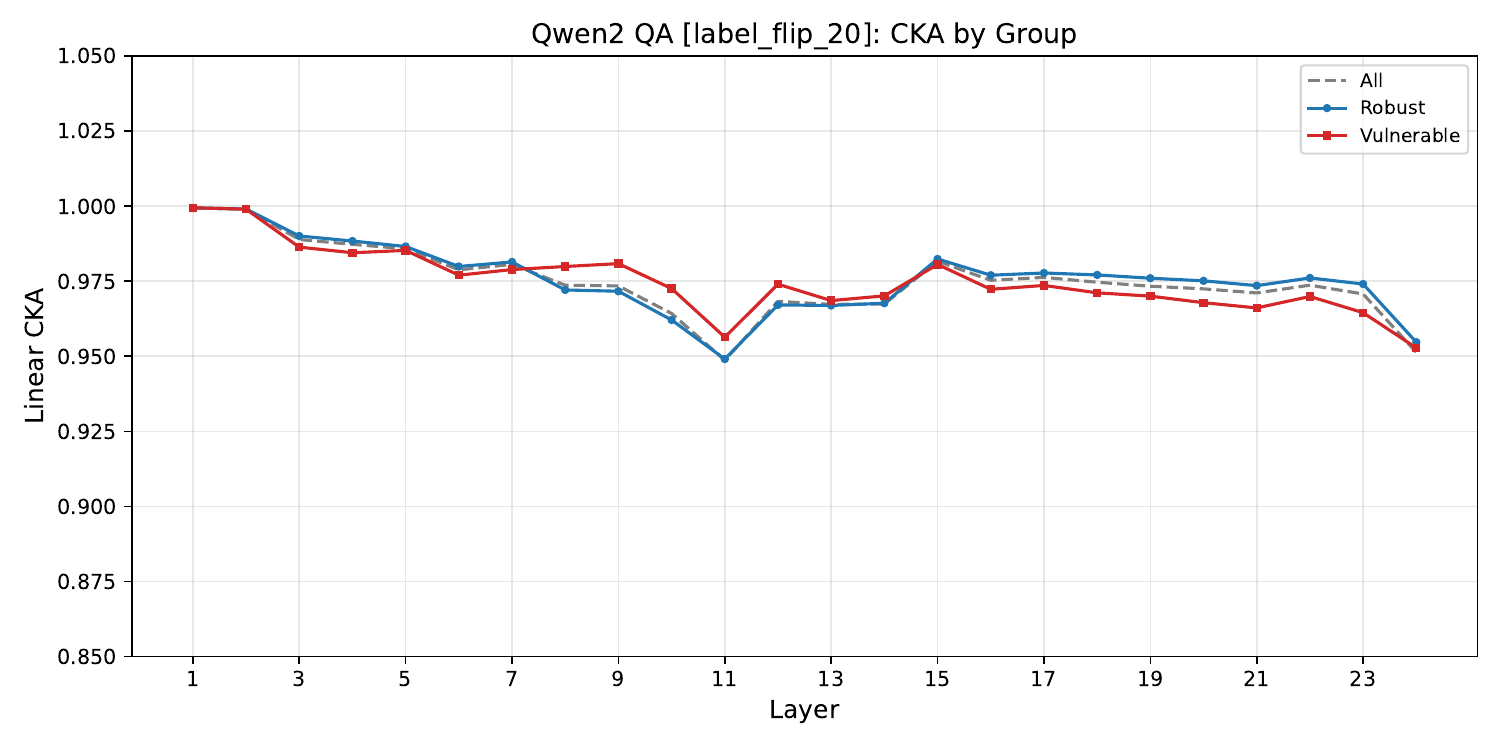}
  \end{subfigure}\hfill
  \begin{subfigure}[t]{0.32\textwidth}
    \includegraphics[width=\textwidth]{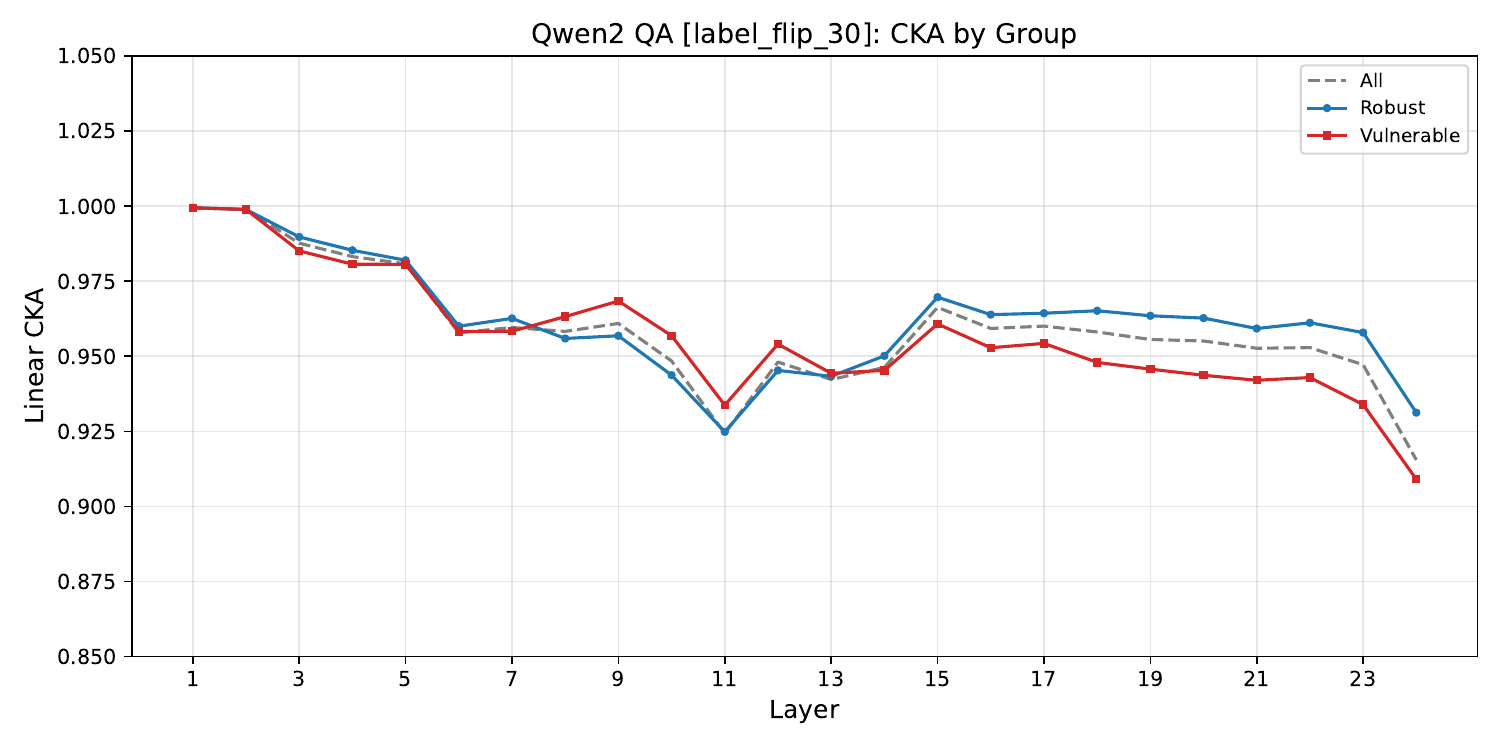}
  \end{subfigure}\hfill
  \begin{subfigure}[t]{0.32\textwidth}
    \includegraphics[width=\textwidth]{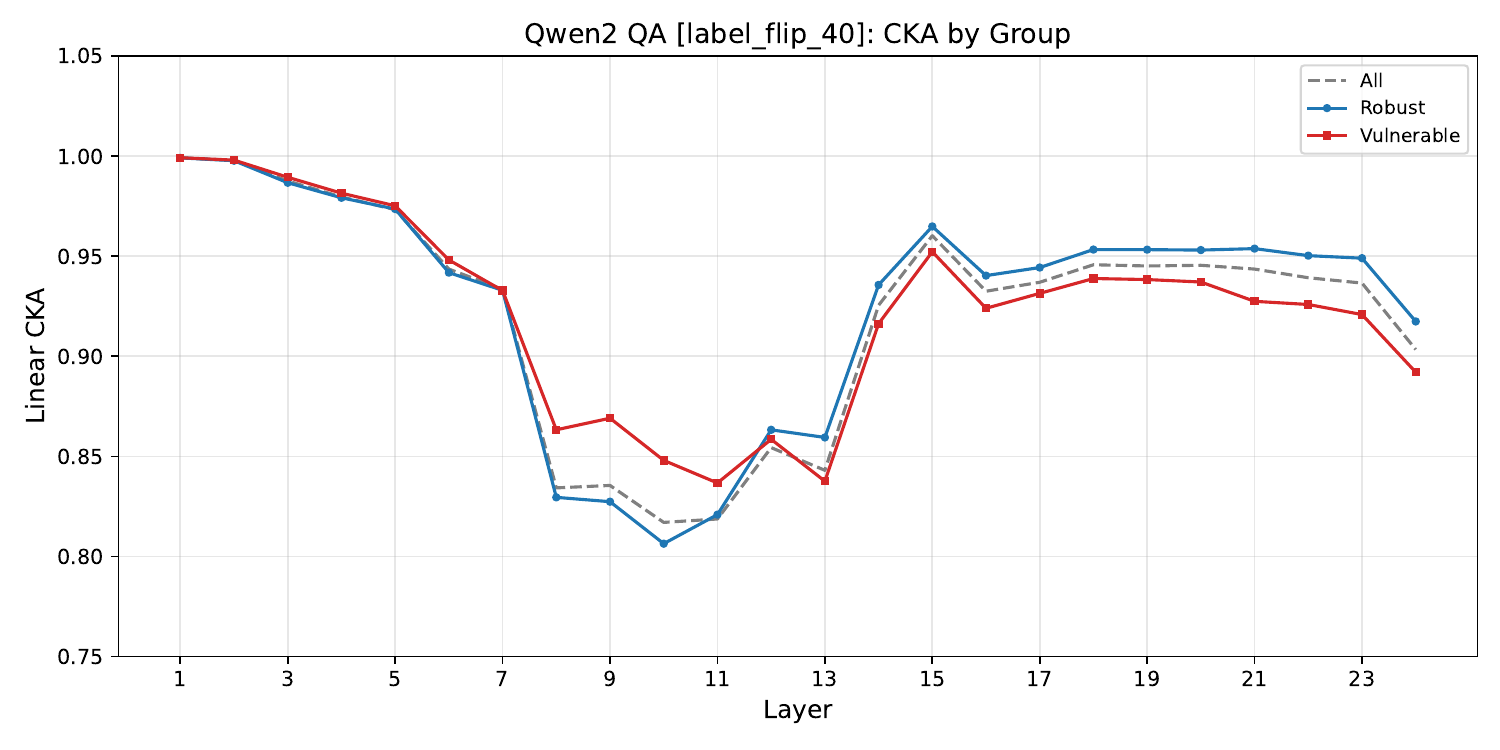}
  \end{subfigure}
  \begin{subfigure}[t]{0.32\textwidth}
    \includegraphics[width=\textwidth]{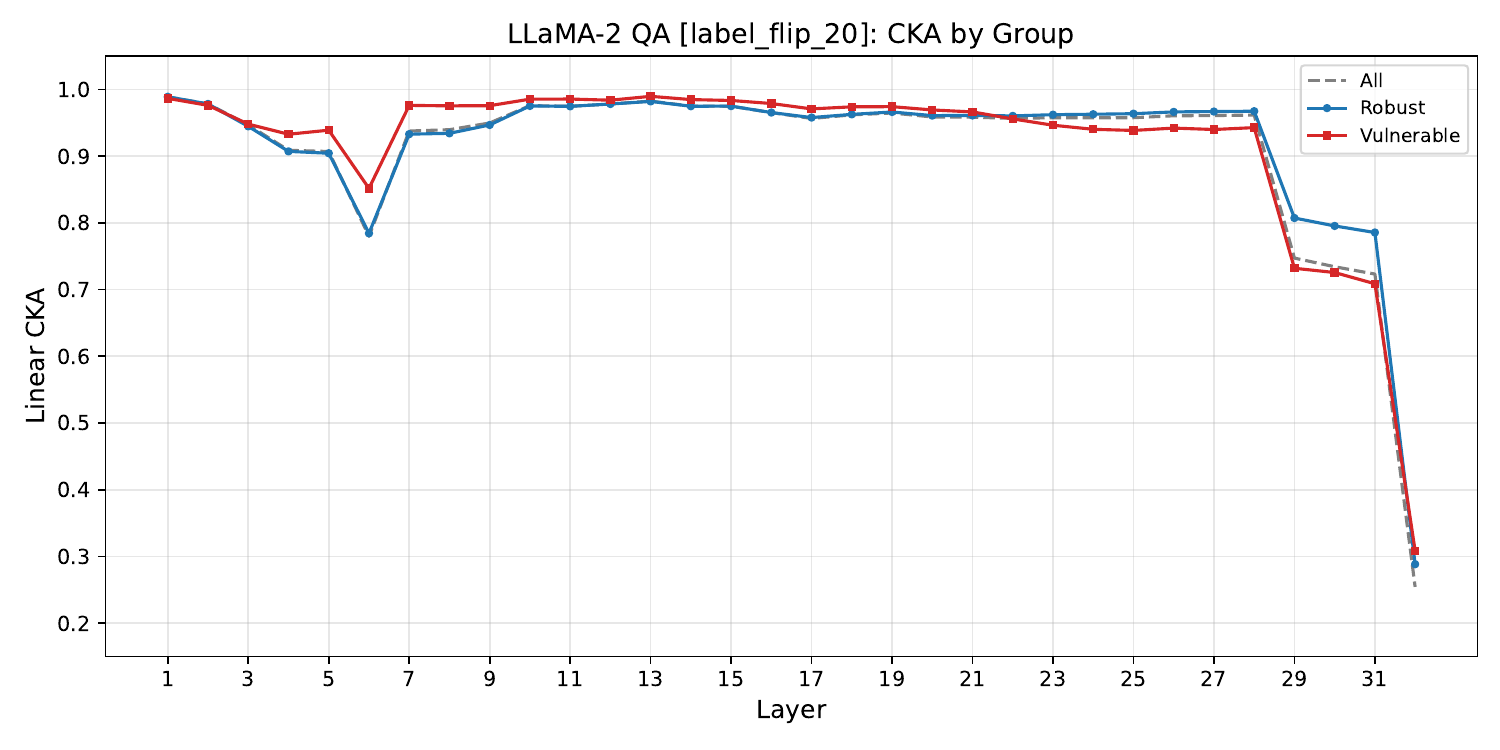}
  \end{subfigure}\hfill
  \begin{subfigure}[t]{0.32\textwidth}
    \includegraphics[width=\textwidth]{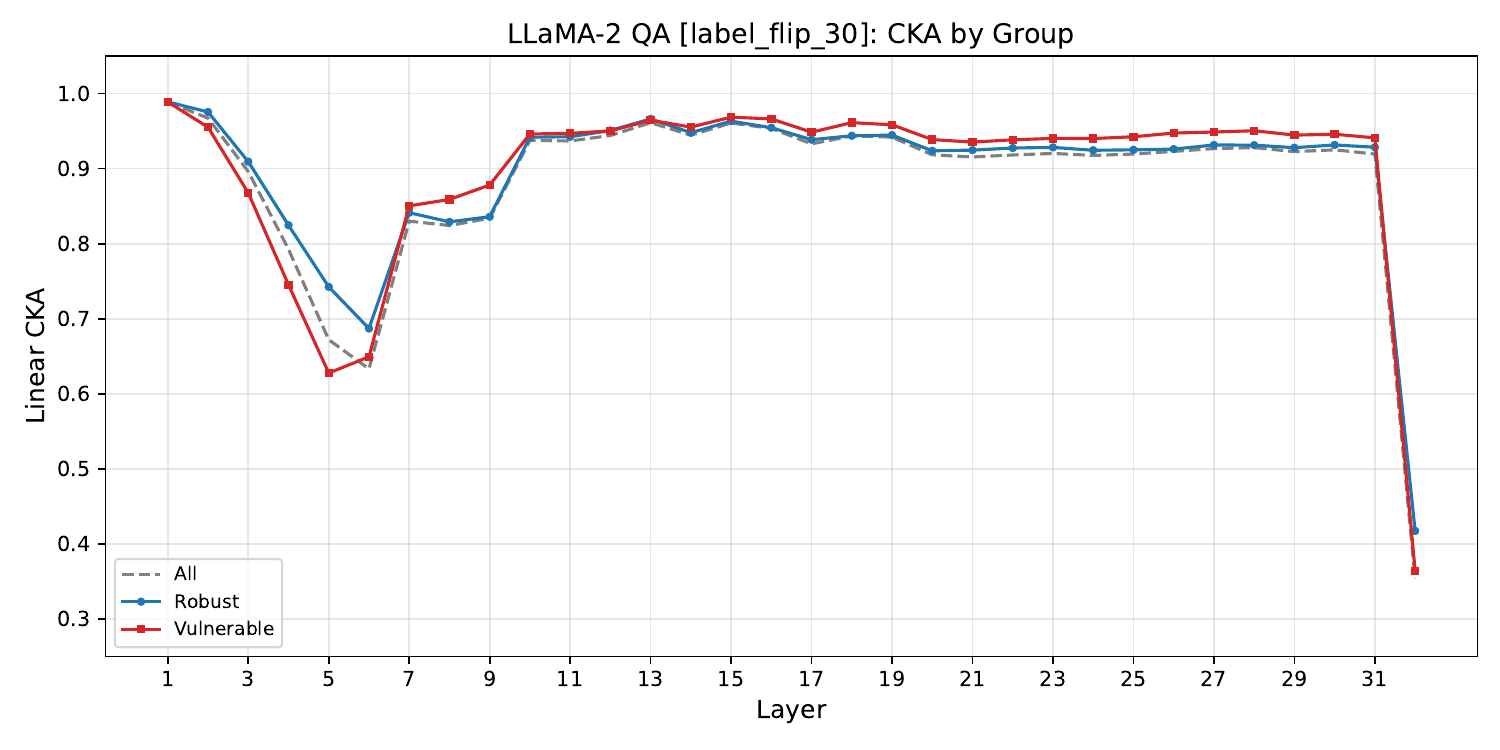}
  \end{subfigure}\hfill
  \begin{subfigure}[t]{0.32\textwidth}
    \includegraphics[width=\textwidth]{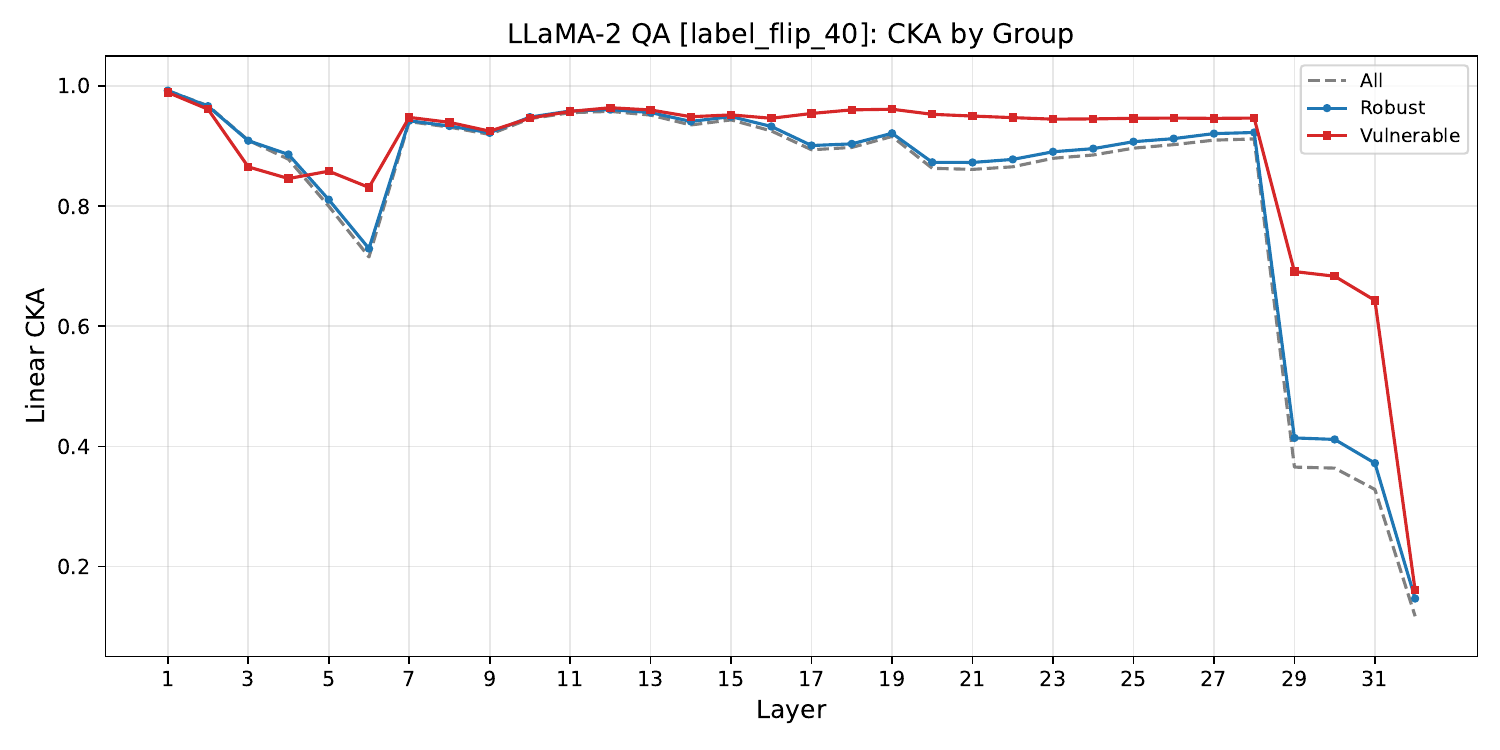}
  \end{subfigure}
    \begin{subfigure}[t]{0.32\textwidth}
    \includegraphics[width=\textwidth]{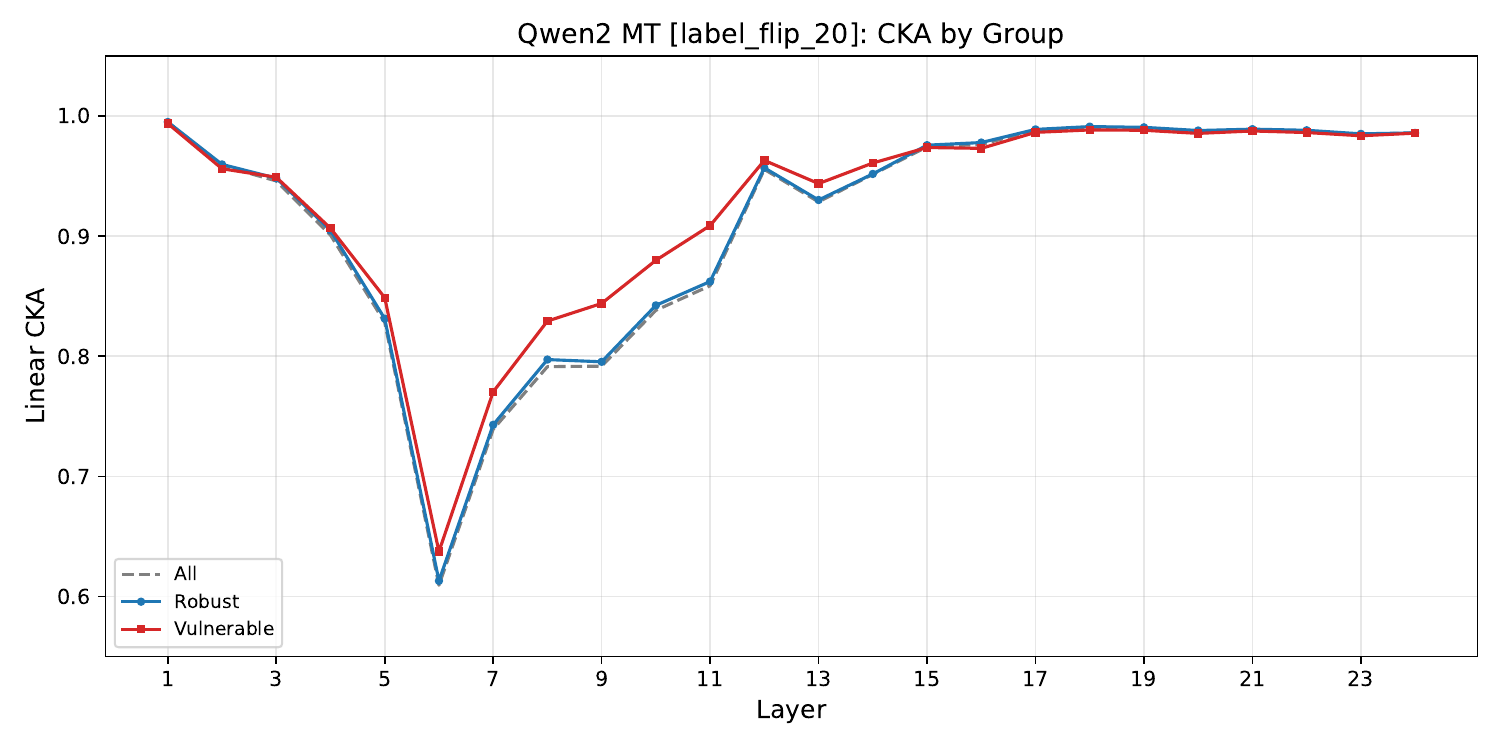}
  \end{subfigure}\hfill
  \begin{subfigure}[t]{0.32\textwidth}
    \includegraphics[width=\textwidth]{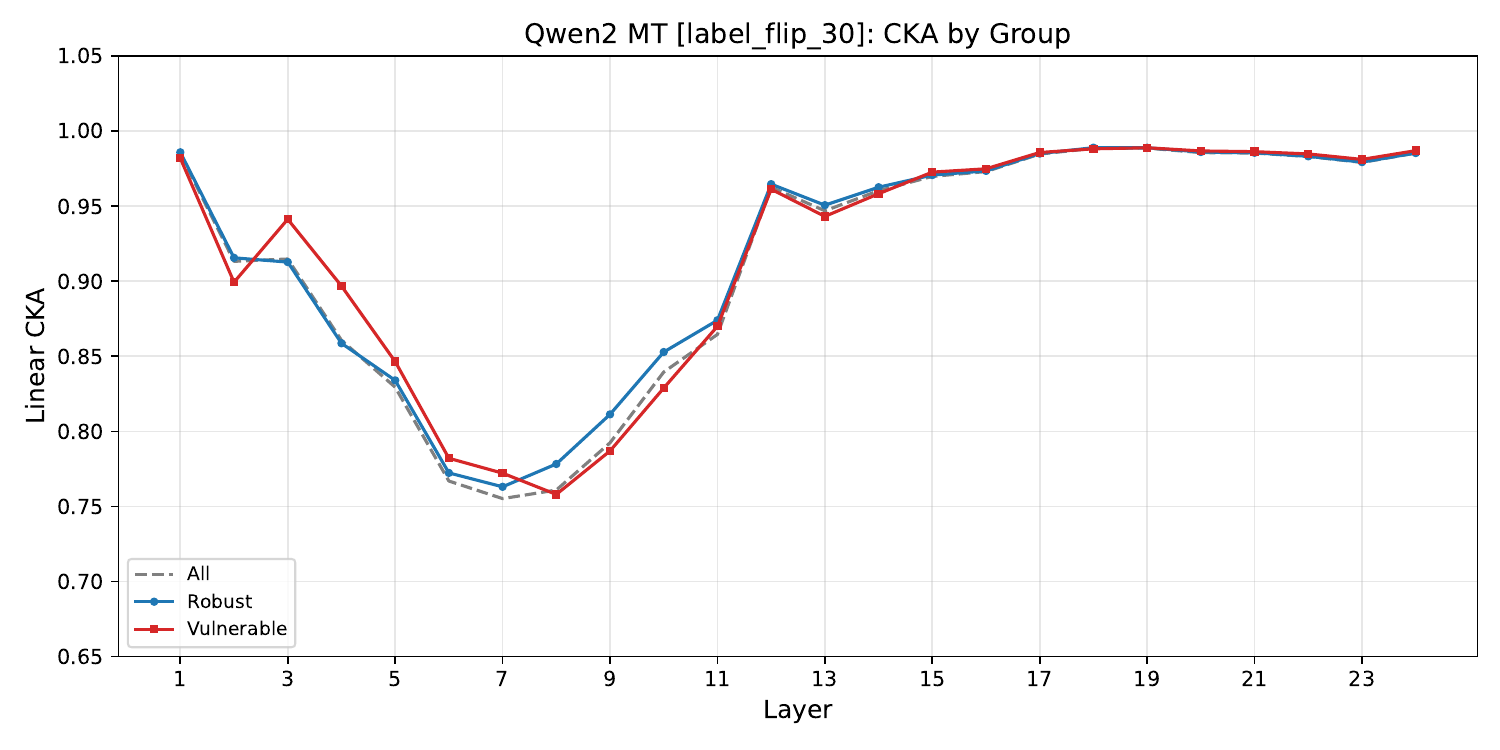}
  \end{subfigure}\hfill
  \begin{subfigure}[t]{0.32\textwidth}
    \includegraphics[width=\textwidth]{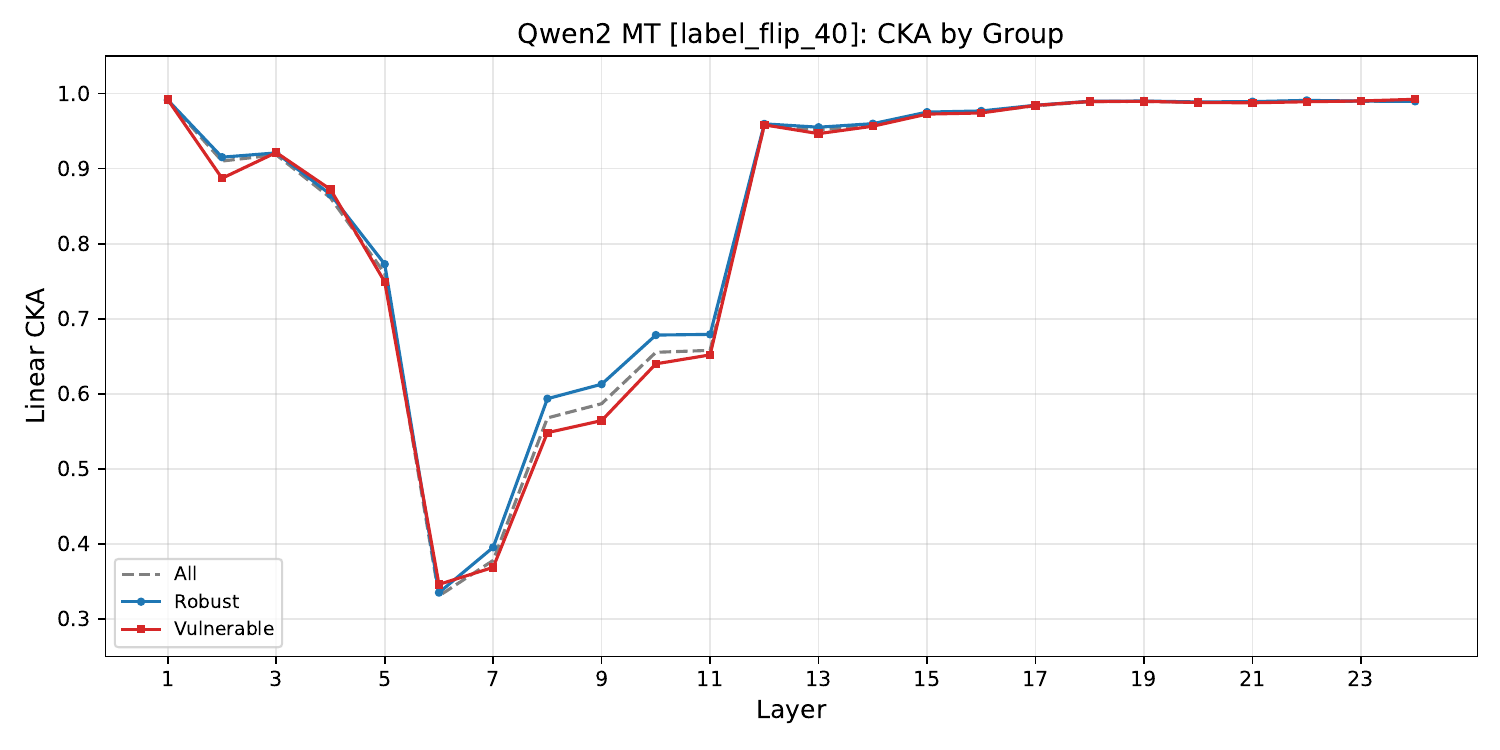}
  \end{subfigure}
  \begin{subfigure}[t]{0.32\textwidth}
    \includegraphics[width=\textwidth]{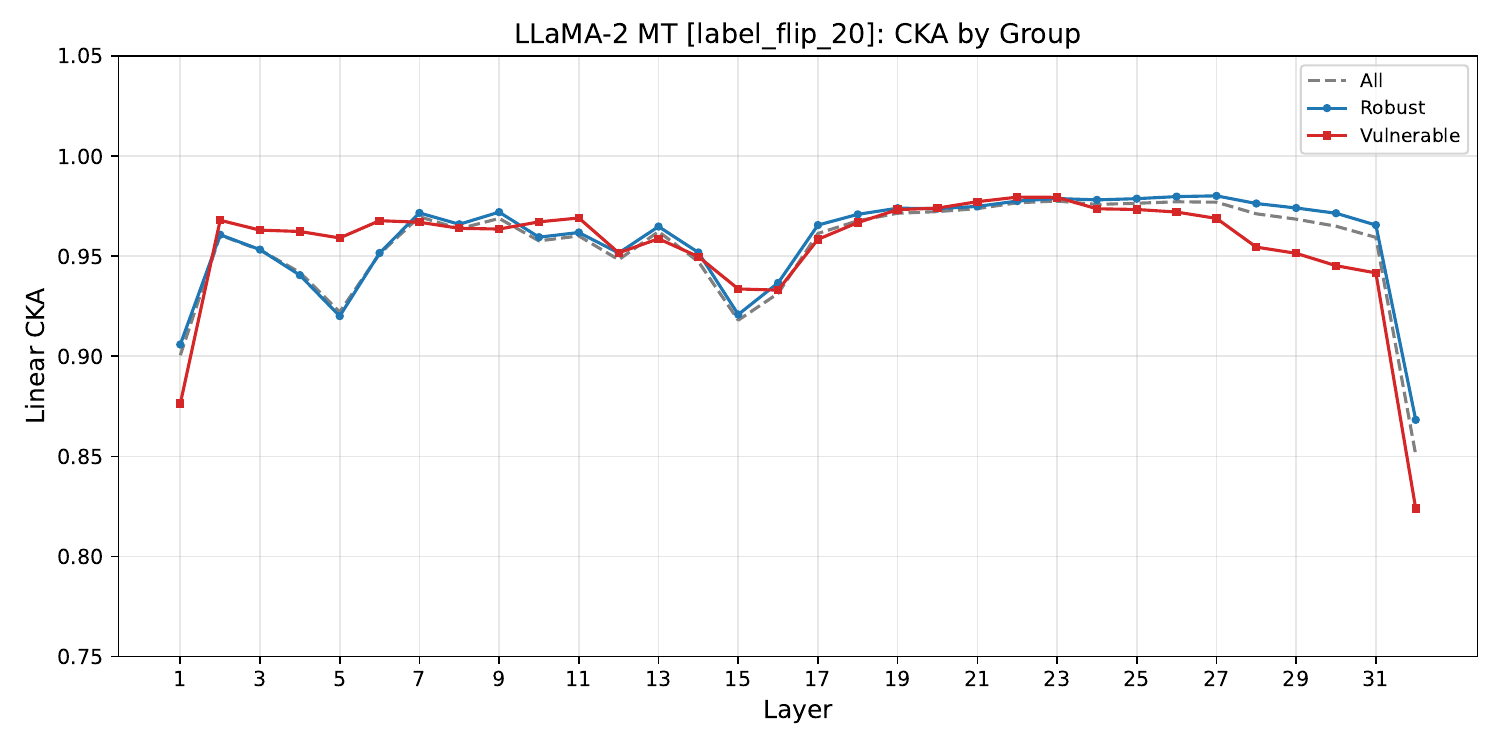}
  \end{subfigure}\hfill
  \begin{subfigure}[t]{0.32\textwidth}
    \includegraphics[width=\textwidth]{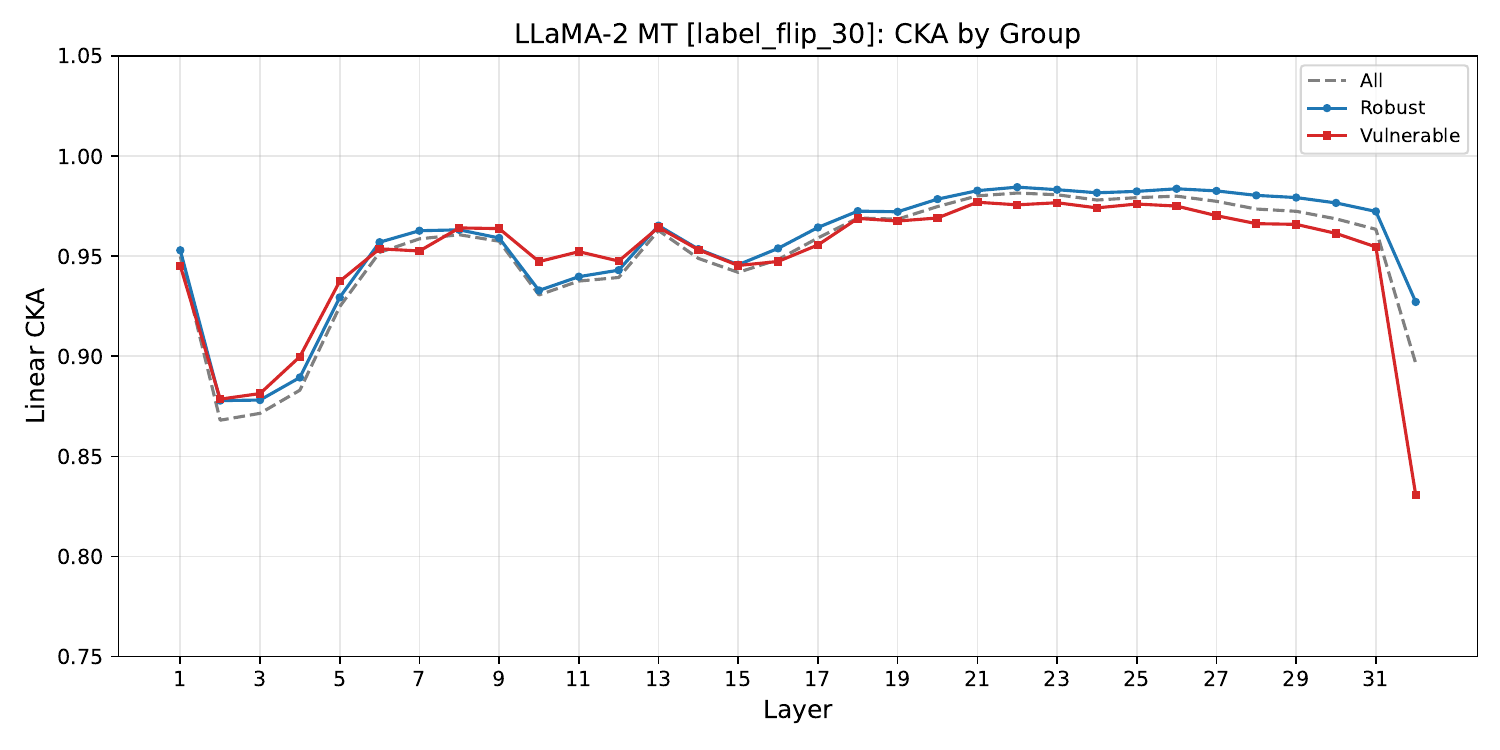}
  \end{subfigure}\hfill
  \begin{subfigure}[t]{0.32\textwidth}
    \includegraphics[width=\textwidth]{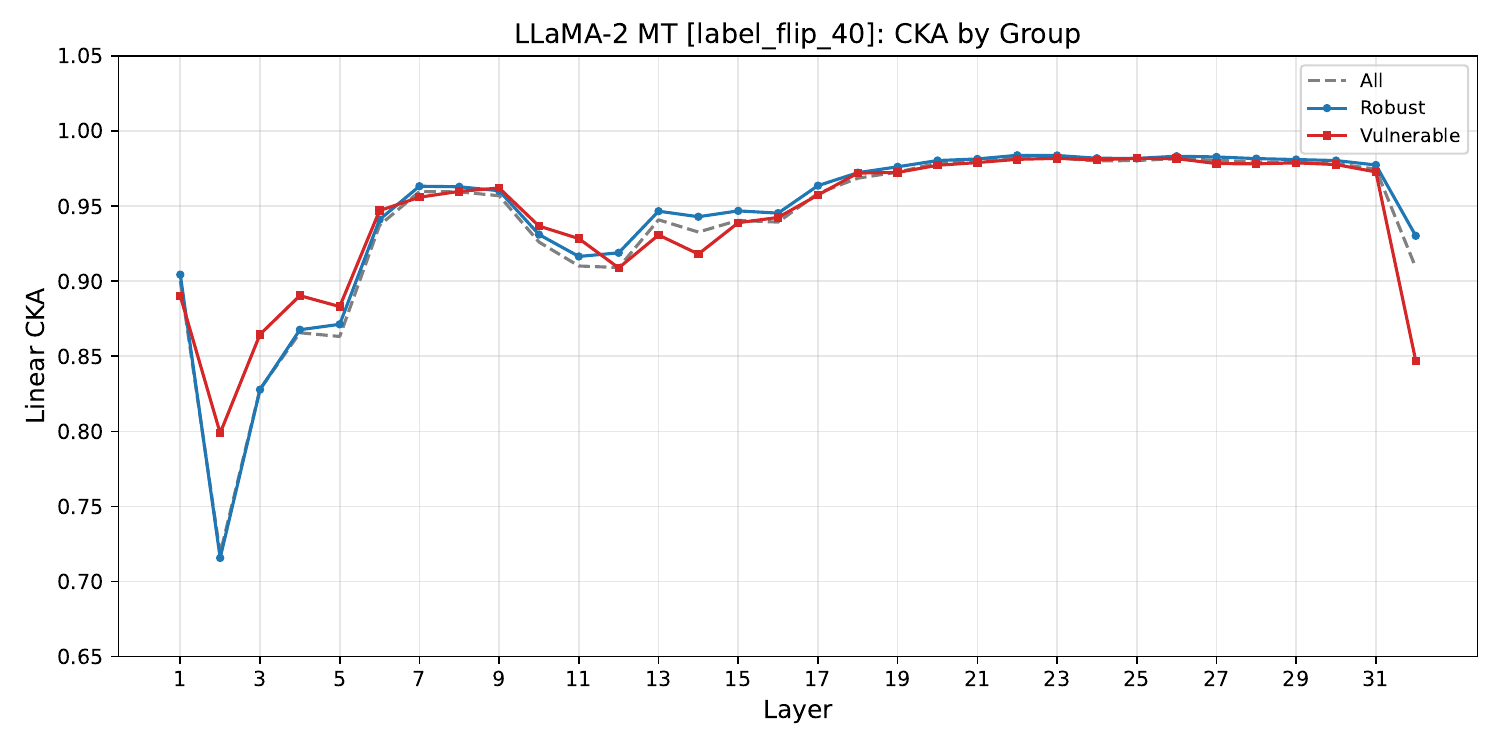}
  \end{subfigure}
  \caption{Robust vs.\ vulnerable stratification: \textbf{Linear CKA} for \textbf{\ac{SC}}, \textbf{\ac{QA}} and \textbf{\ac{MT}} under label-flip noise. CKA captures inter-sample relational structure. 
  }
  \label{fig:stratification_sent_cka}
\end{figure*}

This appendix presents the full stratification results for robust and vulnerable samples across all model–task combinations under label-flip noise. Robust samples are those whose predictions remain unchanged after fine-tuning with noisy data, while vulnerable samples are those whose predictions change.
Figure~\ref{fig:stratification_sent_centered_cosine} shows the centered cosine similarity by group . In most cases, vulnerable samples show lower centered cosine similarity than robust samples at deeper layers, indicating greater representational distortion for samples whose predictions are affected by noise. However, it is interesting to note that even robust samples show non-trivial representational distortion despite their predictions remaining unchanged.
Figure~\ref{fig:stratification_sent_cka} shows the Linear CKA by group. The vulnerable--robust gap is most pronounced for Qwen2 sentiment at 40\% noise, where vulnerable CKA drops to 0.260 compared to 0.612 for robust samples at the final layer. An exception is observed for LLaMA-2 QA, where vulnerable CKA slightly exceeds robust CKA across multiple layers and noise levels, reversing the expected direction. This anomaly may reflect the optimization instability of LLaMA-2 near the critical noise threshold, where small changes in training conditions can lead to very different outcomes, as also suggested by the multi-seed analysis in Appendix~\ref{app:seed_stability}.
Figure~\ref{fig:stratification_qa_mrr_first} shows the first-token Logit Lens MRR by group for QA and MT. Llama-2 vulnerable MRR collapses to 0.365 at the final layer under 40\% noise, compared to 0.740 for robust samples, the largest functional gap observed across all conditions.

\begin{figure*}[p]
  \centering
  \begin{subfigure}[t]{0.32\textwidth}
    \includegraphics[width=\textwidth]{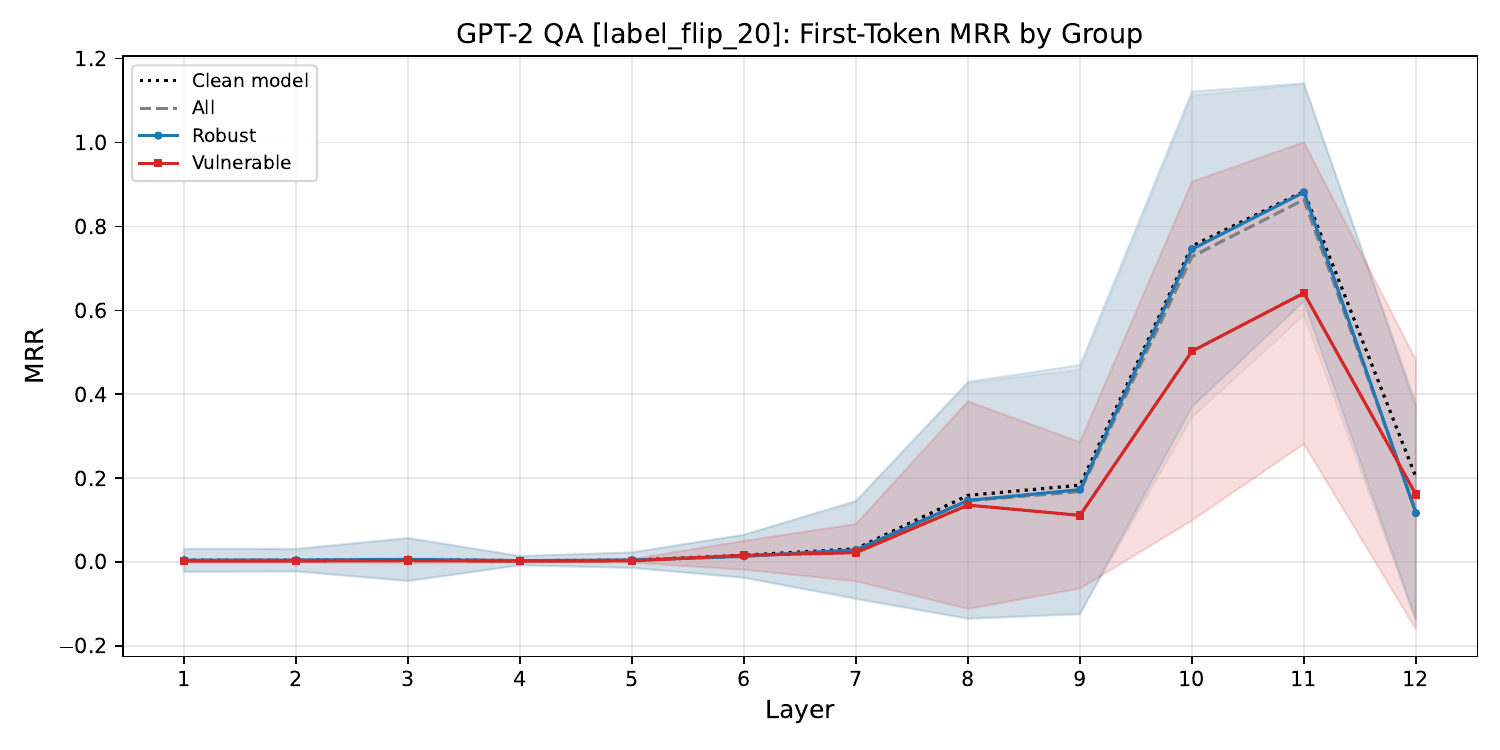}
  \end{subfigure}\hfill
  \begin{subfigure}[t]{0.32\textwidth}
    \includegraphics[width=\textwidth]{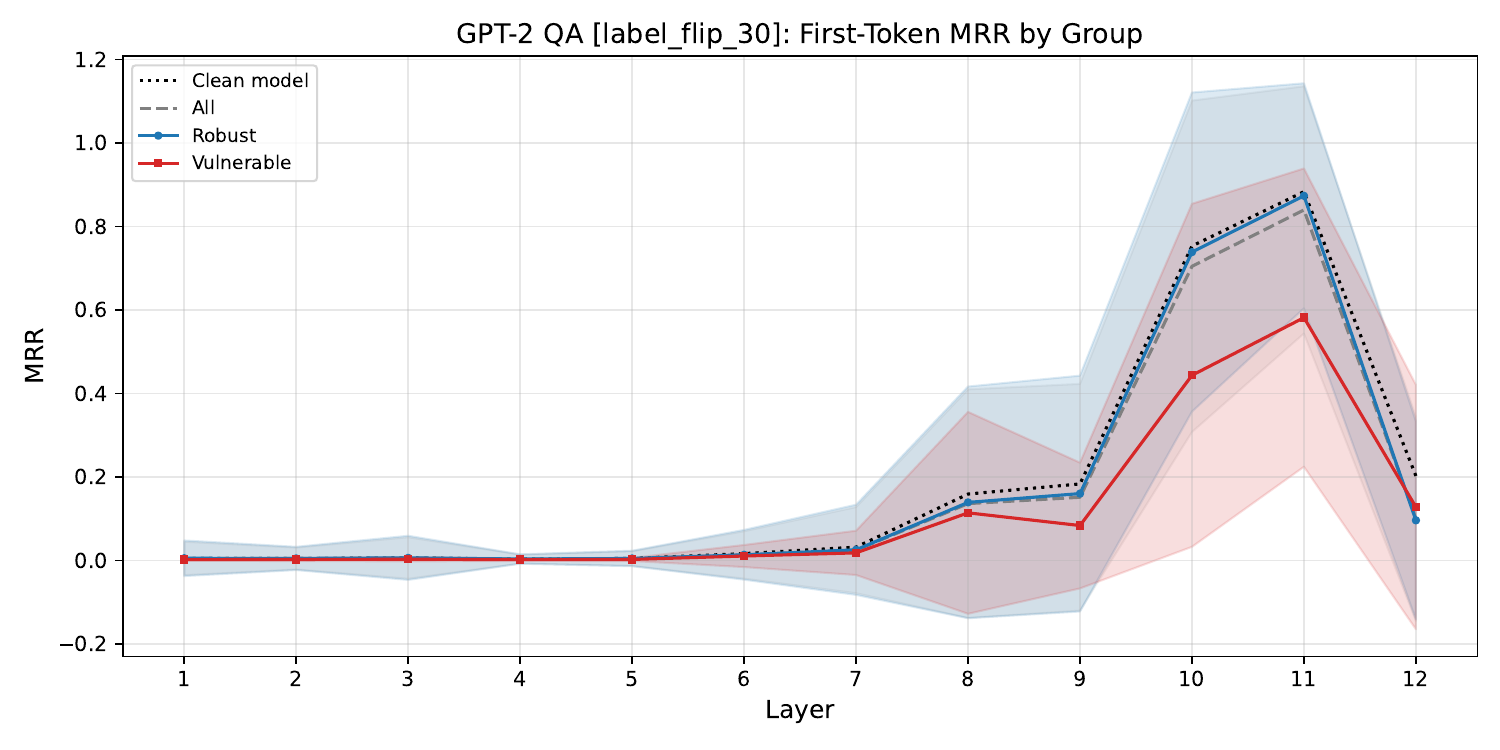}
  \end{subfigure}\hfill
  \begin{subfigure}[t]{0.32\textwidth}
    \includegraphics[width=\textwidth]{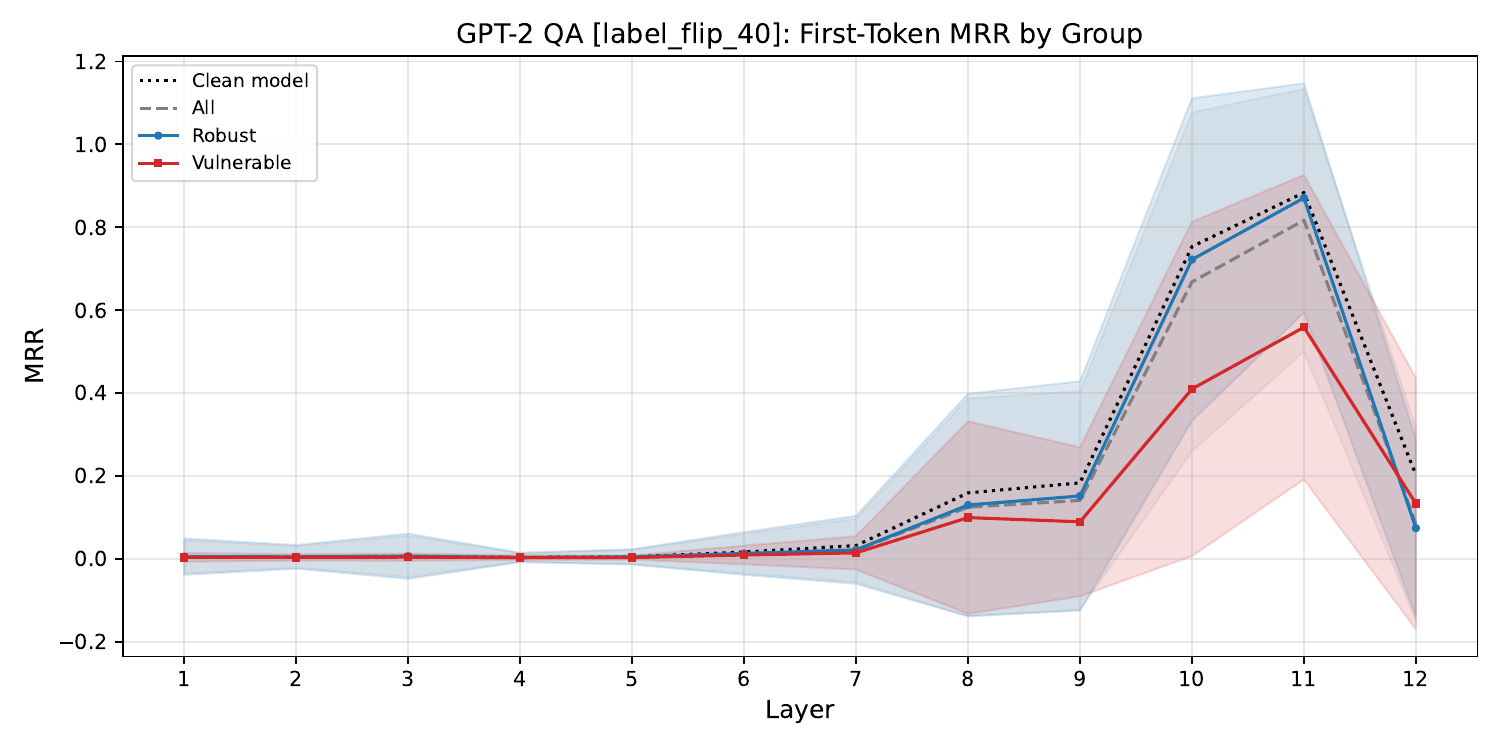}
  \end{subfigure}
  \begin{subfigure}[t]{0.32\textwidth}
    \includegraphics[width=\textwidth]{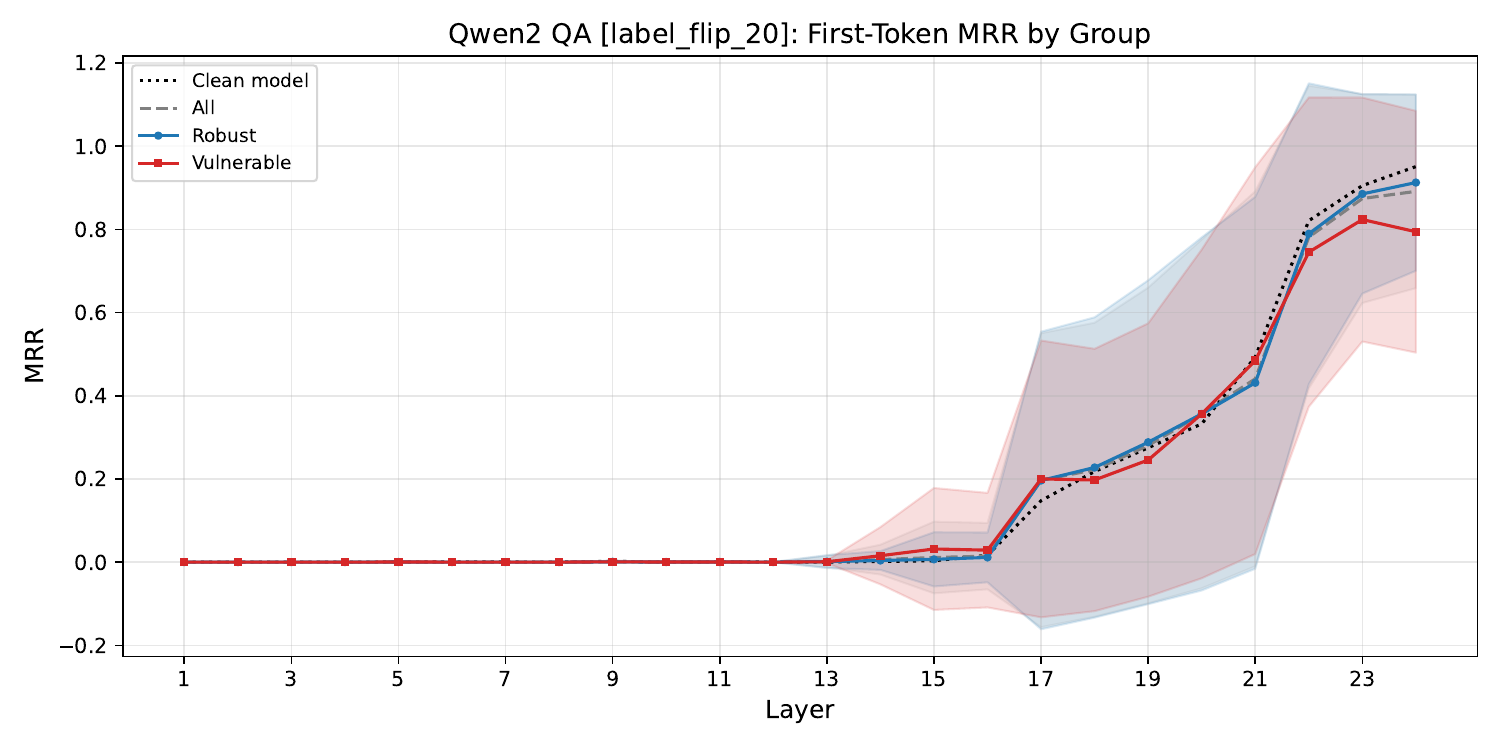}
  \end{subfigure}\hfill
  \begin{subfigure}[t]{0.32\textwidth}
    \includegraphics[width=\textwidth]{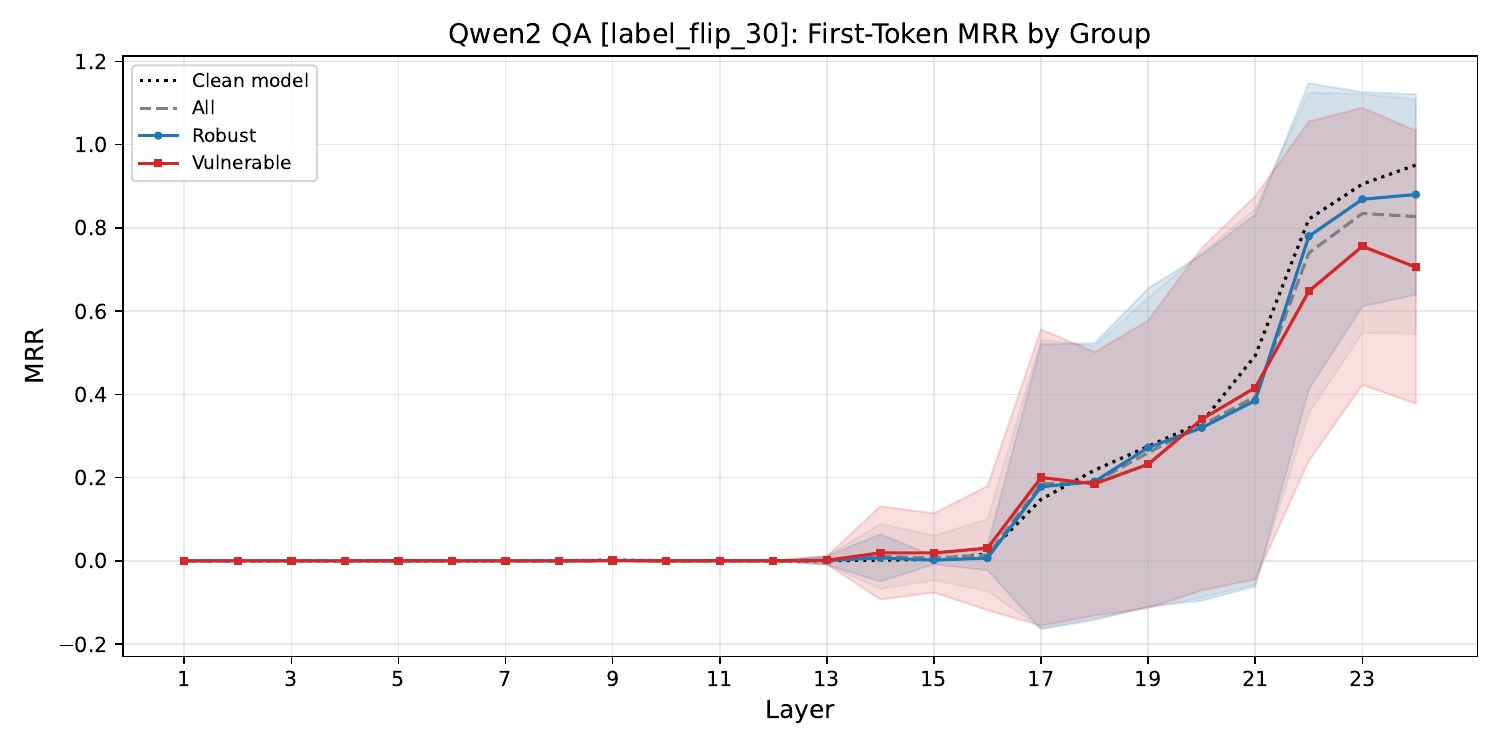}
  \end{subfigure}\hfill
  \begin{subfigure}[t]{0.32\textwidth}
    \includegraphics[width=\textwidth]{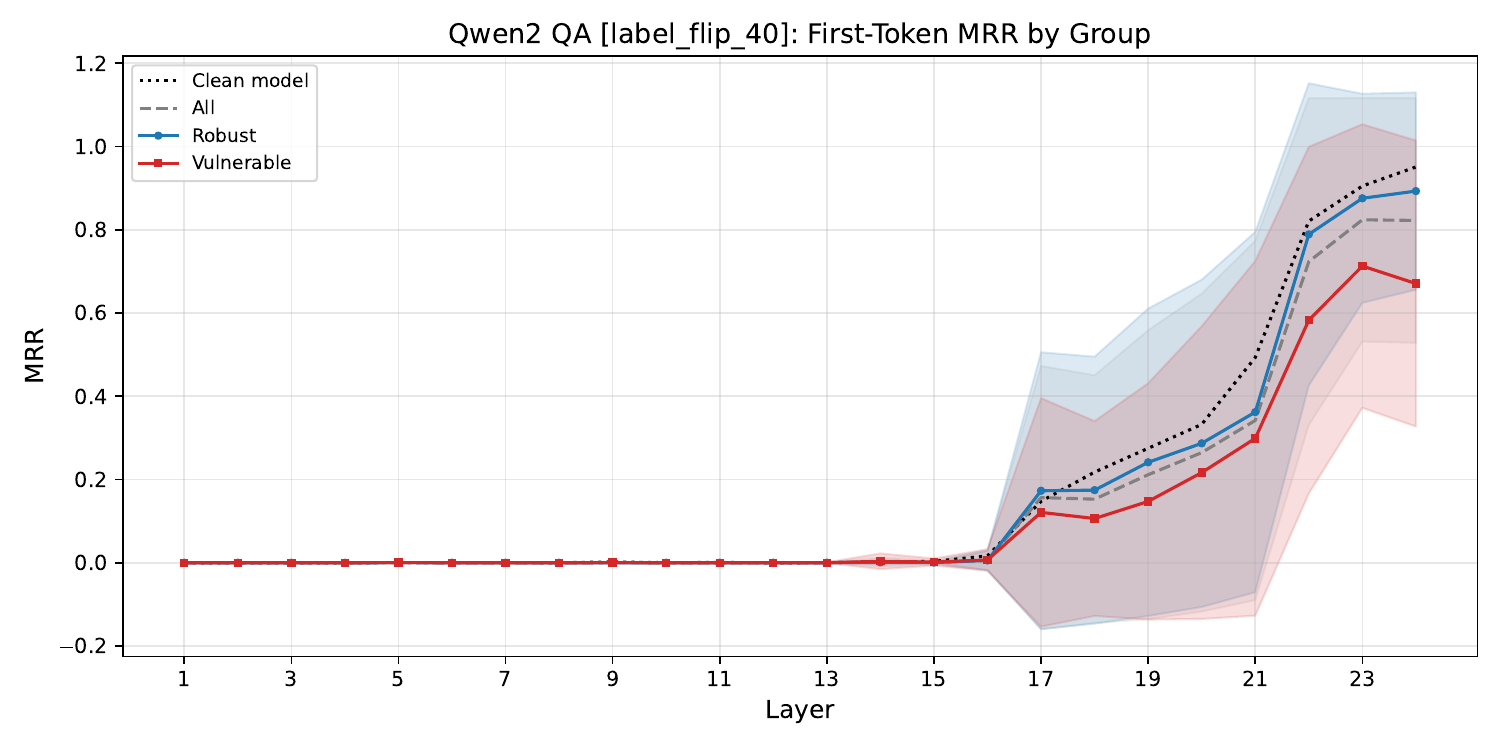}
  \end{subfigure}
  \begin{subfigure}[t]{0.32\textwidth}
    \includegraphics[width=\textwidth]{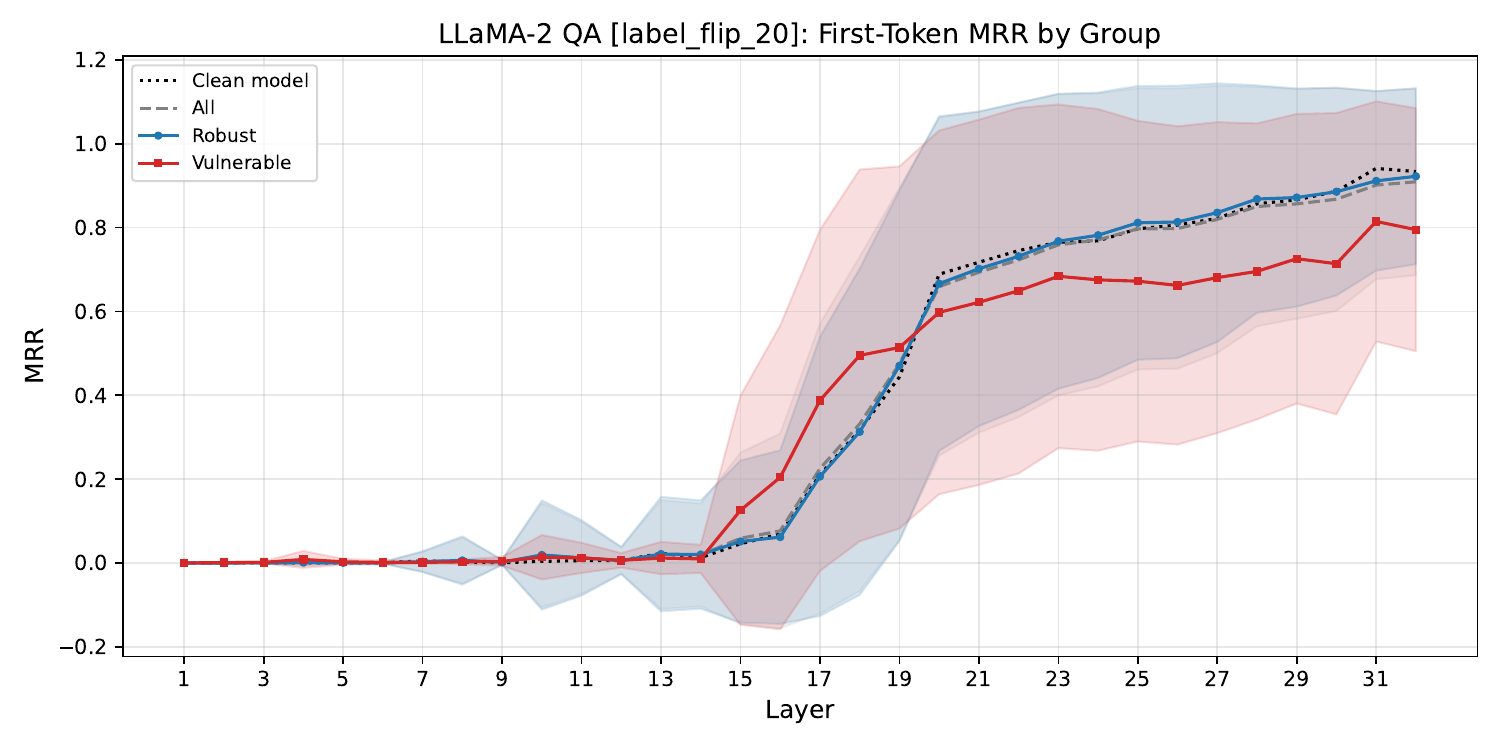}
  \end{subfigure}\hfill
  \begin{subfigure}[t]{0.32\textwidth}
    \includegraphics[width=\textwidth]{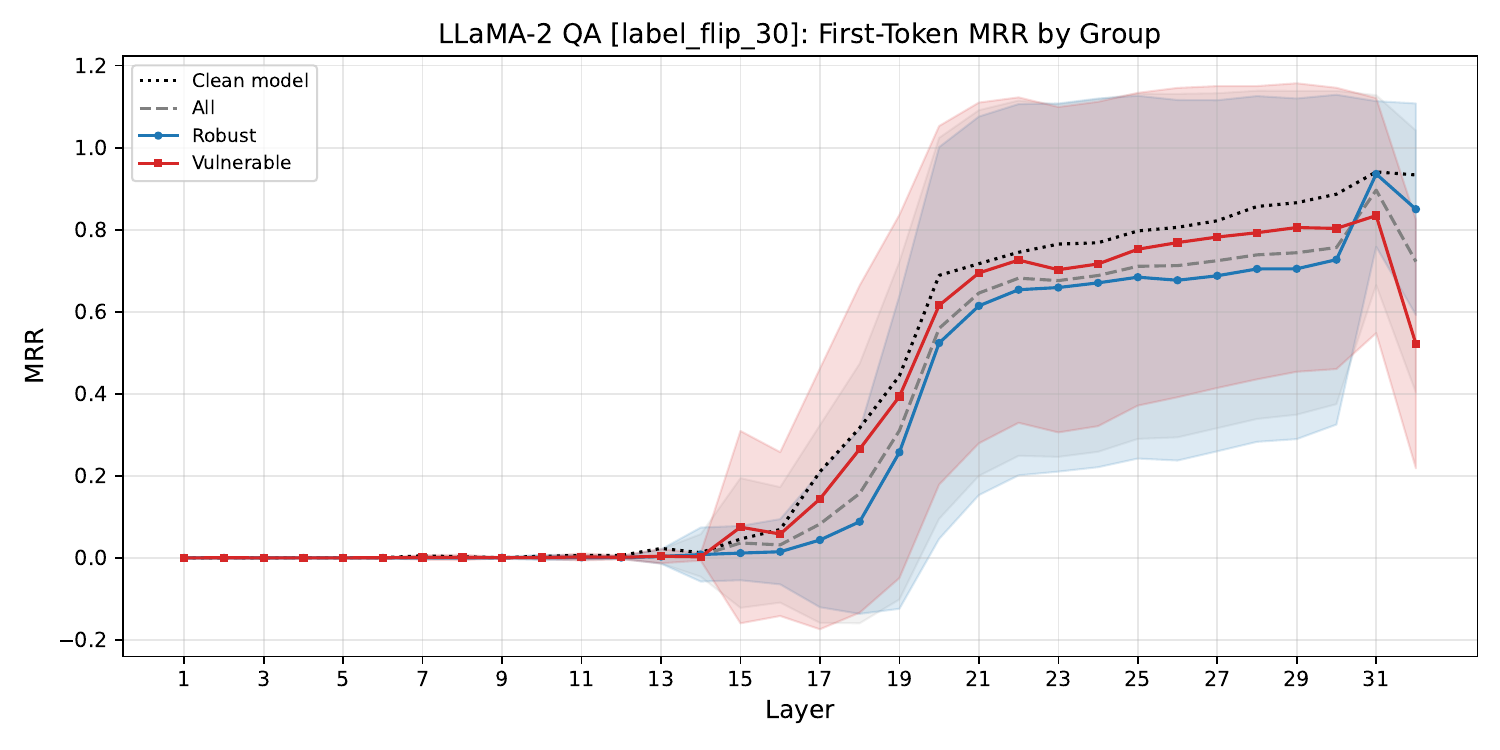}
  \end{subfigure}\hfill
  \begin{subfigure}[t]{0.32\textwidth}
    \includegraphics[width=\textwidth]{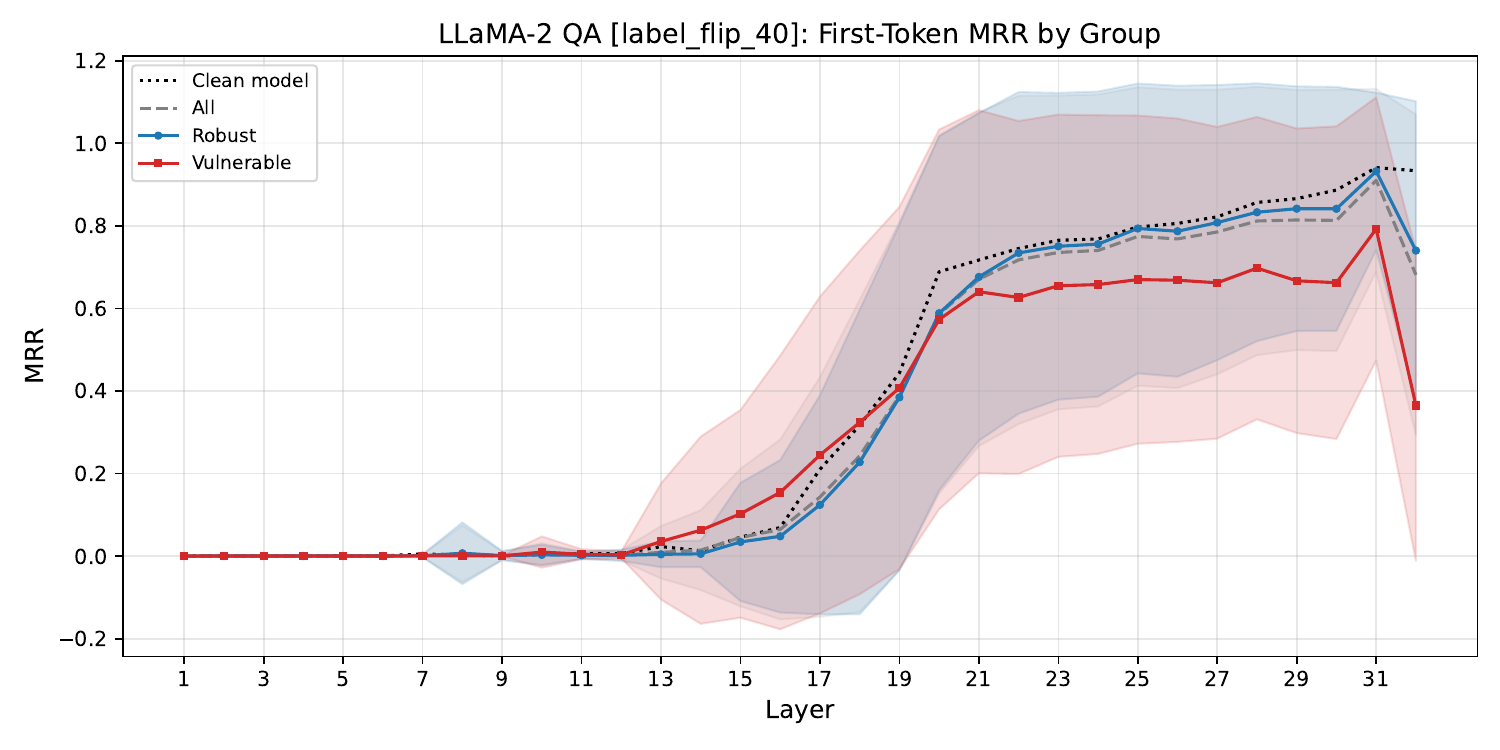}
  \end{subfigure}

   \begin{subfigure}[t]{0.32\textwidth}
    \includegraphics[width=\textwidth]{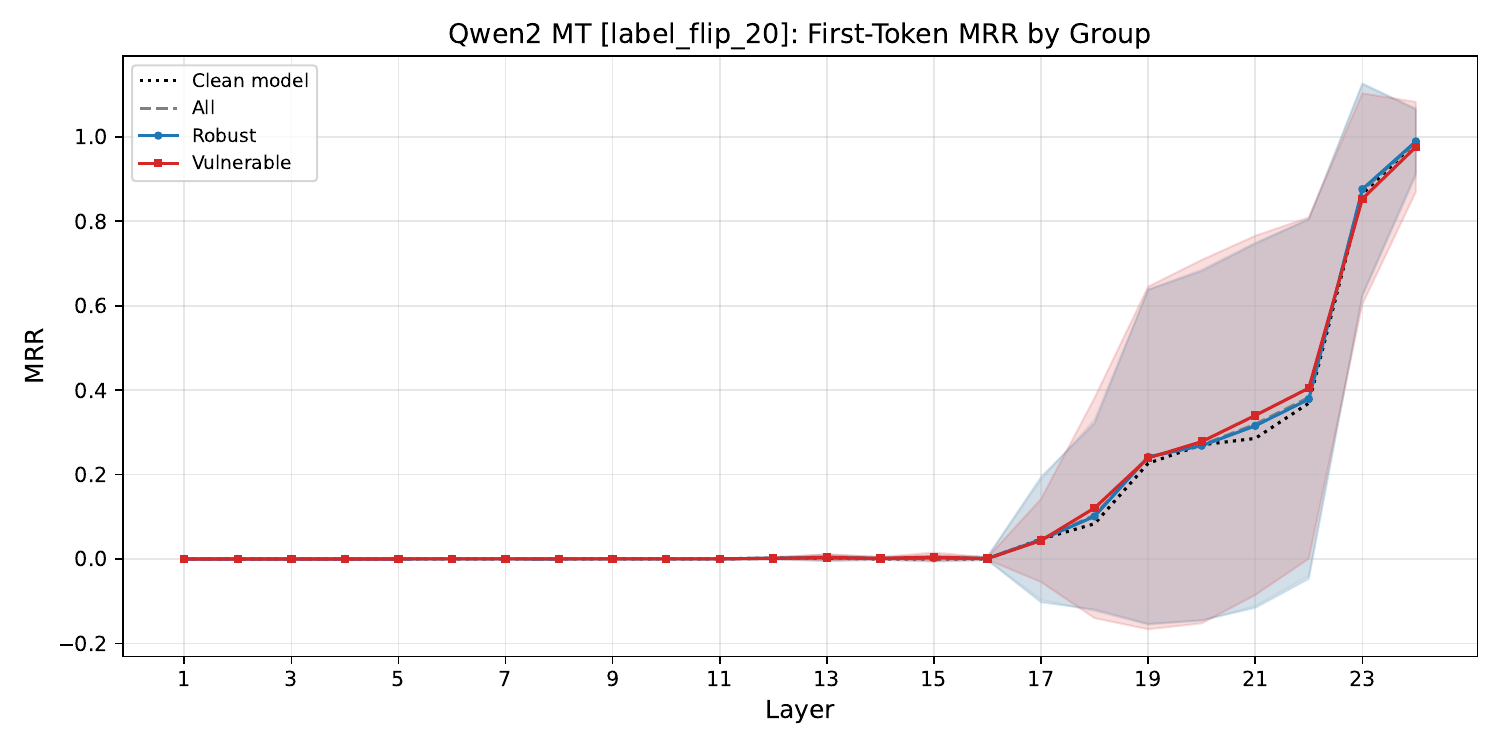}
  \end{subfigure}\hfill
  \begin{subfigure}[t]{0.32\textwidth}
    \includegraphics[width=\textwidth]{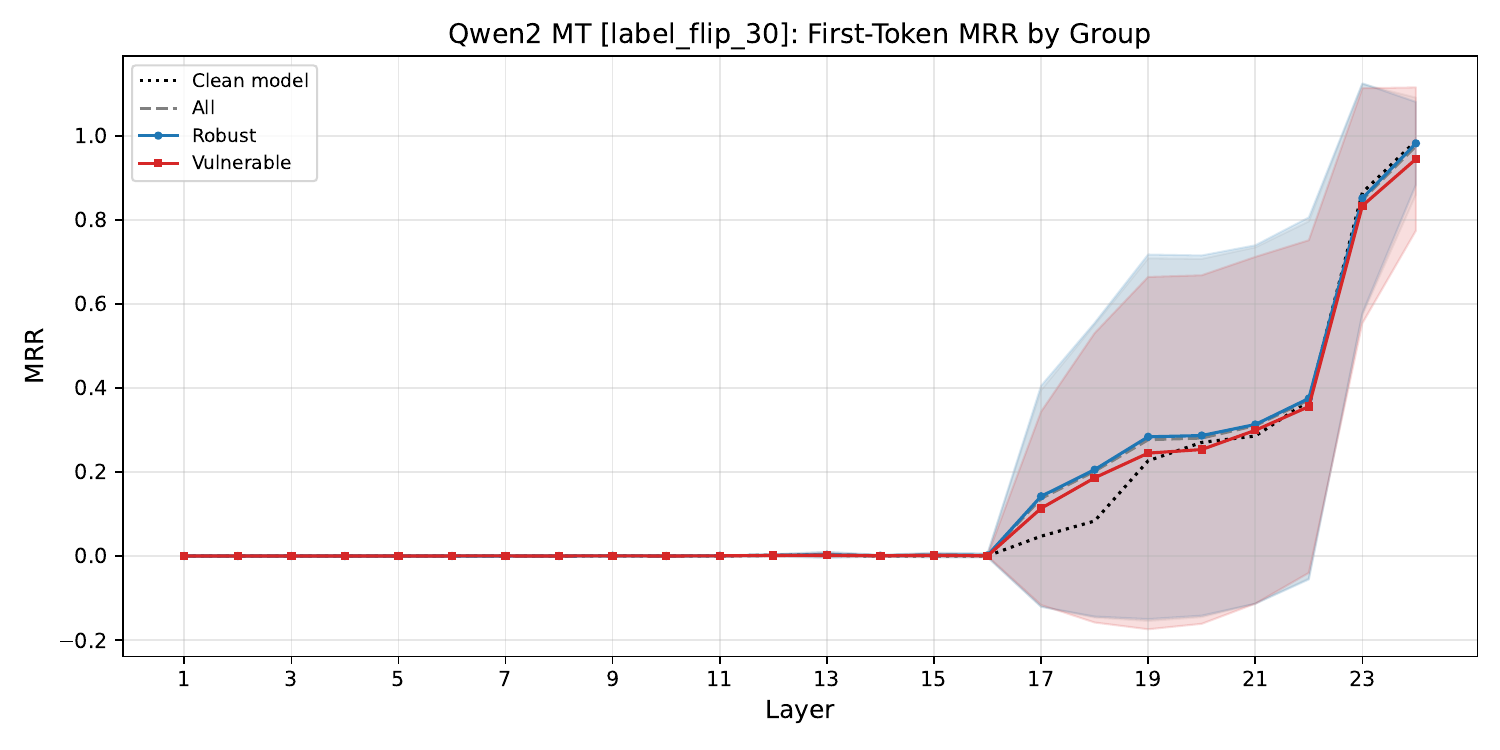}
  \end{subfigure}\hfill
  \begin{subfigure}[t]{0.32\textwidth}
    \includegraphics[width=\textwidth]{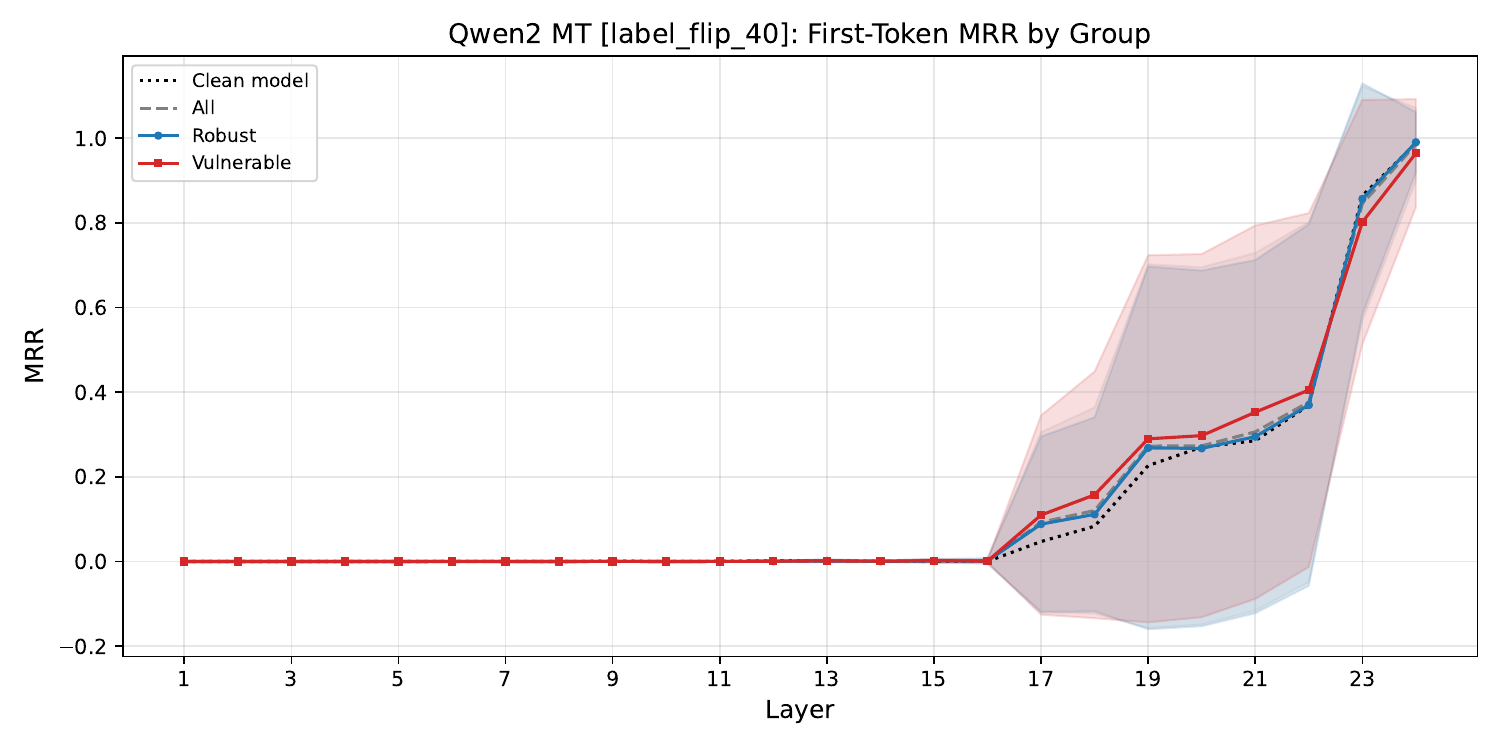}
  \end{subfigure}
  \begin{subfigure}[t]{0.32\textwidth}
    \includegraphics[width=\textwidth]{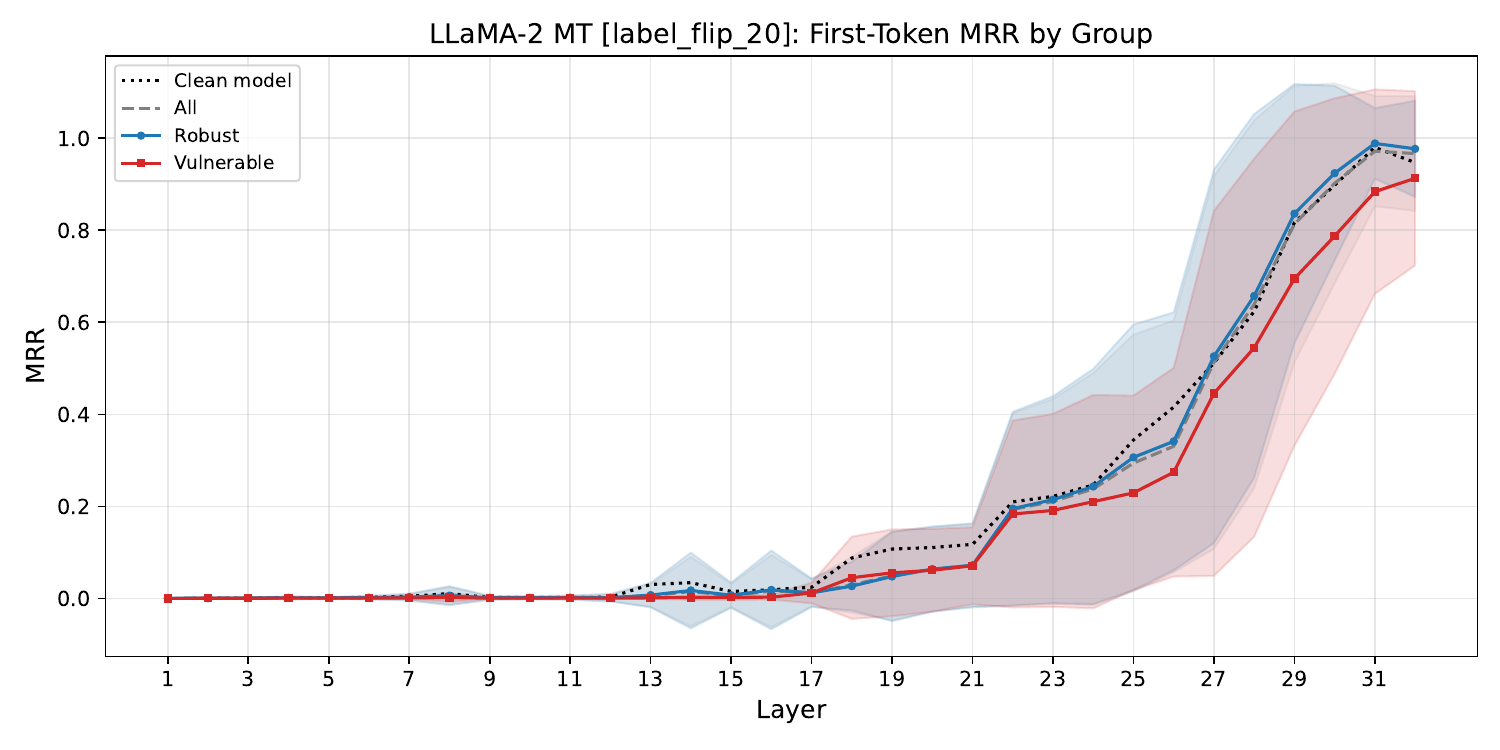}
  \end{subfigure}\hfill
  \begin{subfigure}[t]{0.32\textwidth}
    \includegraphics[width=\textwidth]{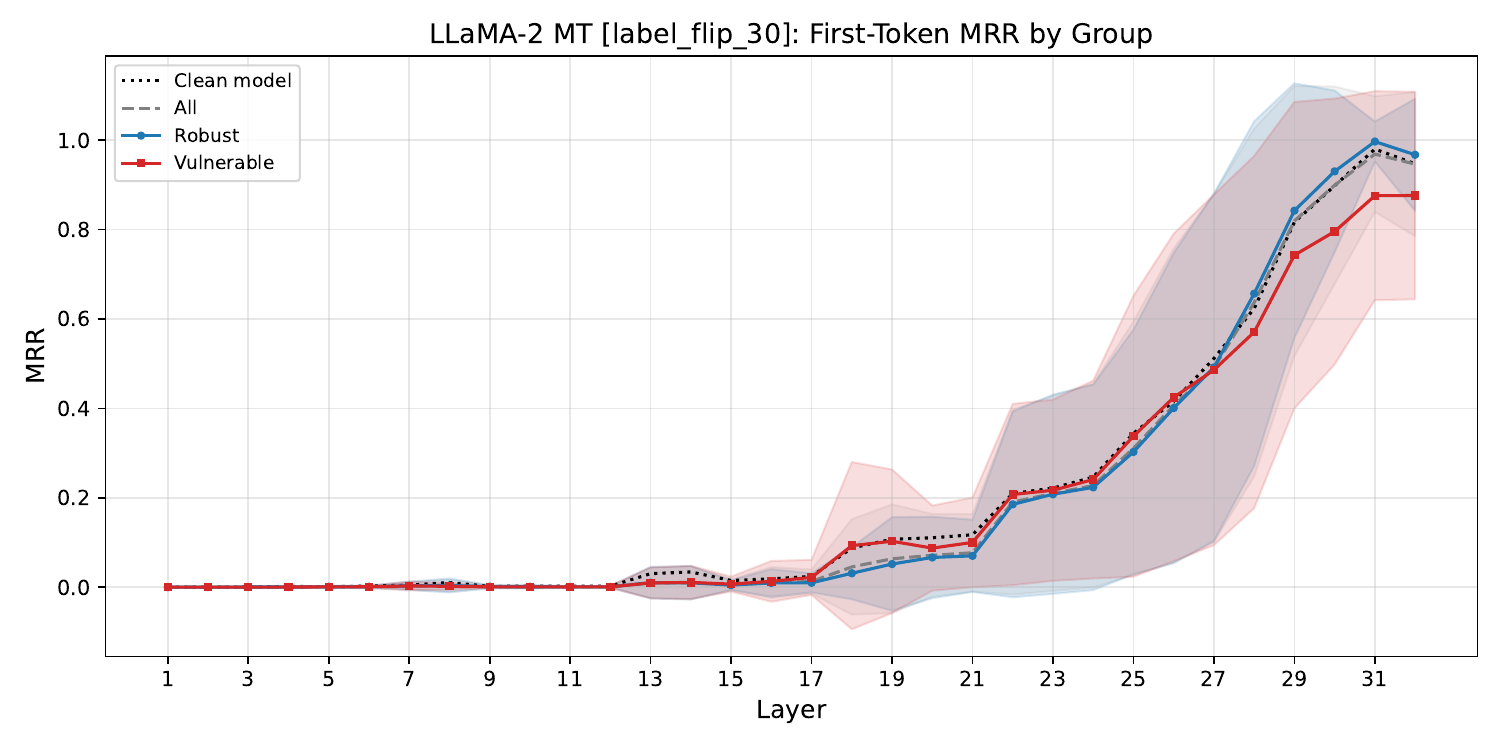}
  \end{subfigure}\hfill
  \begin{subfigure}[t]{0.32\textwidth}
    \includegraphics[width=\textwidth]{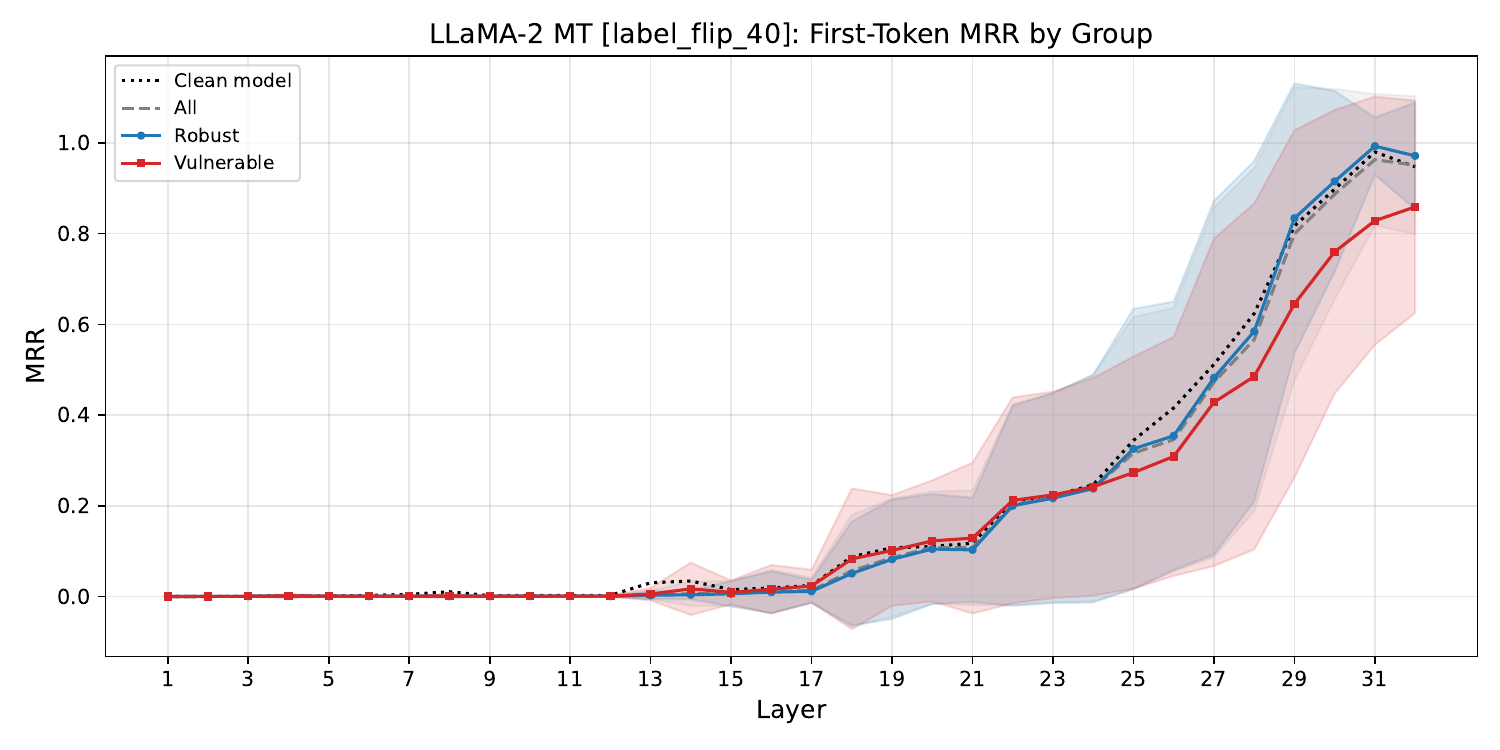}
  \end{subfigure}
  \caption{Robust vs.\ vulnerable stratification: \textbf{first-token Logit Lens MRR} under label-flip noise. 
  MRR measures how well each layer's representation predicts the correct answer token when projected through the language model head. LLaMA-2 vulnerable MRR collapses to 0.365 at the final layer under 40\% noise (vs.\ 0.740 for robust samples), the largest functional gap observed across all conditions. GPT-2 MRR is uniformly low ($<$0.20) for both groups, reflecting its limited QA capability.}
  \label{fig:stratification_qa_mrr_first}
\end{figure*}

\section{LoRA Ablation on GPT-2}
\label{app:lora_ablation}

To verify that differences between GPT-2 (full fine-tuning) and the larger models (QLoRA) reflect model scale rather than the fine-tuning paradigm, we train GPT-2 with LoRA on \ac{SC} under all noise conditions. We compare full fine-tuning and LoRA across noise types and corruption rates.

Key observations:
\begin{itemize}
    \item Label-flip noise is the only type that substantially degrades performance, with influence-based selection causing more damage (75.5\%) than random selection (82.3\%) at 40\% corruption.
    \item \ac{TN} and \ac{GN} have a negligible effect on accuracy regardless of corruption rate or selection strategy.
    \item Full fine-tuning and LoRA produce nearly identical accuracy under label-flip 40\% noise (75.5\% vs.\ 73.9\%), indicating that the fine-tuning method does not drive the representational differences observed across model scales in our main experiments.
\end{itemize}

\section{Multi-Seed Stability Analysis}
\label{app:seed_stability}

\subsection{Task Performance Across Seeds}

To assess the stability of our findings under different random seeds, we train Llama-2 on \ac{SC} with label-flip 40\% noise using three additional seeds (1, 7, 22) beyond the default seed 42.

\begin{table}[t]
\caption{LLaMA-2 sentiment accuracy across random seeds under 40\% label-flip noise. The three seeds produce qualitatively different outcomes: no degradation (seed 1), moderate degradation (seed 22), and catastrophic collapse (seed 7).}
\label{tab:seed_performance}
\centering
\small
\begin{tabular}{lcc}
\toprule
\textbf{Seed} & \textbf{Clean} & \textbf{Label-flip 40\%} \\
\midrule
1  & 94.5\% & 95.7\% (+1.2\%) \\
22 & 94.5\% & 83.5\% ($-$11.0\%) \\
7  & 94.0\% & 27.2\% (collapse) \\
42 & \multicolumn{2}{c}{(default; main experiments)} \\
\bottomrule
\end{tabular}
\end{table}

As shown in Table~\ref{tab:seed_performance}, the three seeds produce qualitatively different outcomes under identical noise conditions: seed 1 shows no degradation, seed 22 shows moderate degradation ($-$11\%), and seed 7 collapses entirely.
This wide variance indicates that LLaMA-2 under 40\% label-flip noise operates near a critical threshold --- the noise level is severe enough that the random initialisation and data ordering determined by the seed can tip the optimisation trajectory toward either a robust or a collapsed solution.
This finding highlights that single-seed evaluations may underestimate the true variance of noise effects.

\subsection{Clean-vs-Clean CKA Baseline}

To establish a ceiling for CKA and confirm that noise-induced CKA drops are not attributable to seed variance alone, we compute CKA between clean models trained with different random seeds.

The clean-vs-clean CKA floor of 0.890 is far above the noise-condition values (0.11--0.42), confirming that the representational changes reported in our main experiments reflect genuine noise effects rather than stochastic training variance.

\section{Five-Token MRR Results}
\label{app:five-mrr}

\begin{figure*}[t]
  \centering
  \begin{subfigure}[t]{0.49\textwidth}
    \includegraphics[width=\textwidth]{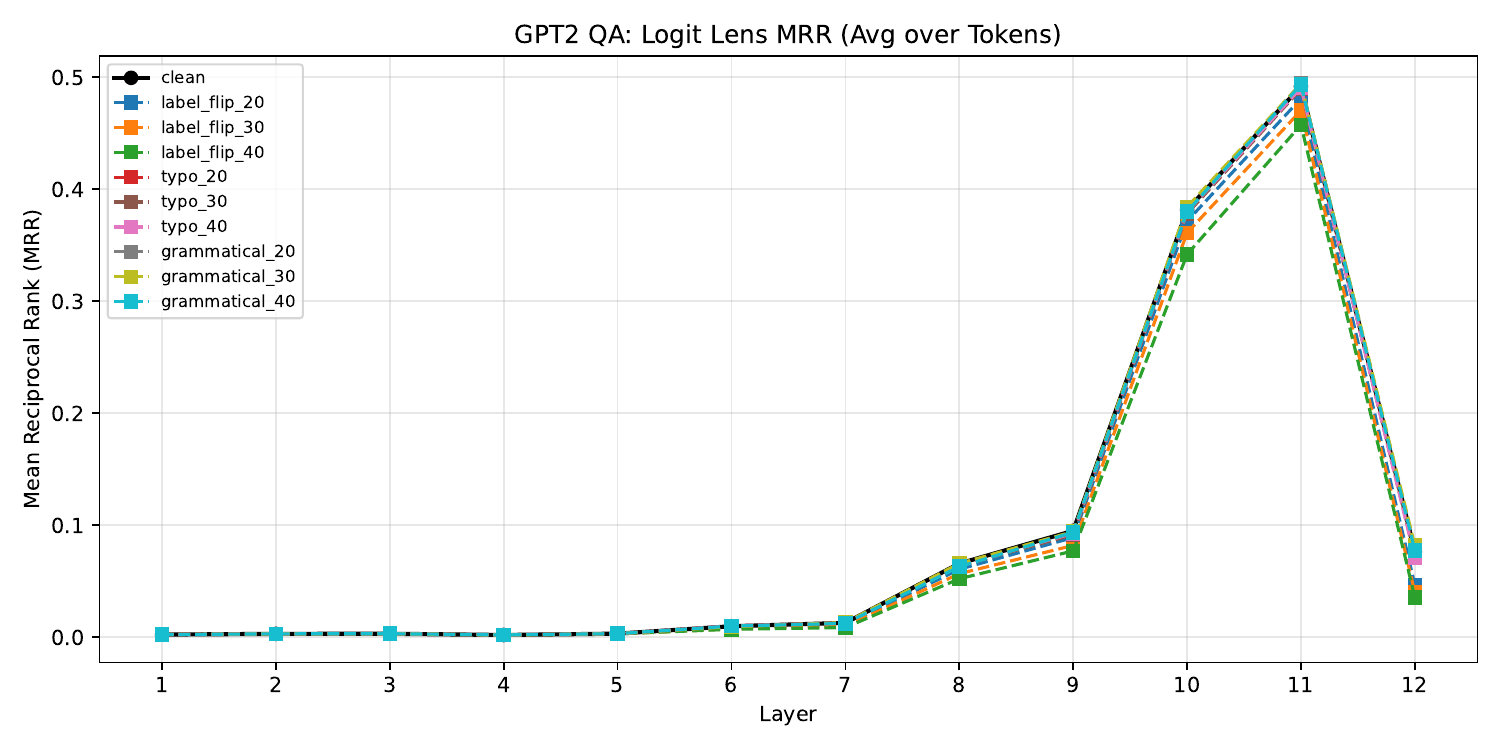}
    \caption{QA - GPT-2 Small}
  \end{subfigure}
  \hfill
  \begin{subfigure}[t]{0.49\textwidth}
    \includegraphics[width=\textwidth]{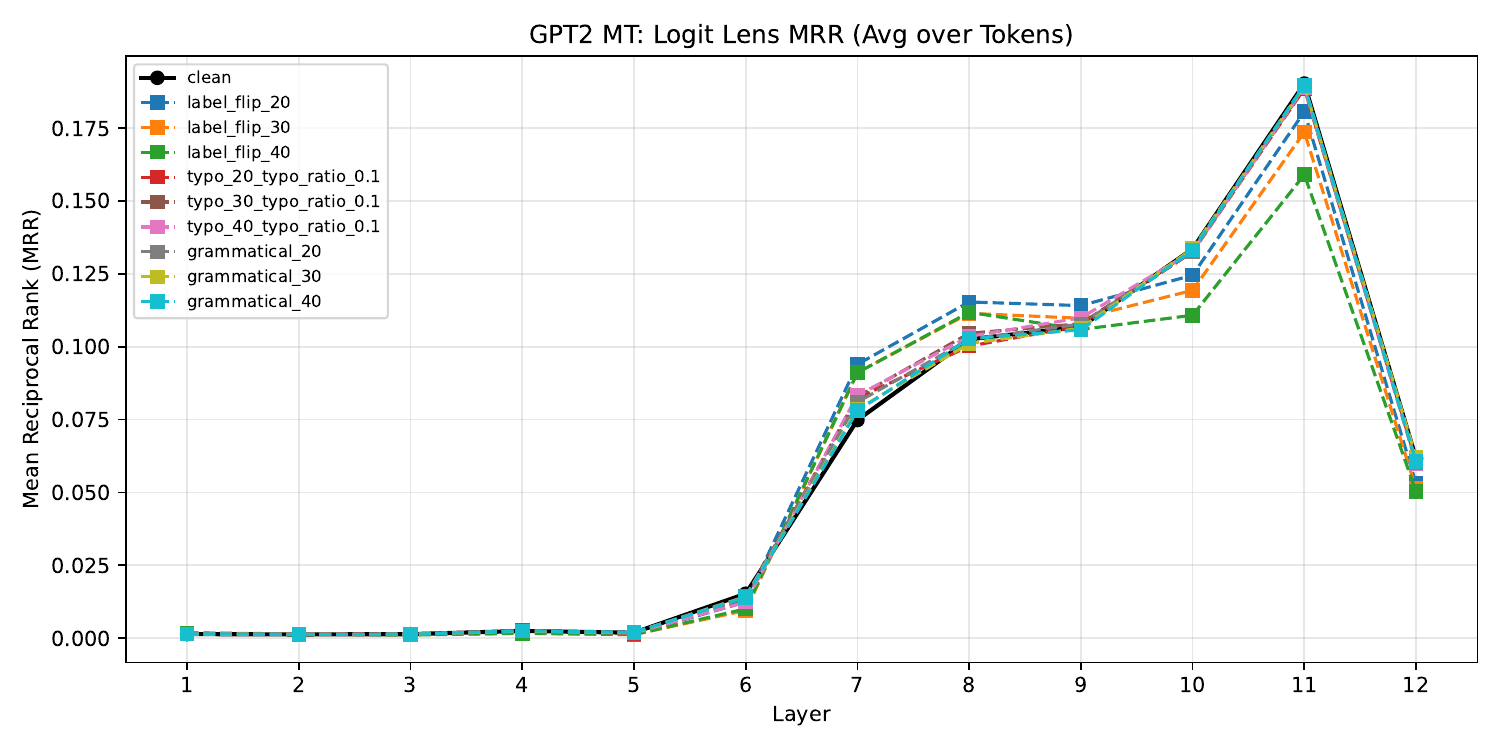}
    \caption{MT - GPT-2 small}
  \end{subfigure}
  \begin{subfigure}[t]{0.49\textwidth}
    \includegraphics[width=\textwidth]{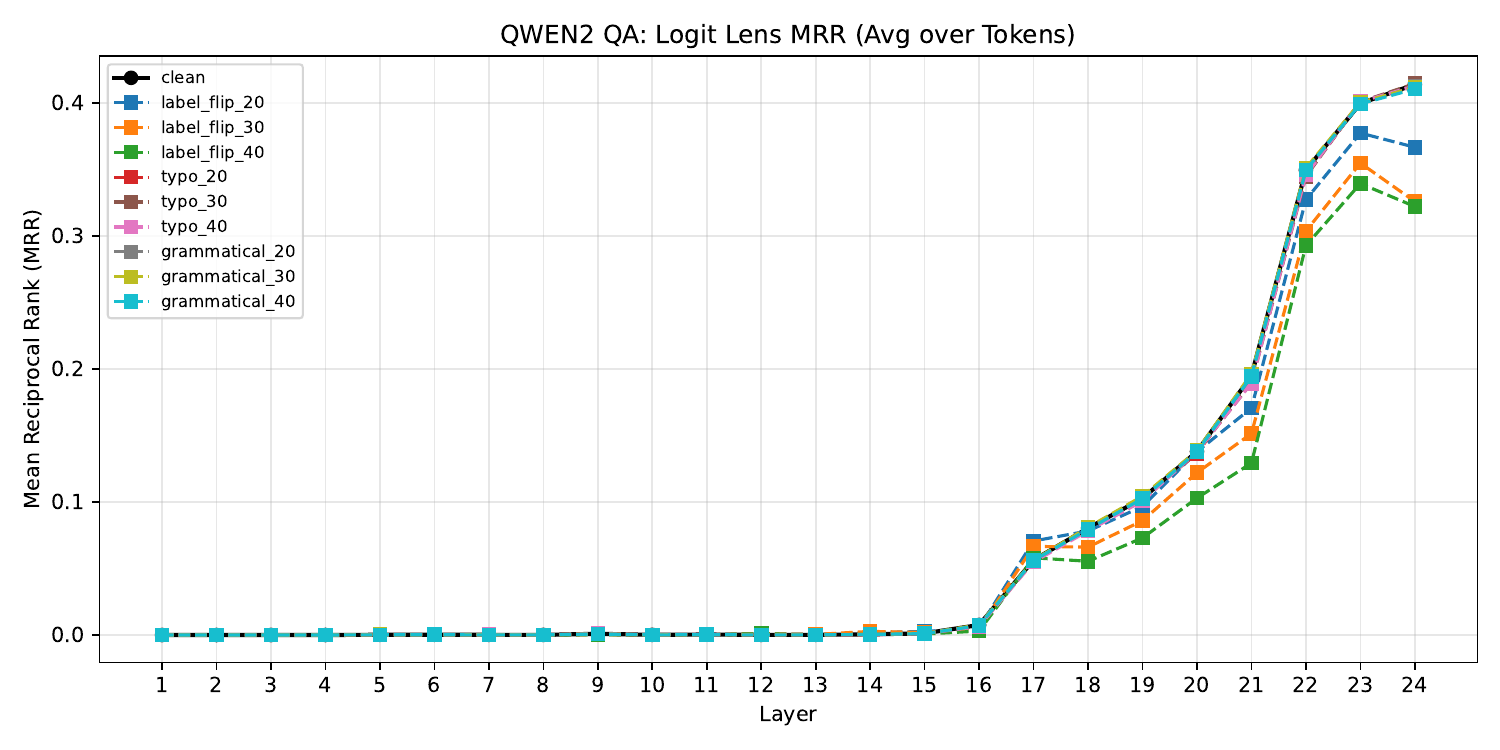}
    \caption{QA - Qwen2-0.5B}
  \end{subfigure}
  \hfill
  \begin{subfigure}[t]{0.49\textwidth}
    \includegraphics[width=\textwidth]{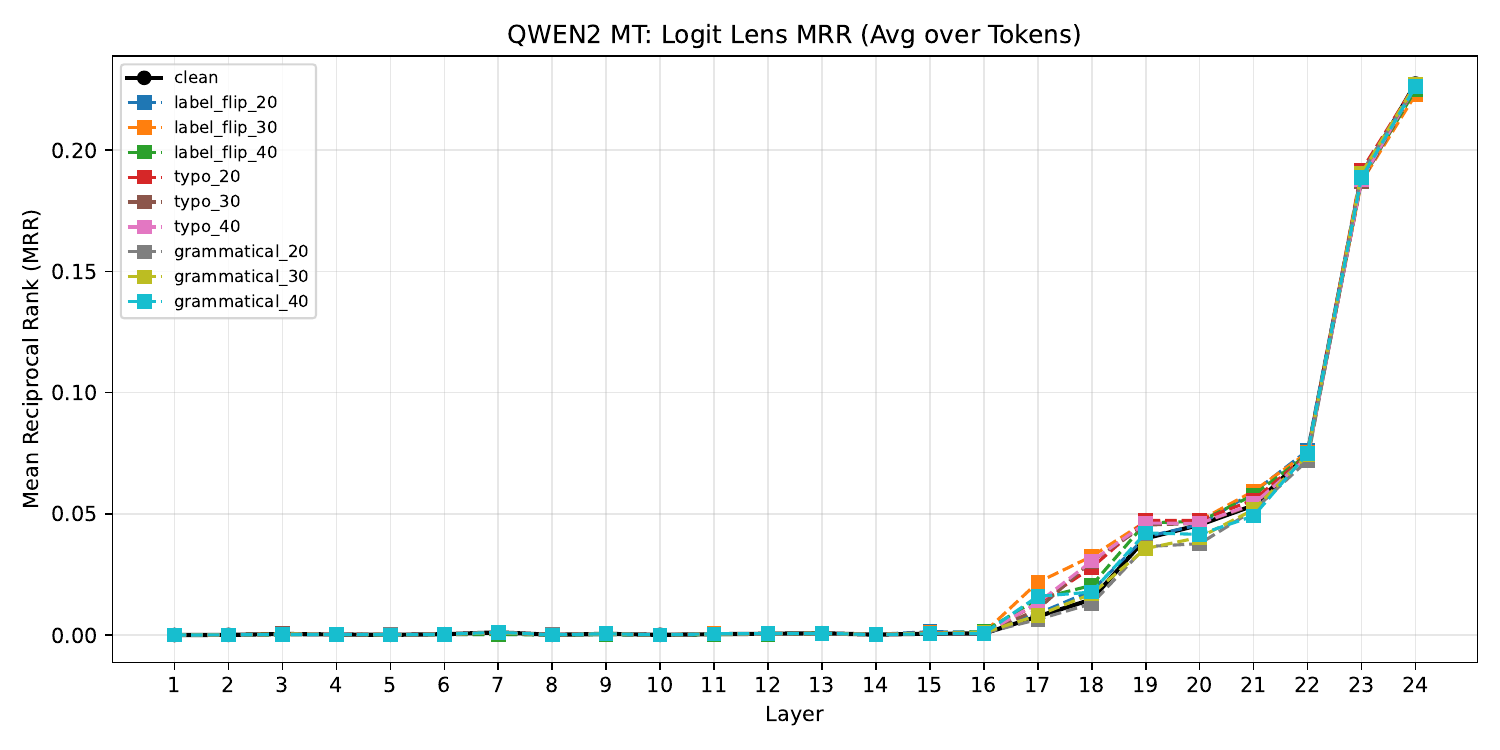}
    \caption{MT - Qwen2-0.5B}
  \end{subfigure}
    \begin{subfigure}[t]{0.49\textwidth}
    \includegraphics[width=\textwidth]{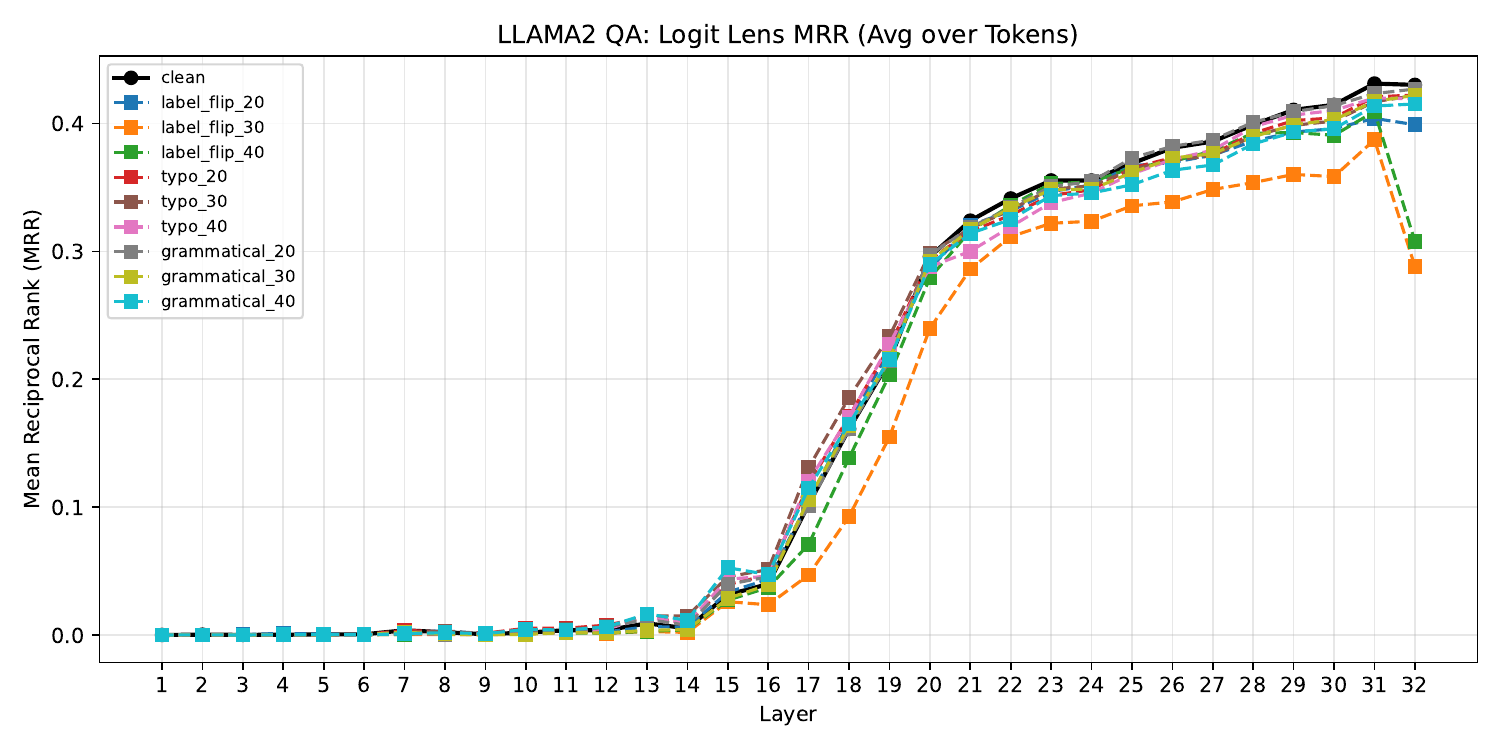}
    \caption{QA - Llama2-7B}
  \end{subfigure}
  \hfill
  \begin{subfigure}[t]{0.49\textwidth}
    \includegraphics[width=\textwidth]{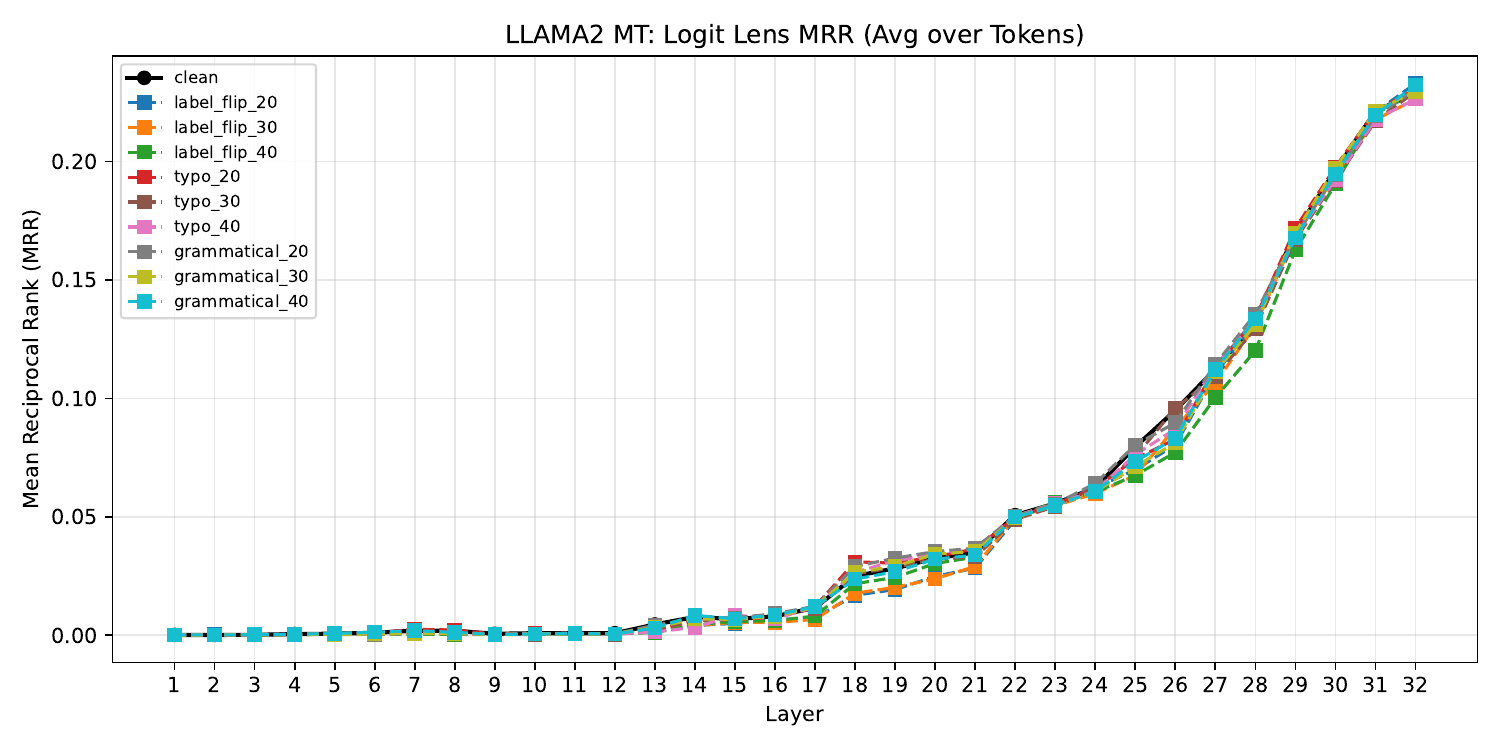}
    \caption{MT - Llama2-7B}
  \end{subfigure}
  
   \caption{Layer-wise top-5 MRR for GPT-2 Small (124M), Qwen-2 (0.5B) and Llama-2 (7B).}
   \label{fig:five_mrr}
\end{figure*}

We extend the first-token MRR analysis from \autoref{sec:layerwise} by computing the average MRR over the first five target tokens. At each layer, the hidden state is projected through the language model head, and we compute the reciprocal rank for each of the five tokens independently using autoregressive generation. The results, shown in \autoref{fig:five_mrr}, are consistent with the first-token MRR patterns reported in the main text.
  
\section{Teacher Forced Five-Token MRR Results}
\label{app:token_acc}

For each evaluation sample, we compute token accuracy under a teacher-forced setting: at each of the first five target positions, the model receives the ground-truth prefix tokens and predicts the next token.
Token accuracy at layer $\ell$ is defined as:
\begin{equation}
    \text{TokAcc}_\ell = \frac{1}{|S|} \sum_{s \in S} \frac{1}{5} \sum_{j=1}^{5} \mathbf{1}\!\left[\argmax p_\ell(\cdot \mid t_{<j}^s) = t_j^s\right]
\end{equation}
where $t_j^s$ is the $j$-th target token for the sample $s$ and $p_\ell(\cdot \mid t_{<j}^s)$ is the distribution obtained by projecting the layer-$\ell$ hidden state through the language model head.

\begin{figure*}[t]
  \centering
  \begin{subfigure}[t]{0.49\textwidth}
    \includegraphics[width=\textwidth]{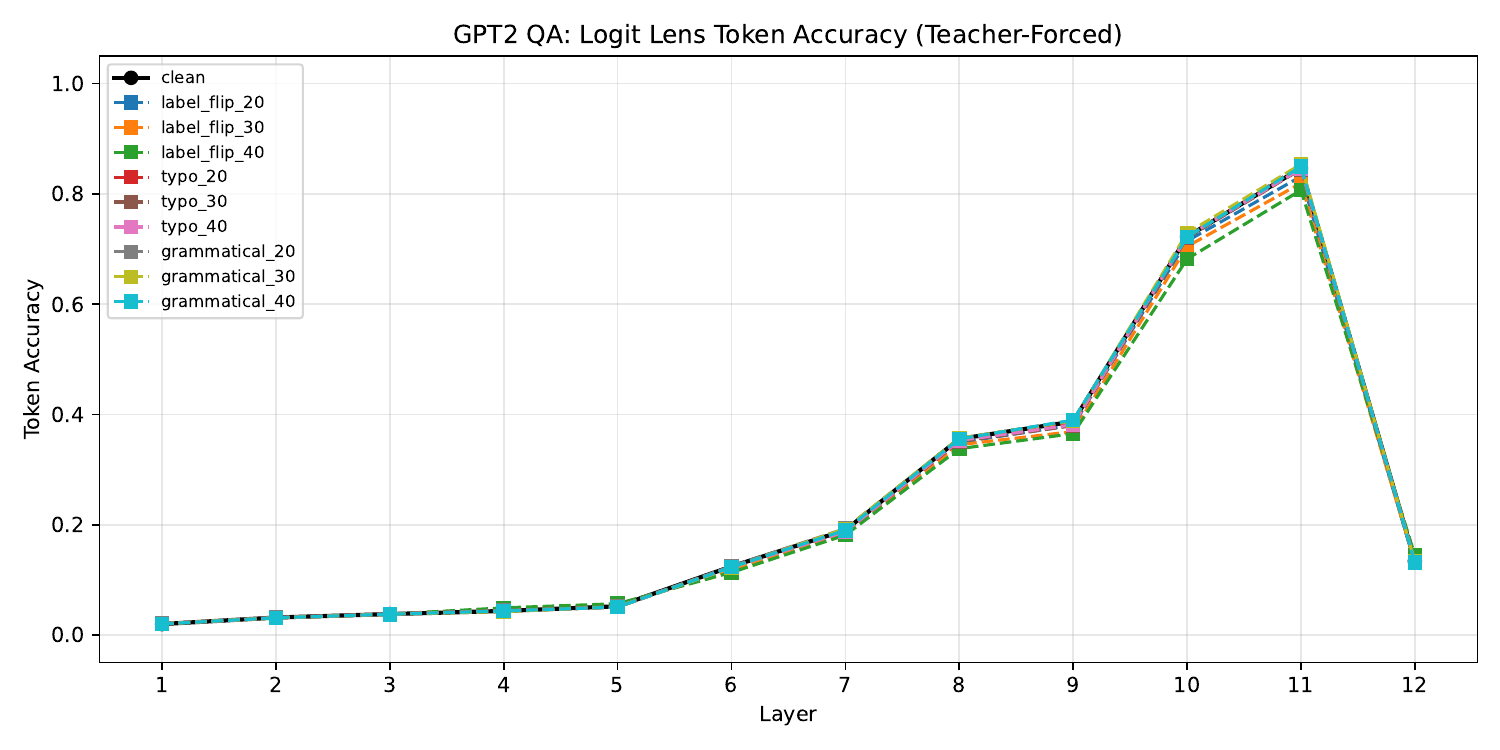}
    \caption{QA - GPT-2 Small}
  \end{subfigure}
  \hfill
  \begin{subfigure}[t]{0.49\textwidth}
    \includegraphics[width=\textwidth]{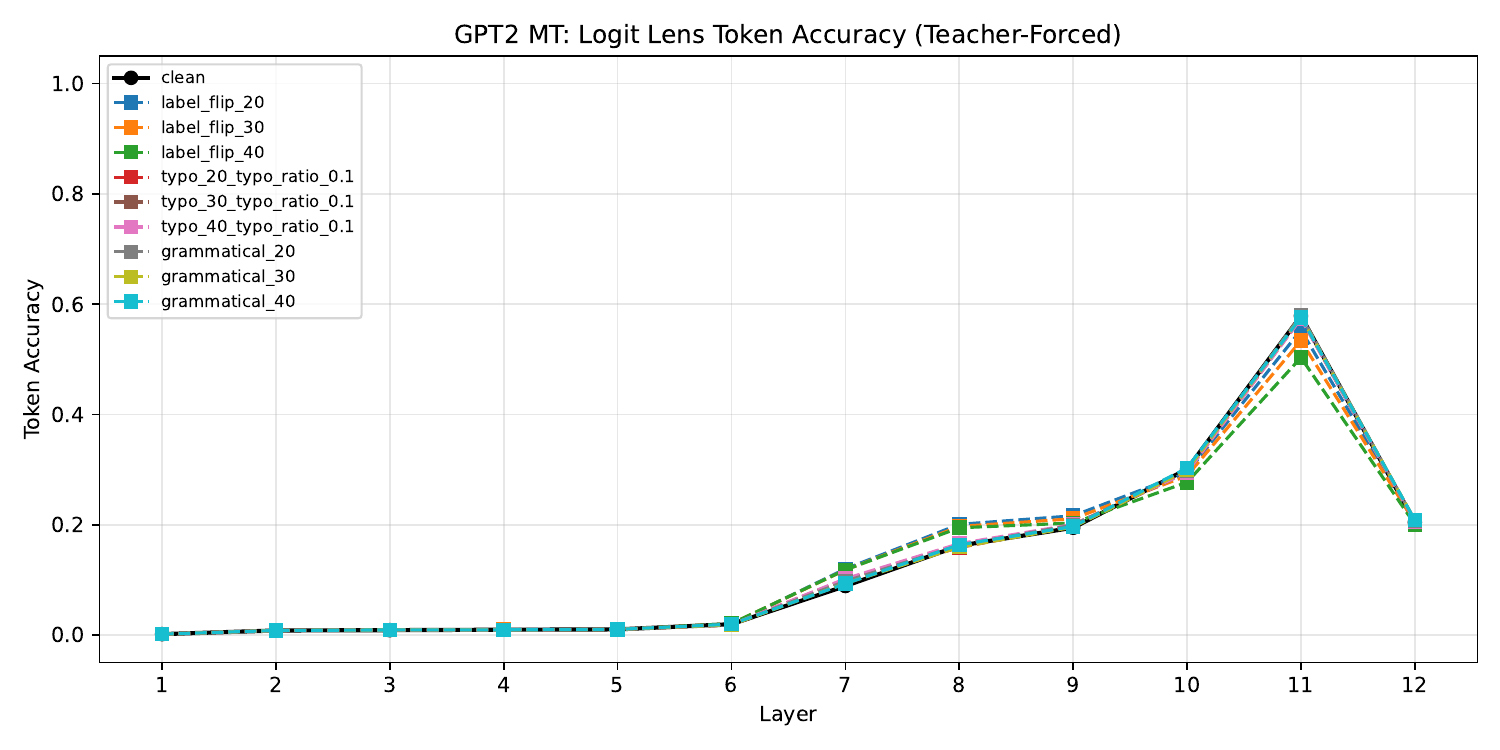}
    \caption{MT - GPT-2 Small}
  \end{subfigure}
  \hfill
  \begin{subfigure}[t]{0.49\textwidth}
    \includegraphics[width=\textwidth]{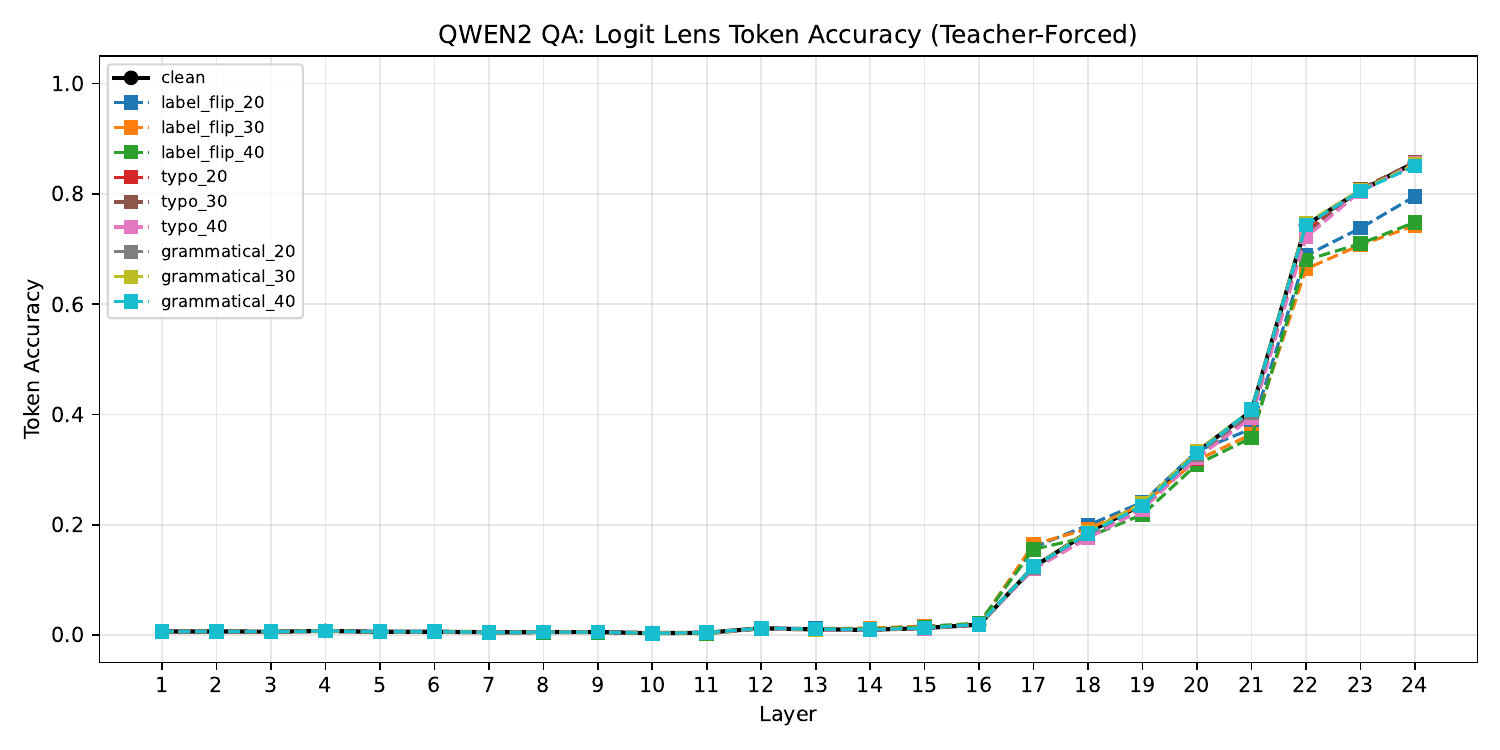}
    \caption{QA - Qwen2-0.5B}
  \end{subfigure}
  \hfill
  \begin{subfigure}[t]{0.49\textwidth}
    \includegraphics[width=\textwidth]{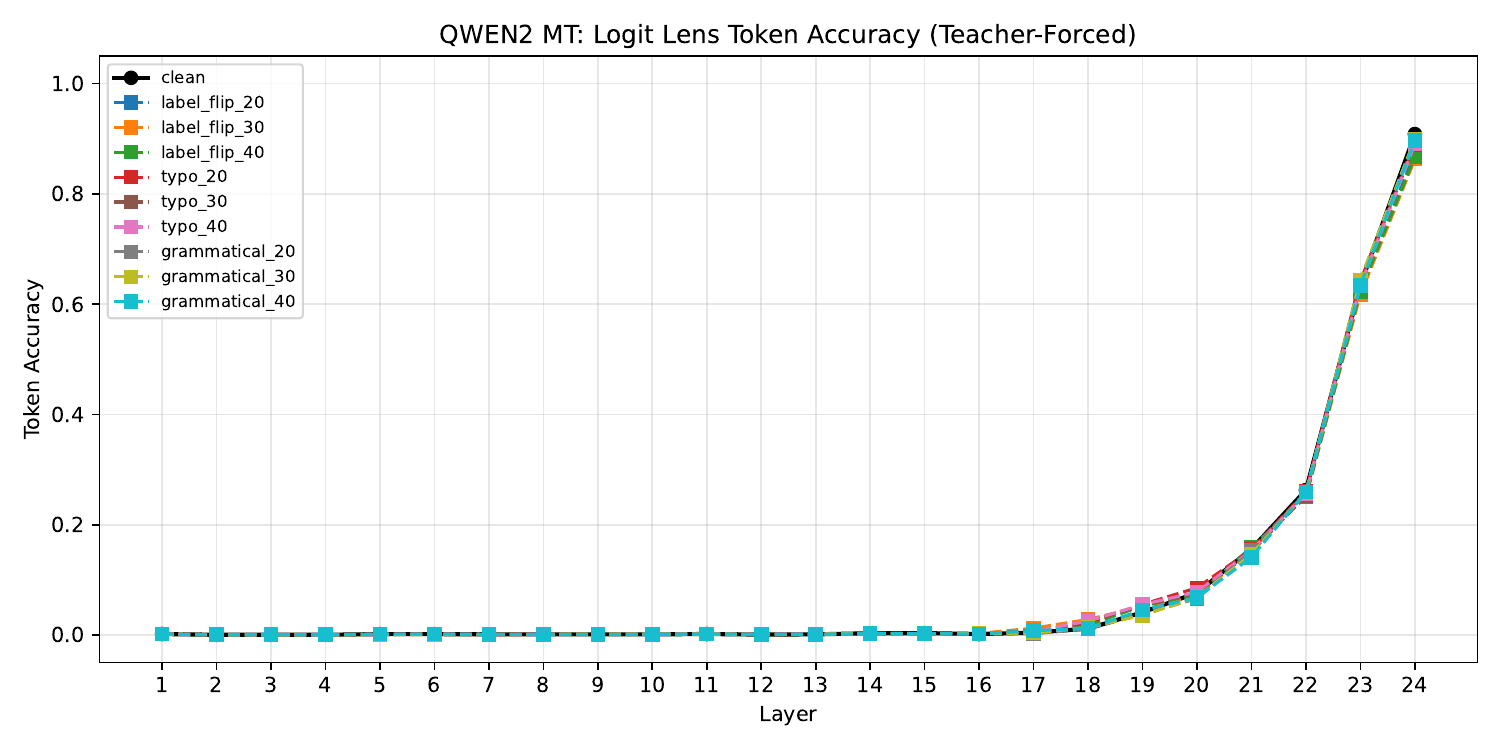}
    \caption{MT - Qwen2-0.5B}
  \end{subfigure}
    \begin{subfigure}[t]{0.49\textwidth}
    \includegraphics[width=\textwidth]{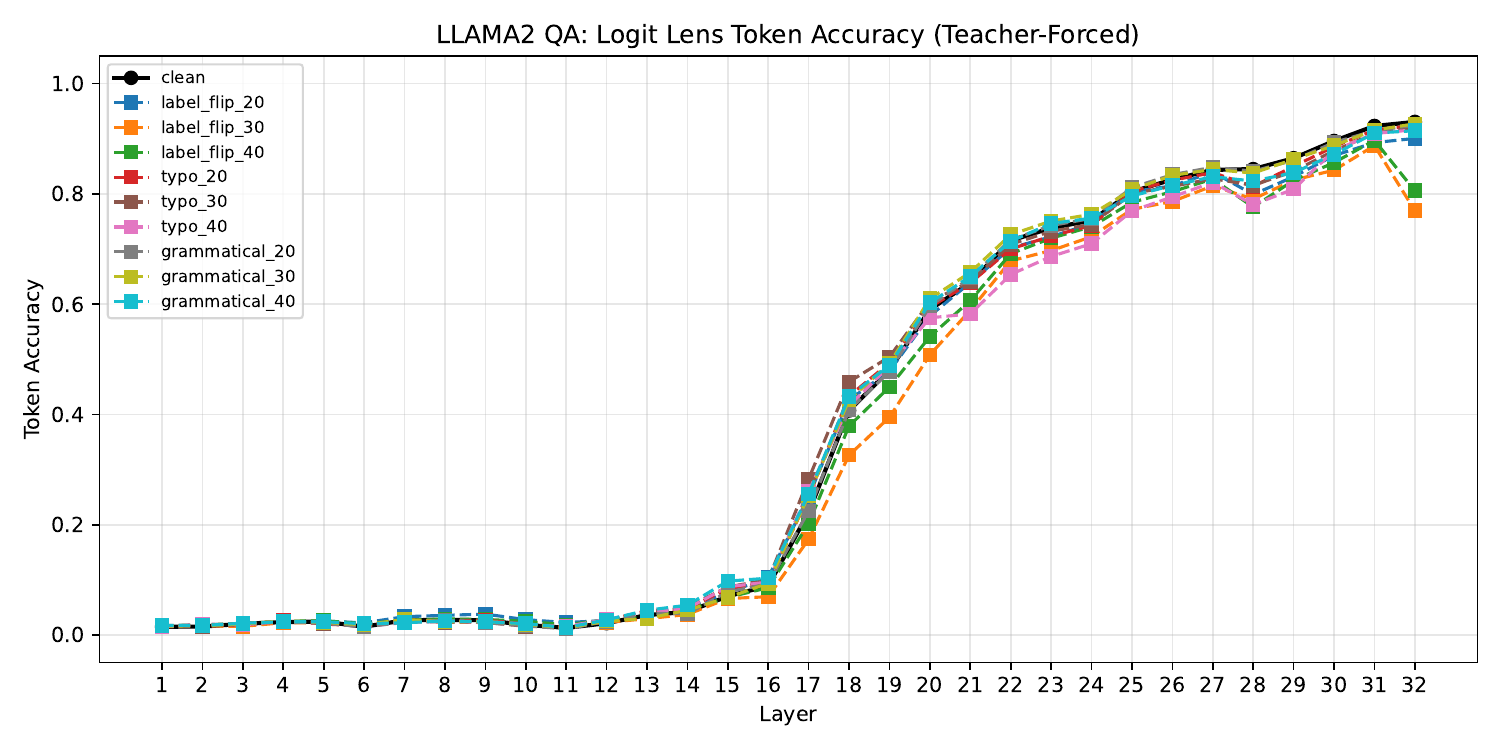}
    \caption{QA - Llama2-7B}
  \end{subfigure}
  \hfill
  \begin{subfigure}[t]{0.49\textwidth}
    \includegraphics[width=\textwidth]{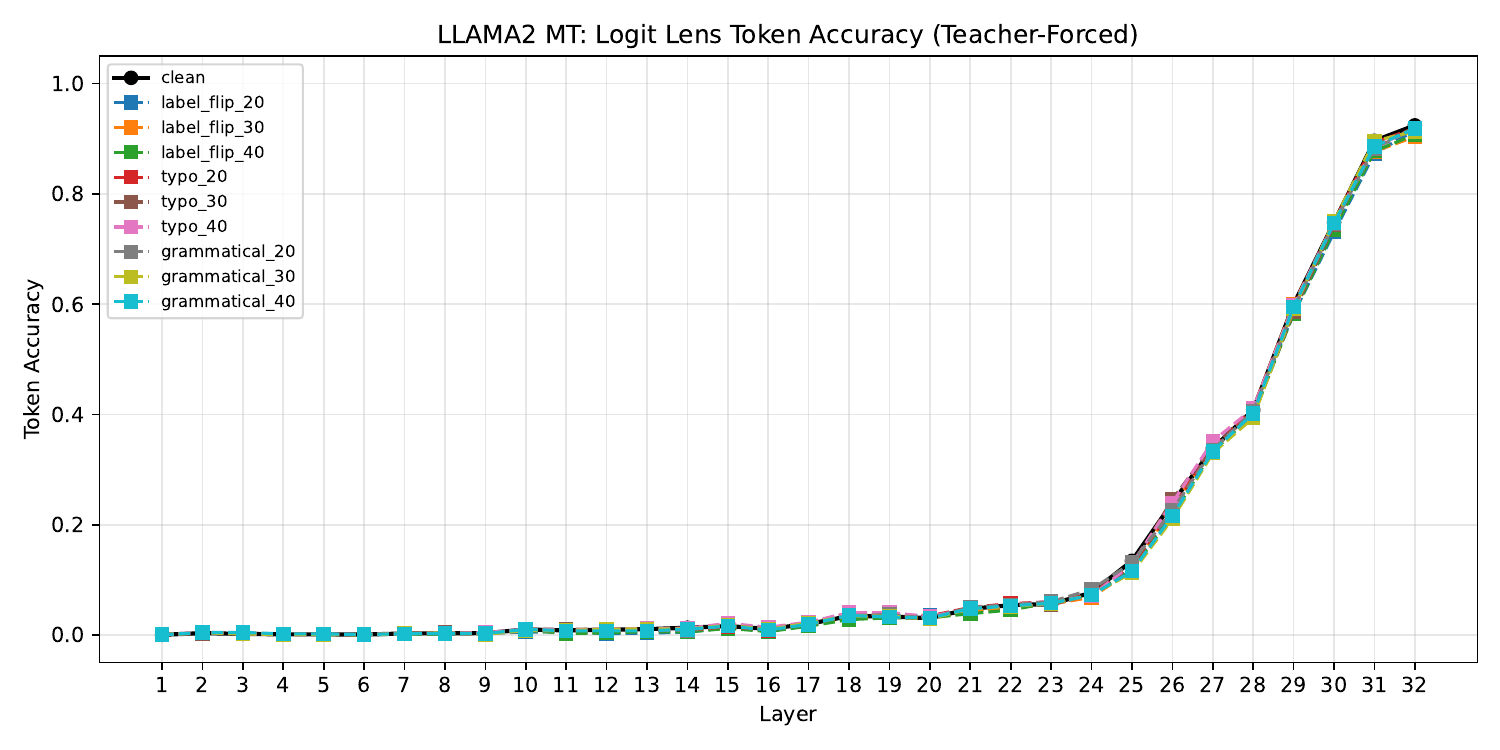}
    \caption{MT - Llama2-7B}
  \end{subfigure}
  \caption{Layer-wise top 5 token accuracy for the 3 models on (a) question answering and 
(b) machine translation under all noise conditions. At each layer, hidden states are projected through 
the LM head, and accuracy is computed as the fraction of the first 5 generated tokens matching the reference.}
  \label{fig:token_acc}
\end{figure*}

This metric complements MRR by providing a binary measure of prediction correctness rather than a rank-based measure, and is less sensitive to near-miss rankings. The results are shown in \autoref{fig:token_acc}.

\end{document}